\documentclass[journal, 10pt,twocolumn]{IEEEtran}

\usepackage{amssymb}
\usepackage[all,arc]{xy}
\usepackage{mathrsfs}
\usepackage{algorithm2e}
\usepackage{algorithmic}
\usepackage{amsbsy}
\usepackage{amsfonts}
\usepackage{amsmath}
\usepackage{amssymb}
\usepackage{amsthm}
\usepackage{array}
\usepackage{authblk}

\usepackage{bbm} 
\usepackage[font=footnotesize,labelfont=bf]{caption}
\usepackage{color}
\usepackage[mathscr]{eucal}
\usepackage{mathrsfs}

\usepackage{float}
\usepackage{hyperref} 
\usepackage{hhline}

\usepackage{stmaryrd}
\usepackage[font=scriptsize,labelfont=scriptsize]{subcaption}
\usepackage{graphicx}
\usepackage{url}
\usepackage{verbatim}
\usepackage{adjustbox}
\usepackage[usenames,dvipsnames,svgnames,table]{xcolor}
\newcommand{\Npatches}{N_{\text{patches}}}

\usepackage{tikz}
\usetikzlibrary{shapes,arrows}
\tikzstyle{decision} = = [rectangle, draw, fill=red!20, 
    text width=10em, text centered, rounded corners, minimum height=2em]
\tikzstyle{block1} = [rectangle, draw, fill=blue!20, 
    text width=20em, text centered, rounded corners, minimum height=2em]
\tikzstyle{block2} = [rectangle, draw, fill=blue!20, 
    text width=8em, text centered, rounded corners, minimum height=2em]
\tikzstyle{line} = [draw, -latex']
\tikzstyle{Loop1} = [rectangle, draw, fill=yellow!20, 
    text width=12em, text centered, rounded corners, minimum height=2em]    
\tikzstyle{Loop2} = [rectangle, draw, fill=yellow!20, 
    text width=15em, text centered, rounded corners, minimum height=2em]

\DeclareMathOperator*{\argmax}{arg\,max}

 \title{Unsupervised Clustering and Active Learning of Hyperspectral Images with Nonlinear Diffusion}

\author{James M. Murphy \and ~ Mauro Maggioni
\IEEEcompsocitemizethanks{
\IEEEcompsocthanksitem J.M. Murphy is with the Department of Mathematics at Tufts University; email: JM.Murphy@tufts.edu\\
\IEEEcompsocthanksitem M. Maggioni is with the Department of Mathematics, Department of Applied Mathematics and Statistics, Institute of Data Intensive Engineering and Science, and the Mathematical Institute of Data Sciences at Johns Hopkins University; email: mauro.maggioni@jhu.edu
}
}
        
\begin{document}

\maketitle


\begin{abstract}The problem of unsupervised learning and segmentation of hyperspectral images is a significant challenge in remote sensing.  The high dimensionality of hyperspectral data, presence of substantial noise, and overlap of classes all contribute to the difficulty of automatically clustering and segmenting hyperspectral images.  We propose an unsupervised learning technique called spectral-spatial diffusion learning (DLSS) that combines a geometric estimation of class modes with a diffusion-inspired labeling that incorporates both spectral and spatial information.  The mode estimation incorporates the geometry of the hyperspectral data by using diffusion distance to promote learning a unique mode from each class.  These class modes are then used to label all points by a joint spectral-spatial nonlinear diffusion process.  A related variation of DLSS is also discussed, which enables active learning by requesting labels for a very small number of well-chosen pixels, dramatically boosting overall clustering results.  Extensive experimental analysis demonstrates the efficacy of the proposed methods against benchmark and state-of-the-art hyperspectral analysis techniques on a variety of real datasets, their robustness to choices of parameters, and their low computational complexity.  
\end{abstract}


\section{Introduction}\label{Intro}

\subsection{Machine Learning for Hyperspectral Data}

Hyperspectral imagery (HSI) has emerged as a significant data source in a variety of scientific fields, including medical imaging \cite{Lu2014}, chemical analysis \cite{Wang2016}, and remote sensing \cite{Eismann2012}.  Hyperspectral sensors capture reflectance at a sequence of localized electromagnetic ranges, allowing for precise differentiation of materials according to their spectral signatures.  Indeed, the power of hyperspectral imagery for material discrimination has led to its proliferation, making manual analysis of hyperspectral data infeasible in many cases.  The large data size of HSI, combined with their high dimensionality, demands innovative methods for storage and analysis.  In particular, efficient machine learning algorithms are needed to automatically process and glean insight from the deluge of hyperspectral data now available.

The problem of \emph{HSI classification}, or supervised segmentation, is to label each pixel in a given HSI as belonging to a particular class, given a training set of labeled samples (pixels) from each class.  A variety of statistical and machine learning techniques have been used for HSI classification, including nearest-neighbor and nearest subspace methods \cite{Ma2010, Li2014},  support vector machines \cite{Melgani2004, Fauvel2008}, neural networks \cite{Ratle2010,Chen2016,Liang2016} and regression methods \cite{Qian2013, Li2013_1}.  These methods are design to perform well especially when the number of labeled training pixels is large.

The process of labeling pixels typically requires an expert and it is costly. This motivates the design of machine learning techniques that require little or no labeled training data. So on the other end of the spectrum from classification, we have the problem of {\em{HSI clustering}}, or {\em{unsupervised}} segmentation, which has the same goal as HSI classification, but no labeled training data is available.  This is considerably more challenging, and is an ill-posed problem unless further assumptions are made, for example about the distribution of the data and how it relates to the unknown labels.  Recent techniques for hyperspectral clustering include those based on particle swarm optimization \cite{Paoli2009}, Gaussian mixture models (GMM) \cite{Acito2003}, nearest neighbor clustering \cite{Cariou2015}, total variation methods \cite{Zhu2017}, density analysis \cite{Chen2017}, sparse manifold models \cite{Elhamifar2011, Elhamifar2013}, hierarchical nonnegative matrix factorization (HNMF) \cite{Gillis2015}, graph-based segmentation \cite{Meng2017}, and fast search and find of density peaks clustering (FSFDPC) \cite{Chen2017, Rodriguez2014, Zhang2016}.  

Another interesting modality is \emph{active learning} for HSI classification.  This is a supervised technique where a small, automatically but carefully chosen set of pixels is labeled, as opposed to the standard supervised learning setting, in which the labels are usually randomly selected.  Active learning can lead to high quality classification results with significantly fewer labeled samples than in the case of randomly selected training data.  Since far fewer training points are available in the active learning setting, the structure of the data may be analyzed with unsupervised learning, in order to decide which data points to query for labels.  Thus, active learning may be understood as a form of \emph{semisupervised learning} that exploits both global structure of the data---learned without supervision---and a small number of supervised training data points.  A variety of active learning methods have been successfully deployed in remote sensing \cite{Tuia2011}, including those based on relevance feedback \cite{Demir2015}, region-based heuristics \cite{Stumpf2014}, exploration-based heuristics \cite{Tuia2011_Using}, belief propagation \cite{Li2013_2}, support vector machines \cite{Tuia2009}, and regression \cite{Li2010}.  

Machine learning for HSI suffers from several major challenges.  First, the dimensionality of the data to be analyzed is high: it is not uncommon for the number of spectral bands in an HSI to exceed $200$.  The corresponding sampling complexity for such a high number of dimensions renders classical statistical methods inapplicable.  Second, clusters in HSI are typically nonlinear in the spectral domain, rendering methods that rely on having linear clusters ineffective.  Third, there is often significant noise and between-cluster overlap among HSI classes, due to the materials being imaged and poor sensing conditions.  Finally, HSI images may be quite large, requiring machine learning methods with computational complexity essentially linear in the number of pixels.  

This article addresses the problems of HSI clustering and, relatedly, active learning, which overcome these significant challenges.  The methods we propose combine density-based methods with geometric learning through diffusion geometry \cite{Coifman2005,Coifman2006} in order to identify class modes.  This information is then used to propagate labels on training data to all data points through a nonlinear process that incorporates both spectral and spatial information.  The use of data-dependent diffusion maps for mode detection significantly improves over current state-of-the-art methods experimentally, and also enjoys robust theoretical performance guarantees \cite{Murphy2018}.  The use of diffusion distances exploits low-dimensional structures in the data, which allows the proposed method to handle data that is high-dimensional but intrinsically low-dimensional, even when nonlinear and noisy.  Moreover, the spectral-spatial labeling scheme takes advantage of the geometric properties of the data, and greatly improves the empirical performance of clustering when compared to labeling based on spectral information alone.  In addition, the proposed unsupervised method assigns to each data point a measure of confidence for the unsupervised label assignment.  This leads naturally to an active learning algorithm in which points with low confidence scores are queried for training labels, which then propagate through the remaining data.  The proposed algorithms enjoy nearly linear computational complexity in the number of pixels in the HSI and in the number of spectral dimensions, thus allowing for its application to large scenes.  Extensive empirical results, including comparisons with many state-of-the-art techniques, for our method applied to HSI clustering and active learning are in Sections \ref{subsec:HyperspectralClustering} and \ref{subsec:ActiveLearning}, respectively.


\subsection{Overview of Proposed Method}
\label{s:overview}
The proposed unsupervised clustering method is provided with data $X=\{x_{n}\}_{n=1}^N\subset\mathbb{R}^{D}$ (for HSI, $N$ = number of pixels and $D$ = number of spectral bands) and the number $K$ of classes, and outputs labels \{$y_{n}\}_{n=1}^{N},$ each $y_{n}\in\{1,\dots,K\}$, by proceeding in two steps:
\begin{itemize}
\item[1.] {\bf{Mode Identification}}: This step consists first in performing density estimation and analyzing the geometry of the data to find $K$ {\em{modes}} $\{x_{i}^{*}\}_{i=1}^{K}$, one for each class.  
\item[2.] {\bf{Labeling Points}}: Once the modes are learned, they are assigned a unique label.  Remaining points are labeled in a manner that preserves spectral and spatial proximity.  
\end{itemize}

By a mode, we mean a point of high density within a class, that is representative of the entire class. We assume $K$ is known, but otherwise we have no access to labeled data; in Section \ref{sec:Future} we discuss a method for estimating $K$.  

One of the key contributions of this article is to measure similarities in the spectral domain not with the widely used Euclidean distance or distances based on angles (correlations) between points, but with \emph{diffusion distance} \cite{Coifman2005,Coifman2006}, which is a data-dependent notion of distance that accounts for the geometry---linear or nonlinear---of the distribution of the data.  The motivation for this approach is to attain robustness with respect to the shape of the distributions corresponding to the different classes, as well as to high-dimensional noise. The modes, suitably defined via density estimation, are robust to noise, and the process we use to pick only one mode per class is based on diffusion distances.  The labeling of the points from the modes respects the geometry of the data, by incorporating proximity in both spectral and spatial domains.  

We model $X$ as samples from a distribution $\mu=\sum_{i=1}^{K}w_{i}\mu_{i},$
where each $\mu_{i}$ corresponds to the probability distribution of the spectra in class $i$, and the nonnegative weights $\{w_{i}\}_{i=1}^{K}$ correspond to how often each class is sampled, and satisfy $\sum_{i=1}^{K} w_{i}=1$.  More precisely, sampling $x\sim\mu$ means first sampling $Z\sim\text{Multinomial}(w_{1},\dots,w_{K})$, then sampling from $\mu_{i}$ conditioned on the event $Z=i\in\{1,\dots,K\}$.  

\subsubsection{Mode Identification}
The computation of the modes is a significant aspect of the proposed method, which we now summarize for a general dataset $X$, consisting of $K$ classes.  The mode identification algorithm outputs a point $x_{i}^{*}$ (``mode'') for each $\mu_{i}$.  We make the assumption that modes of the constituent classes can be characterized as a set of points $\{x_{i}^{*}\}_{i=1}^{K}$ such that
\begin{enumerate}
\item the empirical density of each $x_{i}^{*}$ is relatively high;
\item the diffusion distance between pairs $x_{i}^{*},x_{i'}^{*}$, for $i\neq i'$, is relatively large.
\end{enumerate}
The first assumption is motivated by the fact that points of high density ought to have nearest neighbors corresponding to a single class; the modes should thus produce neighborhoods of points that with high confidence belong to a single class.  However, there is no guarantee that the $K$ densest points will correspond to the $K$ unique classes: some classes may have a multimodal distribution, meaning that the class has several modes, each with potentially higher density than the densest point in another class.  The second assumption addresses this issue, requiring that modes belonging to different distributions are far away in diffusion distance.

\begin{figure}
\centering
\begin{subfigure}{.15\textwidth}
\includegraphics[width=\textwidth]{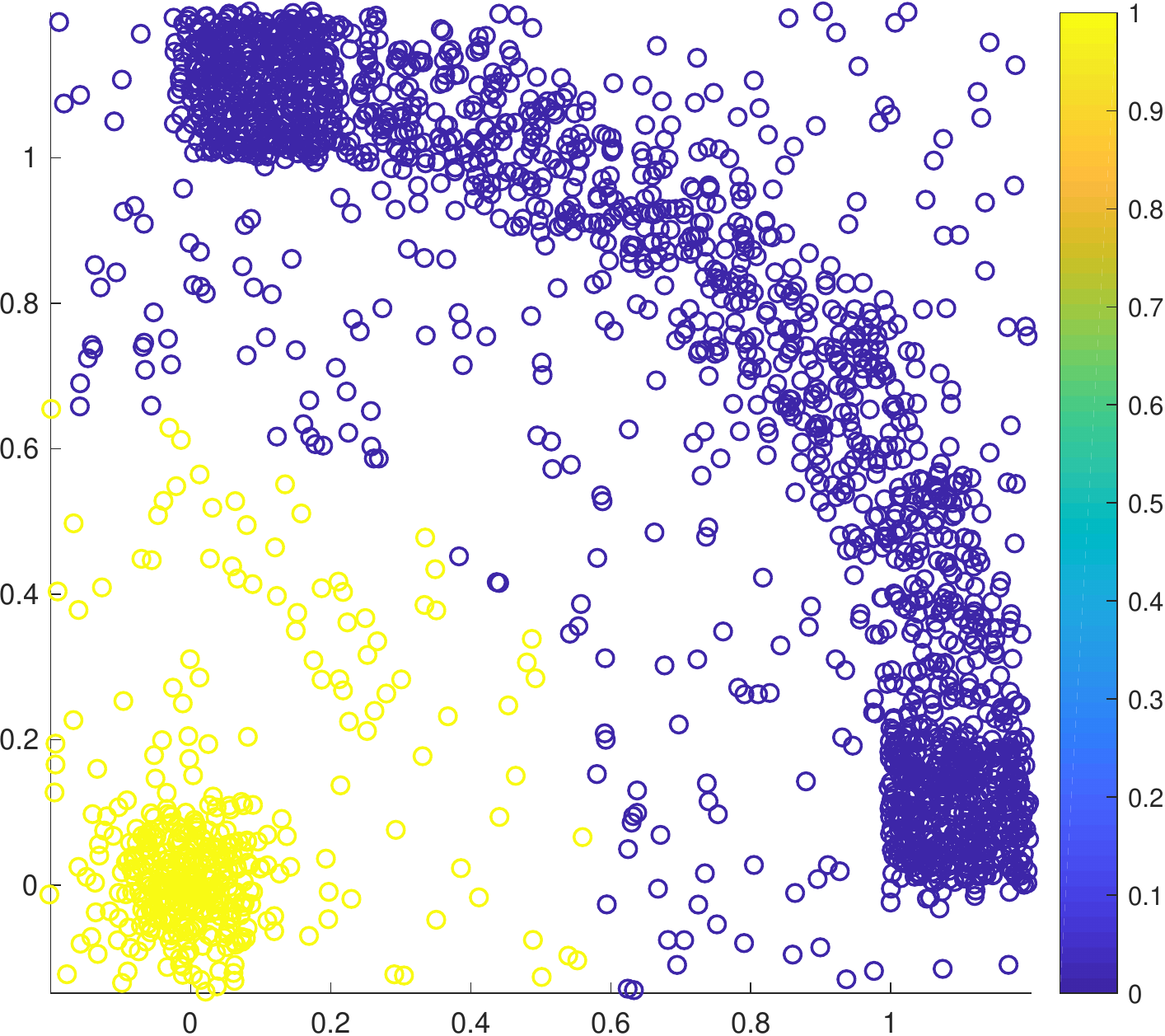}
\caption{Nonlinear, multimodal example data.}
\end{subfigure}
\begin{subfigure}{.15\textwidth}
\includegraphics[width=\textwidth]{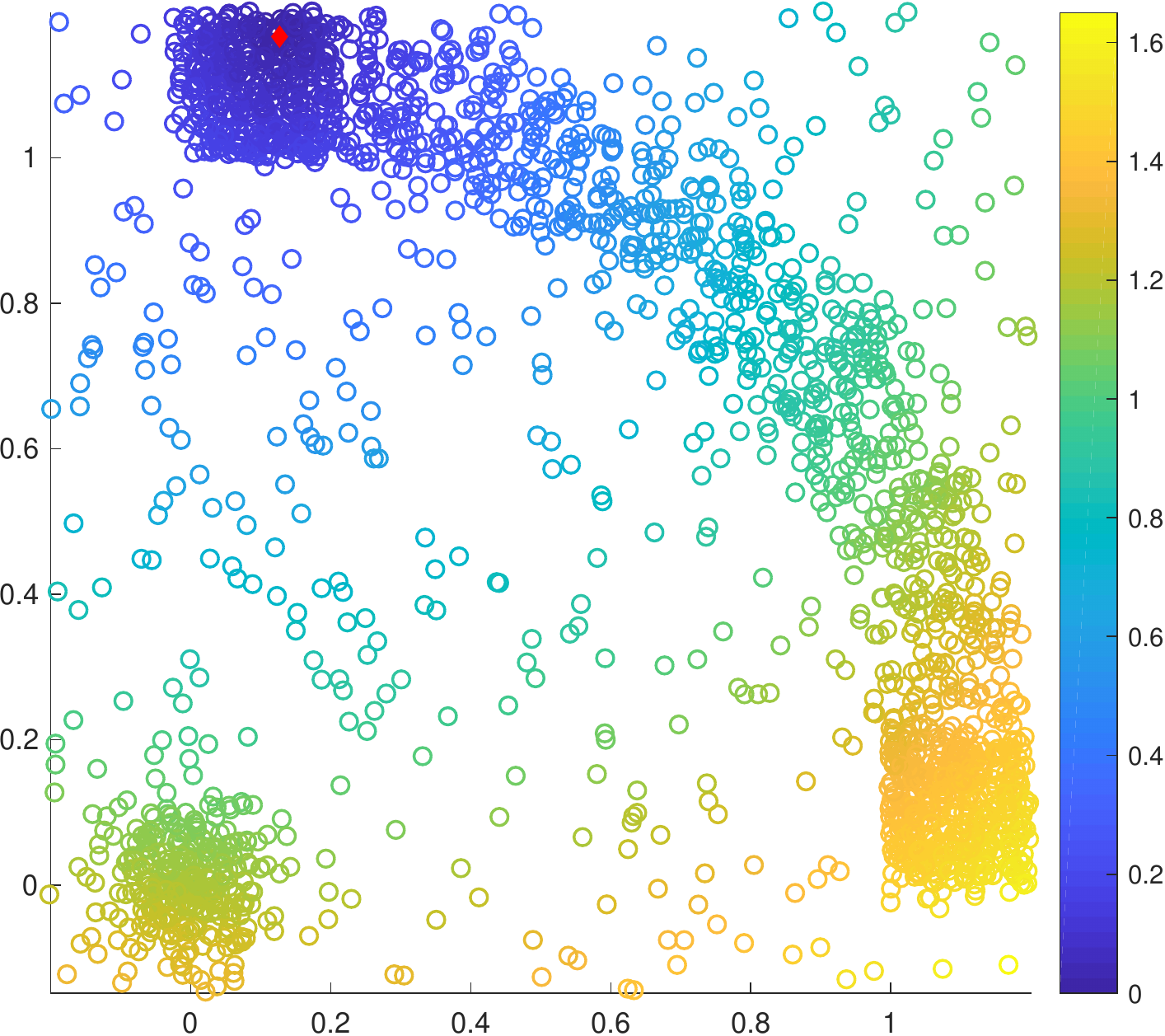}
\caption{Euclidean distance from $(0,1)$.}
\end{subfigure}
\begin{subfigure}{.15\textwidth}
\includegraphics[width=\textwidth]{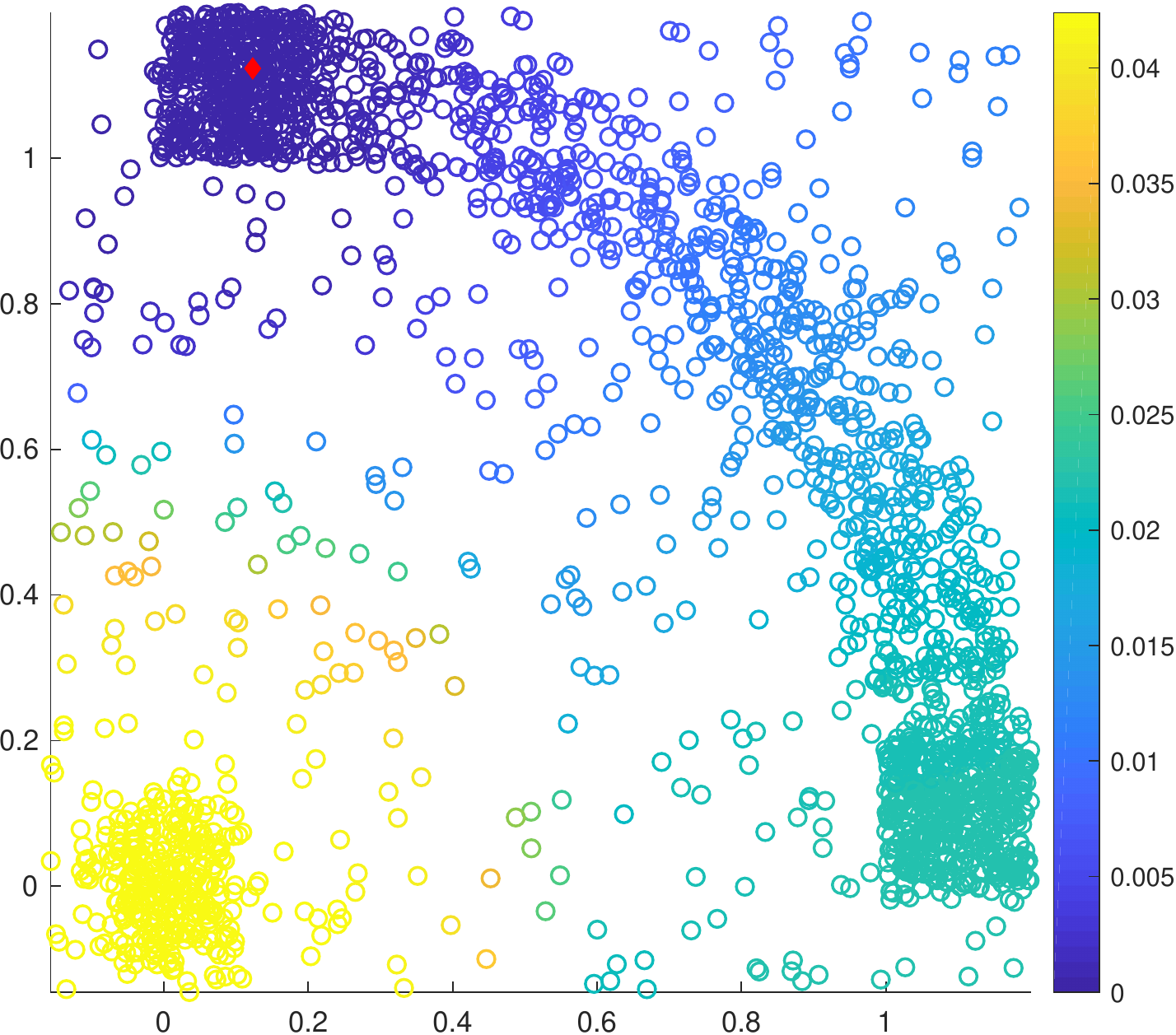}
\caption{Diffusion distance from $(0,1)$.}
\end{subfigure}
\caption{\label{fig:NonlinearComparison}
In this $2$-dimensional example, data is drawn from two distributions $\mu_1$ and $\mu_2$.  $\mu_1$ is a mixture of two isotropic Gaussians with means at $(0,1)$ and $(1,0)$, respectively, connected by a set of points uniformly sampled from a nonlinear, parabolic shape.  $\mu_2$ is an isotropic Gaussian with mean at $(0,0)$. Samples of uniform background noise are added and labeled according to their nearest neighbor among the two clusters.  The data is plotted and colored by cluster in subfigure (a). We plot the distances from the point $(0,1)$ in the Euclidean and diffusion distances in subfigures (b), (c), respectively.  The ``parabolic rectangle'' acts as a ``bridge'' between the two Gaussians and causes the high density regions near $(0,1)$ and $(1,0)$ to be closer in diffusion distance than they would be in the usual Euclidean distance.  The bridge is overcome efficiently with diffusion distance, because there are many paths with short edges connecting the high density regions across this bridge.}
\end{figure}
Enforcing that these modes are far apart in diffusion distance has several advantages over enforcing they are far apart in Euclidean distance.  Importantly, it leads, empirically, to a unique mode from each class.  This is true even when certain classes are multimodal.  Moreover, diffusion distances are robust with respect to the shape of the support of the distribution, and are thus suitable for identifying nonlinear clusters.  An instance of these advantages of diffusion distance is illustrated in the toy example Figure \ref{fig:NonlinearComparison}, with the results of the proposed mode detection algorithm in Figure \ref{fig:ToyExample}. We postpone the mathematical and algorithmic details to Section \ref{subsec:DD}.

\subsubsection{Labeling Points}
At this stage we assume that we found exactly one mode $x_{i}^{*}$ for each class, to which a unique and arbitrary class label is assigned.  The remaining points are now labeled in a two-stage scheme, which takes into account both spectral and spatial information.  It is known that the incorporation of spatial information with spectral information has the potential to improve machine learning of hyperspectral images, compared to using spectral information alone \cite{Fauvel2008,Li2013_2,Zhang2016,Tarabalka2009,Benedetto2012,Fauvel2013,Cahill2014,Cloninger2014,Wang2014,Benedetto2016}.  Spatial information is computed for each pixel by constructing a neighborhood of some fixed radius in the spatial domain, and considering the labels within this neighborhood.  For a given point, let \emph{spectral neighbor} refer to a near neighbor with distances measured in the spectral domain, and let \emph{spatial neighbor} refer to a near neighbor with distances measured in the spatial domain.  

In the first stage, a point is given the same label as its nearest spectral neighbor of higher density, unless that label is sufficiently different from the labels of the point's nearest spatial neighbors, in which case the point is left unlabeled.  This produces an incomplete labeling in which we expect the labeled points to be far from the spectral and spatial boundaries of the classes, since these are points that are unlikely to have conflicting spectral and spatial labels.  The first stage thus labels points using only spectral information, though spatial information may prevent a label from being assigned.

\begin{figure}
\begin{subfigure}{.24\textwidth}
\includegraphics[width=\textwidth]{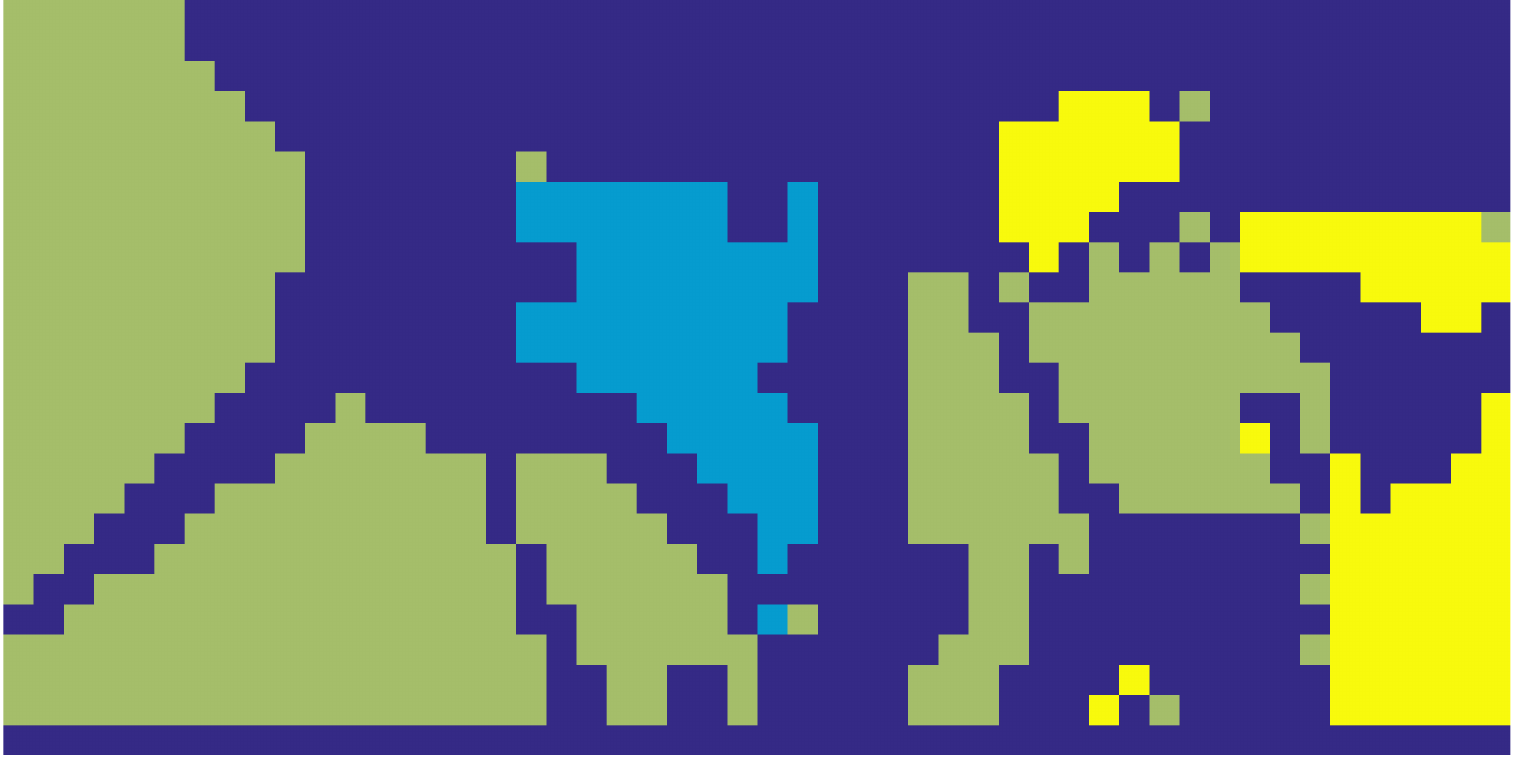}
\subcaption{First stage labeling, using spectral information only.}
\end{subfigure}
\begin{subfigure}{.24\textwidth}
\includegraphics[width=\textwidth]{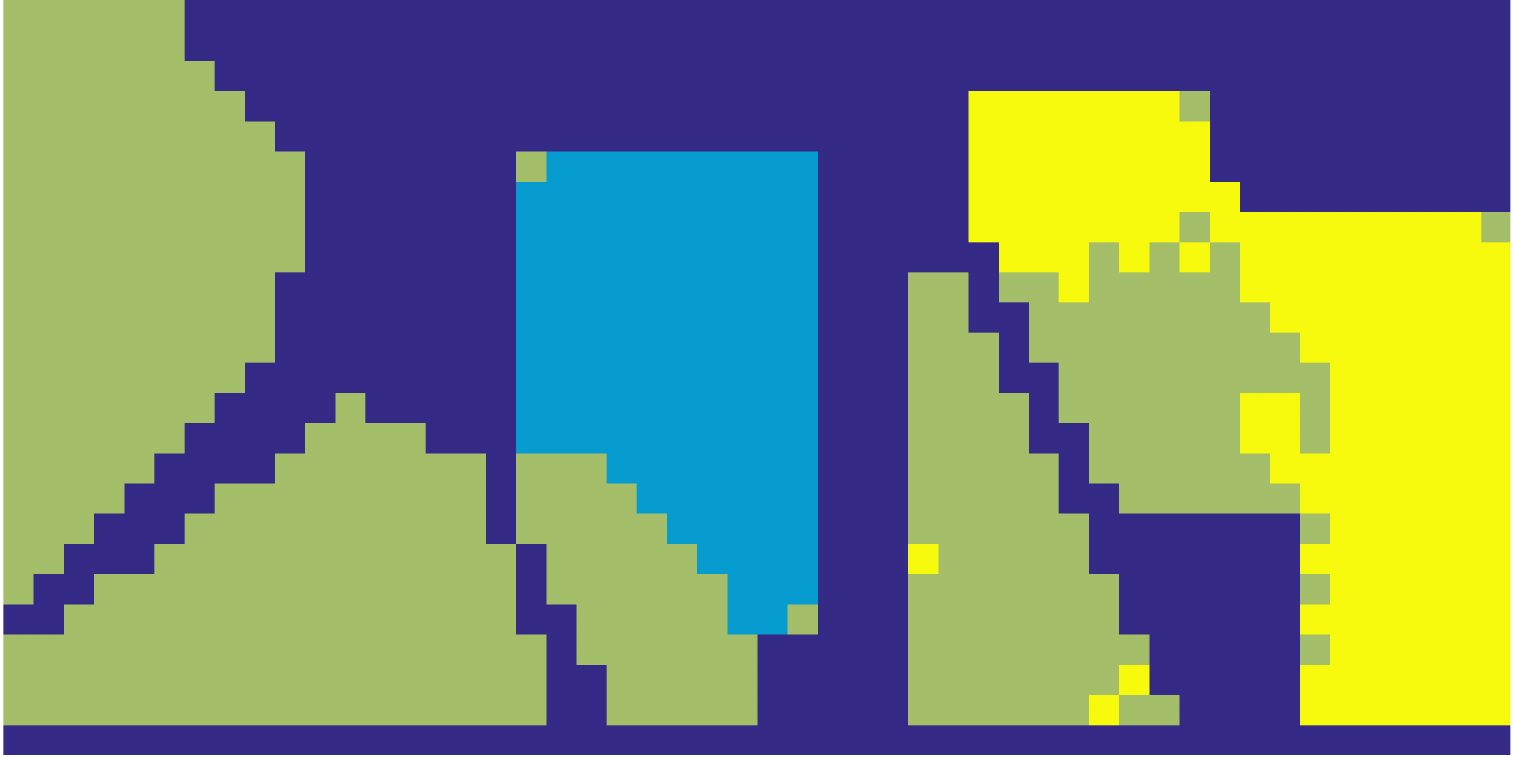}
\subcaption{Second stage, final labeling, using spectral and spatial information jointly.}
\end{subfigure}
\caption{\label{fig:SpectralSpatialLabeling}
An example of the two-stage spectral-spatial labeling process, performed on the Indian Pines dataset used for experiments in Section \ref{subsubsec:IP}.  In subfigure (a), the partial labeling from the first stage is shown.  After mode identification, points are labeled with the same label as their nearest spectral neighbor of higher density, unless that label is different from the consensus label in the spatial domain, in which case a point is left unlabeled.  This leads to points far from the centers of the classes staying unlabeled after the first stage.  In the second stage, unlabeled points are assigned labels by the same rule, unless there is a clear consensus in the spatial domain, in which case the unlabeled point is given the consensus spatial label; the results of this second stage appear in subfigure (b).  For visual clarity, here and throughout the paper, pixels without ground truth (GT) labels are masked out.}
\end{figure}

In the second stage we label each of the points left unlabeled in the first stage, by assigning the \emph{consensus label} of its nearest spatial neighbors (see Section \ref{subsec:AlgorithmDescription}), if it exists, or otherwise the label of its nearest spectral neighbor of higher density. In this way the yet unlabeled points, typically near the spatial and spectral boundaries of the classes, benefit from the spatial information in the already labeled points, which are closer to the centers of the classes.  The second stage thus labels points using both spectral and spatial information.  Figure \ref{fig:SpectralSpatialLabeling} shows an instance of this two-stage labeling process.  

This method of clustering combines the diffusion-based learning of modes with the joint spectral-spatial labeling of pixels and is called \emph{spectral-spatial diffusion learning} (DLSS), detailed in Section \ref{subsec:AlgorithmDescription}.  We contrast it with another novel method we propose, called \emph{diffusion learning (DL)}, in which modes are learned as in DLSS, but the labeling proceeds simply by enforcing that each point has the same spectral label as its nearest spectral neighbor of higher density.  DL therefore disregards spatial information, while DLSS makes significant use of it, particularly in the second stage of the labeling.  Our experiments show that while both DL and DLSS perform very well, DLSS is generally superior.

\subsection{Major Contributions}

We propose a clustering algorithm for HSI with several significant innovations.  First, diffusion distance is proposed to measure distance between high-density regions in hyperspectral data, in order to determine class modes.  Our experiments show that this distance efficiently differentiates between points belonging to the same cluster and points in different clusters.  This correct identification of modes from each cluster is essential to any clustering algorithm incorporating an analysis of modes.  Compared to state-of-the-art fast mode detection algorithms, the proposed method enjoys excellent empirical performance; theoretical performance guarantees are beyond the scope of the present article and will be discussed in a forthcoming article \cite{Murphy2018}.

A second major contribution of the proposed HSI clustering algorithm is the incorporation of spatial information through the labeling process.  Labels for points are determined by diffusing in the spectral domain from the labeled modes, unless spatial proximity is violated.  By not labeling points that would violate spatial regularity, the proposed algorithm first labels points that, with high confidence, are close to the spectral modes of the distributions.  Only after labeling all of these points are the remaining points, further from the modes, labeled.  This enforces a spatial regularity which is natural for HSI, because under mild assumptions, a pixel in an HSI is likely to have the same label as the most common label among its nearest spatial neighbors \cite{Fauvel2008,Li2013_2,Zhang2016,Tarabalka2009,Benedetto2012,Fauvel2013,Cahill2014,Cloninger2014,Wang2014,Benedetto2016}.  In both stages, DLSS takes advantage of the geometry of the dataset by using data-adaptive diffusion processes, greatly improving empirical performance.  The proposed methods are $O(ND\log(N))$ in the number of points ($N$) and ambient dimension of the data ($D$) when the intrinsic dimension of the data is small, and thus have near optimal complexity, suitable for the big data setting.  

A third major contribution is the introduction of an \emph{active learning} scheme based on distances of points to the computed modes.  In the context of active learning, the user is allowed to label only a very small number of points, to be chosen parsimoniously.  We propose an unsupervised method for determining which points to label in the active learning setting.  We note that pixels that are equally far in diffusion distance from their nearest two modes are likely to be near class boundaries, and hence to be the most challenging pixels to label by the proposed unsupervised method.  Our active learning method requires the labels of only the pixels whose distances to their nearest two modes are closest.  The proposed active learning method builds naturally on the fully unsupervised method, since the computation of distances to nearest mode are already computed by the DL and DLSS algorithms, and hence the computational complexity of the proposed active learning method does not differ significantly from the fully unsupervised method.  Our experiments show that this method can dramatically improve labeling accuracy with a number of labels $\ll 1\%$ of the total pixels. This work is detailed in Section \ref{subsec:ActiveLearning}.


\section{Unsupervised Learning Algorithm and Active Learning Variation}\label{sec:Algorithm}

\subsection{Motivating Example and Approach}

A key aspect of our algorithm is the method for identifying the modes of the classes in the HSI data.  This is challenging because of the high ambient dimension of the data, potential overlaps between distributions at their tails, along with differing densities, sampling rates, and distribution shapes.  

\begin{figure}
\centering
\begin{subfigure}{.24\textwidth}
\includegraphics[width=\textwidth]{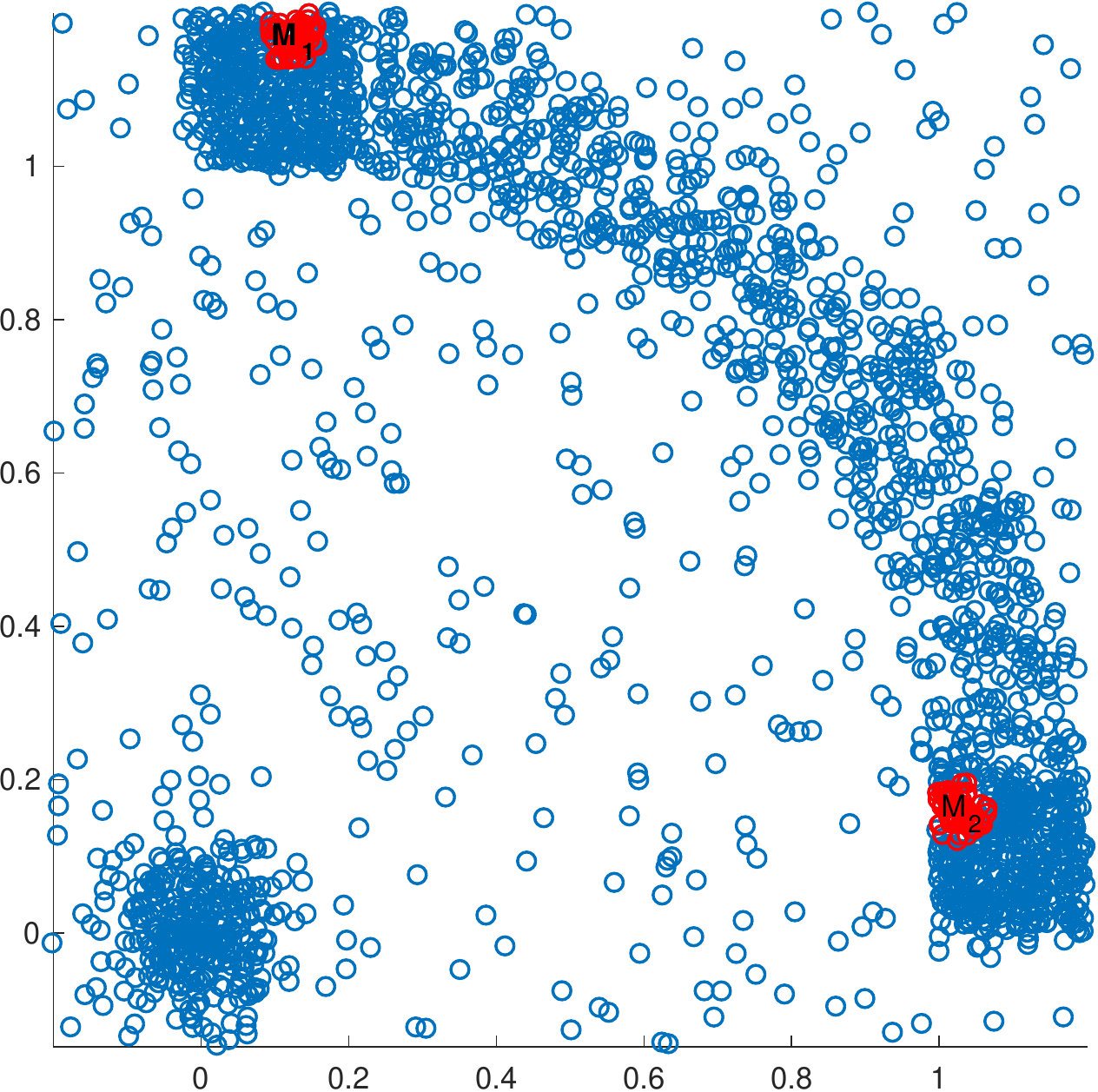}
\caption{Euclidean modes}
\end{subfigure}
\begin{subfigure}{.24\textwidth}
\includegraphics[width=\textwidth]{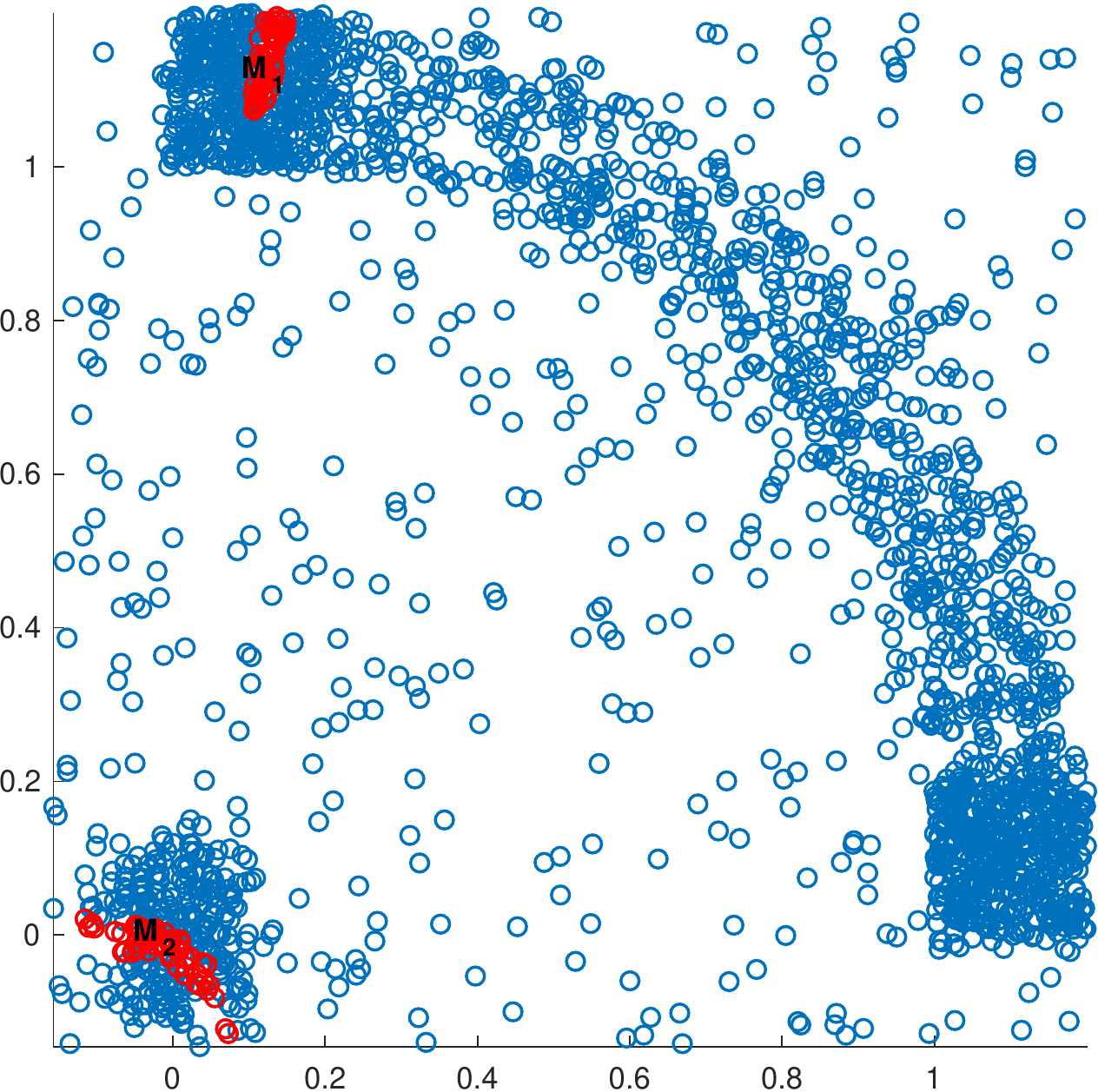}
\caption{Diffusion modes}
\end{subfigure}
\caption{\label{fig:ToyExample}Learned modes with Euclidean distances and diffusion distances.  The Euclidean and diffusion distances from $(0,1)$ are shown in subfigures (b), (c) of Figure \ref{fig:NonlinearComparison}, while the corresponding learned modes are labeled, with nearby points colored red in subfigures (a), (b), of the present figure.  Notice that the proposed diffusion learning method, using diffusion distances, correctly learns $M_{1}, M_{2}$ from different clusters (b), while using Euclidean distances leads assigning both $M_{1}, M_{2}$ to the same cluster (a), which would lead to poor clustering results.}
\end{figure}

\begin{figure}
\centering
\begin{subfigure}[t]{.16\textwidth}
\includegraphics[width=\textwidth]{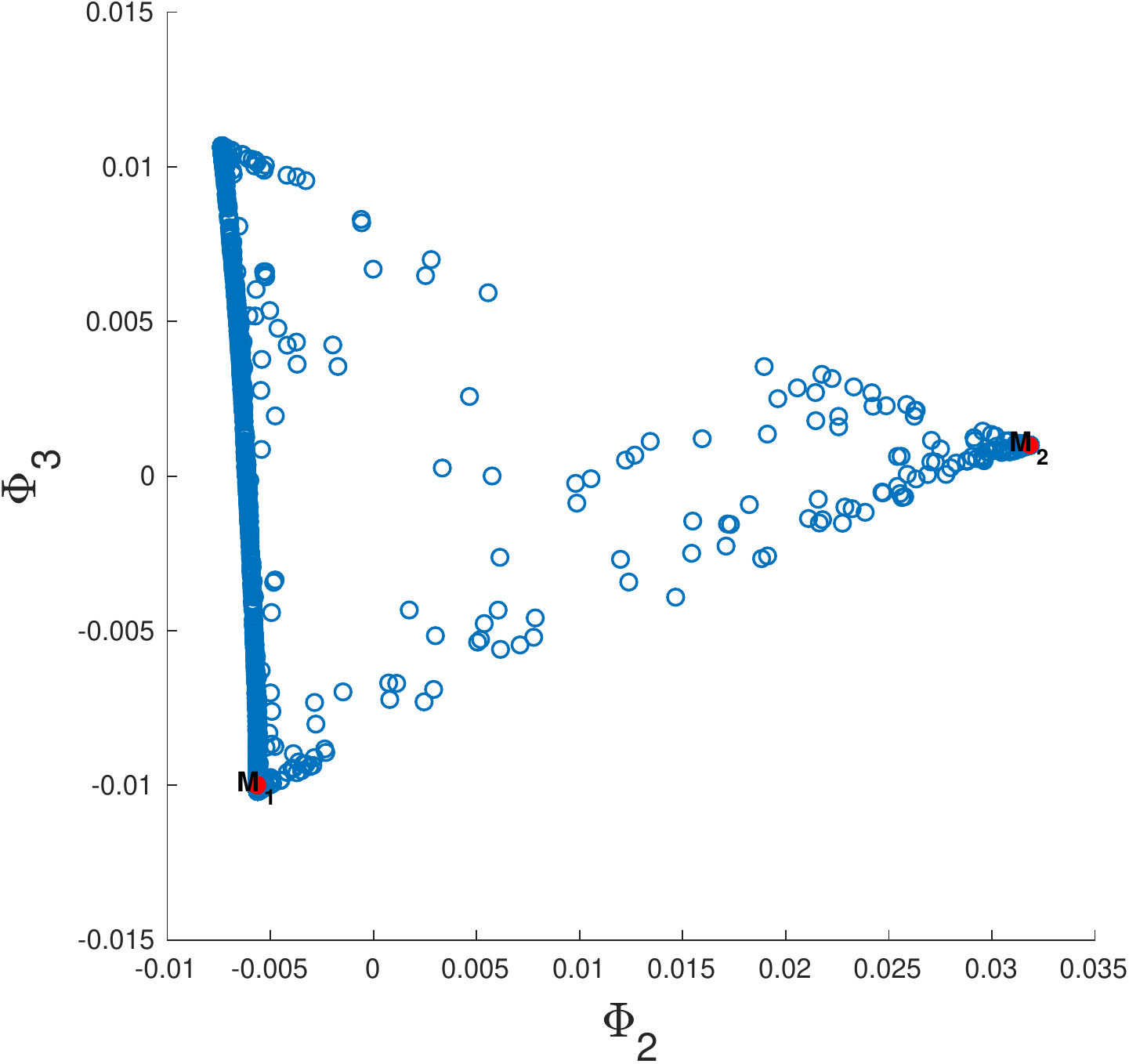}
\caption{Low-dimensional embedding and learned modes.}
\end{subfigure}
\begin{subfigure}[t]{.16\textwidth}
\includegraphics[width=\textwidth]{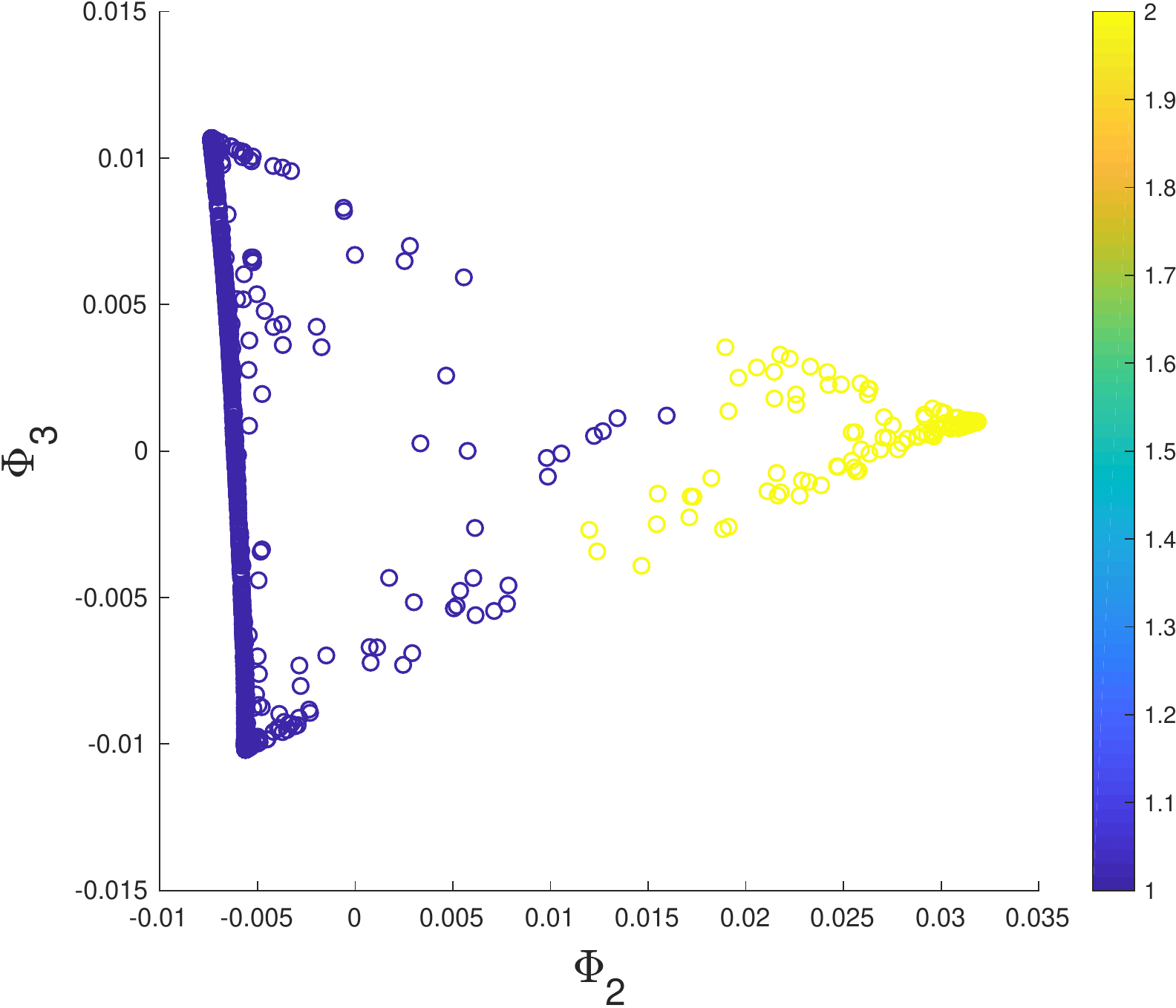}
\caption{Labeling with diffusion distances and learned modes.}
\end{subfigure}
\begin{subfigure}[t]{.15\textwidth}
\includegraphics[width=\textwidth]{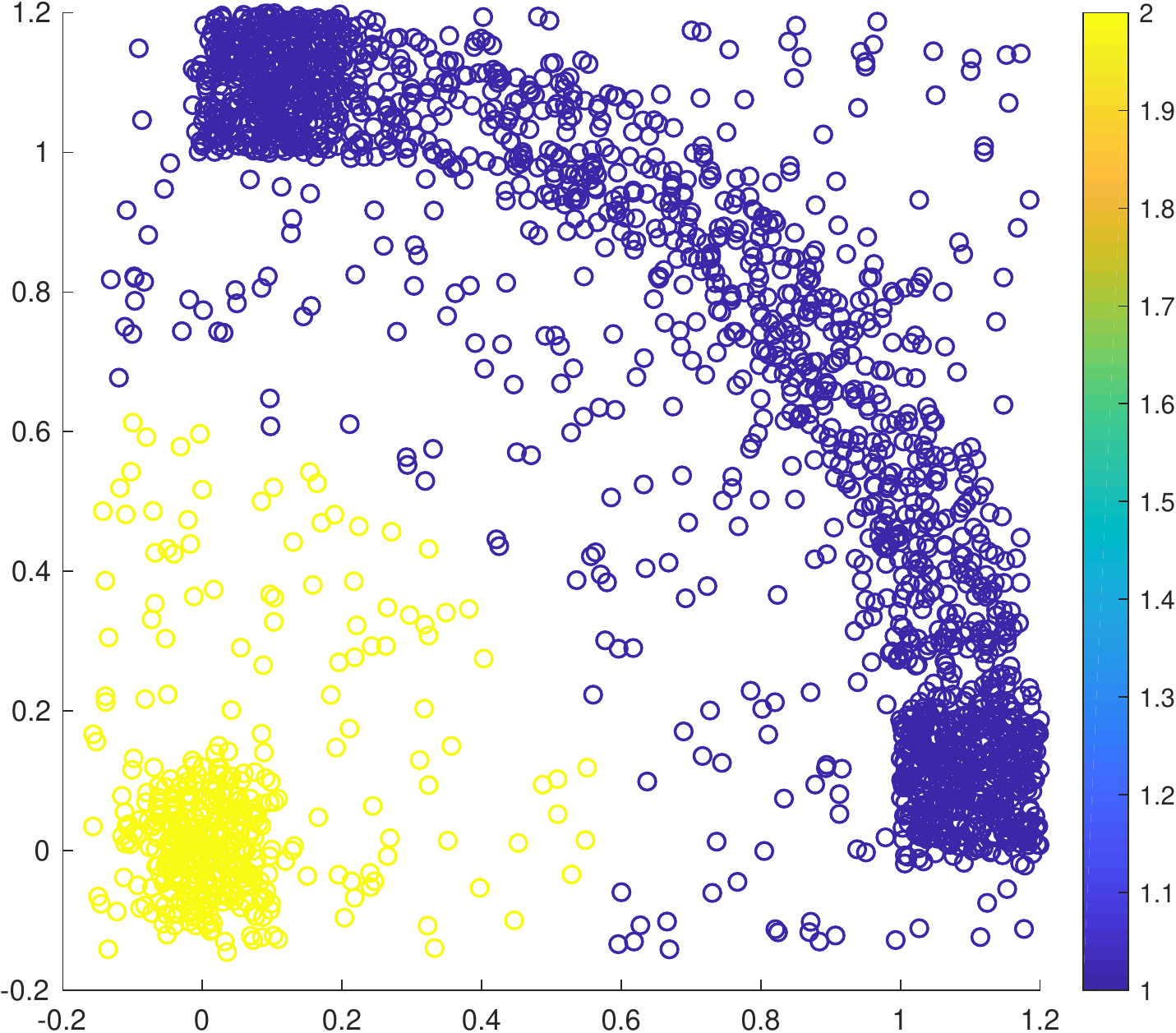}
\caption{Learned labels projected on original data.}
\end{subfigure}
\caption{\label{fig:ToyExample_Learning}In Subfigure (a), the data from Figure \ref{fig:NonlinearComparison} is represented in a new coordinate system, given by the second and third eigenfunctions of a Markov transition matrix.  In this coordinate system, the natural Euclidean distance is equal to the diffusion distance on the original image.  It is seen that the two ends of the parabolic segment are much closer in this embedding than in the original data, owing to the many short paths connecting them.  The learned modes are labeled in this low-dimensional embedding as in Figure \ref{fig:ToyExample}, subfigure (b).  In subfigure (b) of the present figure, points are labeled according to the proposed algorithm based on diffusion distance and the learned modes.  Subfigure (c) shows the labels projected onto the original data, which conforms closely with the cluster structure in the data and the labels in Figure \ref{fig:NonlinearComparison} (a).}
\end{figure}

Consider the simplified example in Figure \ref{fig:ToyExample}, showing the same data set as that in Figure \ref{fig:NonlinearComparison}.  The points of high density lie close to the center of $\mu_2$, and close to the two ends of the support of $\mu_1$.  After computing an empirical density estimate, the distance between high density points is computed.  If Euclidean distance is used to remove spurious modes, i.e. modes corresponding to the same distribution, then the learned modes $M_{1}, M_{2}$ both correspond to $\mu_2$; see subfigure (a) of Figure \ref{fig:ToyExample}.   When diffusion distance is used rather than Euclidean distance, the learned modes $M_{1}, M_{2}$ correspond to two different classes; see subfigure (b) of Figure \ref{fig:ToyExample}.  This is because the modes on the opposite ends of the support of $\mu_2$ are far in Euclidean distance but relatively close in diffusion distance.  Furthermore, the substantial region of low density between the two distributions forces the diffusion distance between them to be relatively large.  This suggests that diffusion distance is more useful than Euclidean distance for comparing high density points for the determination of modes, under the assumption that multimodal regions have modes that are connected by regions of not-too-low density.  The results of the proposed clustering algorithm, as well a low-dimensional representation of diffusion distances, appears in Figure \ref{fig:ToyExample_Learning}.  In the low-dimensional embedding corresponding to diffusion distance coordinates, the parabolic segment is linear and compressed, enabling the correct learning of modes.  Labels are then assigned according to these modes in the diffusion coordinates, which can be projected back onto the original data to yield a clustering of the original data.

\subsection{Diffusion Distance}\label{subsec:DD}

We now present an overview of diffusion distances.  Additional analysis and comments on implementation appear in \cite{Coifman2005,Coifman2006}.  Diffusion processes on graphs lead to a data-dependent notion of distance, known as \emph{diffusion distance}.  This notion of distance has been applied to a variety of application problems, including analysis of stochastic and dynamical systems \cite{Coifman2005,Nadler2006, Coifman2008, Singer2008}, semisupervised learning \cite{Belkin2003,SMC:GeneralFrameworkAdaptiveRegularization}, data fusion \cite{Lafon2006, Czaja2016}, latent variable separation \cite{Lederman2015_1, Lederman2015_2}, and molecular dynamics \cite{Rohrdanz2011, Zheng2011}.  \emph{Diffusion maps} provide a way of computing and visualizing diffusion distances, and may be understood as a type of nonlinear dimension reduction, in which data in a high number of dimensions may be embedded in a low-dimensional space by a nonlinear coordinate transformation.  In this regard, diffusion maps are related to nonlinear dimension reduction techniques such as isomap \cite{Tenenbaum2000}, Laplacian eigenmaps \cite{Belkin2003}, and local linear embedding \cite{Roweis2000}, among several others.

Consider a discrete set $X=\{x_{n}\}_{n=1}^N\subset\mathbb{R}^{D}$.  The diffusion distance \cite{Coifman2005,Coifman2006} between $x,y\in X$, denoted $d_{t}(x,y)$, is a notion of distance that incorporates and is uniquely determined by the underlying geometry of $X$.  The distance depends on a time parameter $t$, which enjoys an interpretation in terms of diffusion on the data.  The computation of $d_{t}$ involves constructing a weighted, undirected graph $\mathcal{G}$ with vertices corresponding to the $N$ points in $X$, and weighted edges given by the $N\times N$ weight matrix
\begin{align}\label{eqn:W}W(x,y):=\begin{cases}e^{-\frac{\|x-y\|_{2}^{2}}{\sigma^{2}}}, & x\in NN_{k}(y) \\ 0, & \text{else} \end{cases}\,,\end{align}for some suitable choice of $\sigma$ and with $NN_{k}(x)$ the set of $k$-nearest neighbors of $y$ in $X$ with respect to Euclidean distance.  A fast nearest neighbors algorithm yields $W$ in quasilinear time in $N$ for $k$ small (see Section \ref{subsec:CC} for details).  
The degree of $x$ is $\deg(x):=\sum_{y\in X}W(x,y).$  

A Markov diffusion, representing a random walk on $\mathcal{G}$ (or $X$) has $N\times N$ transition matrix $P(x,y)={W(x,y)}\big/{\deg(x)}\,.$  For an initial distribution $\mu\in \mathbb{R}^{N}$ on $X$, the vector $\mu P^{t}$ is the probability over states at time $t\ge 0$.  As $t$ increases, this diffusion process on $X$ evolves according to the connections between the points encoded by $P$.  This Markov chain has a stationary distribution $\pi$ s.t. $\pi P=\pi$, given by $\pi(x)={\deg(x)}/{\sum_{y\in X}\deg(y)}$.
The \emph{diffusion distance at time $t$} is 
\begin{equation}
d^{2}_{t}(x,y):=\sum\nolimits_{u\in X} (P^{t}(x,u)- P^{t}(y,u))^{2}{d\mu(u)}/{\pi(u)}\,.
\label{e:diffdist}
\end{equation}  
The computation of $d_{t}(x,y)$ involves summing over all paths of length $t$ connecting $x$ to $y$, so $d_{t}(x,y)$ is small if $x,y$ are strongly connected in the graph according to $P^{t}$, and large if $x,y$ are weakly connected in the graph.  

The eigendecomposition of $P$ allows to derive fast algorithms to compute $d_{t}$: the matrix $ P$ admits a spectral decomposition (under mild conditions, see \cite{Coifman2006}) with eigenvectors $\{\Phi_{n}\}_{n=1}^{N}$ and eigenvalues $\{\lambda_{n}\}_{n=1}^{N}$, where $1=\lambda_{1}\ge |\lambda_{2}|\ge \dots\ge|\lambda_{N}|$.  The diffusion distance \eqref{e:diffdist} can then be written as
\begin{align}\label{eqn:DD_eigen}d_{t}^2(x,y)={\sum\nolimits_{n=1}^{N}\lambda_{n}^{2t}(\Phi_{n}(x)-\Phi_{n}(y))^{2}}\,.\end{align}  
The weighted eigenvectors $\{\lambda_{n}^{t}\Phi_{n}\}_{n=1}^{N}$ are new data-dependent coordinates of $X$, which are in fact close to being geometrically intrinsic \cite{Coifman2005}.  Euclidean distance in these new coordinates is diffusion distance on $\mathcal{G}$.

Diffusion distances are parametrized by $t$, which measures how long the diffusion process on $\mathcal{G}$ has run when the distances are computed.  Small values of $t$ allow a small amount of diffusion, which may prevent the interesting geometry of $X$ from being discovered, but provide detailed, fine scale information. Large values of $t$ allow the diffusion process to run for so long that the fine geometry may be washed out.  In this work an intermediate regime is typically when the diffusion geometry of the data is most useful; in all our experiments we set $t=30$.  The choices of $\sigma,k,t$ in the construction of $W$ are in general important, see Section \ref{subsec:ParameterAnalysis}.

Note that under the mild condition that the underlying graph $\mathcal{G}$ is connected, $|\lambda_n|< 1$ for $n>1$.  Hence, $|\lambda_{n}^{2t}|\ll1$ for large $t$ and $n> 1$, so that the sum (\ref{eqn:DD_eigen}) may approximated by its truncation at some suitable $2\le M\ll N$.  In our experiments, $M$ was set to be the value at which the decay of the eigenvalues $\{\lambda_{n}\}_{n=1}^{N}$ begins to decrease; this is a standard heuristic for diffusion maps. The subset $\{\lambda_{n}^{t}\Phi_{n}\}_{n=1}^{M}$ used in the computation of $d_{t}$ is a dimension-reduced set of diffusion coordinates.  The truncation also enables us to compute only a few eigenvectors, reducing computational complexity, see Section \ref{subsec:CC}.  In this sense, the mapping \begin{align}\label{eqn:DiffusionMapsDR}x\mapsto (\lambda_{1}^{t}\Phi_{1}(x),\lambda_{2}^{t}\Phi_{2}(x),\dots,\lambda_{M}^{t}\Phi_{M}(x))\end{align} is a dimension reduction mapping of the ambient space $\mathbb{R}^{D}$ to $\mathbb{R}^{M}$.


\subsection{Unsupervised HSI Clustering Algorithm Description} \label{subsec:AlgorithmDescription}

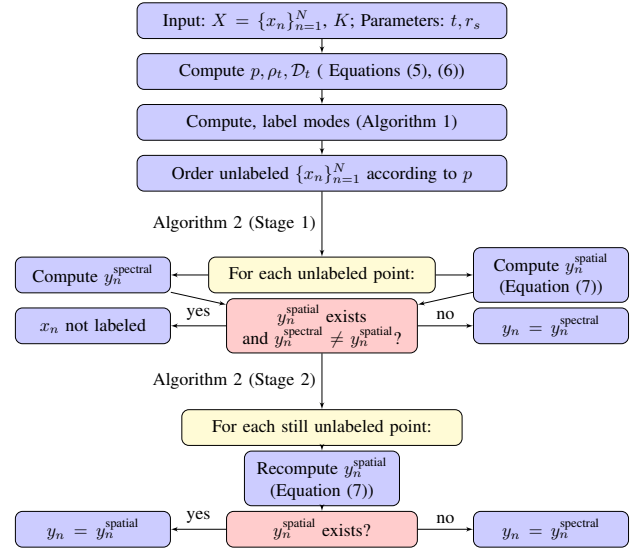
\begin{figure}
\centering
\text{Spectral-Spatial Diffusion Learning - DLSS}\\
\vskip0.2cm
\begin{adjustbox}{max width=0.45\textwidth}
\begin{tikzpicture}[ node distance = 1cm]
    \node [block1] (Init) {Input: $X=\{x_{n}\}_{n=1}^{N}$, $K$; Parameters: $t, r_{s}$};
    \node [block1, below of=Init] (ComputeEstimator) {Compute $p, \rho_{t},\mathcal{D}_{t}$ ( Equations (\ref{eqn:p}), (\ref{eqn:D_t}))};
    \node[block1, below of=ComputeEstimator](ModeLabels){Compute, label modes (Algorithm \ref{alg:modes})};
    \node [block1, below of =ModeLabels](SortDensity){Order unlabeled $\{x_{n}\}_{n=1}^{N}$ according to $p$};
    \node [Loop1, below of = SortDensity, node distance =  2 cm](Iteration1){For each unlabeled point:};
    \node [block2, left of=Iteration1, node distance = 4.5 cm] (SpectralLabel) {Compute $y^{\text{spectral}}_{n}$};
    \node [block2, right of=Iteration1, node distance= 4.5 cm] (SpatialLabel) {Compute $y^{\text{spatial}}_{n}$ (Equation (\ref{eqn:SpatialConsensus}))};
    \node [decision, below of=Iteration1] (LabelAgreement) {$y_{n}^{\text{spatial}}$ exists \\ and $y_{n}^{\text{spectral}}\neq y_{n}^{\text{spatial}}$?};
    \node [block2, left of=LabelAgreement, node distance = 4.5 cm](NoConsensus){$x_{n}$ not labeled};
    \node [block2, right of=LabelAgreement, node distance= 4.5 cm](Consensus){$y_{n}=y_{n}^{\text{spectral}}$};
    \node [Loop2, below of = LabelAgreement, node distance = 2 cm](Iteration2){For each still unlabeled point:};
    \node [block2, below of=Iteration2] (NewSpatialLabel) {Recompute $y^{\text{spatial}}_{n}$ (Equation (\ref{eqn:SpatialConsensus}))};
    \node [decision, below of=NewSpatialLabel] (SpatialLabelExists?) {$y_{n}^{\text{spatial}}$ exists?};
    \node [block2, left of=SpatialLabelExists?, node distance = 4.5 cm](AssignSpatial){$y_{n}=y_{n}^{\text{spatial}}$};
     \node [block2, right of=SpatialLabelExists?, node distance = 4.5 cm](AssignSpectral){$y_{n}=y_{n}^{\text{spectral}}$};

    \path [line] (Init) -- (ComputeEstimator);
    \path[line] (ComputeEstimator)--(ModeLabels);
    \path[line] (ModeLabels)--(SortDensity);
    \path[line] (SortDensity)--node[anchor=east]{Algorithm \ref{alg:labels} (Stage 1)}(Iteration1);
    \path [line] (Iteration1) -- (SpectralLabel);
    \path [line] (Iteration1) -- (SpatialLabel);
    \path [line] (SpectralLabel)--(LabelAgreement);
    \path [line] (SpatialLabel)--(LabelAgreement);
    \path [line](LabelAgreement)-- node[anchor=south]{yes} (NoConsensus);
    \path [line](LabelAgreement)-- node[anchor=south] {no} (Consensus);
    \path [line](LabelAgreement)--node[anchor=east]{Algorithm \ref{alg:labels} (Stage 2)}(Iteration2);
    \path [line] (Iteration2) -- (NewSpatialLabel);
    \path [line] (NewSpatialLabel)--(SpatialLabelExists?);
    \path [line] (SpatialLabelExists?)--node[anchor=south]{yes}(AssignSpatial);
    \path [line] (SpatialLabelExists?)--node[anchor=south]{no}(AssignSpectral);

\end{tikzpicture}
\end{adjustbox}
\caption{\label{fig:Flowchart} Diagram of proposed unsupervised clustering algorithm DLSS.  First, modes are computed.  Second, points are labeled in the two stage algorithm.  Notice that in the first stage, a label may only be assigned based on spectral information, though spatial information may prevent a label from being assigned.  In the second stage, a label may be assigned based on either spectral or spatial information.}
\end{figure}

We now discuss the proposed HSI clustering algorithm in detail; see Figure \ref{fig:Flowchart} for a flowchart representation.
Let $X=\{x_{n}\}_{n=1}^{N}\subset \mathbb{R}^{D}$ be the HSI, and let $K$ be the number of clusters.  As described in Section \ref{s:overview}, our algorithm proceeds in two major steps: mode identification and labeling of points.

The algorithm for learning the modes of the classes is summarized in Algorithm \ref{alg:modes}.  It first computes an empirical density for each point $x_n$ with a kernel density estimator
\begin{align}\label{eqn:p} p(x_{n})={p_{0}(x_{n})}\big/{\sum\nolimits_{m=1}^{N}p_{0}(x_{m})},\end{align}
where
$
p_{0}(x_n)=\sum\nolimits_{x_m\in NN_{k}(x_n)}e^{{-\|x_{n}-x_{m}\|_{2}^{2}}/{\sigma_{1}^{2}}}
$.
Here $\|x_{n}-x_{m}\|_{2}$ is the Euclidean distance in $\mathbb{R}^D$, and $NN_{k}(x_n)$ is the set of $k$-nearest neighbors to $x_n$, in Euclidean distance.  The use of the Gaussian kernel density estimator is standard, enjoying strong theoretical guarantees \cite{Sheather1991, Friedman2001} but certainly other estimators may be used.  In our experiments we set $k=20$, though our method is robust to choosing larger $k$.  The parameter $\sigma_{1}$ in the exponential kernel is set to be one twentieth the mean distance between all points (one could use the median instead in the presence of outliers).  
Once the empirical density $p$ is computed, the modes of the HSI classes are computed in a manner similar in spirit to \cite{Rodriguez2014}, but employing diffusion distances.  We compute the time-dependent quantity $\tilde\rho_{t}$ that assigns, to each pixel, the minimum diffusion distance between the pixel and a point of higher empirical density:
\begin{align*}
\tilde\rho_{t}(x_n)=\begin{cases}\min\limits_{\{p(x_m)\ge p(x_n)\}}\!\!\!\!d_{t}(x_{n},x_{m}), & x_n\neq \argmax_{i} p(x_i) \\ \max_{x_m}d_{t}(x_{n},x_{m}), & x_n = \argmax_{i} p(x_i)\end{cases}\,,
\end{align*}
where $d_{t}(x_{m},x_{n})$ is the diffusion distance between $x_{m}, x_{n}$, at time $t$.  In the following we will use the normalized quantity $\rho_{t}(x_n)=\tilde\rho_t(x_n)/\max_{x_m}\tilde\rho_{t}(x_m)$, which has maximum value $1$.
The modes of the HSI are computed as the points $x_{1}^{*},\dots,x_{K}^{*}$ yielding the $K$ largest values of the quantity
\begin{align}\label{eqn:D_t}\mathcal{D}_{t}(x_n)=p(x_n)\rho_{t}(x_n)\,.\end{align}
Such points should be both high density and far in diffusion distance from any other higher density points, and can therefore be expected to be modes of different cluster distributions.  This method provably detects modes correctly under certain distributional assumptions on the data \cite{Murphy2018}.

\RestyleAlgo{algoruled}
 \LinesNumbered
 
\begin{algorithm}
	\caption{Geometric Mode Detection Algorithm}
	 \label{alg:modes}
  	\emph{Input}: $X, K$; $t$.\\
  	Compute the empirical density $p(x_{n})$ for each $x_n\in X$.\\
  	Compute $\{\rho_{t}(x_{n})\}_{n=1}^{N}$, the diffusion distance from each point to its nearest neighbor in diffusion distance of higher empirical density, normalized.\\
  	Set the learned modes $\{x_{i}^{*}\}_{i=1}^{K}$ to be the $K$ maximizers of $\mathcal{D}_{t}(x_{n})=p(x_{n})\rho_{t}(x_{n})$.\\
  	\emph{Output:} $\{x_{i}^{*}\}_{i=1}^{K}, \{p(x_{n})\}_{n=1}^{N}, \{\rho_{t}(x_{n})\}_{n=1}^{N}$.
\end{algorithm}

Once the modes are detected, each is given a unique label.  All other points are labeled using these mode labels in the following two-stage process, summarized in Algorithm \ref{alg:labels}.  In the first stage, running in order of decreasing empirical density, the \emph{spatial consensus label} of each point is computed by finding all labeled points within distance $r_{s}\ge 0$ in the spatial domain of the pixel in question; call this set $NN^{s}_{r_{s}}(x_{n})$.  If one label among $NN^{s}_{r_{s}}$ occurs with relative frequency $>1/2$, that label is the spatial consensus label.  Otherwise, no spatial consensus label is given.  In detail, let  $L^{\text{spatial}}_{n}=\{y_{m} \ | \ x_{m}\in NN^{s}_{r_{s}}(x_{n}), x_{m}\neq x_{n}\}$ denote the labels of the spatial neighbors within radius $r_{s}$.  Then the spatial consensus label of $x_i$ is 
\begin{align}\label{eqn:SpatialConsensus}
y_{i}^{\text{spatial}}=\begin{cases}k, & \frac{|\{y_{n}| y_{n}=k, \ y_{n}\in L^{\text{spatial}}_{n}\}|}{|L_{n}^{\text{spatial}}|}>\frac12,\\ 0\ (\text{no label}), & \text{else}.\end{cases}
\end{align}
After a point's spatial consensus label is computed, its \emph{spectral label} is computed as its nearest neighbor in the spectral domain, measured in diffusion distance, of higher density.  The point is then given the overall label of the spectral label unless the spatial consensus label exists (i.e. is $\neq0$ in \eqref{eqn:SpatialConsensus}) and differs from the spatial consensus label.  In this case, the point in question remains unlabeled in the first stage.  Note that points that are unlabeled are considered to have label 0 for the purposes of computing the spatial consensus label, so in the case that most pixels in the spatial neighborhood are unlabeled, the spatial consensus label will be 0.  Hence, only pixels with many labeled pixels in their spatial neighborhood can have a consensus spatial label.  In this first stage, a label is only assigned based on spectral information, though the spatial information may prevent a label from being assigned.  

Upon completion of the first stage, the dataset will be partially labeled; see Figure \ref{fig:SpectralSpatialLabeling}. In the second stage, an unlabeled point is given the label of its spatial consensus label, if it exists, or otherwise the label of its nearest spectral neighbor of higher density.  Thus, in the second stage, a label is assigned based on joint spectral-spatial information.

\RestyleAlgo{algoruled}
\LinesNumbered
\begin{algorithm}[h]
\SetAlCapSkip{5em}

  \caption{Spectral-Spatial Labeling Algorithm\label{alg:labels}}
  \emph{Input:} $\{x_{i}^{*}\}_{i=1}^{K}, \{p(x_{n})\}_{n=1}^{N}$, $\{\rho_{t}(x_{n})\}_{n=1}^{N}$;  $r_{s}$.\\
  Assign each mode a unique label.\\
  \emph{Stage 1}: Iterating through the remaining unlabeled points in order of decreasing density among unlabeled points, assign each point the same label as its nearest spectral neighbor (in diffusion distance) of higher density, unless the spatial consensus label exists and differs, in which case the point is left unlabeled.\\
  \emph{Stage 2}: Iterating in order of decreasing density among unlabeled points, assign each point the consensus spatial label, if it exists, otherwise the same label as its nearest spectral neighbor of higher density.  \\
  \emph{Output:} Labels $\{y_{n}\}_{n=1}^{N}$.
  
\end{algorithm}

Points of high density are likely to be labeled according to their spectral properties.  The reasons for this are twofold.  First, these points are likely to be near the centers of distributions, and hence are likely to be in spatially homogeneous regions.  Second, points of high density are labeled before points of low density, so it is unlikely for high density points to have many labeled points in their spatial neighborhoods.  This means that the spatial consensus label is unlikely to exist for these points.  Conversely, points of low density may be at the boundaries of the classes, and are hence more likely to be labeled by their spatial properties.  The incorporation of spatial information into machine learning for HSI is justified by the fact that HSI images typically show some amount of spatial regularity, in that if a pixel's nearest spatial neighbors all have the same class label, it is likely that the pixel has this same label, compared to the case in which the pixel's nearest spatial neighbors have random labels \cite{Fauvel2008,Li2013_2,Zhang2016,Tarabalka2009,Benedetto2012,Fauvel2013,Cahill2014,Cloninger2014,Wang2014,Benedetto2016}. 
The spatial information regularizes and improves performance, but it cannot take the place of the spectral information, as shall be seen in Section \ref{subsec:SpaceParameter}: the spectral information is more discriminative than the spatial information, and is the more important of the two.

The proposed method, combining Algorithms \ref{alg:modes}, \ref{alg:labels} is called \emph{spectral-spatial diffusion learning (DLSS)}.  In our experimental analysis, the significance of the spectral-spatial labeling scheme is validated by comparing DLSS against a simpler method, called \emph{diffusion learning (DL)}.  This method learns class modes as in Algorithm \ref{alg:modes}, but labels all pixels simply by requiring each point have the same label as its nearest spectral neighbor of higher density.  The expectation is that DLSS will generally outperform DL, due to the former's incorporation of spatial data; this is confirmed by our experiments.


\subsection{Active Learning DLSS Variation}\label{subsec:ActiveLearning}

Both the DL and DLSS methods are unsupervised.  We now present a variation of the DLSS method for active learning of hyperspectral images, where a few well-chosen pixels are automatically selected for labeling.  The DLSS method labels points beginning with the learned class modes, and mistakes tend to be made on points that are near the class boundaries; in the active learning scheme the algorithm will ask for the labels of the points whose distances from their nearest two modes are closest.  That is, points whose nearest mode is ambiguous will be labeled using training data, and all other points will be labeled as in the DLSS algorithm.

More precisely, we fix a time $t$, and for each pixel $x_{n}$, let $x_{n_{1}}^{*}, x_{n_{2}}^{*}$ be the two modes closest to $x_{n}$ in diffusion distance $d_t$.  We compute the quantity 
\begin{align}\label{eqn:ActiveQuantity}F_{t}(x_{n})=|d_{t}(x_{n},x_{n_{1}}^{*})-d_{t}(x_{n},x_{n_{2}}^{*})|.\end{align}If $F_{t}(x_{n})$ is close to 0, then there is substantial ambiguity as to the nearest mode to $x_{n}$.  Suppose the user is afforded the labels of exactly $L$ points.  Then the $L$ labels requested in our active learning regime are the $L$ minimizers of $F_{t}$.  The proposed active learning scheme is summarized in Algorithm \ref{alg:ActiveLearning}. To evaluate performance, we consider a range of $L$ values in our experiments. The active learning setting is most interesting when $\alpha=L/N$ is very small, where $N$ is the total number of pixels in the image.

\RestyleAlgo{algoruled}
\LinesNumbered
\begin{algorithm}[ht]
  \caption{Active Learning with DLSS\label{alg:ActiveLearning}}
  \emph{Input:} $X,K$;  $t,r_{s},L$.\\
  Compute the modes of the data using Algorithm \ref{alg:modes}.\\
  Give each mode a unique label.\\
  Compute, for each point $x_{n}$, $F_{t}(x_{n})$ as in \eqref{eqn:ActiveQuantity}.\\
  Label the $L$ minimizers of $F_{t}$ with ground truth labels.\\
  Label the remaining, unlabeled points as in steps 3, 4 in Algorithm \ref{alg:labels}.\\
  \emph{Output:}  Labels $\{y_{n}\}_{n=1}^{N}$.\\
\end{algorithm}

Note that the active learning algorithm can be iterated, by labeling points then recomputing the quantity \eqref{eqn:ActiveQuantity} to determine the most challenging points after some labels have been introduced \cite{Murphy2018_iterative}.  


\section{Experiments}\label{sec:Experiments}

\subsection{Algorithm Evaluation Methods and Experimental Data}

We consider several HSI datasets to evaluate the proposed unsupervised (Algorithms \ref{alg:modes}, \ref{alg:labels}) and active learning (Algorithm \ref{alg:ActiveLearning}) algorithms.  For evaluation in the presence of ground truth (GT), we consider three quantitative measures, besides visual performance, namely:
\begin{enumerate}

\item  \emph{Overall Accuracy (OA)}:  Total number of correctly labeled pixels divided by the total number of pixels. This method values large classes more than small classes.

\item \emph{Average Accuracy (AA)}:  The average, over classes, of the OA of each class.  This method values small classes and large classes equally.  

\item \emph{Cohen's }$\kappa$-\emph{statistic} ($\kappa$): A measurement of agreement between two labelings, corrected for random agreement \cite{Banerjee1999}.  Letting $a_{o}$ be the observed agreement between the labeling and the ground truth and $a_{e}$ the expected agreement between a uniformly random labeling and the ground truth, $\kappa={(a_{o}-a_{e})}/{(1-a_{e})}$.  $\kappa=1$ corresponds to perfect overall accuracy, $\kappa\le0$ corresponds to labels no better than what is expected from random guessing.  
\end{enumerate}
In order to perform quantitative analysis with these metrics and make consistent visual comparisons, the learned clusters are aligned with ground truth, when available.  More precisely, let $S_{K}$ be the set of permutations of $\{1,2,\dots,K\}$.  Let $\{C_{i}\}_{i=1}^{K}$ be the clusters learned from one of the clustering methods, and let $\{C_{i}^{GT}\}_{i=1}^{K}$ be the ground truth clusters.  Cluster $C_{i}$ is assigned label $\hat{\eta}_{i}\in\{1,2,\dots,K\}$, with $\hat{\eta} = \argmax_{\eta=(\eta_{1},\dots,\eta_{K})\in S_{K}} \sum_{i=1}^{K}|C_{\eta_{i}}\cap C_{i}^{GT}|$.  We remark that while this alignment method maximizes the overall accuracy of the labeling and is most useful for visualization, better alignments for maximizing $AA$ and $\kappa$ may exist. 

We consider $4$ real HSI datasets to shed light on strengths and weaknesses of the proposed algorithm.  These datasets are standard, have ground truth, and are publicly available\footnote{\url{http://www.ehu.eus/ccwintco/index.php?title=Hyperspectral_Remote_Sensing_Scenes}}.  Experiments with active learning are performed for these same real HSI datasets with Algorithm \ref{alg:ActiveLearning}.  Additional experiments on synthetic and real HSI data are available, for conciseness, only in an appendix in the online preprint version.  

Note that some images are restricted to subsets in the spatial domain, which is noted in their respective subsections.  This is because unsupervised methods for HSI struggle with data containing a large number of classes, due to variation within classes and similarity between certain end-members of different classes \cite{Zhu2017}.  Hence, the Indian Pines, Pavia, and Kennedy Space Center datasets are restricted to reduce the number of classes and achieve meaningful clusters.  The Salinas A dataset is considered in its entirety.  The ground truth, when available, is often incomplete, i.e. not all pixels are labeled.  For these datasets, labels are computed for all data, then the pixels with ground truth labels are used for quantitative and visual analysis.  The number of class labels in the ground truth images were used as parameter $K$ for all clustering algorithms, though the proposed method automatically estimates the number of clusters; see Section \ref{sec:Future}.  Grayscale images of the projection of the data onto its first principal component and images of ground truth (GT), colored by class, for the Indian Pines, Pavia, Salinas A, and Kennedy Space Center datasets are in Figures \ref{fig:IP}, \ref{fig:Pavia}, \ref{fig:SalinasA}, and \ref{fig:KSC}, respectively.  The projection onto the first principal component of the data is presented as a simple visual summary of the data, though it washes out the subtle information presented in individual bands.

Since the proposed and comparison methods are unsupervised, experiments are performed on the entire dataset, including points without ground truth labels.  The labels for pixels without ground truth are not accounted for in the quantitative evaluation of the algorithms tested.  Note that additional experiments, not shown, were performed, using only the data with ground truth labels.  These experiments consisted in restricting the HSI to the pixels with labels, which makes the clustering problem significantly easier.  Quantitative results were uniformly better for all datasets and methods in these cases; the relative performances of the algorithms on a given dataset remained the same.


\subsection{Comparison Methods}
\label{s:comparisonmethods}

We consider a variety of benchmark and state-of-the-art methods of HSI clustering for comparison.  First, we consider the classic \emph{$K$-means} algorithm \cite{Friedman2001} applied directly to $X$.  This method is not expected to perform well on HSI data, due to the non-spherical shape of clusters, high dimensionality, and noise, all well-known problems for $K$-means.  Several {\emph{dimension reduction}} methods to reduce the dimensionality of the data, while preserving important discriminatory properties of the classes, as well as increasing the signal-to-noise ratio in the projected space, are also used as benchmarks for comparison with the proposed method.  These methods first reduce the dimension of the data from $D$ to $K_{GT}\ll D$, where $K_{GT}$ is the number of classes, then run $K$-means on the reduced data.  We consider linear dimension reduction via {\em{principal component analysis} (PCA)}; {\em{independent component analysis} (ICA)} \cite{Comon1994, Hyvarinen2000}, using the fast implementation \cite{Hyvarinen1999}\footnote{\url{https://www.cs.helsinki.fi/u/ahyvarin/papers/fastica.shtml}}; and {\em{random projections}} via Gaussian random matrices, shown to be efficient in highly simplified data models \cite{Dasgupta2000, Candes2006}.  

We also consider more computationally intensive methods for benchmarking.  \emph{DBSCAN} \cite{Ester1996} is a popular density-based clustering method, that although highly parameter-dependent, has proved useful for a variety of unsupervised tasks.  \emph{Spectral clustering} (SC) \cite{Ng2001, VonLuxburg2007} has been applied with success in classification and clustering HSI \cite{Fauvel2013}.  The spectral embedding consists of the top $K_{GT}$ row-normalized eigenvectors of the normalized graph Laplacian $L$; in this features space $K$-means is then run (see Section \ref{subsec:ParameterAnalysis}).  We also cluster with {\em{Gaussian mixture models}} (GMM) \cite{Acito2003, Manolakis2001, Kraut2001}, with parameters determined by expectation maximization (EM).  

Finally we consider several recent, state-of-the-art clustering methods: \emph{sparse manifold clustering and embedding (SMCE)} \cite{Elhamifar2011, Elhamifar2013}\footnote{\url{http://vision.jhu.edu/code/}}, which fits the data to low-dimensional, sparse structures, and then applies spectral clustering; \emph{hierarchical clustering with non-negative matrix factorization (HNMF)} \cite{Gillis2015}\footnote{\url{https://sites.google.com/site/nicolasgillis/code}}, which has shown excellent performance for HSI clustering when the clusters are generated from a single endmember; a graph-based method based on the Mumford-Shah segmentation \cite{Mumford1989}\cite{Meng2017}, related to spectral clustering, and called \emph{fast Mumford-Shah (FMS)} in this article (we use a highly parallelized version\footnote{\url{http://www.ipol.im/pub/art/2017/204/?utm_source=doi}}); \emph{fast search and find of density peaks clustering} (FSFDPC) algorithm  \cite{Rodriguez2014}, which has been shown effective in clustering a variety of data sets.  


\subsection{Relationship Between Proposed Method and Comparison Methods}

The FSFDPC method has similarities with the mode estimation aspect of our work, in that both algorithms attempt to learn the modes of the classes via a density-based analysis, as described in, for example, \cite{Chen2017, Rodriguez2014}.  Our method is quite different, however:  the proposed measure of distance between high density points is not Euclidean distance, but diffusion distance \cite{Coifman2005,Coifman2006}, which is more adept at removing spurious modes, due to its incorporation of the geometry of the data.  This phenomenon is illustrated in Figures \ref{fig:NonlinearComparison},\ref{fig:ToyExample}.  The assignment of labels from the modes is also quite different, as diffusion distances are used to determine spectral nearest neighbors, and spatial information is accounted for in our DLSS algorithm.  FSFDPC, in contrast, assigns to each of the modes its own class label, and to the remaining points a label by requiring that each point has the same label as its Euclidean nearest neighbor of higher density.  This means that FSFDPC only incorporates spectral information measured in Euclidean distance, disregarding spatial information. The benefits of both using diffusion distances to learn modes, and incorporating spatial proximities into the clustering process are very significant, as the experiments demonstrate.  

Both FSFDPC and the proposed algorithm have some similarities to DBSCAN which, however, performs poorly for data with clusters of differing densities, and is highly sensitive to its parameters.  Note that FSFDPC was in fact proposed to improve on these drawbacks of DBSCAN \cite{Rodriguez2014}.  

The proposed DLSS and DL algorithms also share commonalities with spectral clustering, SMCE, and FMS in that these comparison methods compute eigenvectors of a graph Laplacian in order to develop a nonlinear notion of distance.  This is related to computing the eigenvectors of the Markov transition matrix in the computation of diffusion maps.  The proposed method, however, directly incorporates density into the detection of modes, which allows for more robust clustering compared to these methods, which work by simply applying $K$-means to the eigenvectors of the graph Laplacian.  Moreover, our technique does not rely on any assumption about sparsity (unlike SMCE), and is completely invariant under distance-preserving transformations (it shares this property with SMCE), which could be useful if different imaging modalities (e.g. compressed modalities) were used.  

Additionally, our approach is connected to semisupervised learning techniques on graphs, where initial given labels are propagated by a diffusion process to other vertices (points); see \cite{SMC:GeneralFrameworkAdaptiveRegularization} and references therein. Here of course we have proceeded in an unsupervised fashion, replacing initial given labels by estimated modes of the clusters.


\subsection{Summary of Proposed and Comparison Methods}

The experimental methods are summarized in Table \ref{tab:Methods}.  The two novel methods we proposed are the full spectral-spatial diffusion learning method (DLSS), as well as a simplified diffusion learning method (DL).  We note that several algorithms were not implemented by the authors of this article: publicly available libraries were used when available.  Links to these libraries are noted where appropriate.

\begin{table}[h]
\begin{adjustbox}{max width=.49\textwidth}
\begin{tabular}{| l | c | c |}\hline
Method & D.R. & Metric \\ \hline
$K$-means on full dataset &  No & Euclidean   \\ \hline
$K$-means on PCA reduced dataset  & Yes & Euclidean  \\ \hline
$K$-means on ICA reduced dataset  & Yes & Euclidean  \\ \hline
$K$-means on data reduced by random projections  & Yes & Euclidean \\ \hline
DBSCAN \cite{Ester1996} & No & Euclidean \\ \hline
Spectral clustering \cite{VonLuxburg2007}  & Yes & Spectral \\ \hline
Gaussian mixture models  & No & Euclidean \\ \hline
Sparse manifold clustering and embedding \cite{Elhamifar2011, Elhamifar2013}  & Yes & Spectral \\ \hline
Hierarchical NMF \cite{Gillis2015}  & No & Euclidean \\ \hline
Fast Mumford Shah \cite{Meng2017} & No & Spectral \\ \hline
FSFDPC  \cite{Rodriguez2014}  & No & Euclidean \\ \hline
\textbf{Diffusion learning (DL)} & Yes & Diffusion  \\ \hline
\textbf{Spectral-spatial diffusion learning (DLSS)}  & Yes & Diffusion \\ \hline
\end{tabular}
\end{adjustbox}
\caption{\label{tab:Methods}Methods used for experimental analysis, along with whether the method employs dimensionality reduction and which metric is used to compared points. The methods proposed in this article appear in bold.  Note that the proposed methods employ dimension reduction, as illustrated in (\ref{eqn:DiffusionMapsDR}).}
\end{table}

All experiments and subsequent analyses, except those involving FMS, were performed in MATLAB running on a 3.1 GHz Intel 4-Core i7 processor with 16 GB of RAM; code to reproduce all results is available on the authors' website\footnote{\url{http://www.math.jhu.edu/~jmurphy/}}.  

\subsection{Unsupervised HSI Clustering Experiments}
\label{subsec:HyperspectralClustering}

\subsubsection{Indian Pine Dataset}\label{subsubsec:IP}

\begin{figure}
\centering
\includegraphics[width=.24\textwidth]{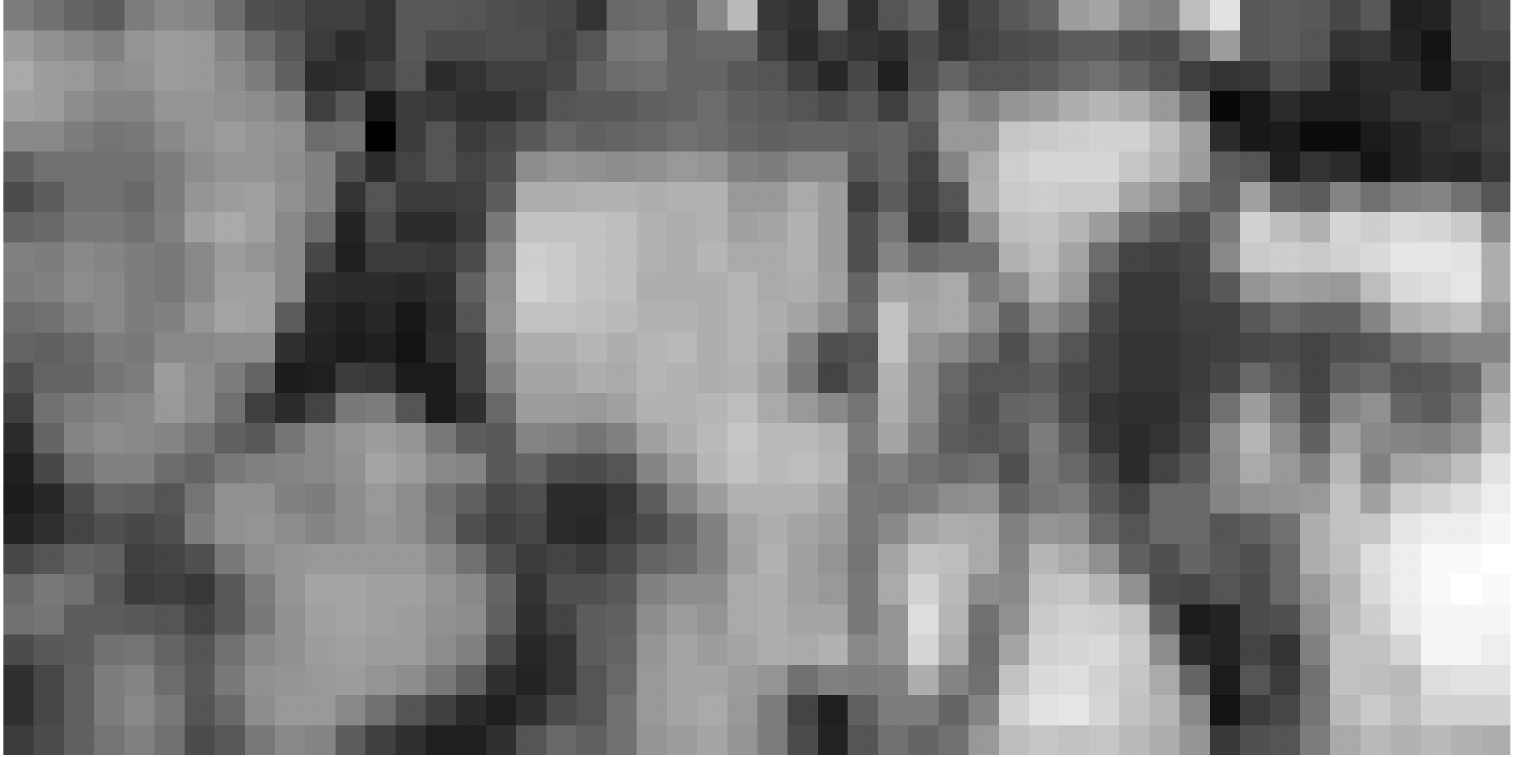}
\includegraphics[width=.24\textwidth]{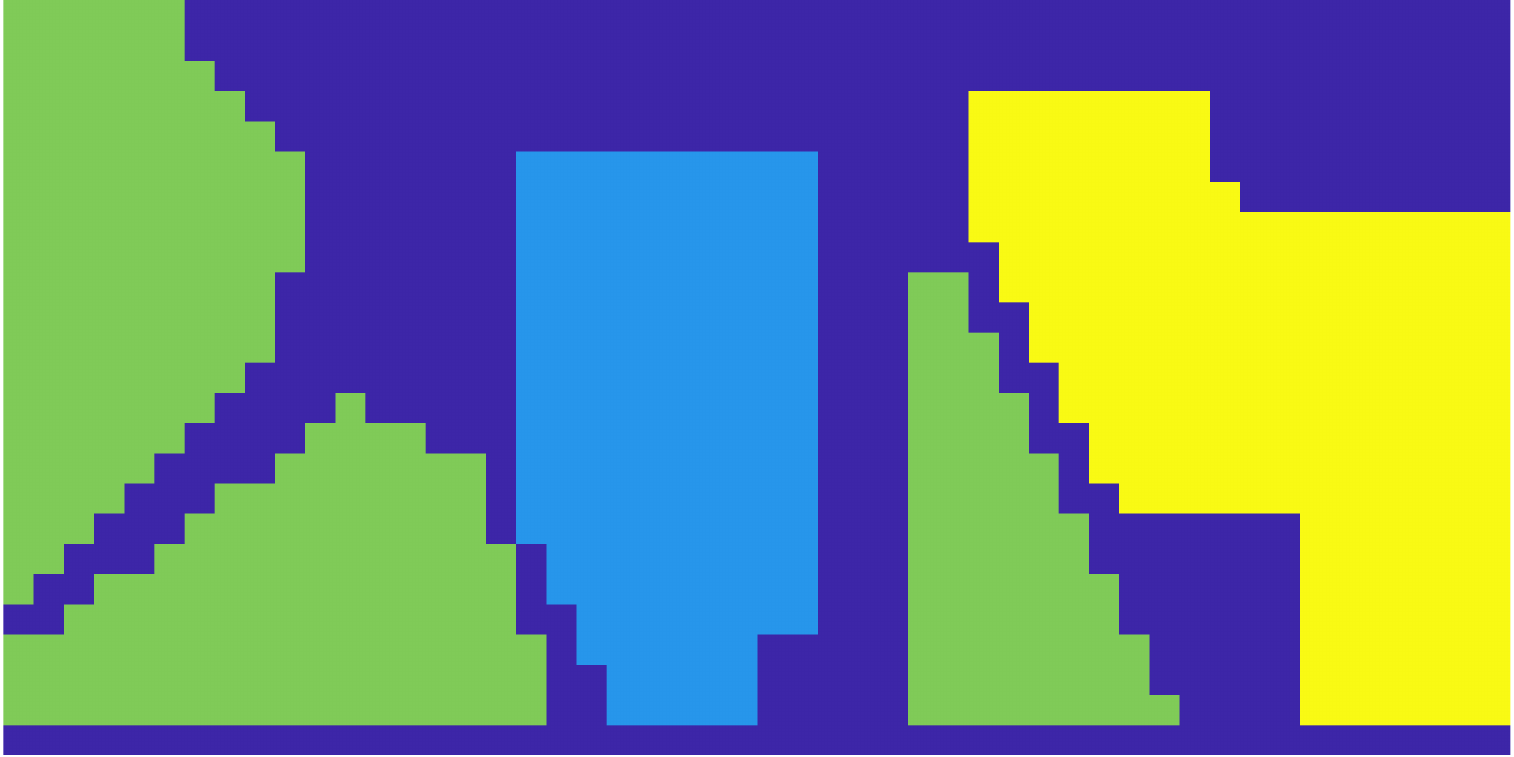}
\caption{\label{fig:IP}The Indian Pines data is a $50\times 25$ subset of the full Indian Pines dataset.  It contains 3 classes, one of which is not well-localized spatially.  The dataset was captured in 1992 in Northwest IN, USA by the AVRIS sensor.  The spatial resolution is 20m/pixel.  There are $200$ spectral bands. Left: projection onto the first principal component of the data; right: ground truth (GT).} 
\end{figure}

\begin{figure}
\centering
\begin{subfigure}{ .09\textwidth}
\includegraphics[width=\textwidth]{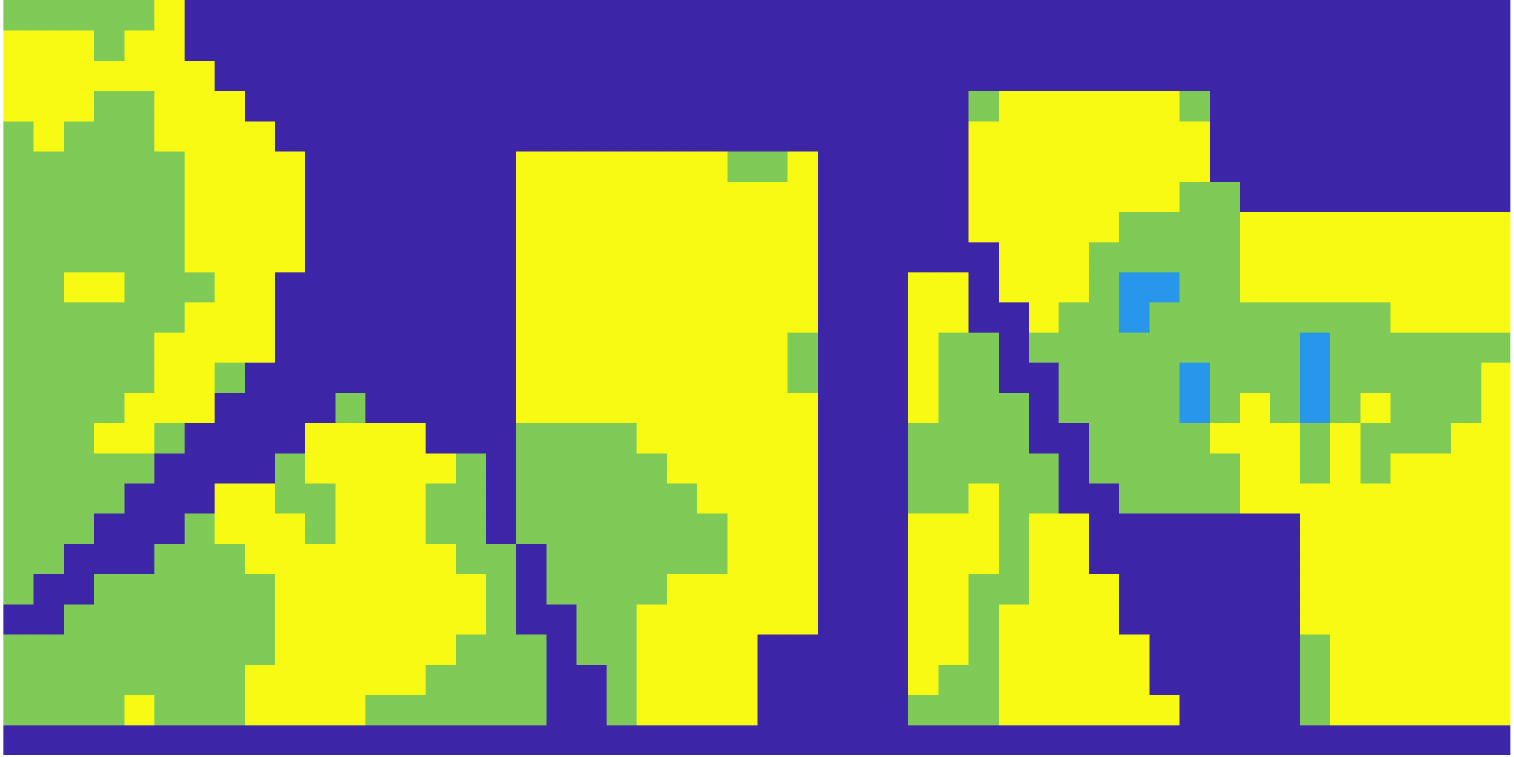}
\caption{$K$-means}
\end{subfigure}
\begin{subfigure}{ .09\textwidth}
\includegraphics[width=\textwidth]{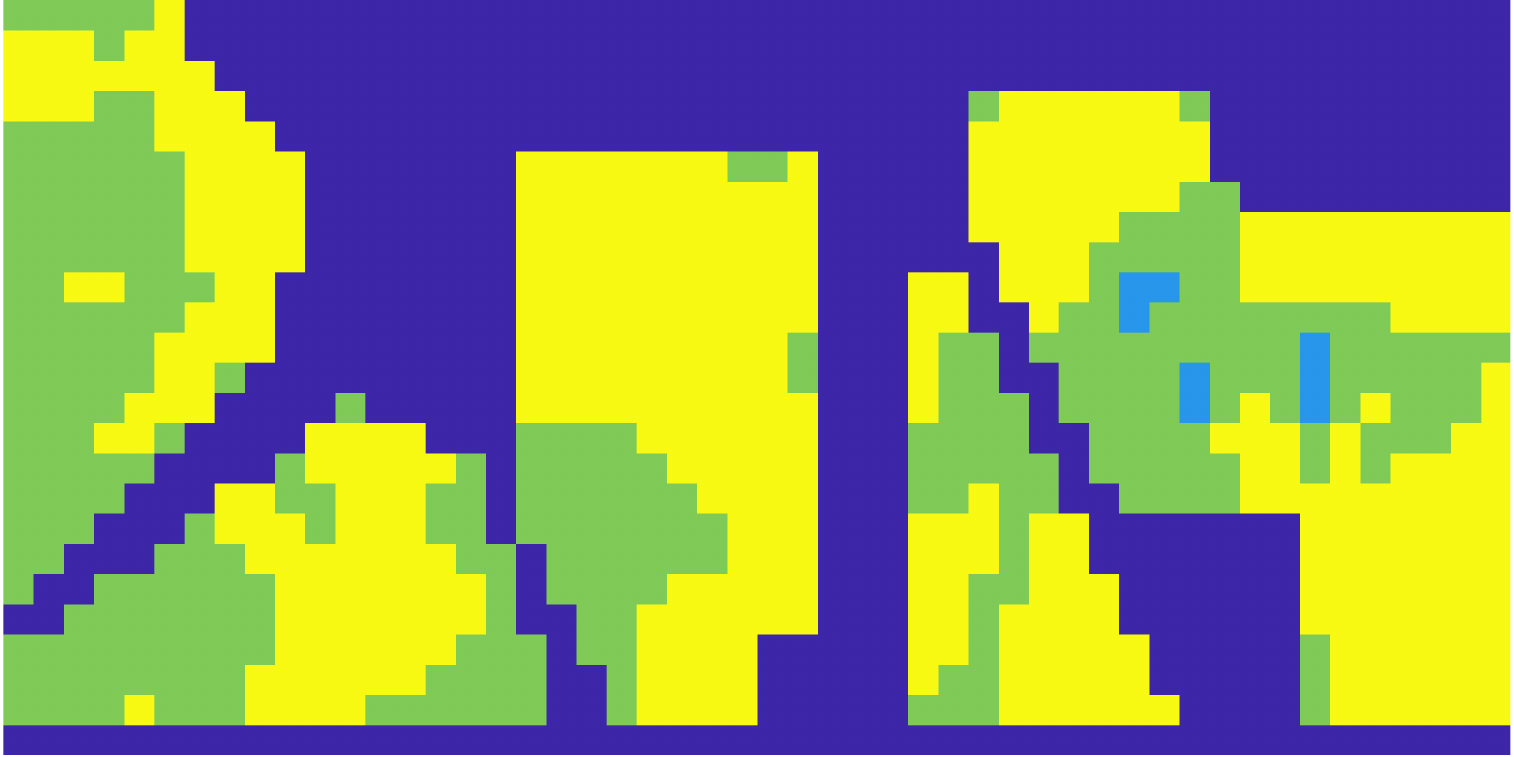}
\caption{PCA+$K$M}
\end{subfigure}
\begin{subfigure}{ .09\textwidth}
\includegraphics[width=\textwidth]{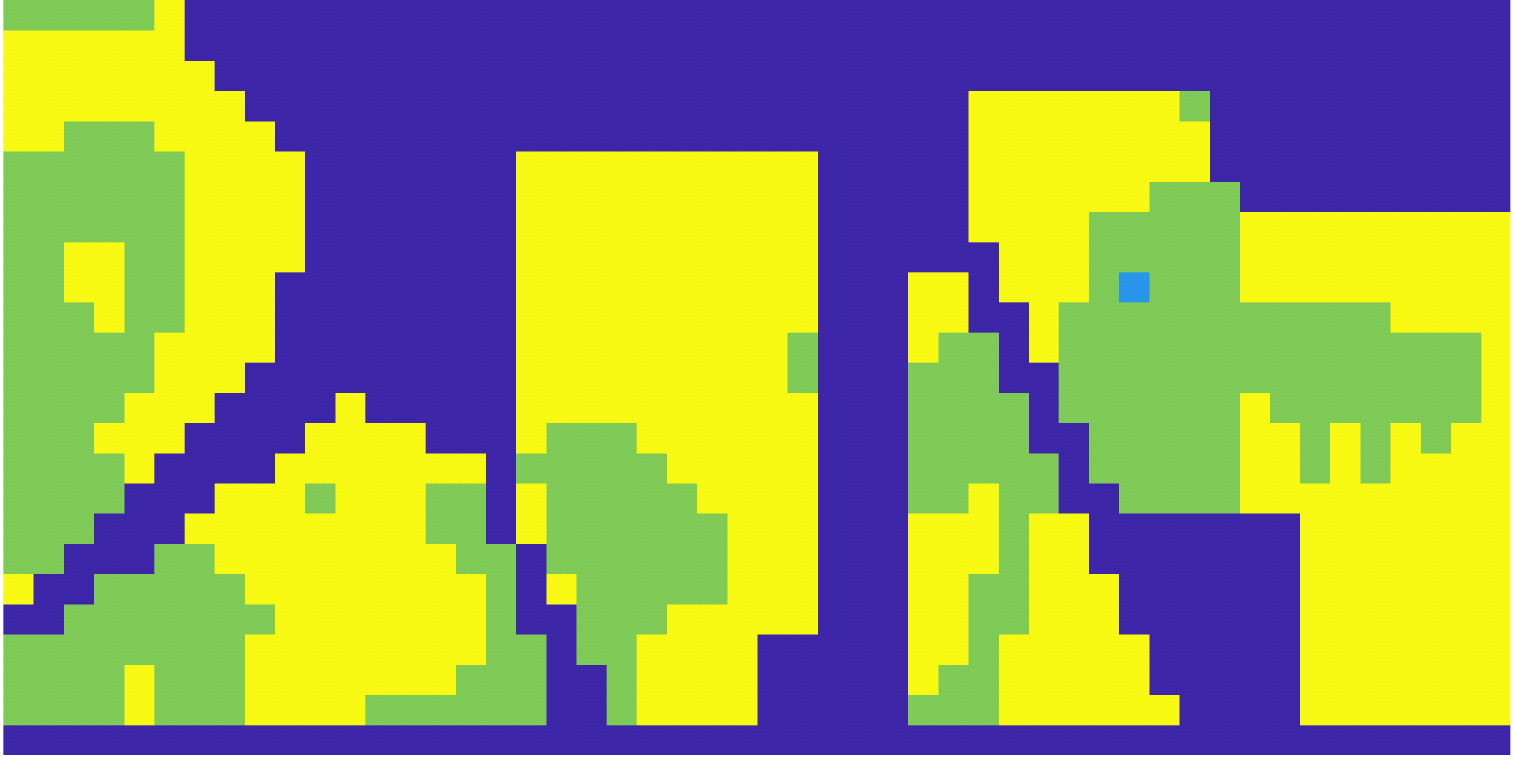}
\caption{ICA+$K$M}
\end{subfigure}
\begin{subfigure}{ .09\textwidth}
\includegraphics[width=\textwidth]{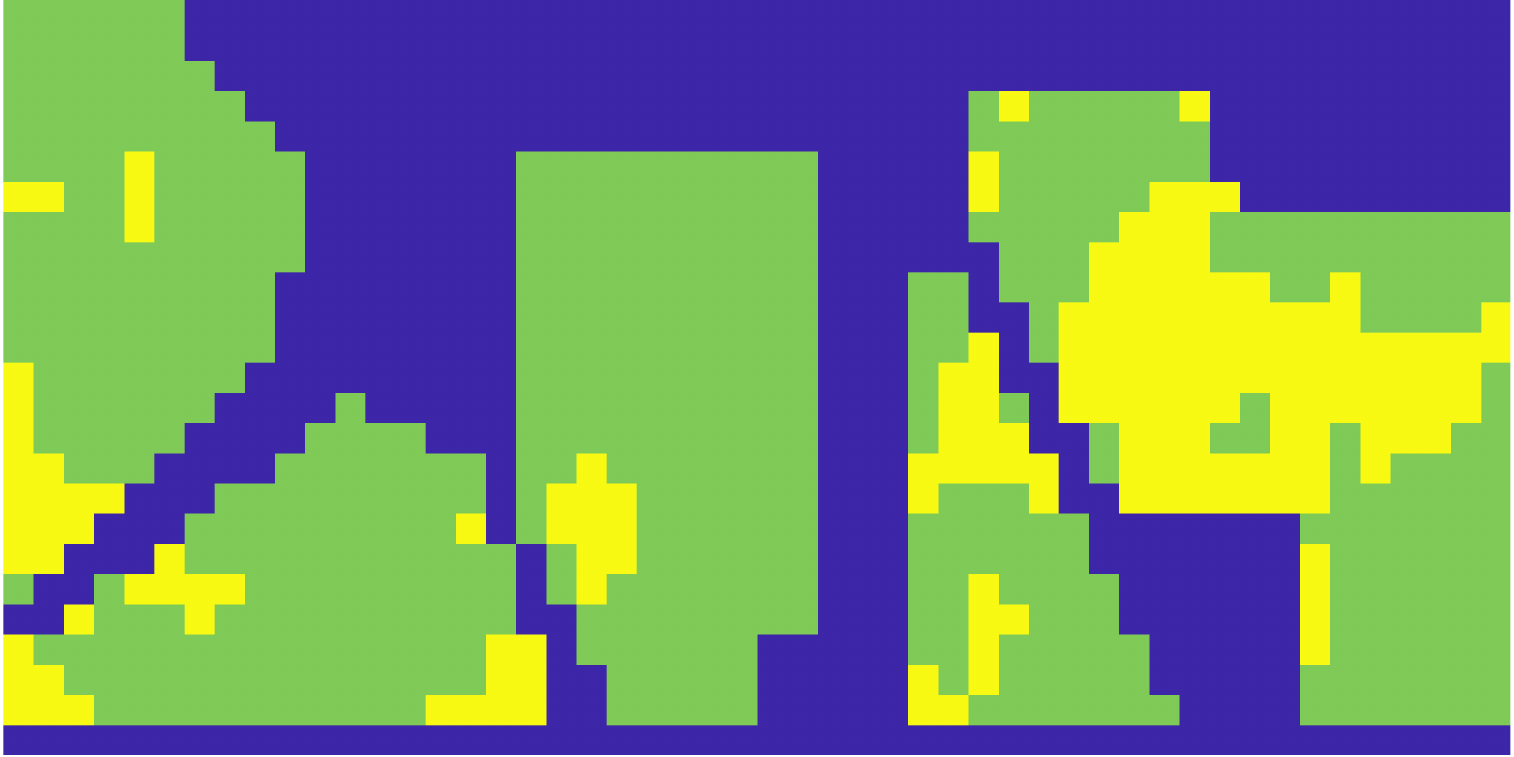}
\caption{RP+$K$M}
\end{subfigure}
\begin{subfigure}{ .09\textwidth}
\includegraphics[width=\textwidth]{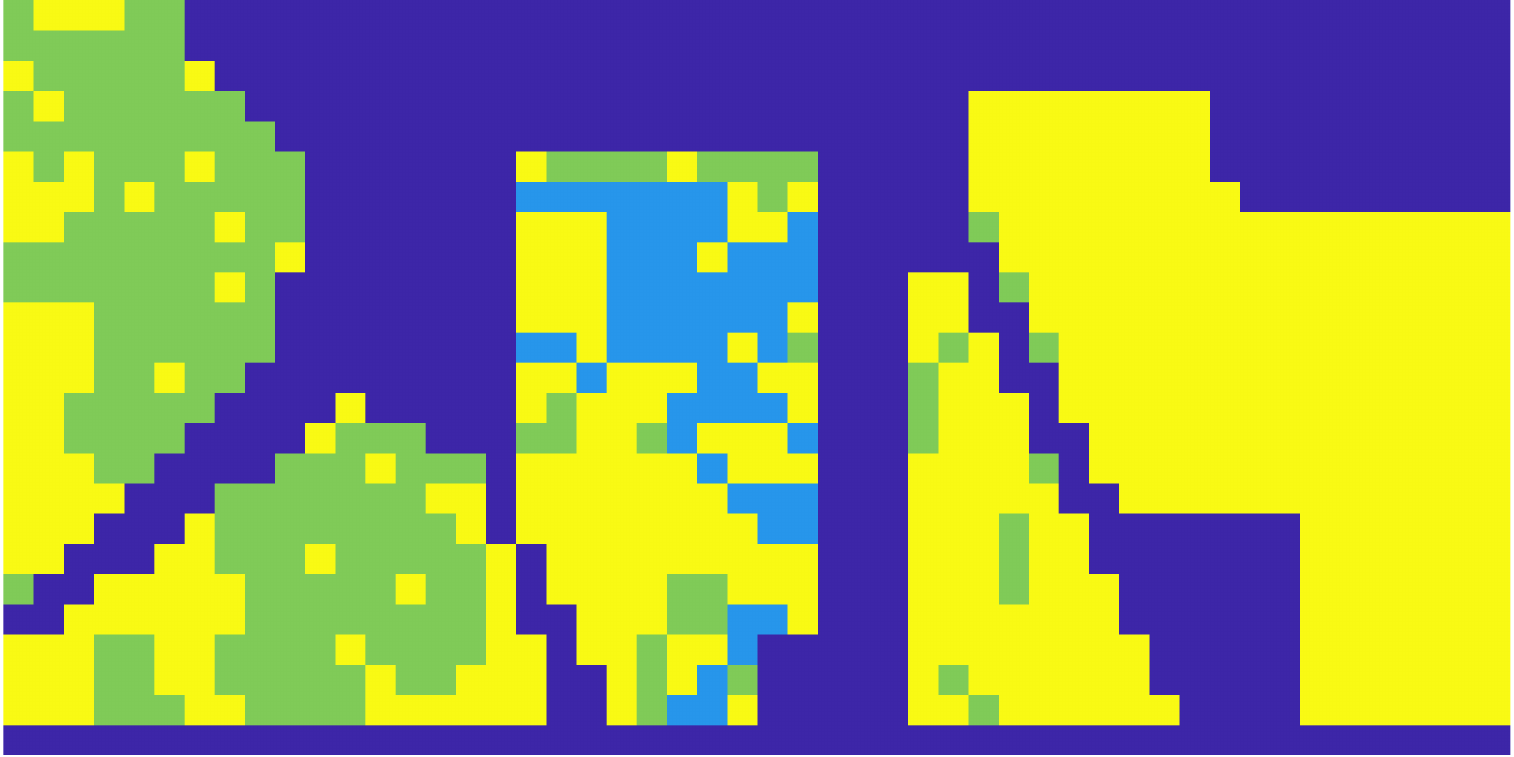}
\caption{DBSCAN}
\end{subfigure}
\begin{subfigure}{ .09\textwidth}
\includegraphics[width=\textwidth]{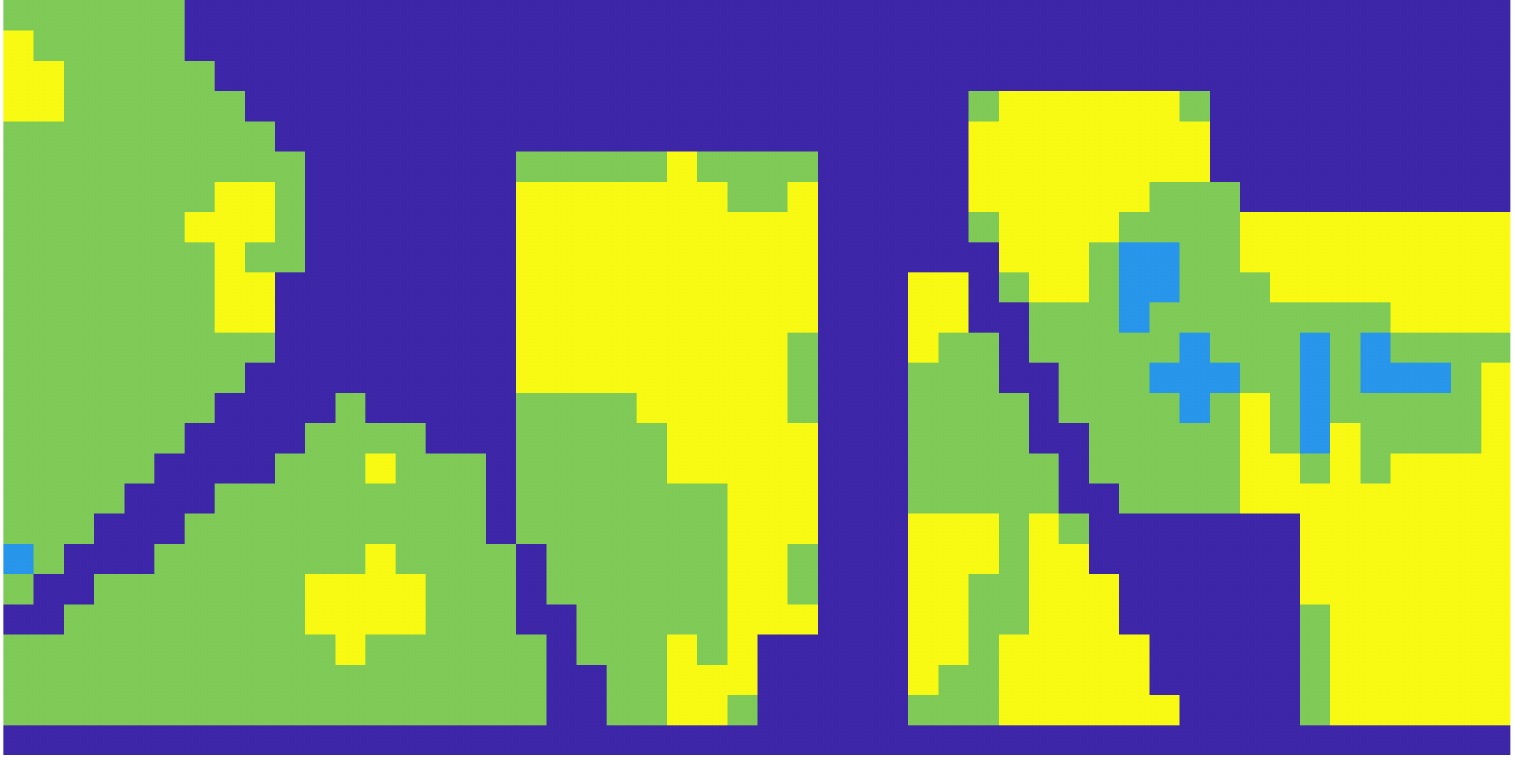}
\caption{SC}
\end{subfigure}
\begin{subfigure}{.09\textwidth}
\includegraphics[width=\textwidth]{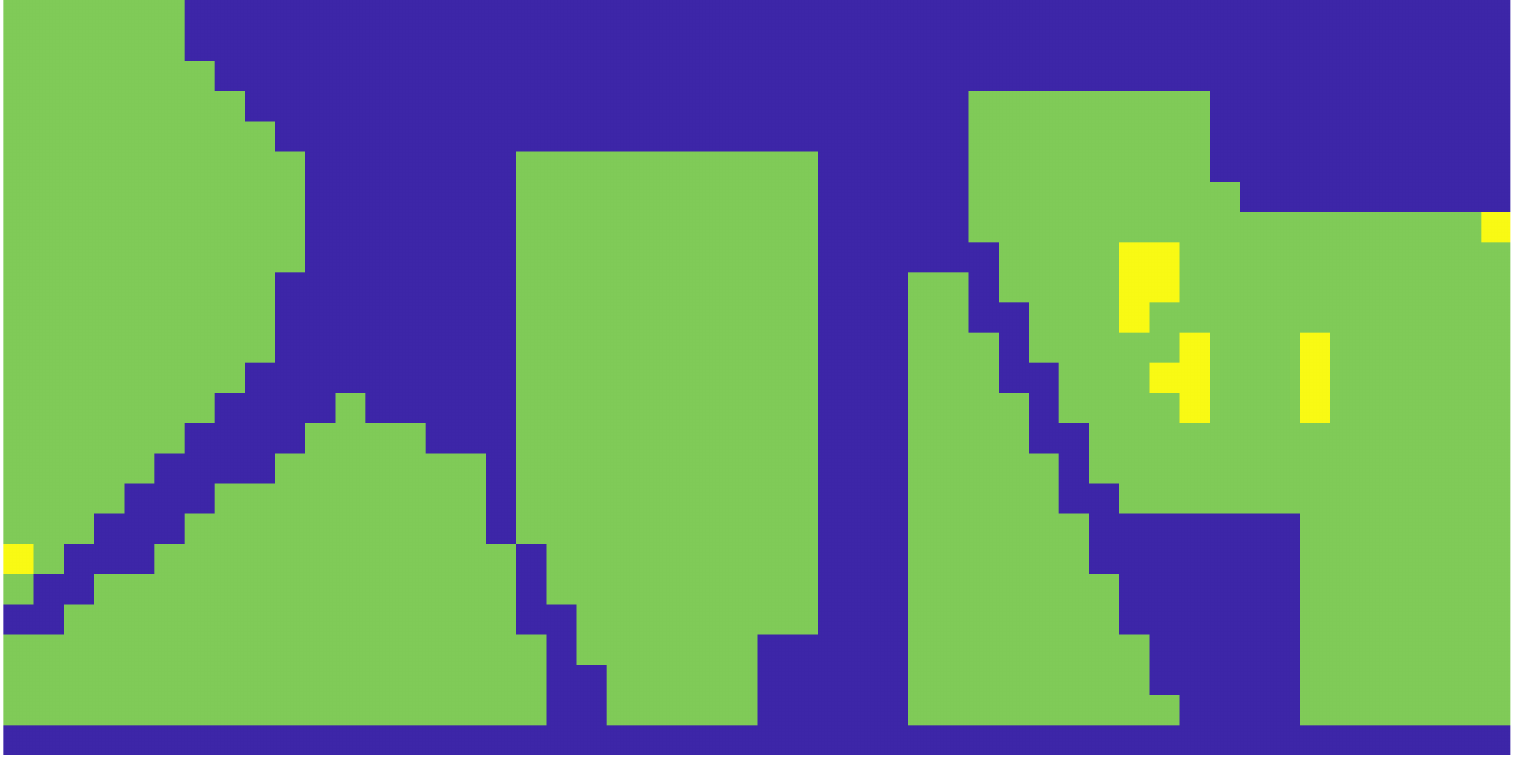}
\caption{GMM}
\end{subfigure}
\begin{subfigure}{.09\textwidth}
\includegraphics[width=\textwidth]{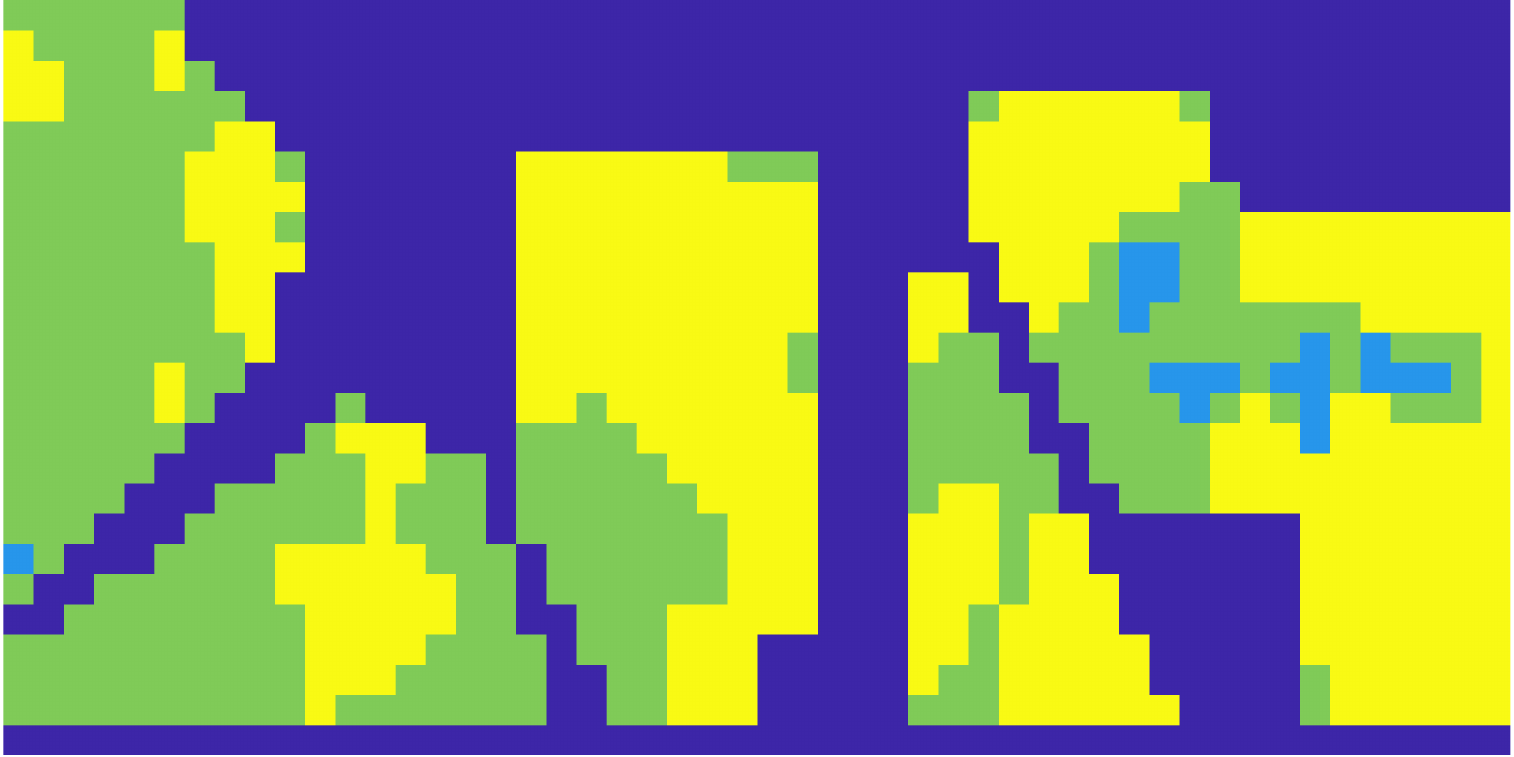}
\caption{SMCE}
\end{subfigure}
\begin{subfigure}{.09\textwidth}
\includegraphics[width=\textwidth]{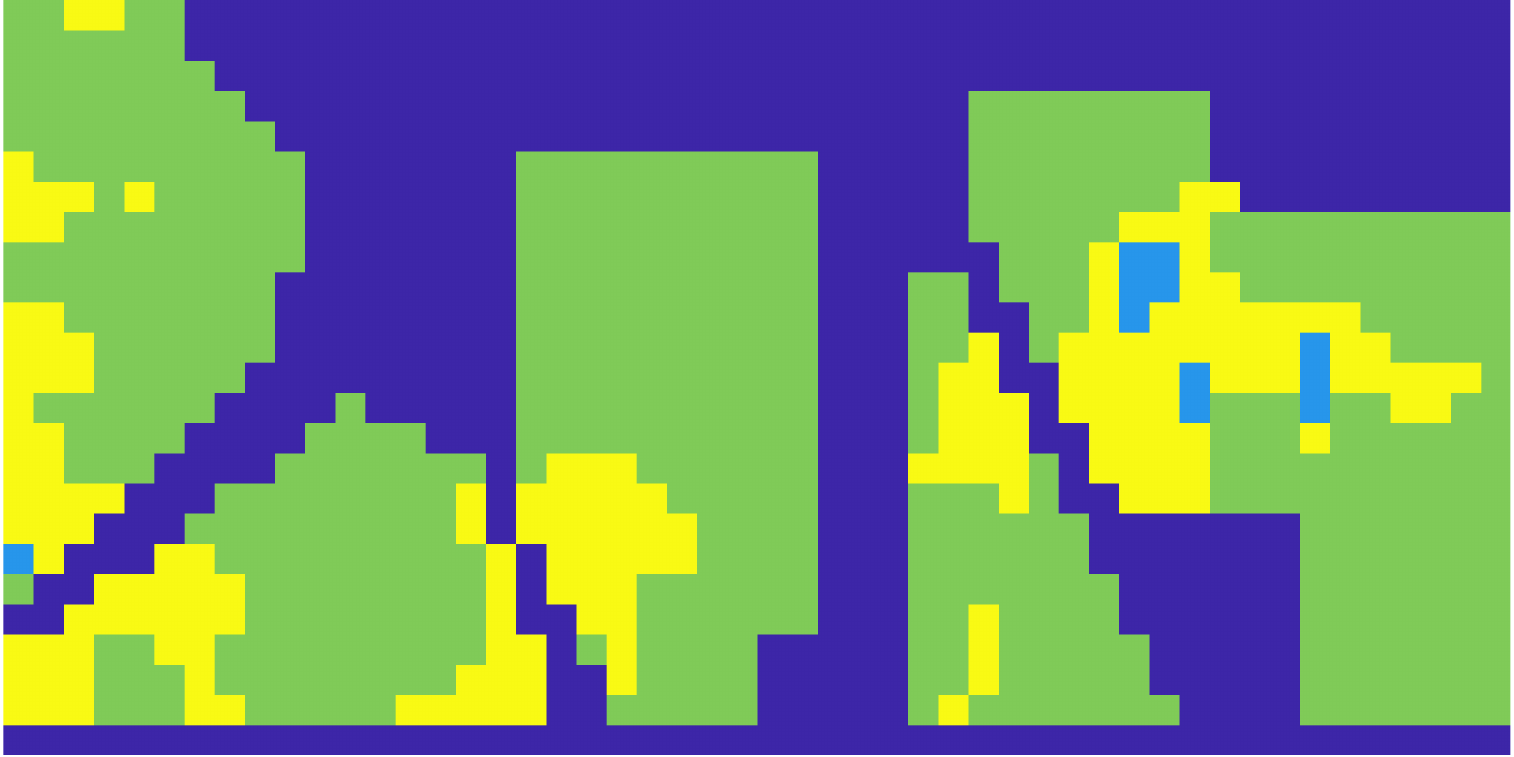}
\caption{HNMF}
\end{subfigure}
\begin{subfigure}{.09\textwidth}
\includegraphics[width=\textwidth]{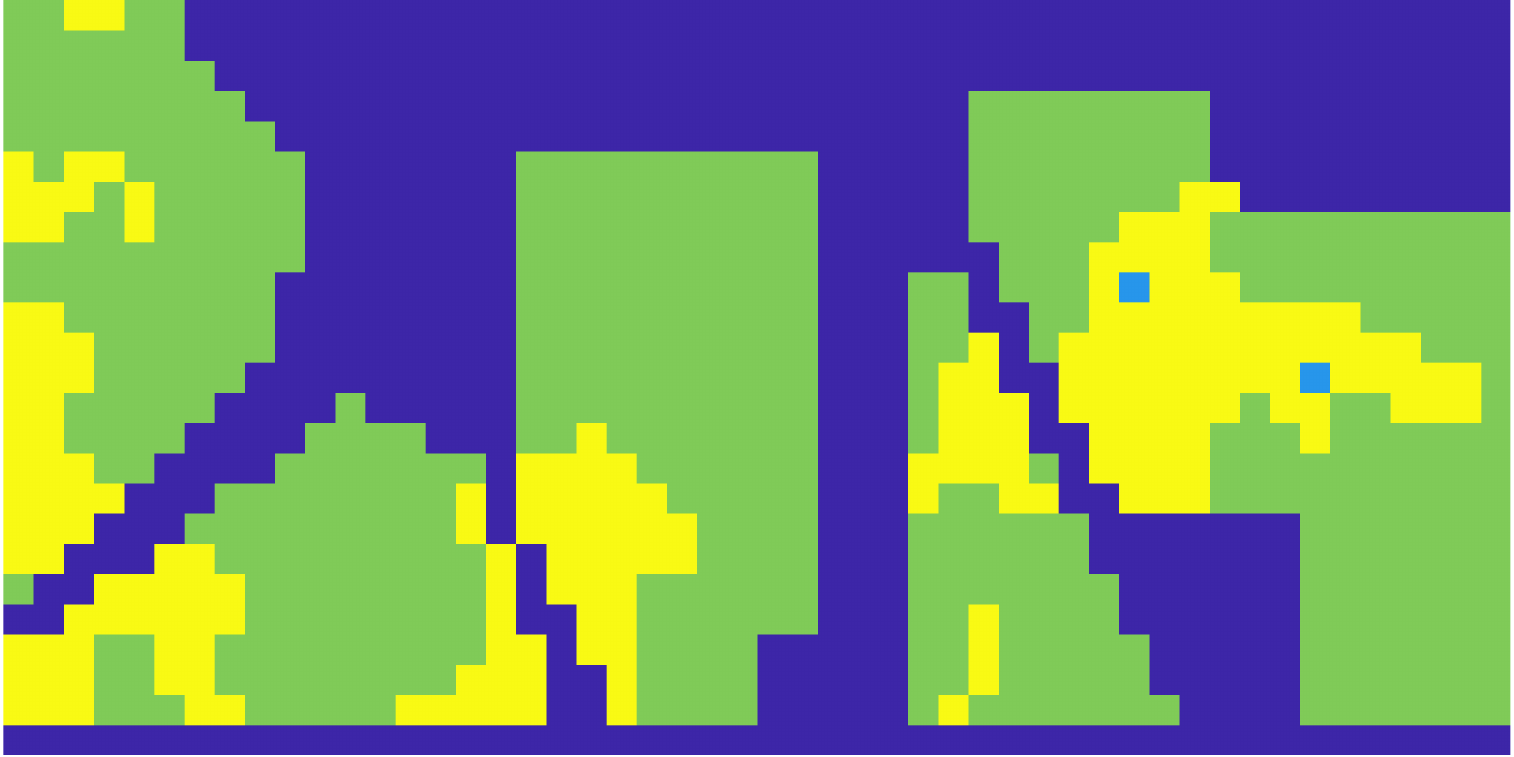}
\caption{FMS}
\end{subfigure}
\begin{subfigure}{ .09\textwidth}
\includegraphics[width=\textwidth]{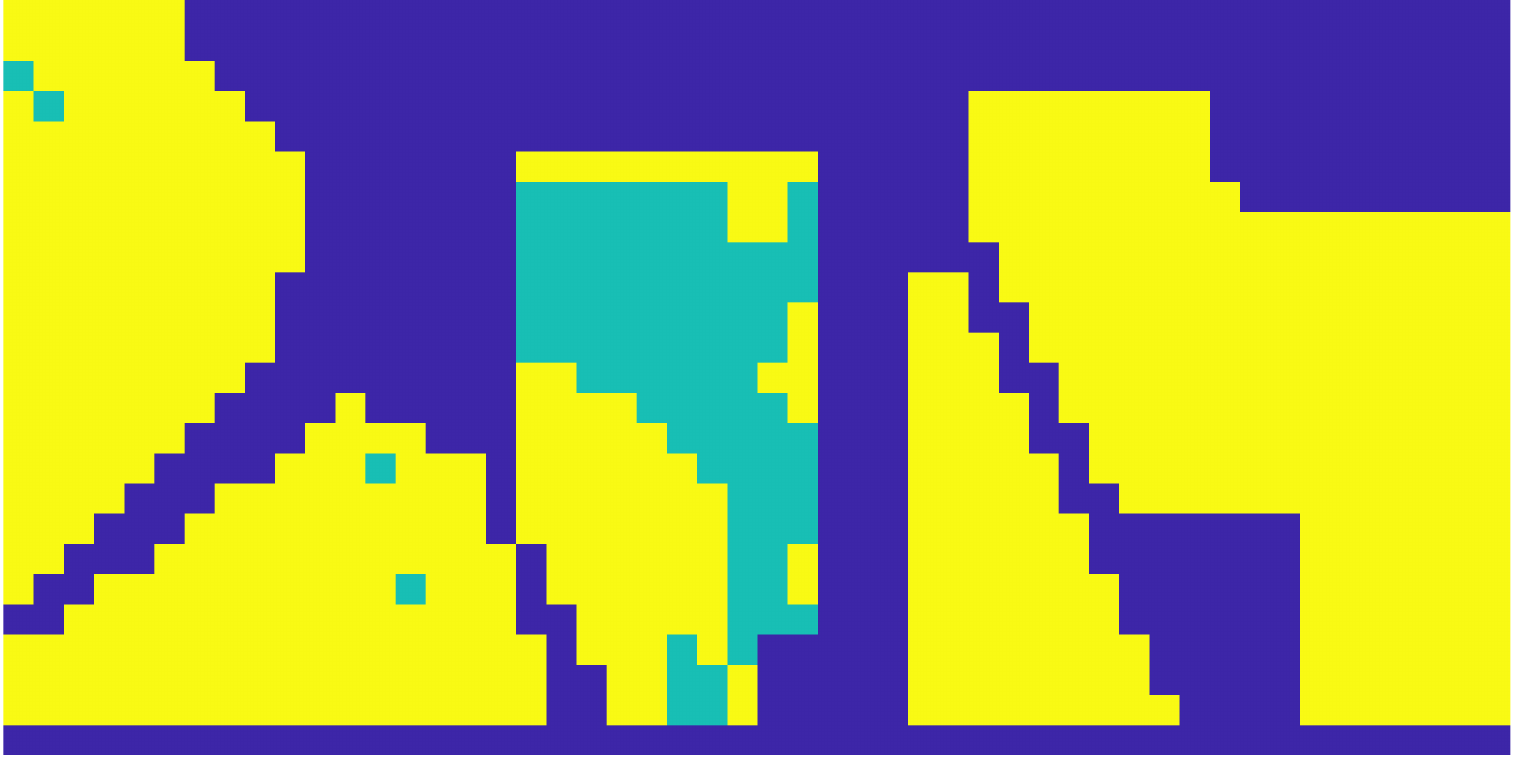}
\caption{FSFDPC}
\end{subfigure}
\begin{subfigure}{ .09\textwidth}
\includegraphics[width=\textwidth]{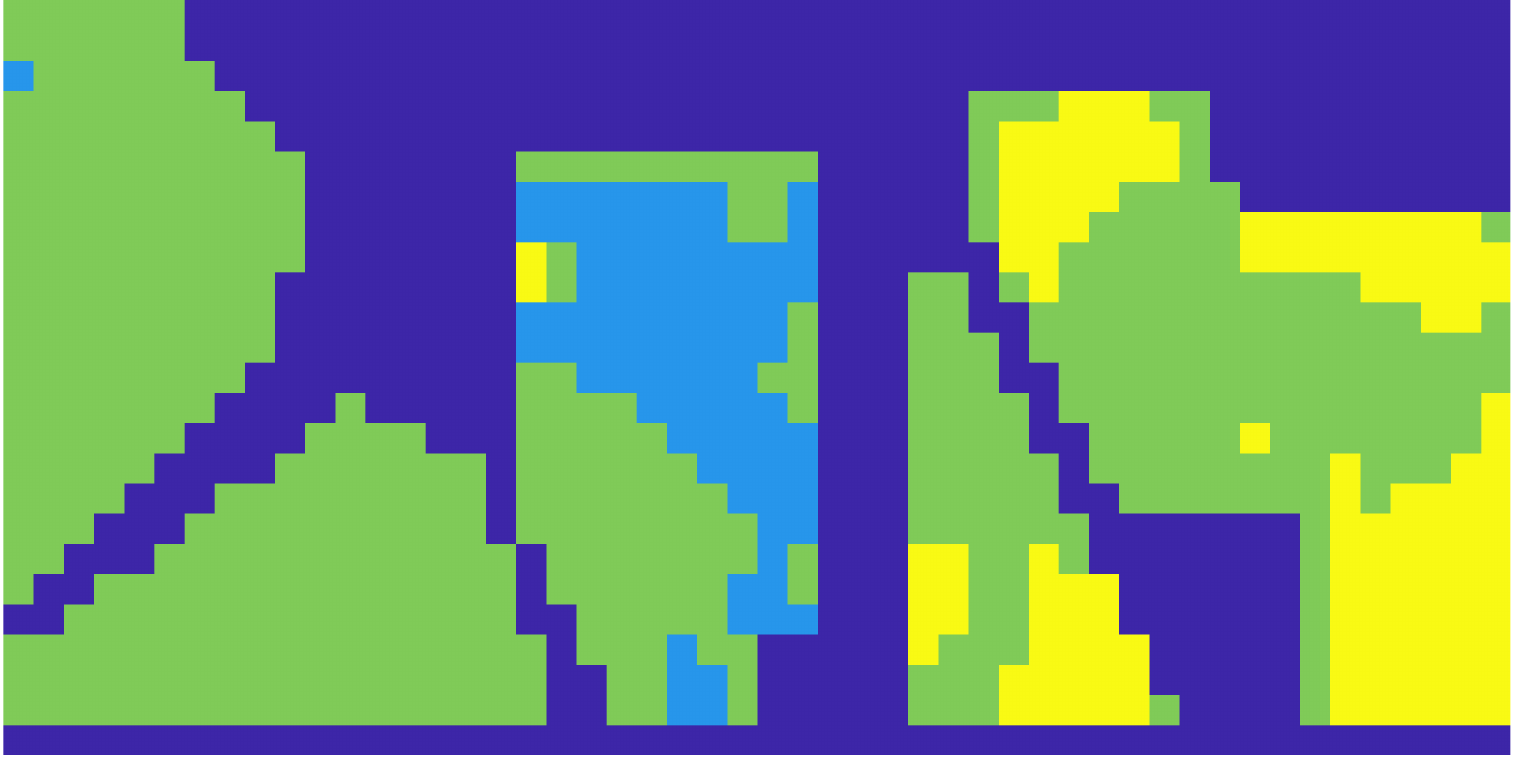}
\caption{DL}
\end{subfigure}
\begin{subfigure}{ .09\textwidth}
\includegraphics[width=\textwidth]{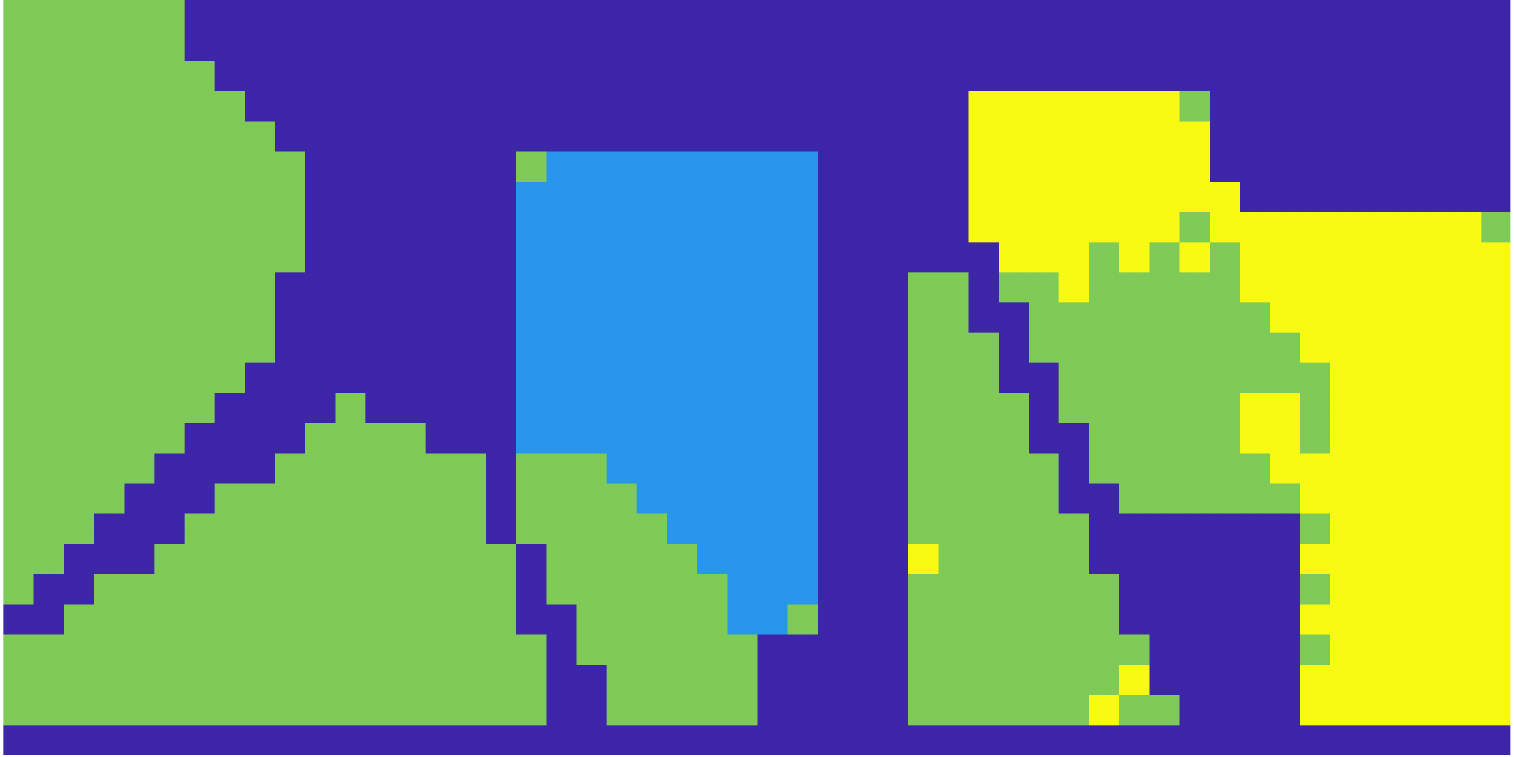}
\caption{DLSS}
\end{subfigure}
\begin{subfigure}{ .09\textwidth}
\includegraphics[width=\textwidth]{Images/IP_Images/IP_GT-crop.pdf}
\caption{GT}
\end{subfigure}
\caption{\label{fig:ResultsIP}Clustering results for Indian Pines dataset.  The impact of the spectral-spatial labeling scheme is apparent, as the labels for the DLSS method are more spatially regular than those of the DL method.  Note that the regions of difference between DL and DLSS are primarily near boundaries of classes and in very small interior regions.  Near the boundaries of classes, pixels are likely to be far from the spectral class cores, and hence are more likely to be labeled based on spatial properties.  The small interior regions are unlikely to be formed under the DLSS labeling regime, since these regions consist of points whose spectral label differs from their spatial consensus label.  The simplified DL method performs second best, and in particular outperforms FSFDPC, which performs well among the comparison methods.} 
\end{figure}

The Indian Pines dataset used for experiments is a subset of the full Indian Pines datasets, consisting of three classes that are difficult to distinguish visually; see Figure \ref{fig:IP}.  This dataset is expected to be challenging due to the lack of clear separation between the classes.  Results for Indian Pines appear in Figure \ref{fig:ResultsIP} and Table \ref{tab:Summary}.

The proposed methods, DL and DLSS, perform the best, with DLSS strongly outperforming the rest.  For the average accuracy statistic, DBSCAN performs as well as DL, indicating that the clusters for this data are likely of comparable empirical density.  The use of diffusion distances for mode detection and determination of spectral neighbors is evidently useful, as DL significantly outperforms FSFDPC, which has among the best quantitative performance of the comparison methods.  Moreover, the use of the proposed spectral-spatial labeling scheme DLSS clearly improves over spectral-only labeling DL:  as seen in Figure \ref{fig:ResultsIP}, DLSS correctly labels many small interior regions that DL labels incorrectly.  


\subsubsection{Pavia Dataset}

\begin{figure}
\centering
\includegraphics[width=.24\textwidth]{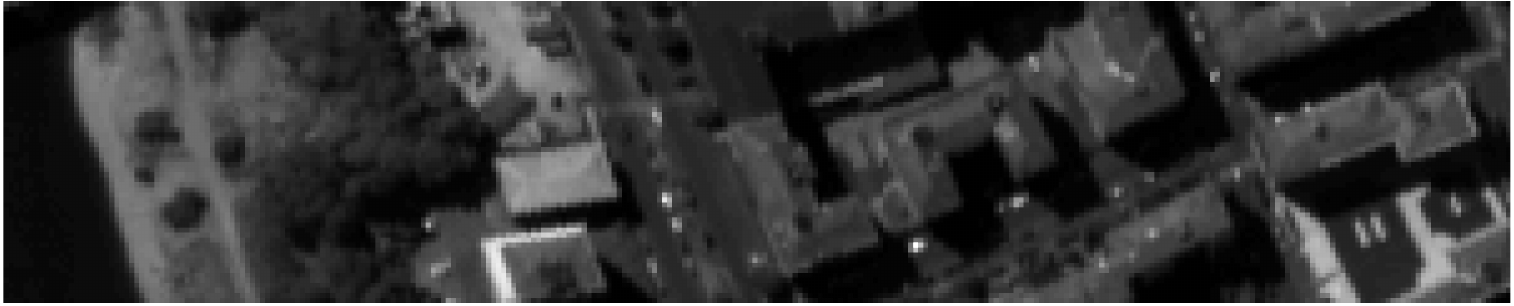}
\includegraphics[width=.24\textwidth]{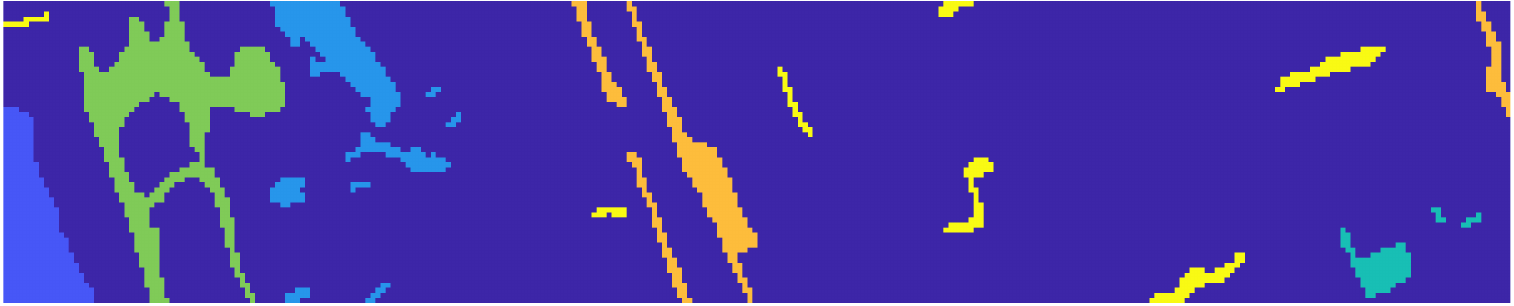}
\caption{\label{fig:Pavia}The Pavia data is a $270\times 50$ subset of the full Pavia dataset.  It contains 6 classes, some of which are not well-localized spatially.  The dataset was captured by the ROSIS sensor during a flight over Pavia, Italy.  The spatial resolution is 1.3 m/pixel.  There are 102 spectral bands.  Left: projection onto the first principal component of the data; right: ground truth (GT).} 
\end{figure}

The Pavia dataset used for experiments consists of a subset of the original dataset, and contains six classes, with one of them spread out across the image.  As can be seen in Figure \ref{fig:Pavia}, the yellow class is small and diffuse, which is expected to add challenge to this example.  Results appear in Table \ref{tab:Summary}.  Visual results appear in the online preprint version of this article.

The proposed methods give the best results, which also provide evidence of the value of both the diffusion learning stage and the spectral-spatial labeling scheme.  The proposed DLSS algorithm makes essentially only two errors: the yellow-green class is slightly mislabeled, and the blue-green class in the bottom right is labeled completely incorrectly.  However, both of these errors are made by all algorithms, often to a greater degree.  Among the comparison methods, SMCE performs best; classical spectral clustering also performs well.  


\subsubsection{Salinas A Dataset}
\label{subsubsec:SalinasA}

\begin{figure}[b]
\centering
\includegraphics[width=.24\textwidth]{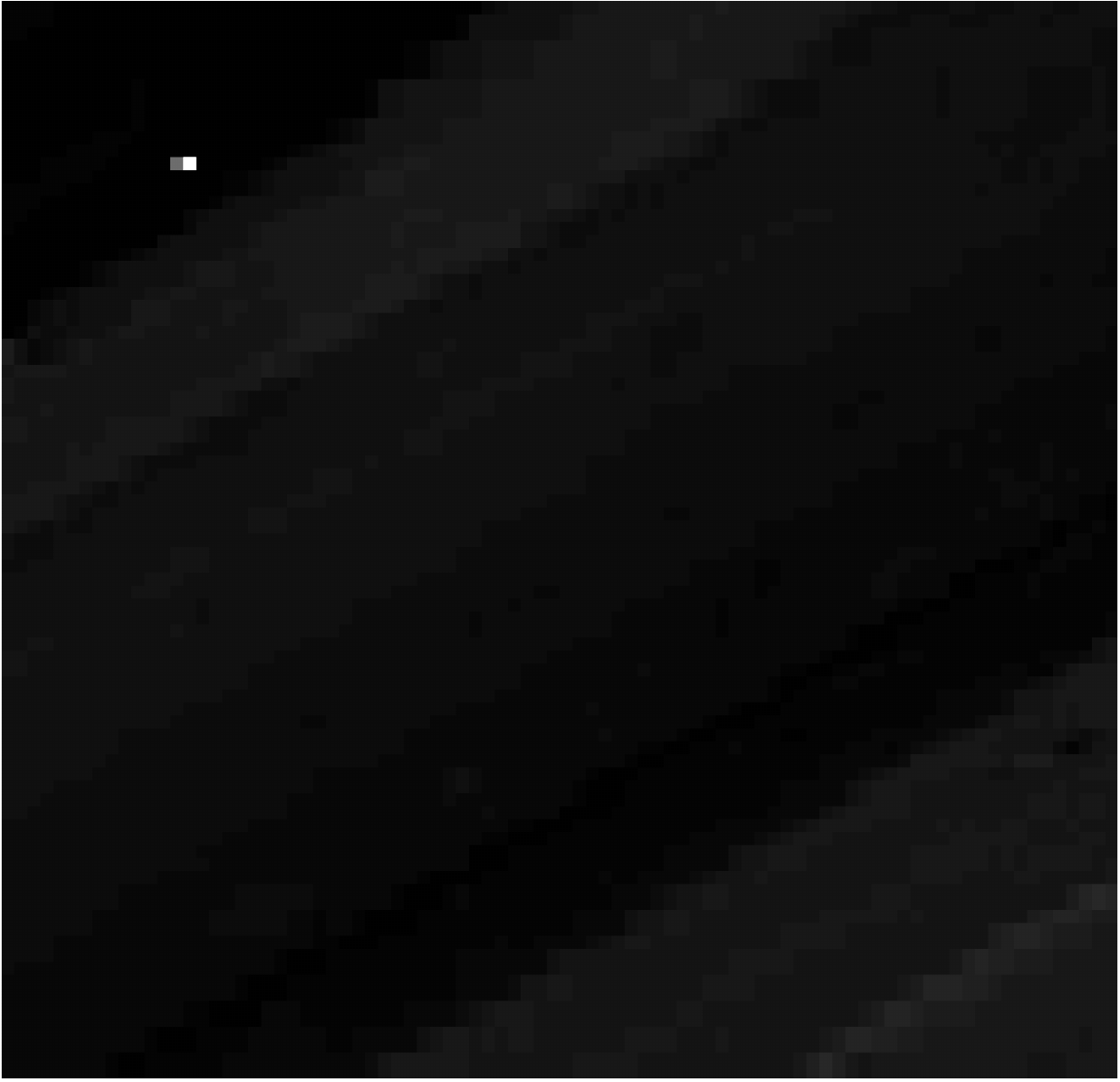}
\includegraphics[width=.24\textwidth]{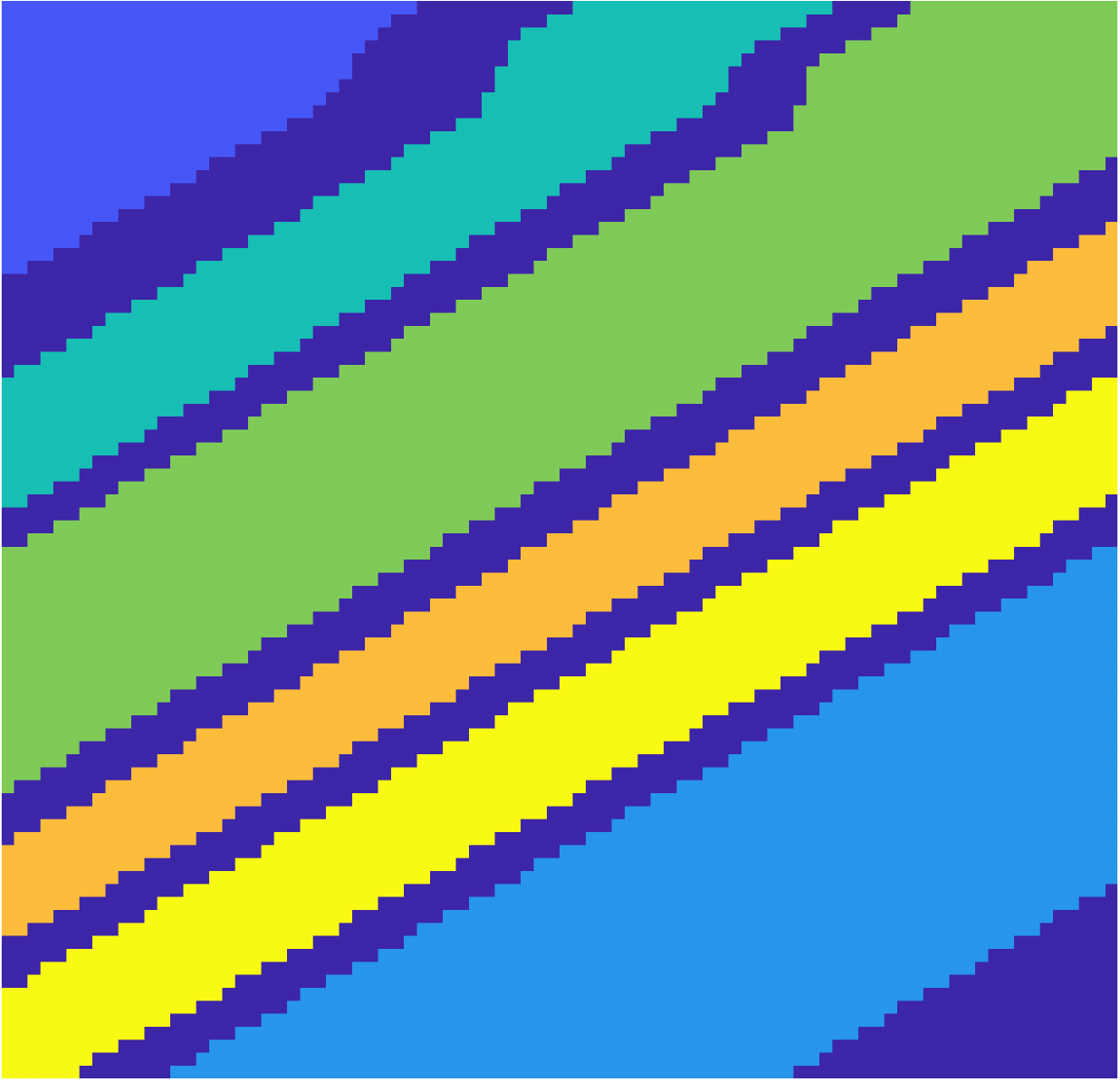}
\caption{\label{fig:SalinasA}The Salinas A data consists of the full $86\times 83$ HSI.  It contains 6 classes, all of which are well-localized spatially.  The dataset was captured over Salinas Valley, CA, by the AVRIS sensor.  The spatial resolution is 3.7 m/pixel.  The image contains 224 spectral bands.  Left: projection onto the first principal component of the data; right: ground truth (GT).} 
\end{figure}

The Salinas A dataset (see Figure \ref{fig:SalinasA}) consists of 6 classes arrayed diagonally.  Some pixels in the original images have the same values, so some small Gaussian noise (variance $< 10^{-3})$ was added as a preprocessing step to distinguish these pixels.  
\begin{figure}
\centering
\begin{subfigure}{.09\textwidth}
\includegraphics[width=\textwidth]{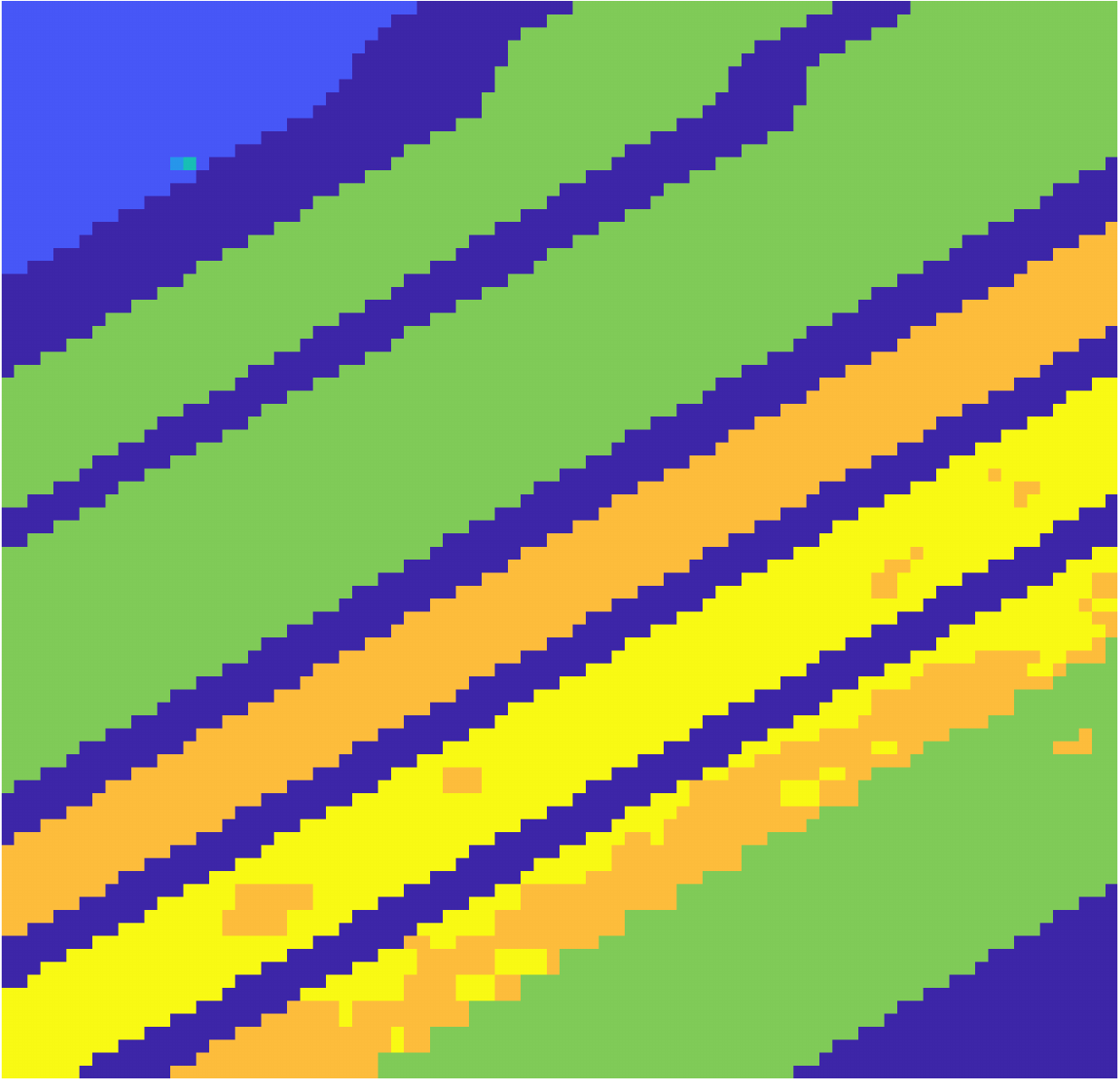}
\caption{$K$-means}
\end{subfigure}
\begin{subfigure}{.09\textwidth}
\includegraphics[width=\textwidth]{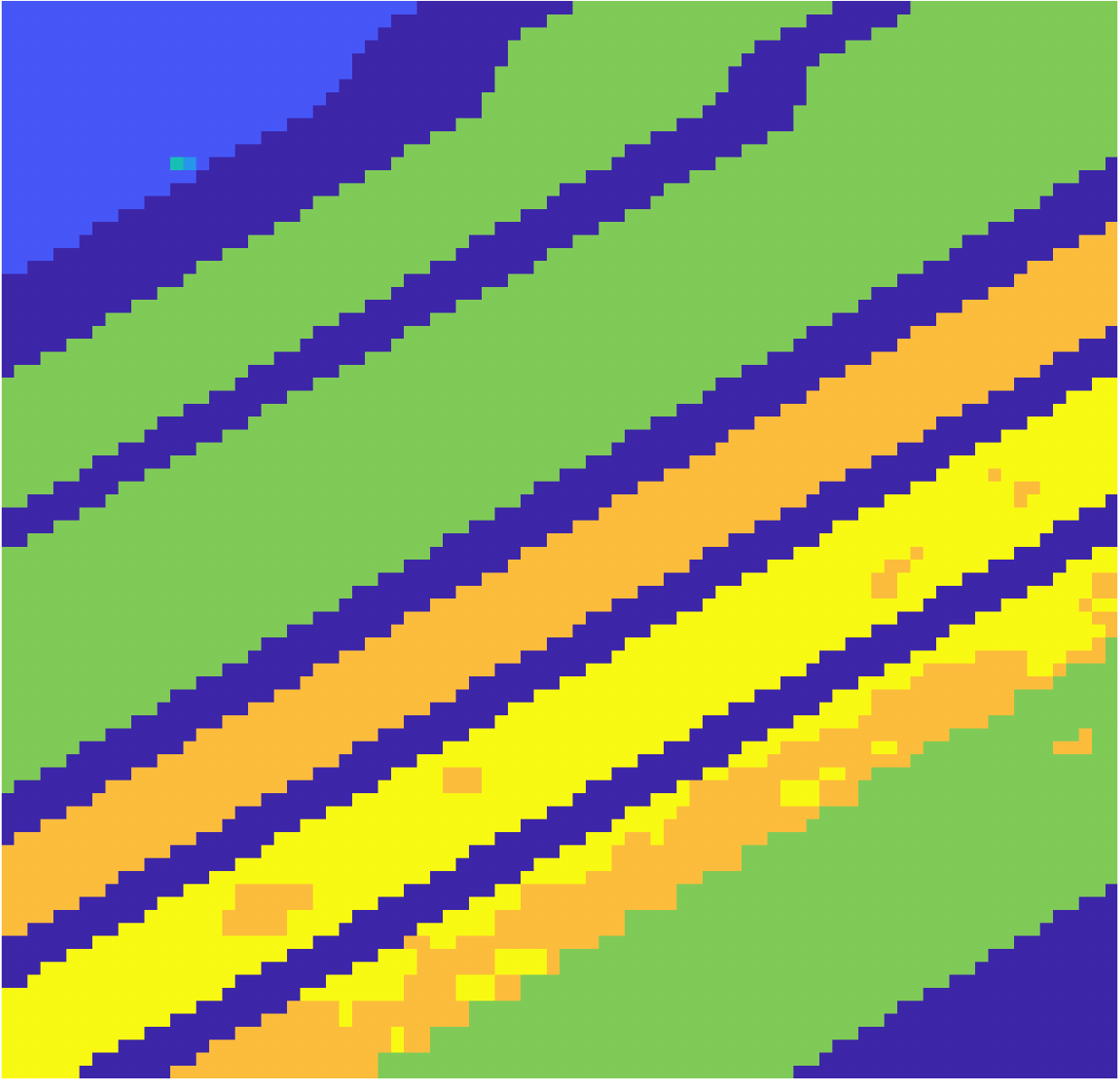}
\caption{PCA+$K$M}
\end{subfigure}
\begin{subfigure}{.09\textwidth}
\includegraphics[width=\textwidth]{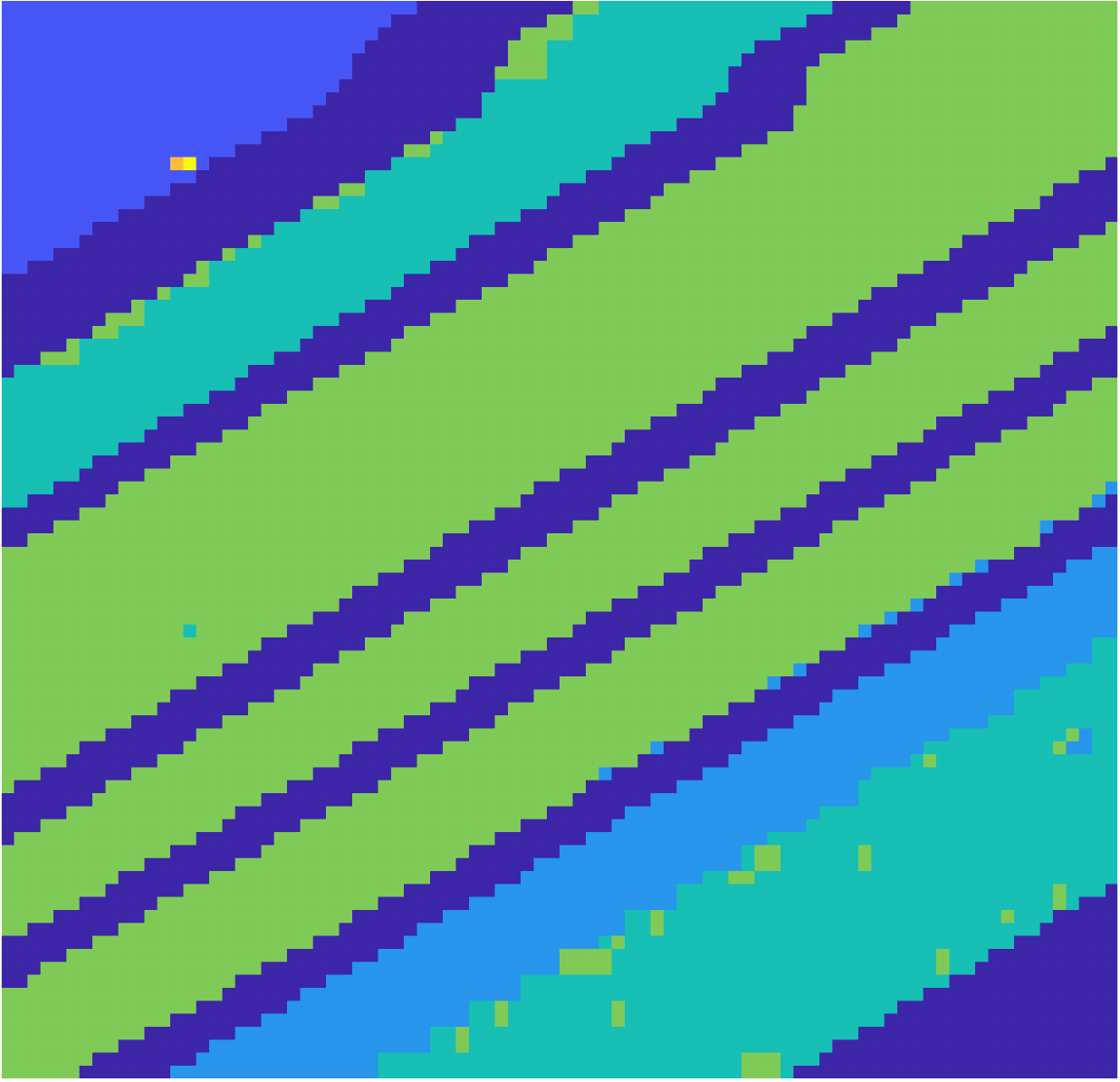}
\caption{ICA+$K$M}
\end{subfigure}
\begin{subfigure}{.09\textwidth}
\includegraphics[width=\textwidth]{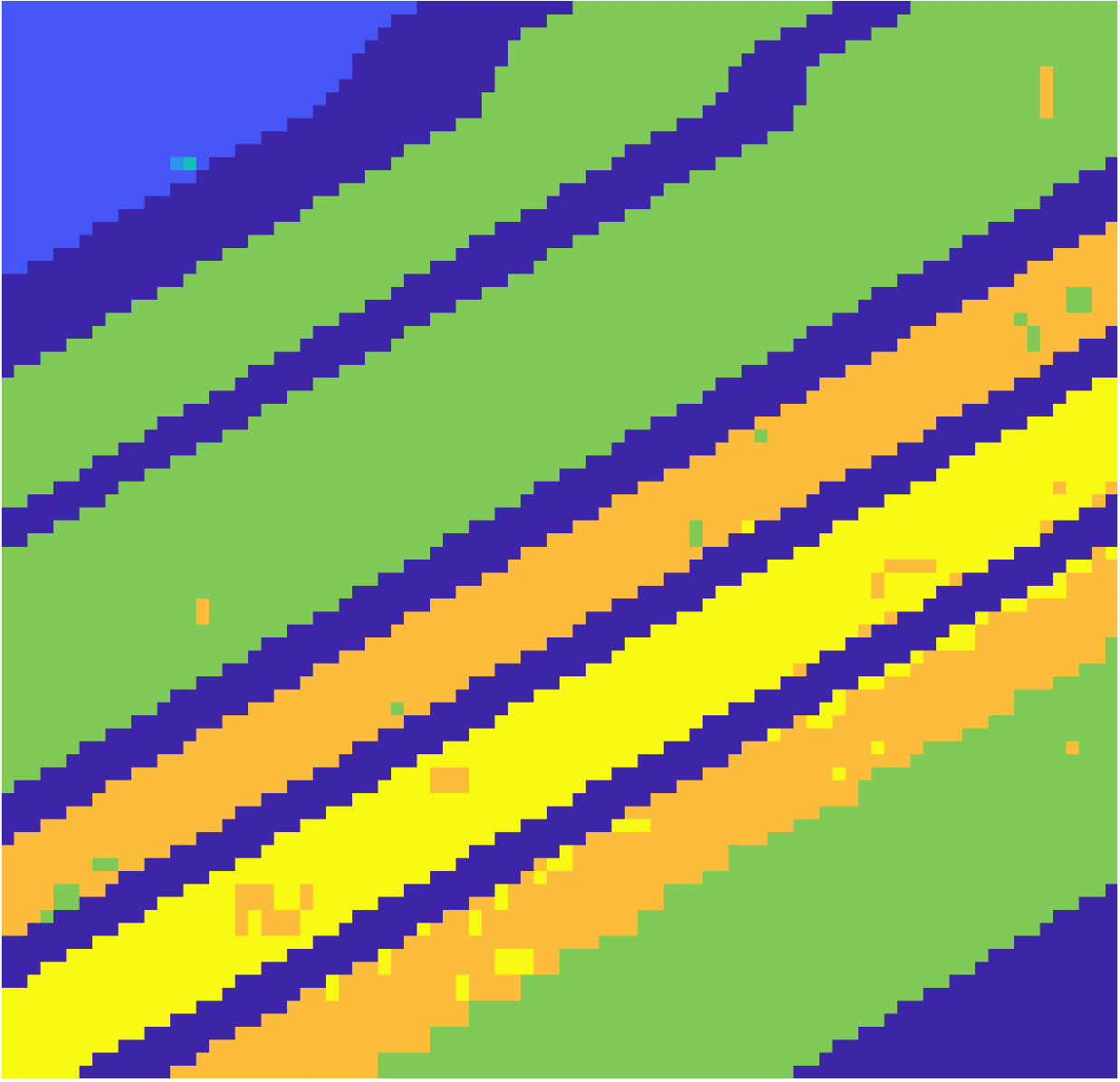}
\caption{RP+$K$M}
\end{subfigure}
\begin{subfigure}{ .09\textwidth}
\includegraphics[width=\textwidth]{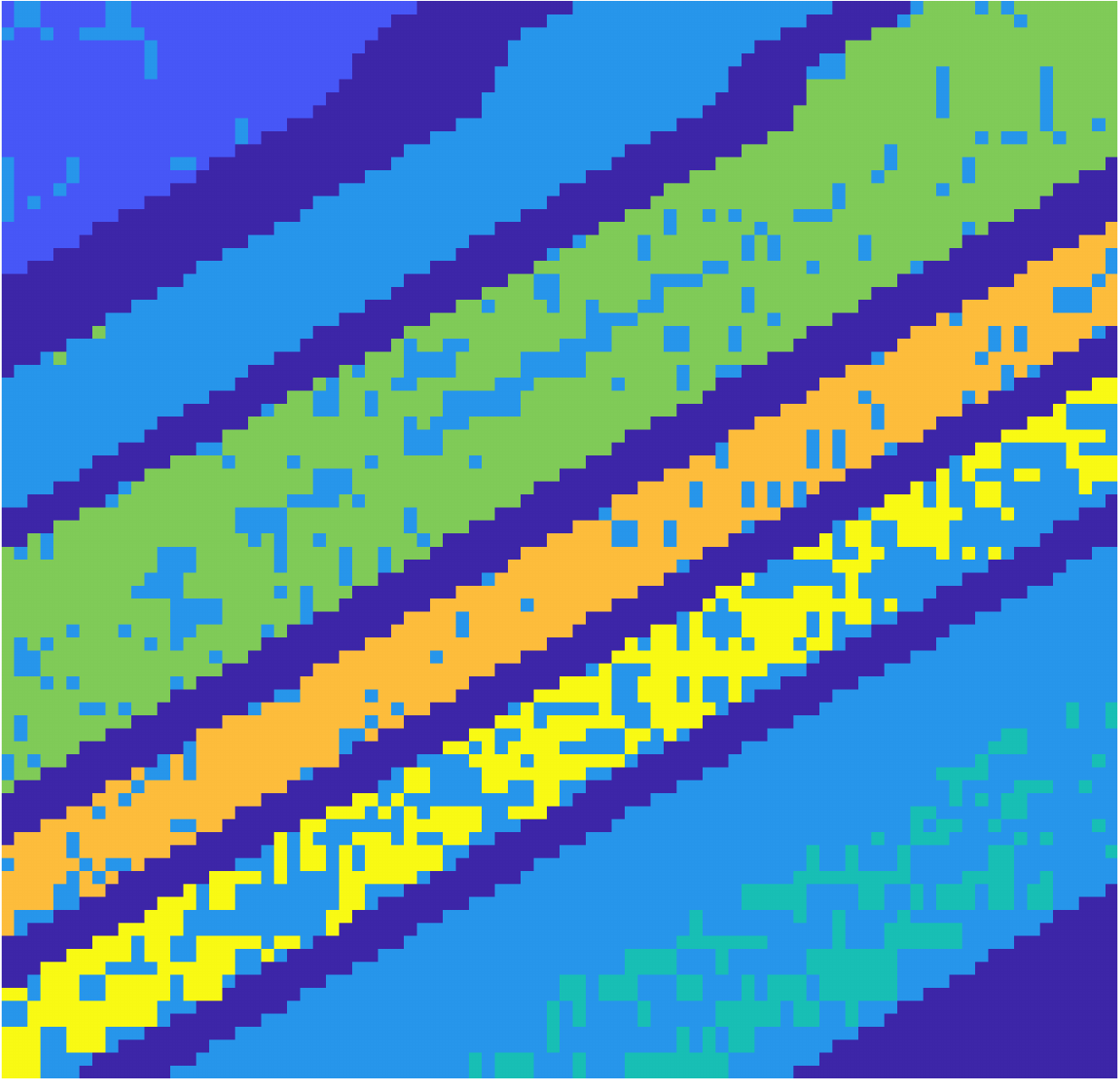}
\caption{DBSCAN}
\end{subfigure}
\begin{subfigure}{.09\textwidth}
\includegraphics[width=\textwidth]{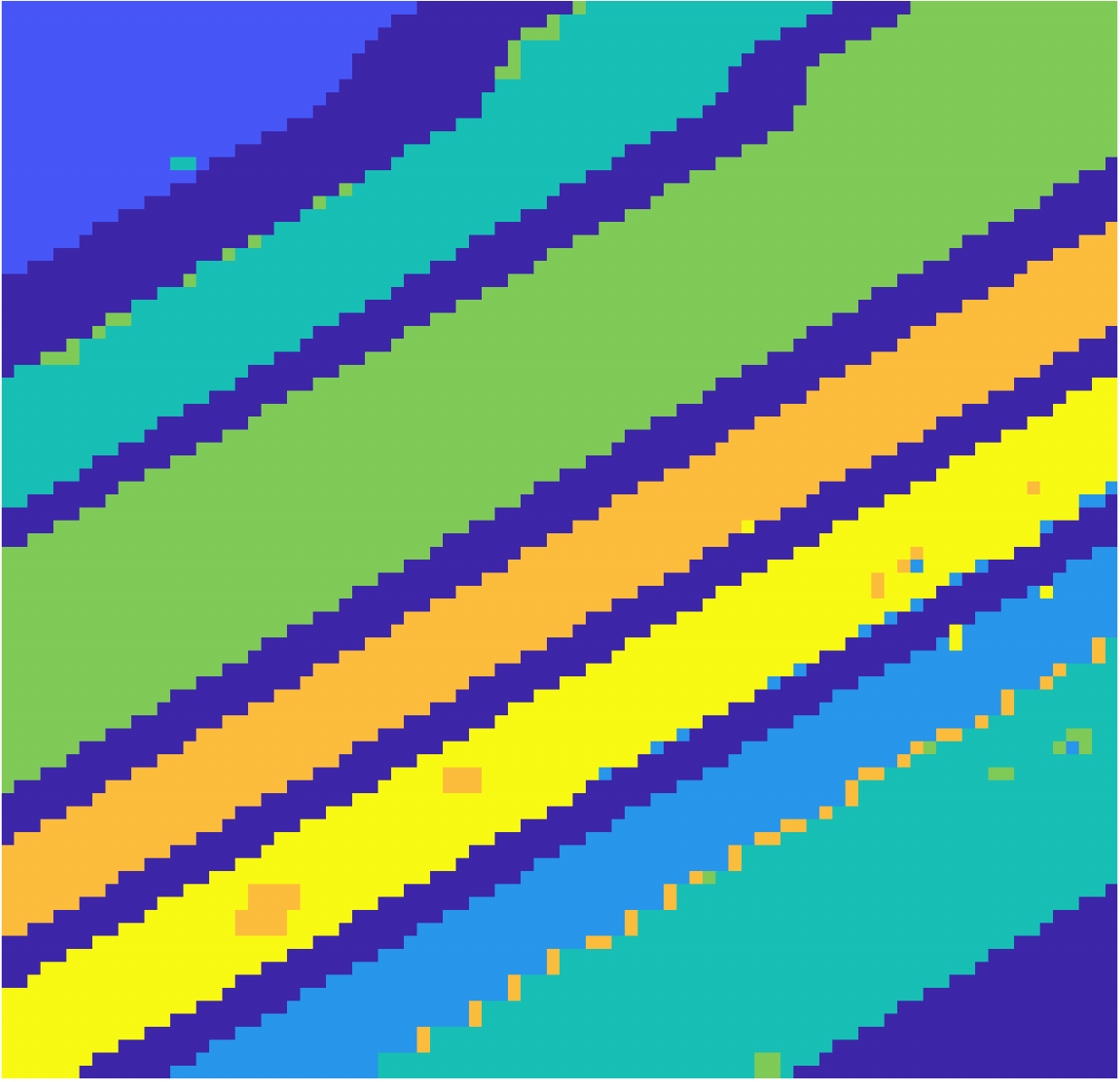}
\caption{SC}
\end{subfigure}
\begin{subfigure}{.09\textwidth}
\includegraphics[width=\textwidth]{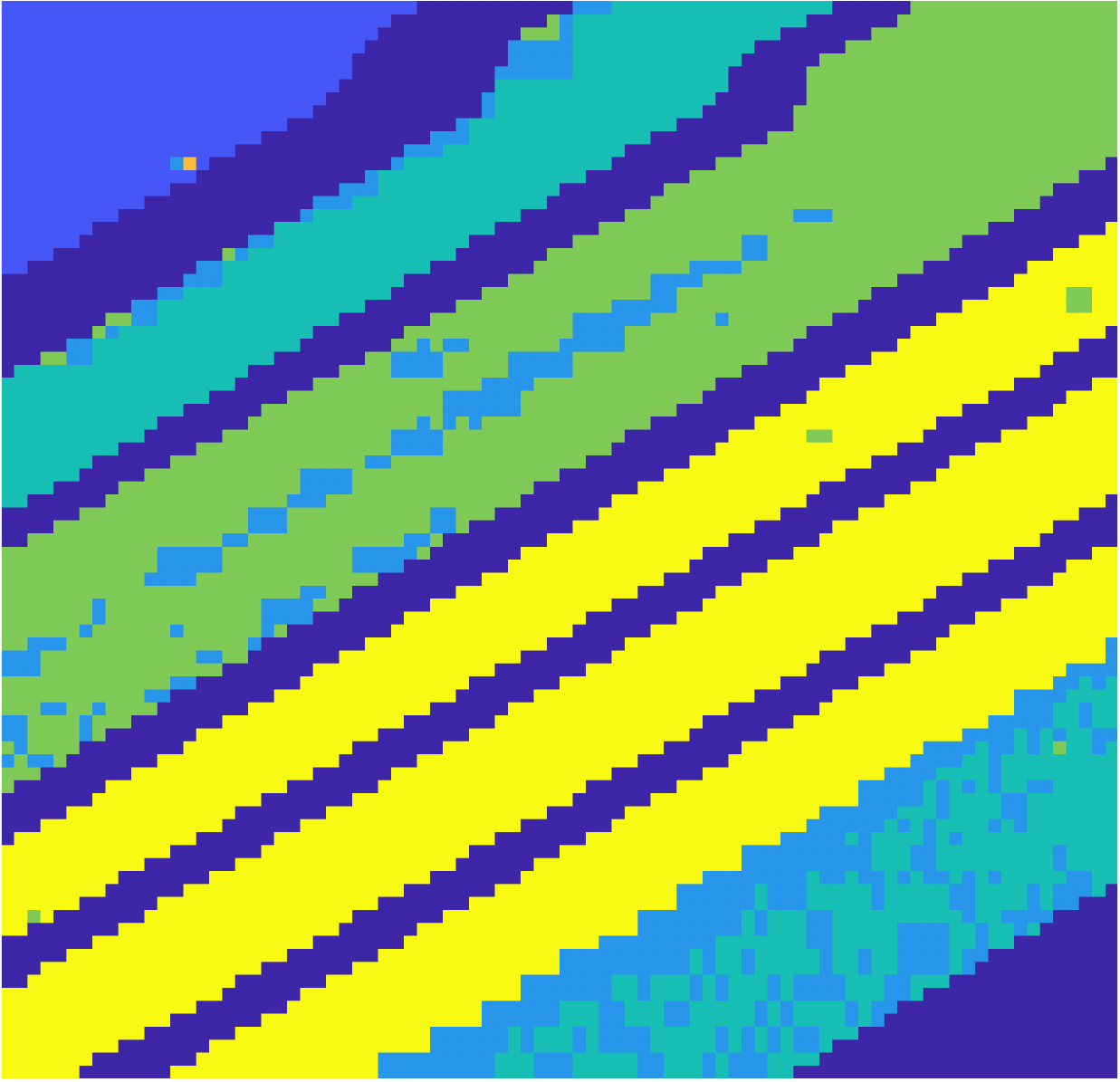}
\caption{GMM}
\end{subfigure}
\begin{subfigure}{.09\textwidth}
\includegraphics[width=\textwidth]{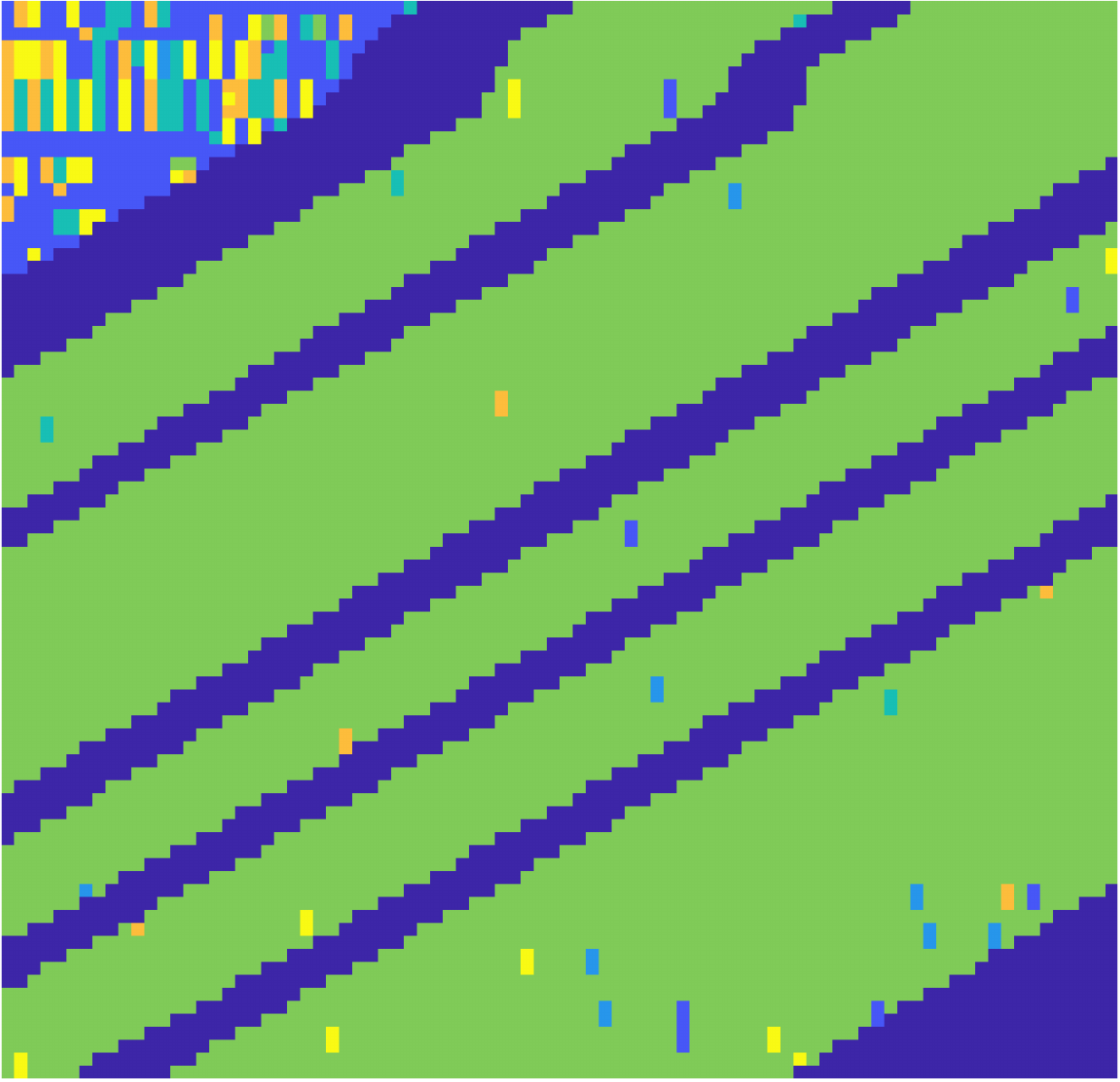}
\caption{SMCE}
\end{subfigure}
\begin{subfigure}{.09\textwidth}
\includegraphics[width=\textwidth]{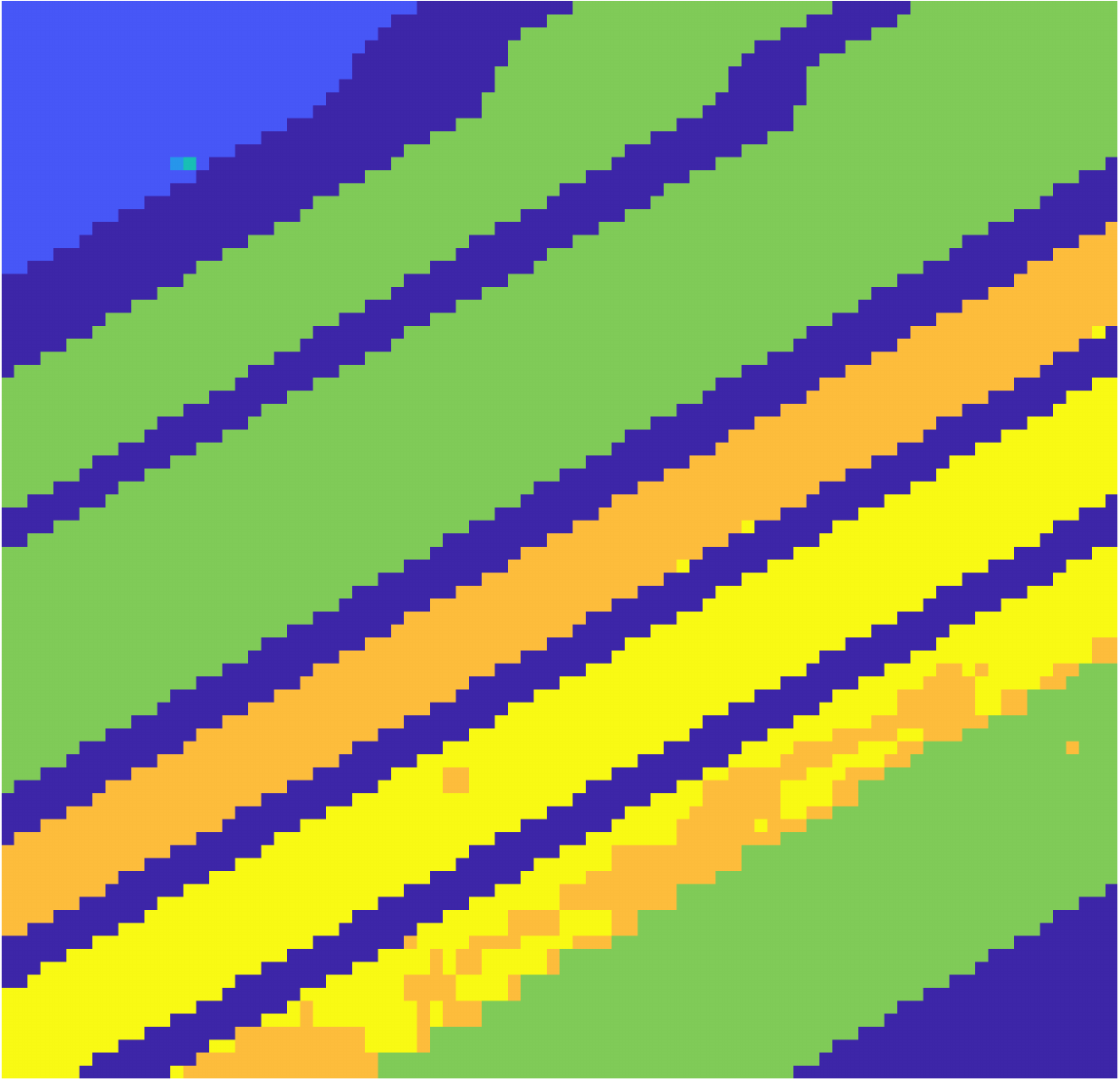}
\caption{HNMF}
\end{subfigure}
\begin{subfigure}{ .09\textwidth}
\includegraphics[width=\textwidth]{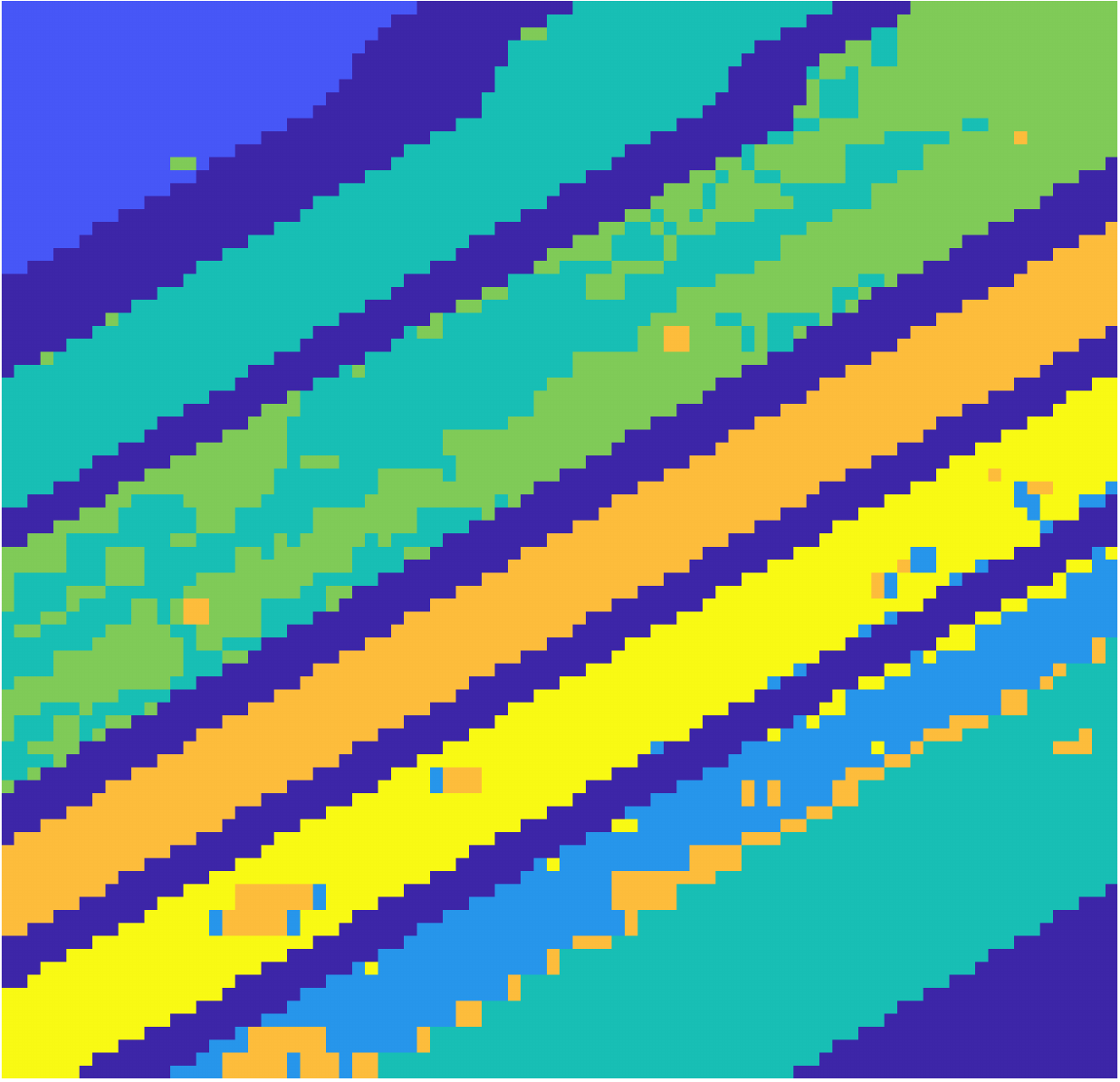}
\caption{FMS}
\end{subfigure}
\begin{subfigure}{ .09\textwidth}
\includegraphics[width=\textwidth]{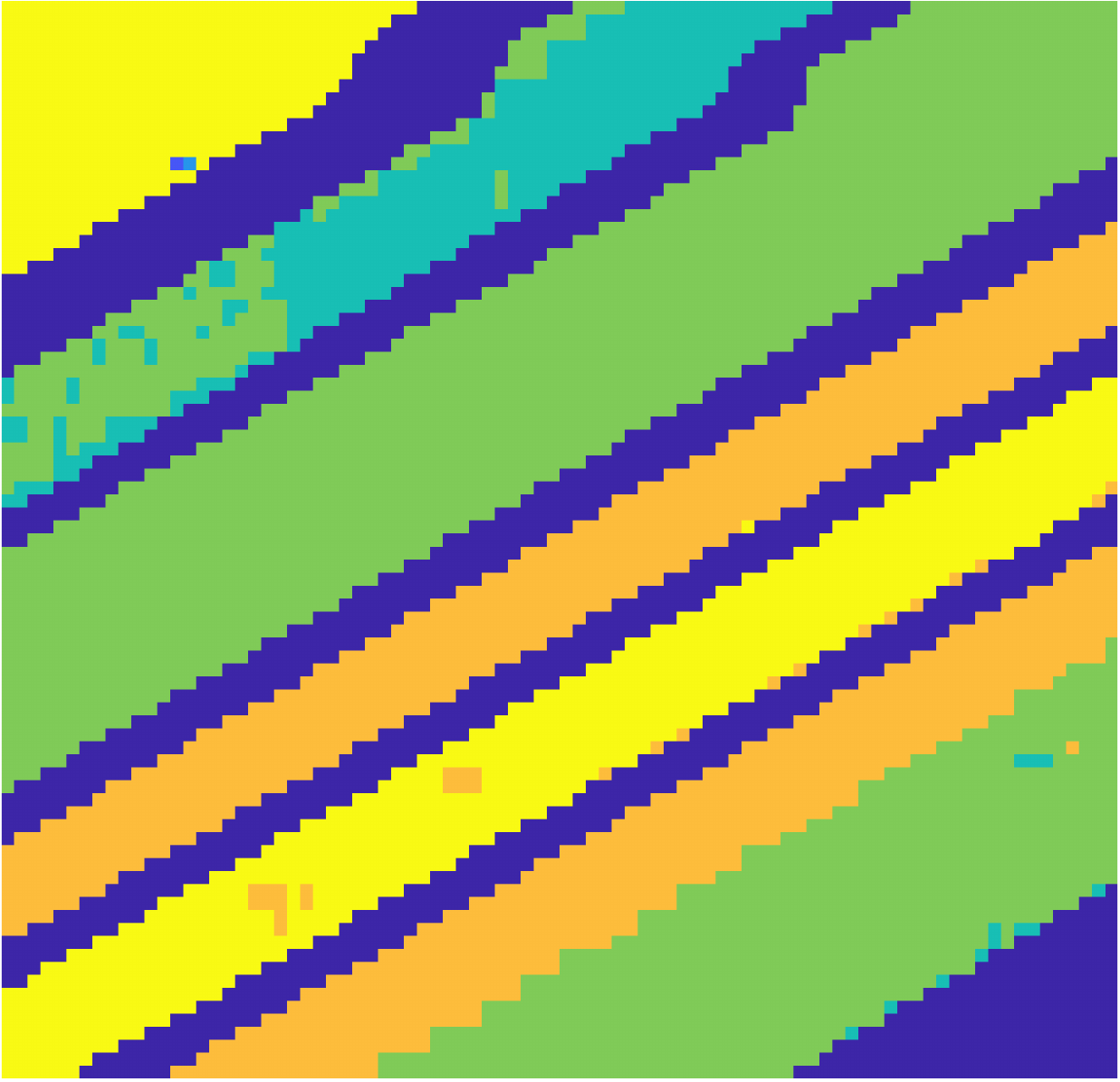}
\caption{FSFDPC}
\end{subfigure}
\begin{subfigure}{ .09\textwidth}
\includegraphics[width=\textwidth]{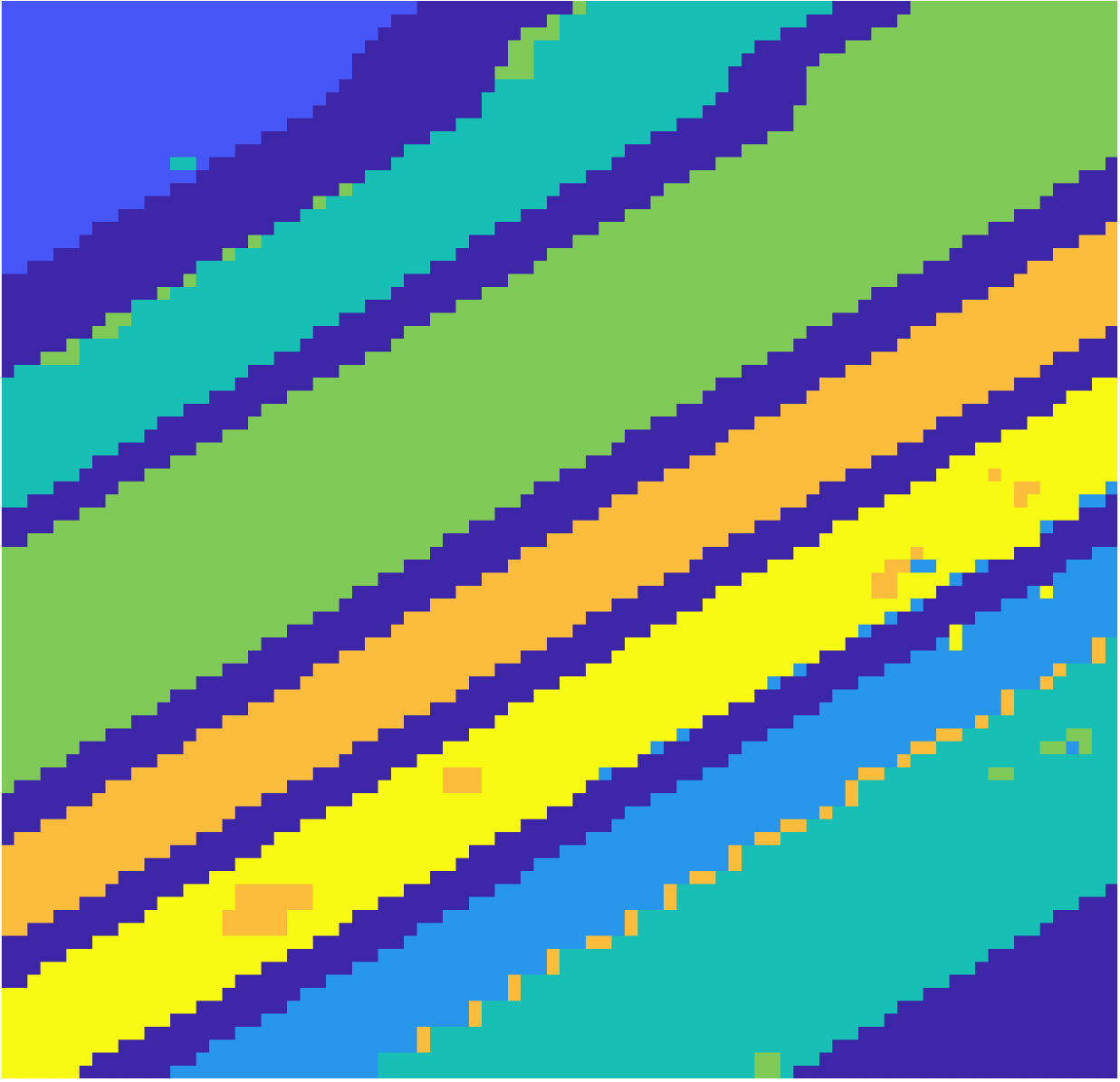}
\caption{DL}
\end{subfigure}
\begin{subfigure}{ .09\textwidth}
\includegraphics[width=\textwidth]{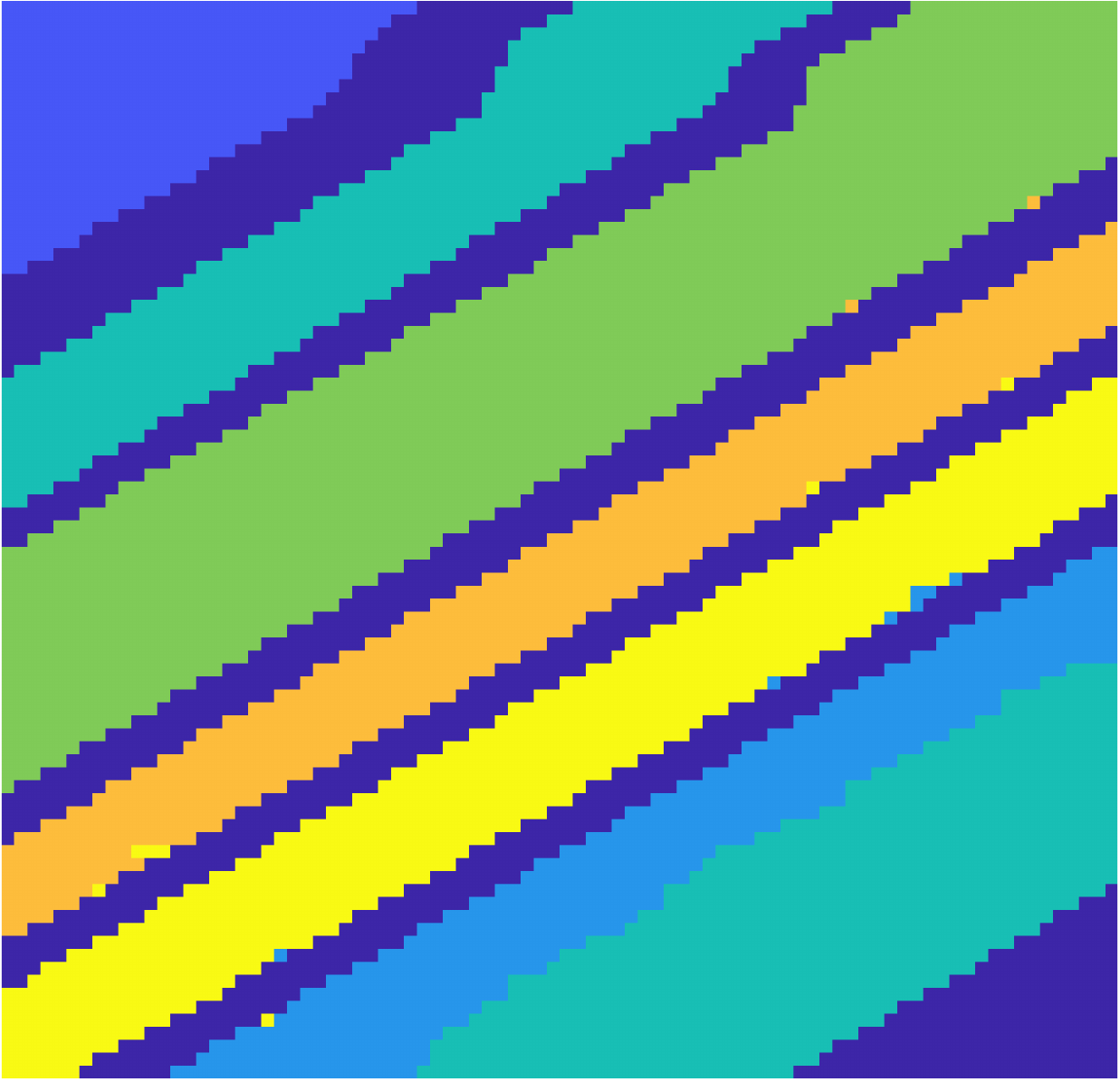}
\caption{DLSS}
\end{subfigure}
\begin{subfigure}{ .09\textwidth}
\includegraphics[width=\textwidth]{Images/SalinasA_Images/SalinasA_GT-crop.pdf}
\caption{GT}
\end{subfigure}
\caption{\label{fig:ResultsSalinasA}Clustering results for Salinas A dataset.  The proposed method, DLSS performs best, with the simplified DL method and benchmark spectral clustering also performing well.  Notice that the spectral-spatial labeling scheme removes some of the mistakes in the yellow cluster, and also improves the labeling near some class boundaries.  However, it is not able to fix the mislabeling of the light blue cluster in the lower right.  Indeed, all methods split the cluster in the lower right of the image, indicating the challenging aspects of this dataset for unsupervised learning.} 
\end{figure}
For this dataset, the proposed DLSS method performs best, with the only error made in splitting the bottom right cluster into two pieces, an error made by all algorithms.  The simpler DL method also performs well, as does the benchmark spectral clustering algorithm.  Comparing the labeling for DL and DLSS, the small regions of mislabeled pixels in DL are correctly labeled in DLSS, because these pixels are likely of low empirical density, and hence benefit from being labeled based on both spectral and spatial similarity, not spectral similarity alone.  However, some pixels correctly classified by DL were labeled incorrectly by DLSS, indicating that the spatial proximity condition enforced in DLSS may not lead to improved results for every pixel.  Details on this, and how to tune the size of the neighborhood with which spatial consensus labels are computed, are given in Section \ref{subsec:SpaceParameter}.

\subsubsection{Kennedy Space Center Data Set}
\label{subsubsec:KSC}

\begin{figure}[b]
\centering
\includegraphics[width=.24\textwidth]{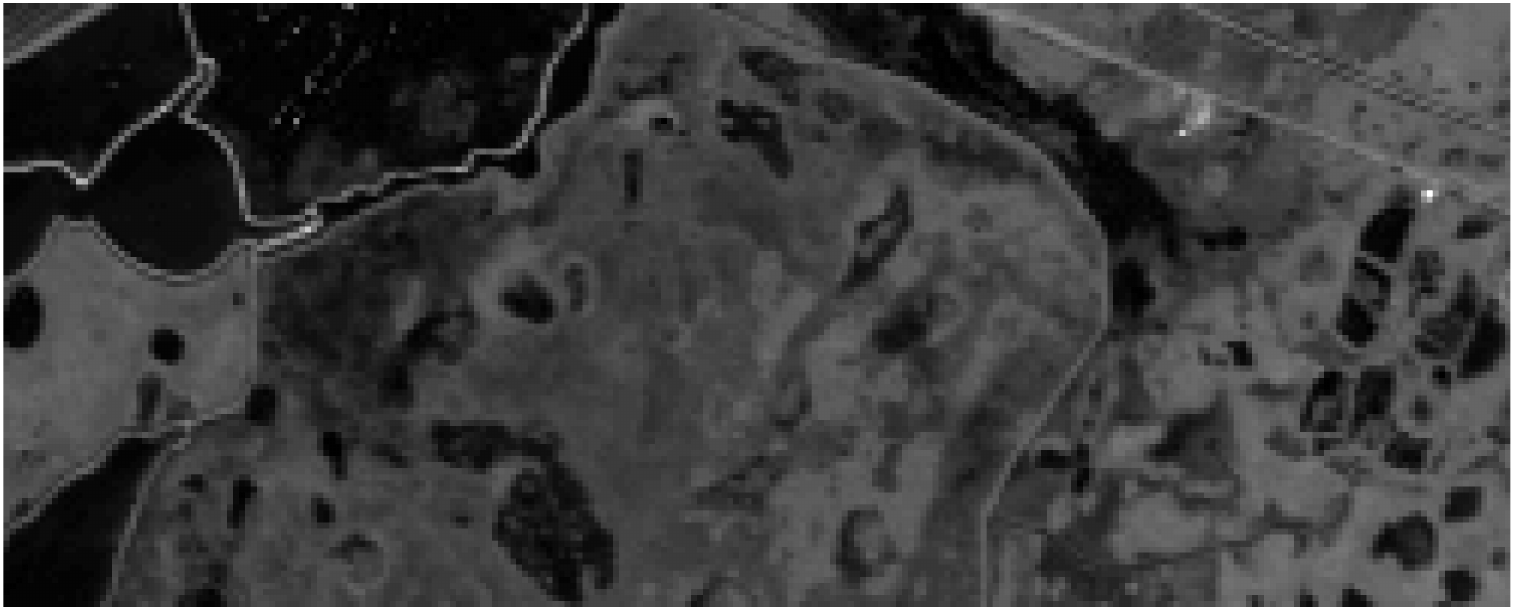}
\includegraphics[width=.24\textwidth]{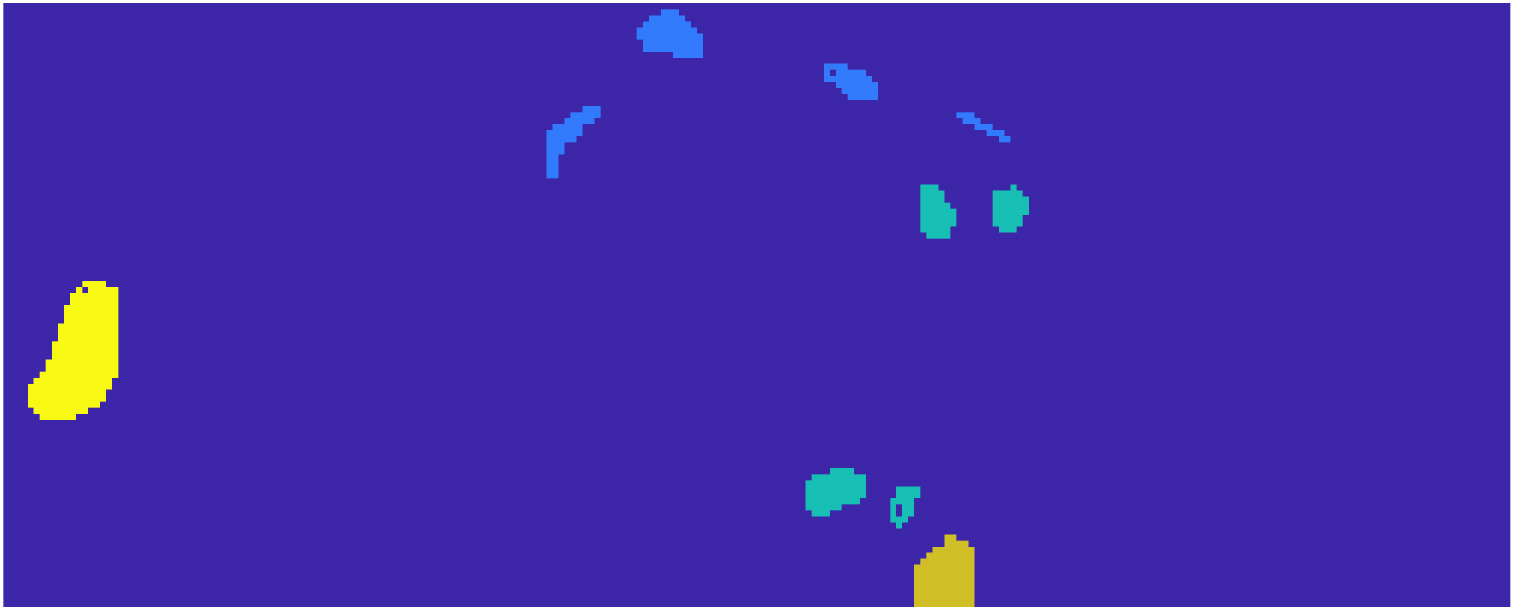}
\caption{\label{fig:KSC}The Kennedy Space Center data is a $250\times 100$ subset of the full Kennedy Space Center dataset.  It contains 4 classes, some of which are not well-localized spatially.  The scene was captured with the NASA AVIRIS instrument over the Kennedy Space Center (KSC), Florida, USA.  The spatial resolution is 18 m. After removing low signal-to-noise-ratio and water-absorption bands, the dataset consists of 176 bands.  Left: projection onto the first principal component of the data; right: ground truth (GT).} 
\end{figure}

The Kennedy Space Center dataset used for experiments consists of a subset of the original dataset, and contains four classes.  Figure \ref{fig:KSC} illustrates the first principal component of the data, as well as the labeled ground truth, which consists of the examples of four vegetation types which dominate the scene.  Results appear in Table \ref{tab:Summary}.
The proposed methods yield the best results, noting that the FMS method also performs well.  Most linear methods, such as $K$-means with linear dimension reduction or NMF, perform poorly, suggesting that nonlinear methods are needed for this data.  Spectral clustering performs much better than the linear methods.  We note that spatial information for this dataset is less helpful than for the Indian Pines and Pavia datasets.

\subsubsection{Overall Comments on Clustering}
Quantitative results for the clustering experiments appear in Table \ref{tab:Summary}.  We see that the DLSS method performs best among all metrics for all datasets.  The DL method generally performs second best, though DBSCAN, spectral clustering, and SMCE occasionally perform comparably to DL.  It is notable that DL outperforms FSFDPC, which uses a similar labeling scheme, but computes modes with Euclidean distances, rather than diffusion distances.  This provides empirical evidence for the need to use nonlinear methods of measuring distances for HSI. 

\begin{figure*}
\centering
\begin{adjustbox}{max width=\textwidth}
\begin{tabular}{| c | c | c | c | c | c | c | c | c | c  | c | c | c ||}\hline
Method & OA I.P. & AA I.P. & $\kappa$ I.P. & OA P. & AA P. & $\kappa$ P. & OA S.A. & AA S.A. & $\kappa$ S.A. & OA  K.S.C. & AA K.S.C. & $\kappa$ K.S.C.\\ \hline
$K$-means & 0.43 & 0.38 & 0.09 & 0.78 & 0.62 &  0.72 & 0.63 & 0.66 & 0.52 & 0.36 & 0.25 & 0.01 \\ \hline
PCA+$K$-means & 0.43  & 0.38  & 0.10 &  0.78 & 0.62 & 0.72 & 0.63 &  0.66 &  0.52 & 0.36  & 0.25 & 0.01 \\ \hline
ICA+$K$-means & 0.41 & 0.36 & 0.06 &   0.67 &  0.55 & 0.58 & 0.57 &  0.56 & 0.44  & 0.36 & 0.25 & 0.01 \\ \hline
RP+$K$-means & 0.51  & 0.51 & 0.26 &  0.76 &  0.61 &  0.70 &  0.63 &  0.66 &  0.53 & 0.60  & 0.50 & 0.43 \\ \hline
DBSCAN & 0.63  & \underline{0.62}  & 0.43 & 0.73 & 0.72 &  0.66  & 0.71 & 0.71 &  0.63 & 0.36 & 0.25 & 0.01 \\ \hline
SC & 0.54 & 0.45  & 0.24 &   0.82  &  0.76  &  0.77 & \underline{0.83} &  \underline{0.88} &  \underline{0.80} & 0.62  & 0.52  & 0.44\\ \hline
GMM & 0.44 & 0.35  & 0.02 & 0.64 & 0.59 & 0.55 &  0.64 & 0.61 & 0.55 & 0.42 & 0.31 & 0.10 \\ \hline
SMCE & 0.52 & 0.45  & 0.22 & 0.83 &  0.77 &  0.79 &   0.47 & 0.42 &  0.30  & 0.36 & 0.26 & 0.01 \\ \hline
HNMF & 0.41 & 0.32  & -0.02 & 0.72 &   0.74 & 0.66 & 0.63 & 0.66 & 0.53   & 0.36 & 0.25 & 0.00 \\ \hline
FMS & 0.57 & 0.50  & 0.27 &  0.77 & 0.64 & 0.69 & 0.70 & 0.81 & 0.65 & 0.74 & 0.70 & 0.65 \\ \hline
FSFDPC  & 0.58 & 0.51 &  0.26 &  0.78 &  0.75 &  0.73 &  0.63 &  0.61 & 0.54  &  0.36 & 0.25 & 0.00 \\ \hline
DL & \underline{0.67} & \underline{0.62} & \underline{0.44} & \underline{0.85} & \underline{0.78} & \underline{0.81} & \underline{0.83} &  \underline{0.88}  & 0.79 & \underline{0.81} & \underline{0.72} & \underline{0.74} \\ \hline
DLSS & \textbf{0.85}  & \textbf{0.82}  & \textbf{0.75}  & \textbf{0.94} & \textbf{0.83} & \textbf{0.93} & \textbf{0.85} & \textbf{0.90} & \textbf{0.81} & \textbf{0.83} & \textbf{0.73}  & \textbf{0.76} \\ \hline
\end{tabular}

\end{adjustbox}
\caption{\label{tab:Summary}Summary of quantitative analyses of real HSI clustering; best results are in bold, second best are underlined.  The datasets have been abbreviated as I.P. (Indian Pines), P. (Pavia), S.A. (Salinas A), and K.S.C. (Kennedy Space Center).  Generally the proposed diffusion methods offer the strongest overall performance, particular DLSS.  In all cases, DL outperforms FSFDPC, indicating the importance of using diffusion distances over Euclidean distances for HSI clustering.}

\end{figure*}

\subsection{Active Learning}\label{subsec:ActiveLearning}

\begin{figure}
\centering
\begin{subfigure}{ .24\textwidth}
\includegraphics[width=\textwidth]{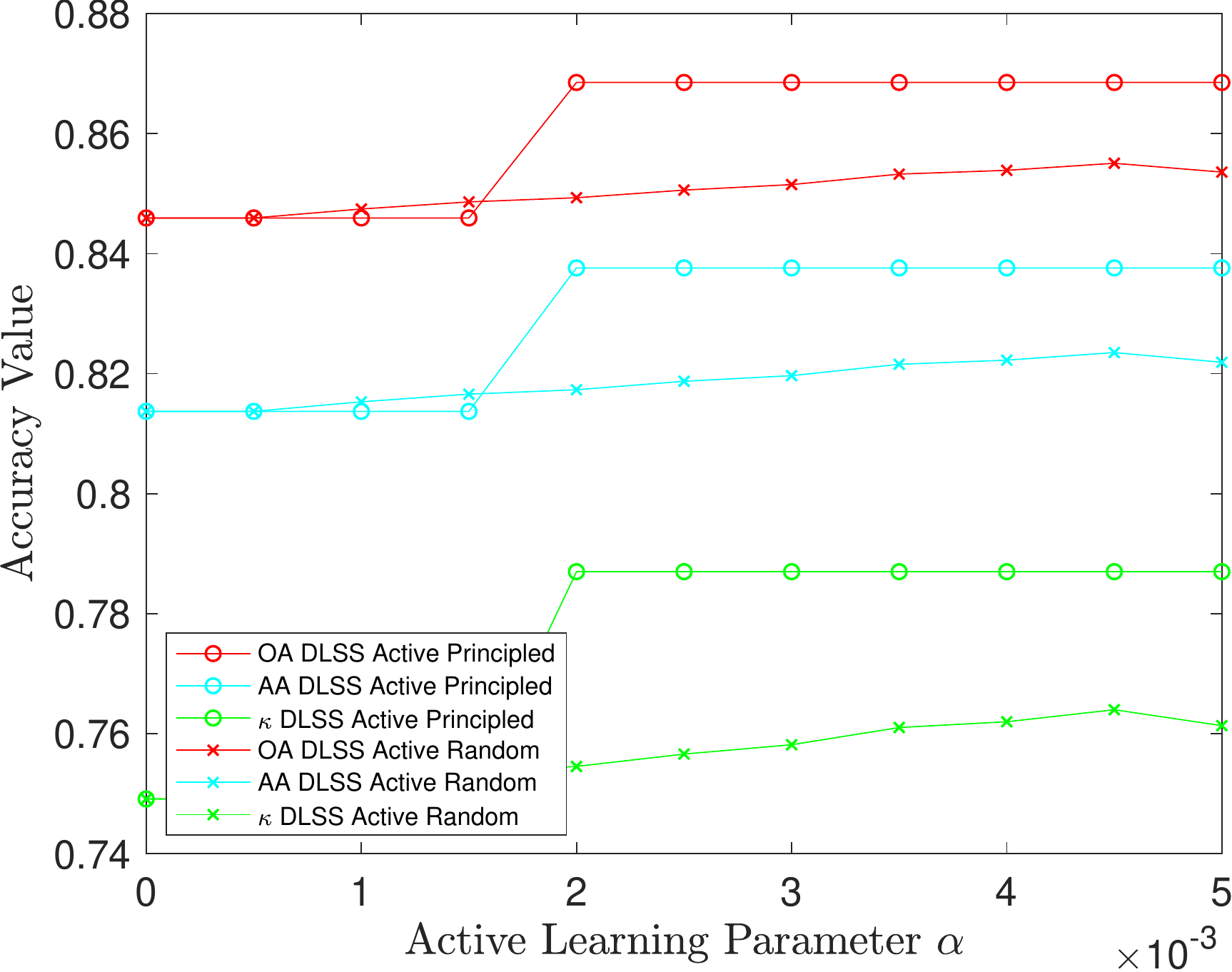}
\subcaption{Indian Pines}
\end{subfigure}
\begin{subfigure}{ .24\textwidth}
\includegraphics[width=\textwidth]{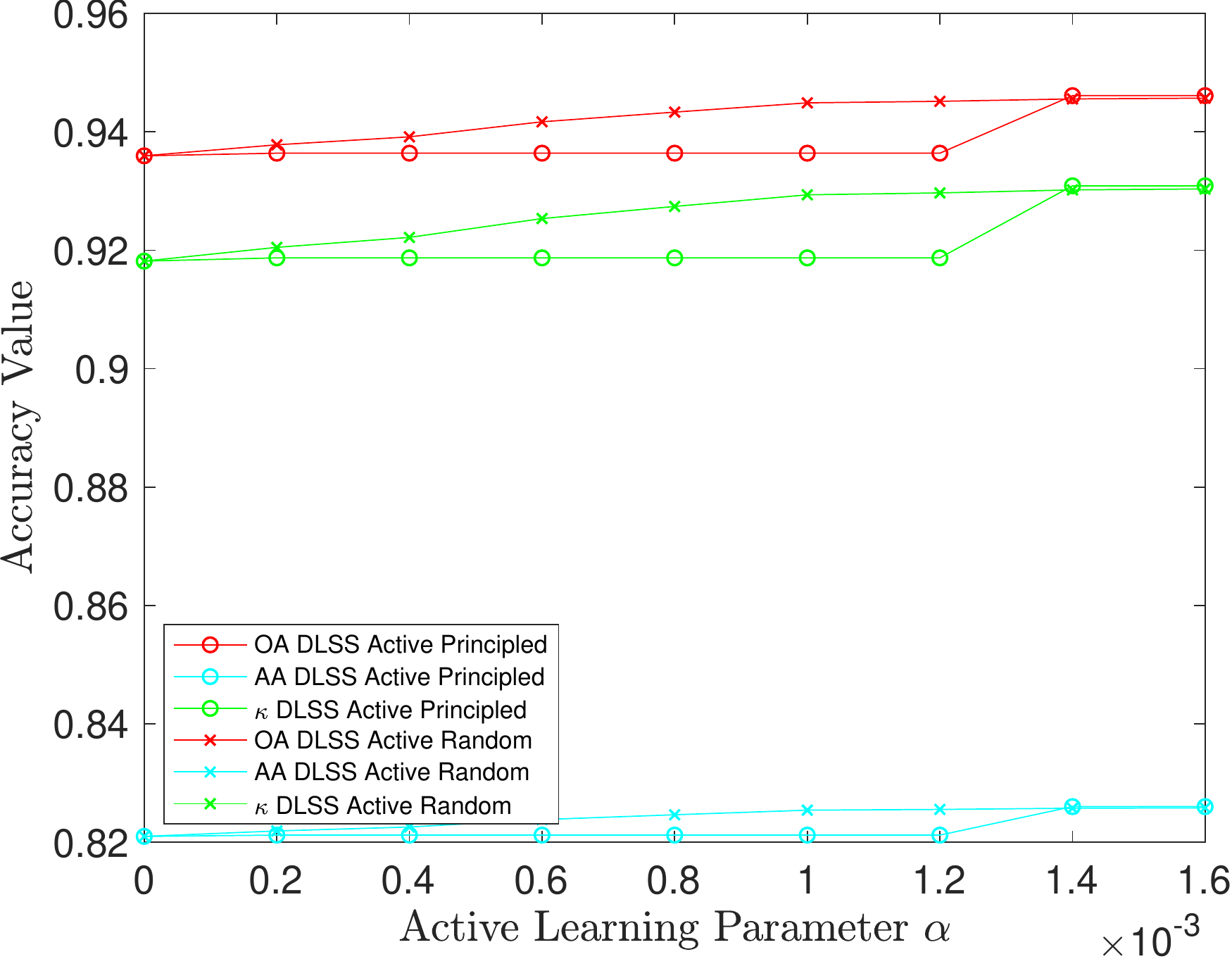}
\subcaption{Pavia}
\end{subfigure}
\begin{subfigure}{.24\textwidth}
\includegraphics[width=\textwidth]{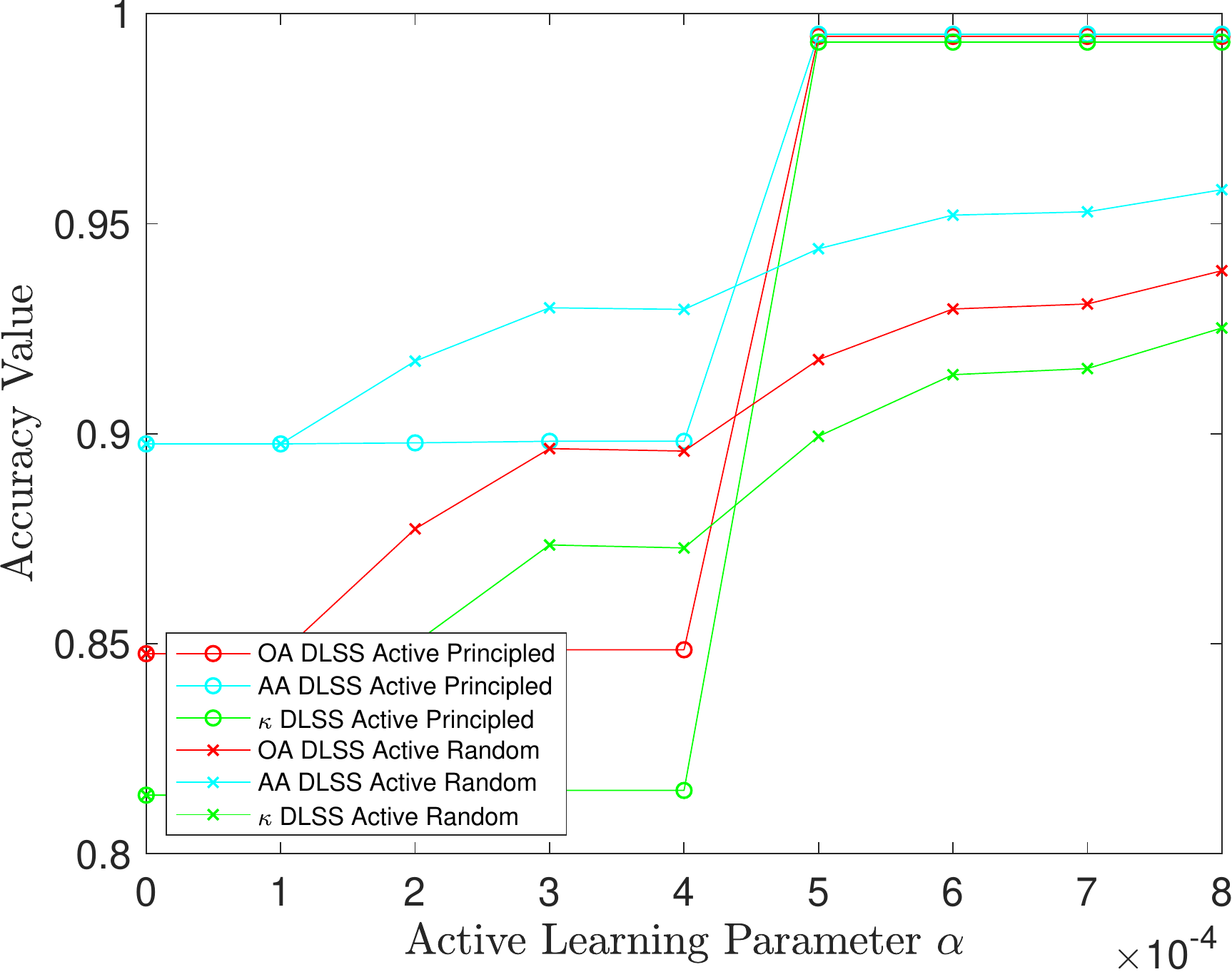}
\subcaption{SalinasA}
\end{subfigure}
\begin{subfigure}{.24\textwidth}
\includegraphics[width=\textwidth]{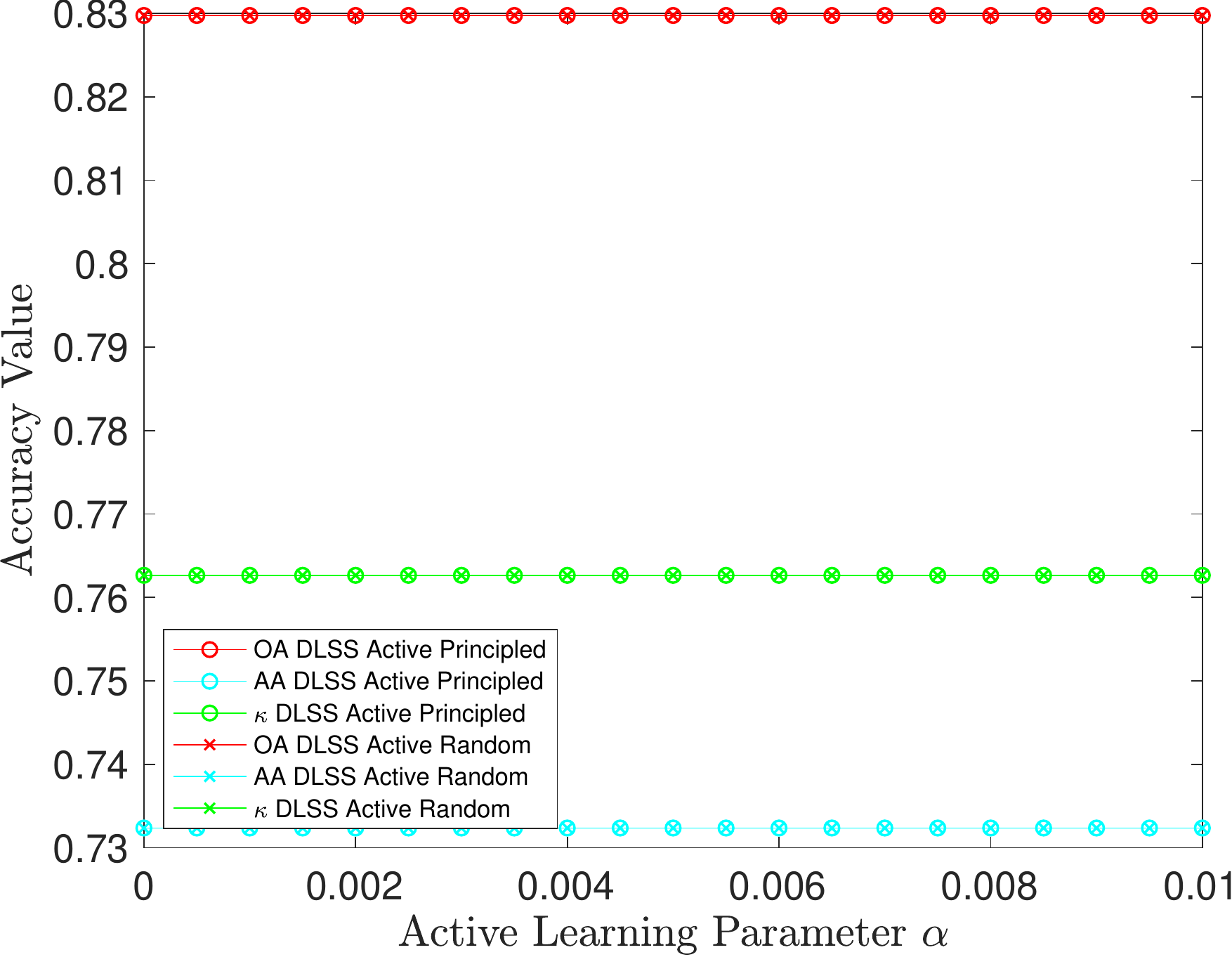}
\subcaption{KSC}
\end{subfigure}
\caption{\label{fig:ActiveLearningAnalysis} Active learning parameter analysis.  The $x$-axis denotes the parameter $\alpha$.  As $\alpha$ increases, more labeled pixels are introduced.  All measures of accuracy are monotonic increasing in $\alpha$, and a small increase can lead to a huge jump in accuracy, as seen in the Indian Pines and Salinas A datasets.  We see that randomly selecting points has a more incremental impact on improving accuracy than the principled approach, and may require a very large number of labels to achieve the performance achieved by active learning with a small number of labels.  Many iterations of randomly selected points were used and averaged to produce the plots.} 
\end{figure}

To evaluate our proposed active learning method, Algorithm \ref{alg:ActiveLearning}, the same $4$ labeled HSI datasets were clustered with increasing the percentage $\alpha$ of labeled points, chosen as in Algorithm \ref{alg:ActiveLearning}. Note that $\alpha=0$ corresponds the the unsupervised DLSS algorithm.  The empirical results for this active learning scheme appear in Figure \ref{fig:ActiveLearningAnalysis}.  We also consider selecting the labeled points uniformly at random; we hypothesize our principled approach will be superior to random sampling.  
The plots indicate that the proposed active learning can produce dramatic improvements in labeling with very few training labels.  Indeed, an improvement in overall accuracy from $85\%$ to $87\%$ for Indian Pines can be achieved with only 3 labels.  Even more dramatic is the Salinas A dataset, in which 3 labeled points improves the overall accuracy from $84\%$ to $99.5\%$.  The Pavia dataset enjoys some improved performance, though the random labels do about as well as the principled labels, and Kennedy Space Center dataset labelings are not affected by the small collection of labeled points.  In the case of Pavia, however, the overall accuracy was already very large, so active learning seems not needed for this data set.  Note that our principled scheme is generally superior to using randomly selected labeled points, which leads to a more gradual improvement in accuracy, compared to the huge gains that can be seen with the proposed principled method.  

It is interesting to compare our active learning results to a state-of-the-art \emph{supervised} method.  We consider the supervised HSI classification with edge preserving filtering method (EPF) \cite{Kang2014} algorithm, which combines a support vector machine with an analysis of spectral-spatial probability maps to label points.  Using a publicly available implementation \footnote{\url{http://xudongkang.weebly.com/}}, we ran this algorithm using $1\%$ and $5\%$ of points as training data, generated as a uniformly random sample over all labeled points.  10 experiments were ran on each of the four datasets considered, with results averaged.  Quantitative results are shown in Table \ref{tab:SupervisedComparison}.  The supervised results are generally superior to the results achieved by the unsupervised DL and DLSS method.  The proposed active learning, however, is able to achieve the same performance on the Salinas A dataset using two orders of magnitude fewer points.  This is because the proposed active learning method only uses training points for pixels that are considered especially important, whereas the EPF algorithm trains on a random subset of points.  Moreover, when only $1\%$ of training points are used, our active learning DLSS method with $.2\%$ of training points used outperforms the EPF method on the Indian Pines, Salinas A, and Kennedy Space Center datasets.  This indicates the promise of the proposed active learning method, as it is able to outperform a state-of-the-art supervised method in the regime in which a low proportion of training points is available.  

\begin{figure*}
\centering
\begin{adjustbox}{max width=\textwidth}
\begin{tabular}{| c | c | c | c | c | c | c | c | c | c  | c | c | c ||}\hline
Method & OA I.P. & AA I.P. & $\kappa$ I.P. & OA P. & AA P. & $\kappa$ P. & OA S.A. & AA S.A. & $\kappa$ S.A. & OA  K.S.C. & AA K.S.C. & $\kappa$ K.S.C.\\ \hline
DLSS (unsupervised) & 0.85  & 0.82  & 0.75  & 0.95 & 0.83 & 0.93 & 0.85 & 0.90 & 0.81 & 0.83 & 0.73 & 0.76 \\ \hline
Active learning, .2\% training  & 0.87 & 0.84 & 0.79 &  0.95  &  0.83 & 0.93 & 1.00 &  1.00 & 1.00 & 0.83 & 0.73 & 0.76 \\ \hline
EPF, 1\% training & 0.49 & 0.33 & 0.16 & 0.99 & 0.99 & 0.99 & 0.97 & 0.97 & 0.96 & 0.51 & 0.37 & 0.31 \\ \hline
EPF, 5\% training & 0.82  & 0.86  & 0.72 & 0.99& 0.99 & 0.99 &  1.00 &  1.00 & 1.00  & 0.98 & 0.98 & 0.98 \\ \hline
\end{tabular}
\end{adjustbox}
\caption{\label{tab:SupervisedComparison}We compare the state-of-the-art supervised classification algorithm, EPF, with the unsupervised DLSS algorithm and DLSS active learning variation.  We see that the active learning method with only $.2\%$ of pixels used for training outperforms EPF with $1\%$ training labels on the Indian Pines, Salinas A, and Kennedy Space Center datasets.  Moreover, the active learning method with $.2\%$ labels performs comparably to or better than EPF with $5\%$ training labels on the Indian Pines and Salinas A datasets.}

\end{figure*}

\subsection{Parameter Analysis}
\label{subsec:ParameterAnalysis}

We now discuss the parameters used in all methods, starting with those used for all comparison methods, and then discussing the two key parameters for the proposed method: diffusion time $t$ and radius size $r_{s}$ for the computation of the spatial consensus label in Algorithm \ref{alg:labels}.  For experimental parameters except these, a range of parameters were considered, and those with best empirical results were used. 

All instances of the $K$-means algorithm are run with $100$ iterations, with $10$ random initializations each time, and number of clusters $K$ equal to the known number of classes in the ground truth.  Each of the linear dimension reduction techniques, PCA, ICA, and random projection, embeds the data into $\mathbb{R}^{K}$, where $K$ is the number of clusters.  DBSCAN is highly dependent on several parameters, and a grid search was used on each dataset to select optimal parameters.  Note that this means DBSCAN was optimized specifically for each dataset, while other methods used a fixed set of parameters across all experiments.  Spectral clustering is run by computing a weight matrix as in $(\ref{eqn:W})$, with $k=100$ and $\sigma=1$.  The top $K$ eigenvectors are then normalized to have Euclidean norm $1$, then used as features with $K$-means.

Among the state-of-the-art methods, HNMF uses the recommended settings listed in the available online toolbox\footnote{\url{https://sites.google.com/site/nicolasgillis/code}}.  For FSFDPC, the empirical density estimate is computed as described in Section \ref{subsec:AlgorithmDescription}, with a Gaussian kernel and 20 nearest neighbors.  For SMCE, the sparsity parameter was set to be 10, as suggested in the online toolbox\footnote{\url{http://www.vision.jhu.edu/code/}}.  The FMS algorithm depends on several key parameters; grid search was implemented, and empirically optimal parameters with respect to a given dataset were used.  Note that this means FMS was, like DBSCAN, optimized specifically for each dataset, while other methods used a fixed set of parameters across all experiments.  

For the proposed algorithm, the same parameters for the density estimator as described above are used, in order to make a fair comparison with FSFDPC.  Moreover, in the construction of the graph used to compute diffusion distances, we use the same construction as in spectral clustering and SMCE, again to make fair comparisons.  The remaining parameters, diffusion time and spatial radius, were set to $30$ and $3$, respectively, for all experiments.  We justify these choices and analyze their robustness in the following subsections.  

\subsubsection{Diffusion Time $t$}\label{subsec:TimeParameter}
The most important parameter when using diffusion distances $d_t(x,y)$ is the time parameter $t\ge0$, see eqn. $(\ref{eqn:DD_eigen})$.  The larger $t$ is, the smaller the contribution of the smaller eigenvalues in the spectral computation of $d_{t}$.  Allowing $t$ to vary, connections in the dataset are explored by allowing the diffusion process to evolve.  For small values of $t$, all points appear far apart because the diffusion process has not reached far, while for large values of $t$, all points appear close together because the diffusion process has run for so long that it has dissipated over the entire state space.  In general, the interesting choice of $t$ is moderate, which allows for the data geometry to be discovered, but not washed out in long-term.  

\begin{figure}
\centering
\begin{subfigure}{ .24\textwidth}
\includegraphics[width=\textwidth]{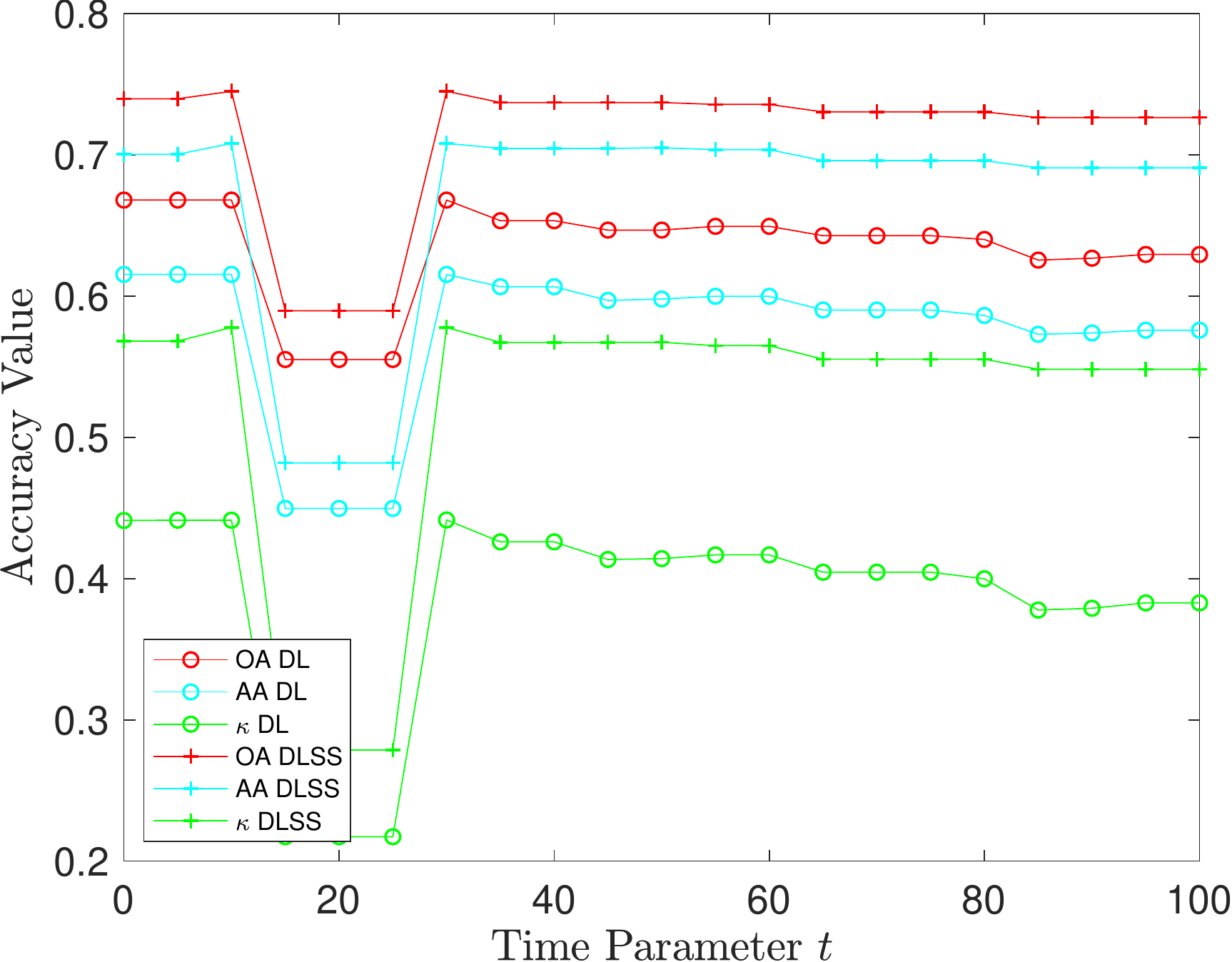}
\subcaption{Indian Pines}
\end{subfigure}
\begin{subfigure}{ .24\textwidth}
\includegraphics[width=\textwidth]{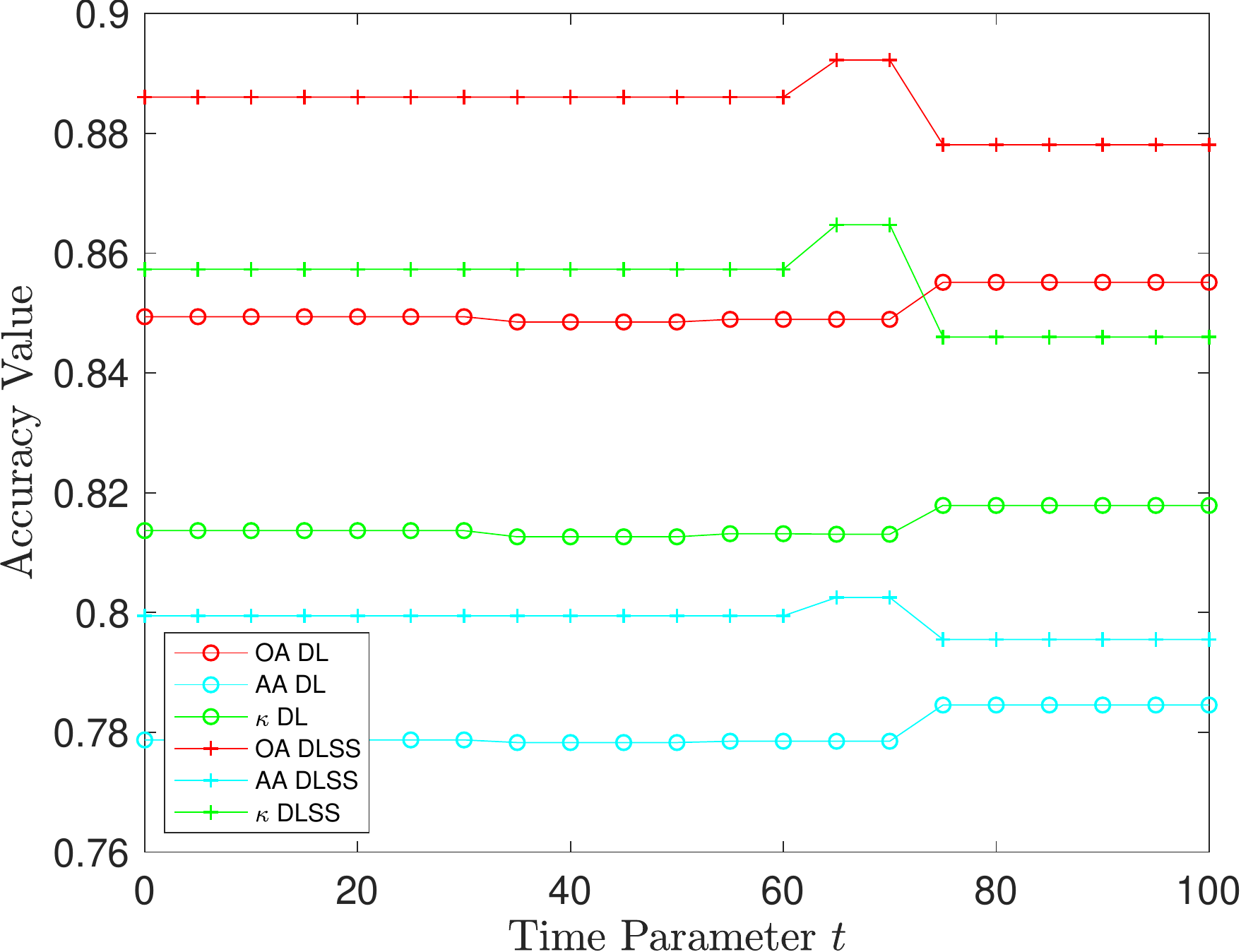}
\subcaption{Pavia}
\end{subfigure}
\begin{subfigure}{.24\textwidth}
\includegraphics[width=\textwidth]{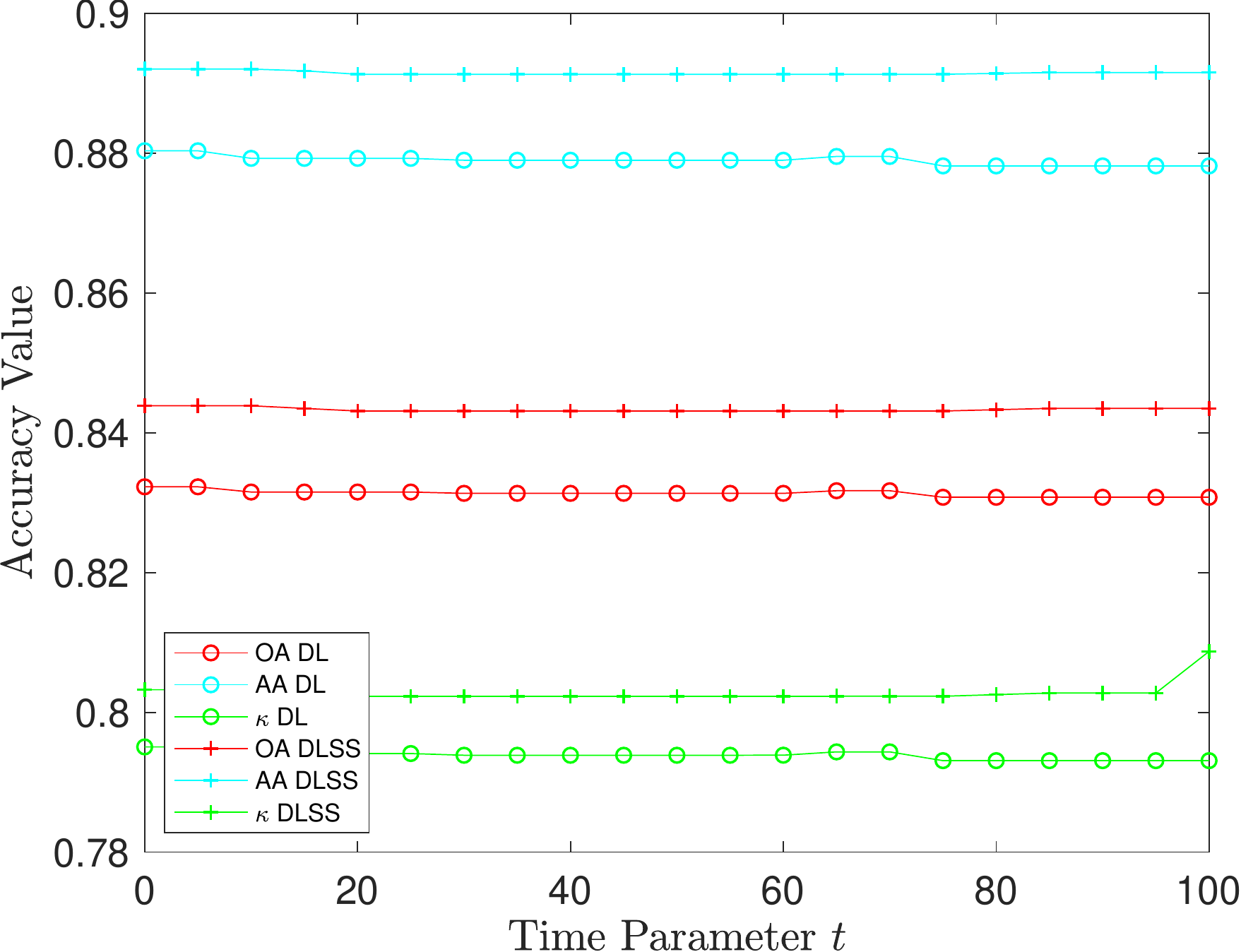}
\subcaption{SalinasA}
\end{subfigure}
\begin{subfigure}{.24\textwidth}
\includegraphics[width=\textwidth]{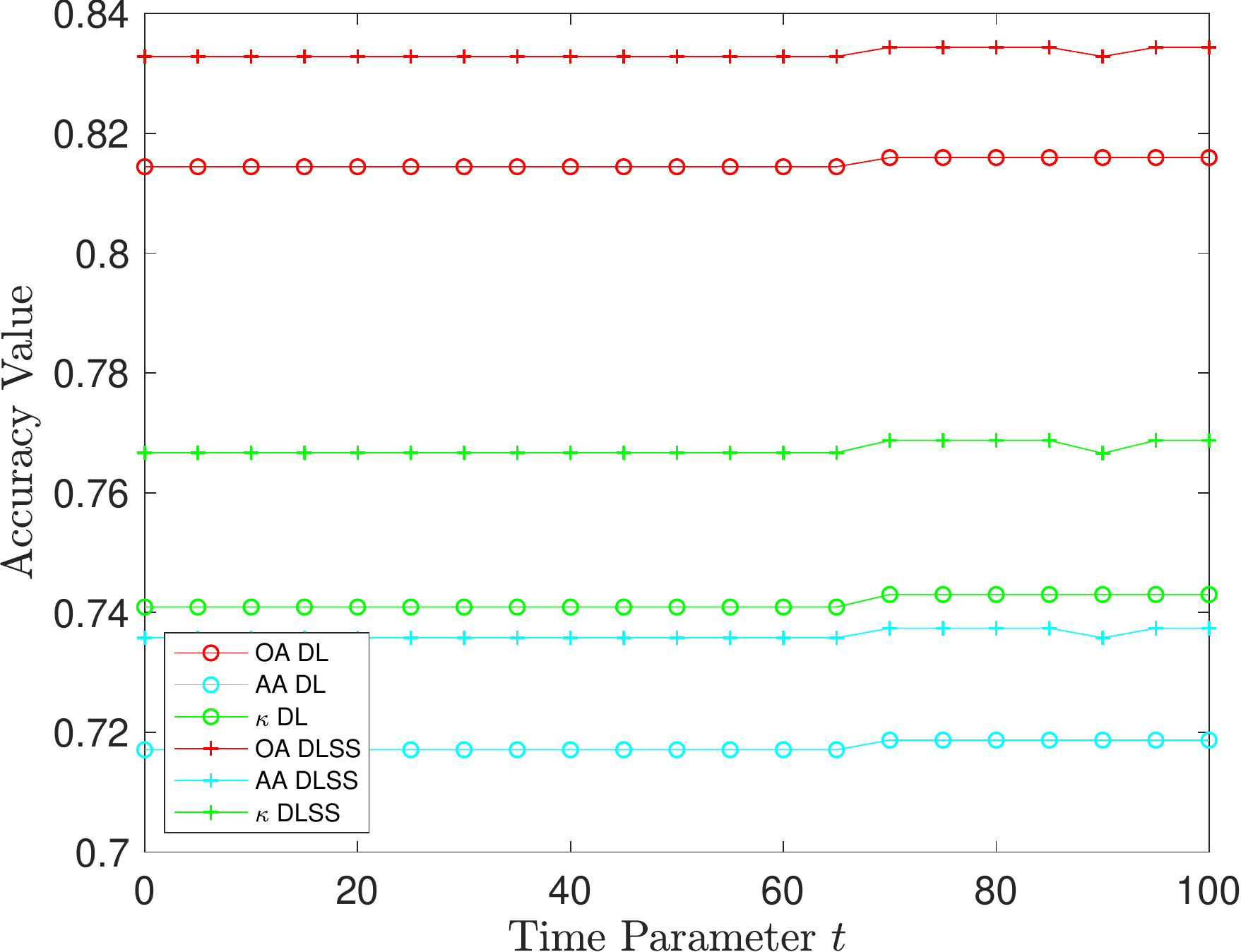}
\subcaption{KSC}
\end{subfigure}
\caption{\label{fig:TimeParameterAnalysis} Time parameter analysis for the four real datasets.  In general, the time parameter has little impact on performance of the proposed algorithm.  As suggested by these plots, $t=30$ is used for all ex periments.} 
\end{figure}

In Figure \ref{fig:TimeParameterAnalysis}, all the accuracy measures for $t$ in $[0,100]$ are displayed. 
The behavior is robust with respect to time.  For Indian Pines, performance is largely constant, except for a dip from time $t=15$ to $t=25$.  For the Pavia, Salinas A, and Kennedy Space Center examples, the performance is invariant with respect to the diffusion time.  We conclude that a large range of $25\le t\le 65$ or $t\ge 75$  would have led to the same empirical results as our choice $t=30$.
  
\subsubsection{Spatial Diffusion Radius}\label{subsec:SpaceParameter}
\begin{figure}
\centering
\begin{subfigure}{ .24\textwidth}
\includegraphics[width=\textwidth]{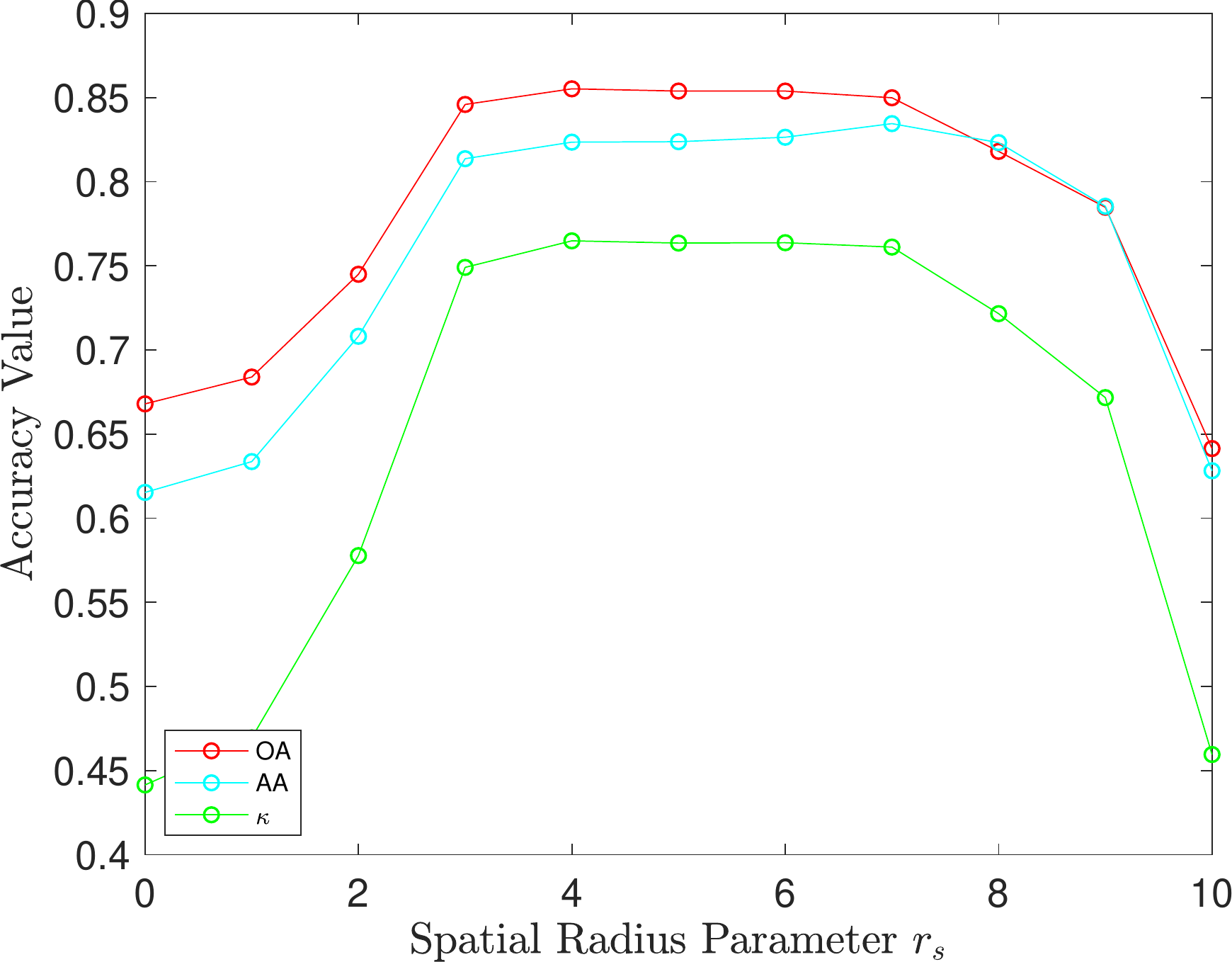}
\subcaption{Indian Pines}
\end{subfigure}
\begin{subfigure}{ .24\textwidth}
\includegraphics[width=\textwidth]{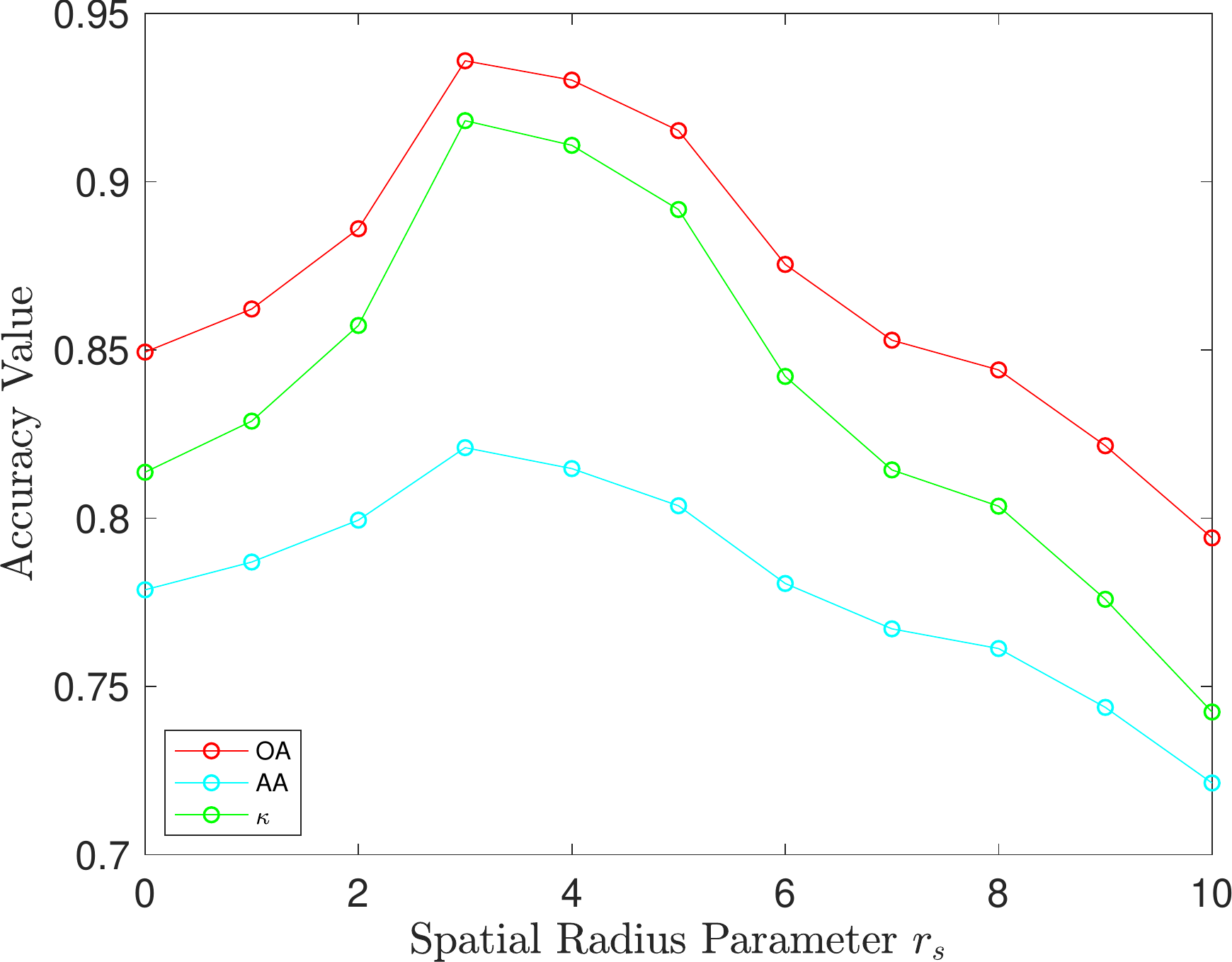}
\subcaption{Pavia}
\end{subfigure}
\begin{subfigure}{.24\textwidth}
\includegraphics[width=\textwidth]{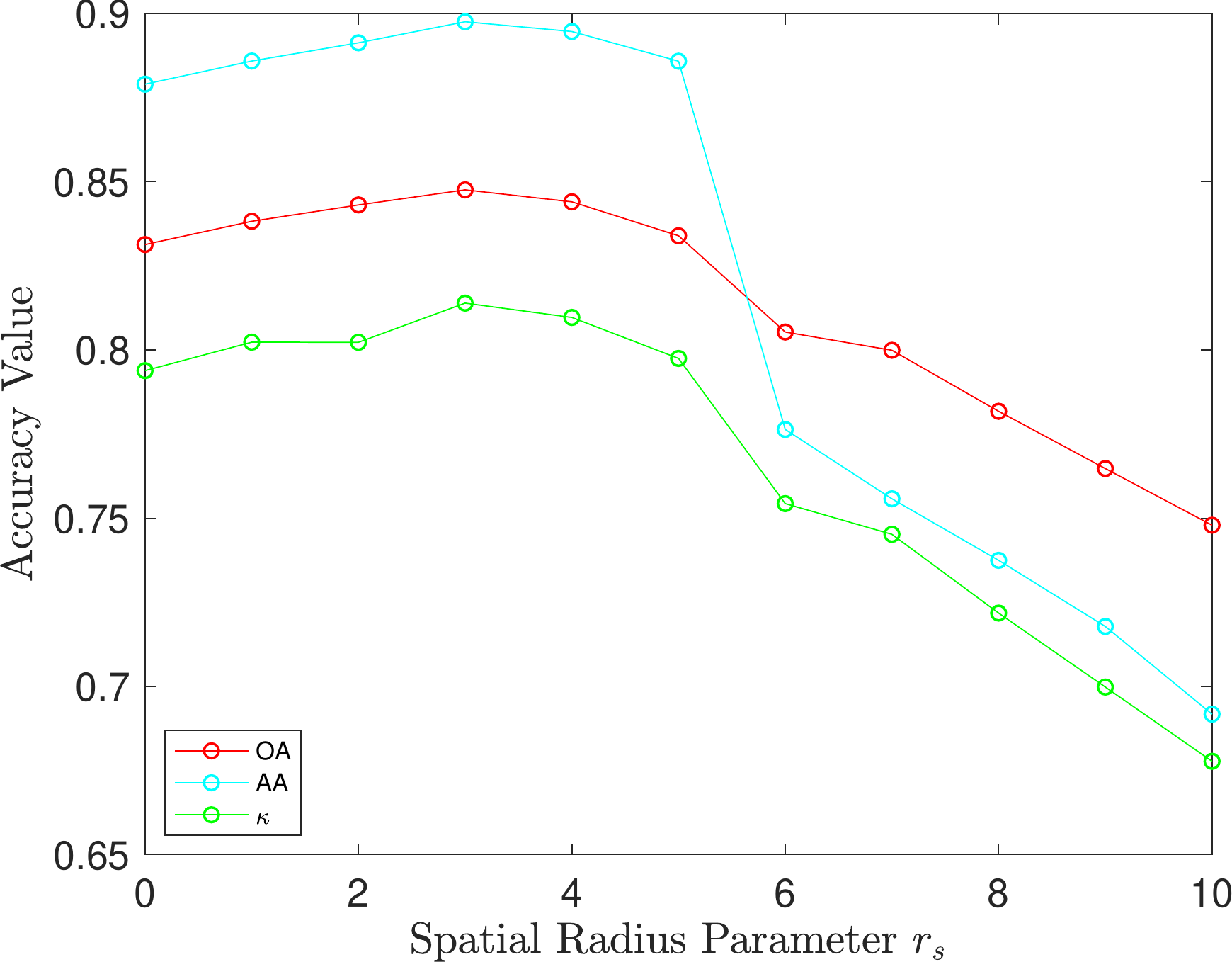}
\subcaption{Salinas A}
\end{subfigure}
\begin{subfigure}{.24\textwidth}
\includegraphics[width=\textwidth]{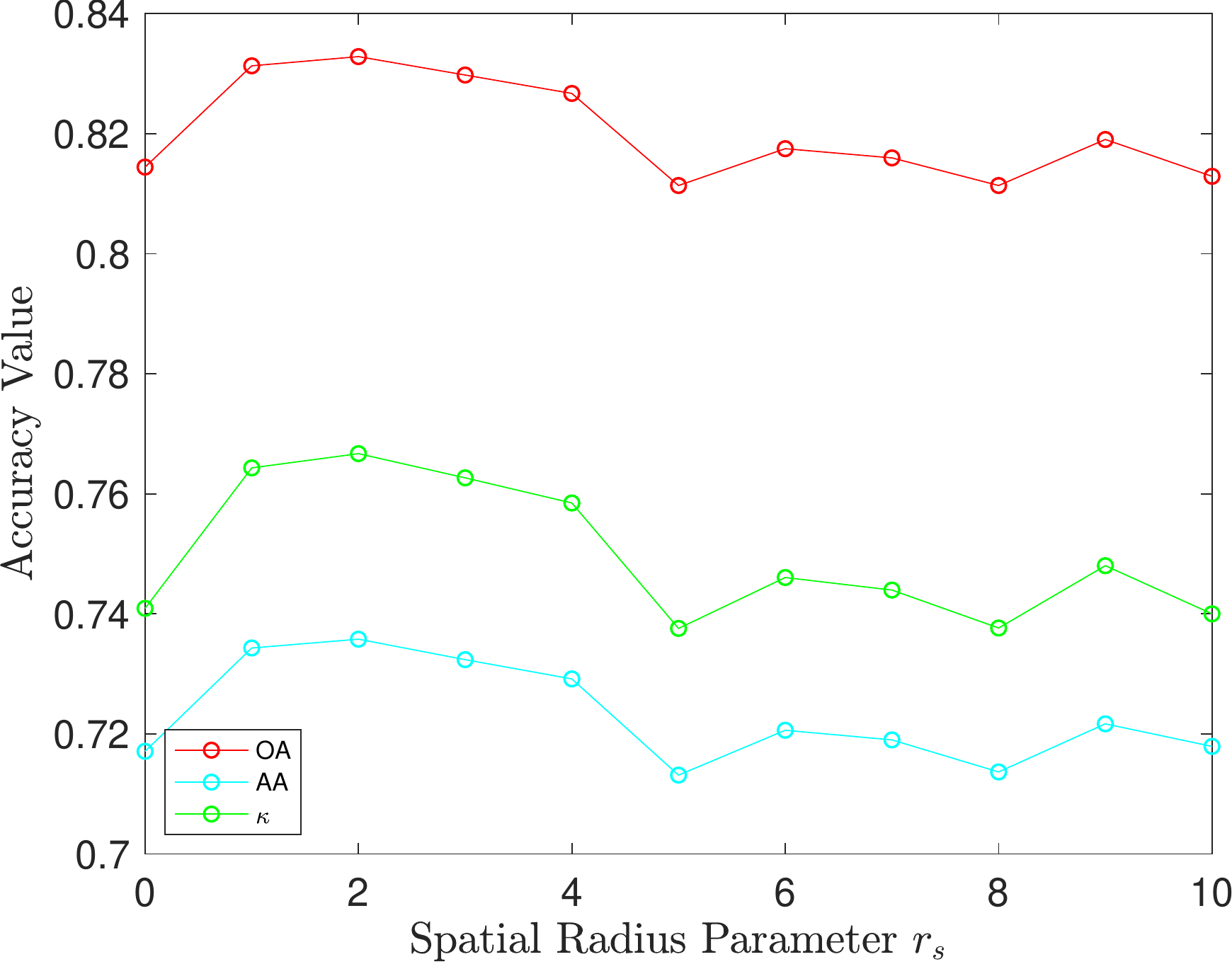}
\subcaption{KSC}
\end{subfigure}
\caption{\label{fig:SpaceParameterAnalysis} Space parameter analysis for DLSS.  For each curve, increasing the radius of the neighborhood in which the spatial consensus label is computed improves quantitative performance up until a certain point, after which performance decays.  The point at which the decay sets in differs for each example.  These plots suggest a spectral-spatial tradeoff: spectral and spatial information must be balanced to achieve empirically optimal clustering.} 
\end{figure}

The spatial consensus radius $r_s$ can also impact the performance of the proposed DLSS algorithm.  Recall that this is the distance in the spatial domain used to compute the spatial consensus label (see Section \ref{subsec:AlgorithmDescription} and definition \eqref{eqn:SpatialConsensus}).  If $r_{s}$ is too small, insufficient spatial information is incorporated;  if $r_{s}$ is too large, the spectral information becomes drowned out.  
All measures of accuracy for each dataset for $r_{s}$ in $[0,10]$ appear in Figure \ref{fig:SpaceParameterAnalysis}.
We see a trade-off between spectral and spatial information, suggesting that $r_{s}$ should take a moderate value sufficiently greater than 0 but less than 10.  This trade-off can be interpreted in the following way: empirically optimal results are achieved when both spectral and spatial information contribute harmoniously, and results deteriorate when one or the other dominates.  We choose $r_{s}=3$ for all experiments, though other choices would give comparable (or sometimes better) quantitative results for the datasets considered.  

We note that the role of the spatial radius is analogous to the role of a regularization parameter for a general statistical learning problem.  Taken too small, the problem is insufficiently regularized, leading to poor results.  Taken too large, the regularization term dominates the fidelity term, leading also to poor results.  In particular, the geometric regularity of the clusters in the spatial domain determine how large $r_{s}$ may be taken while still preserving the spectral information.  If the clusters are convex and not too elongated, then taking $r_{s}$ large is reasonable.  On the other hand, if the classes are very irregular spatially, for example highly non-convex or elongated, choosing $r_{s}$ too large will wash out the spectral information which is generally more discriminative than the spatial information, resulting in inaccurate clustering.

\subsection{Large Scale Experiments}\label{sec:LargeScaleExperiments}

The results of Section \ref{subsec:HyperspectralClustering} analyzed subsets of larger images, in order to reduce the number of classes to allow for effective unsupervised learning \cite{Zhu2017}.  In order to evaluate the robustness of these results, we performed experiments in which the full HSI scenes were subdivided into small patches with fewer classes, then each patch---with a smaller number of classes than the total scene---were clustered.  The results on individual patches may be used as the basis for a statistical evaluation of the performance of each clustering method.  For the Indian Pines, Pavia, and Kennedy Space Center datasets, experiments for the entire dataset, suitably partitioned into smaller patches, were performed, with DLSS again performing best among all studied methods.  Note that Salinas A had only 6 classes, and was considered in its entirety.  The Indian Pines data set was partitioned into 24 rectangular patches of equal size; Pavia into 50 rectangular patches of equal size, and Kennedy Space Center into 25 patches of equal size.  On each piece that contained some non-trivial ground truth, all clustering algorithms were ran.  A series of statistical tests on the differences in performance were then executed as follows.  For a pair of methods---denoted method $i$ and $j$--- let $OA_{k}^{i}, OA_{k}^{j}$ be the overall accuracy of methods $i$ and $j$ on patch $k$, and let $\Delta_{k}^{i,j}=OA_{k}^{i}-OA_{k}^{j}$.  The sample mean difference in error between methods $i$ and $j$ across the different patches is $\overline{\Delta^{i,j}}=\sum_{k=1}^{\Npatches}\Delta_{k}^{i,j}/\Npatches$, where $\Npatches$ is the total number of patches with ground truth.  It is of interest to investigate whether $\overline{\Delta_{i,j}}$ can be inferred to be different from 0.  In order to perform a statistical test, the sample standard deviation of difference between methods $i,j$ is computed as $\sigma^{i,j}=\sqrt{\sum_{k}(\Delta_{k}^{i,j}-\overline{\Delta^{i,j}})^2/(\Npatches-1)}$.  Then, the null hypothesis that $\overline{\Delta_{i,j}}=0$ may be tested against the alternative hypothesis that $\overline{\Delta_{i,j}}\neq0$ by performing a two-sided $t$-test \cite{Wasserman2013_All} with $\Npatches-1=72$ degrees of freedom.  The normalized $t$-scores for the $j$ corresponding to the DLSS method and $i$ running through all other methods are reported in Table \ref{tab:StatisticalAnalysis}.  The test confirms that for all methods $i$, the hypothesis that DLSS $(j=13)$ does not significantly differ from method $i$ ($\overline{\Delta_{i,13}}=0$) is rejected in favor of the alternative hypothesis that DLSS significantly differs from method $i$ $(\overline{\Delta_{i,13}}\neq 0$) at the $95\%$ level.  This provides evidence that DLSS performs competetively with benchmark and state-of-the-art HSI clustering algorithms across HSI with a variety of land cover types and complexity.  Note that the values are $\Delta_{i,j}^{k}$ are not independent for different $k$, due to correlations across images.  However, the $t$-test still provides a powerful method for inferring statistical significance in this case, despite this theoretical assumption not being satisfied.  

\begin{figure}
\centering
\begin{subfigure}{.09\textwidth}
\includegraphics[width=\textwidth]{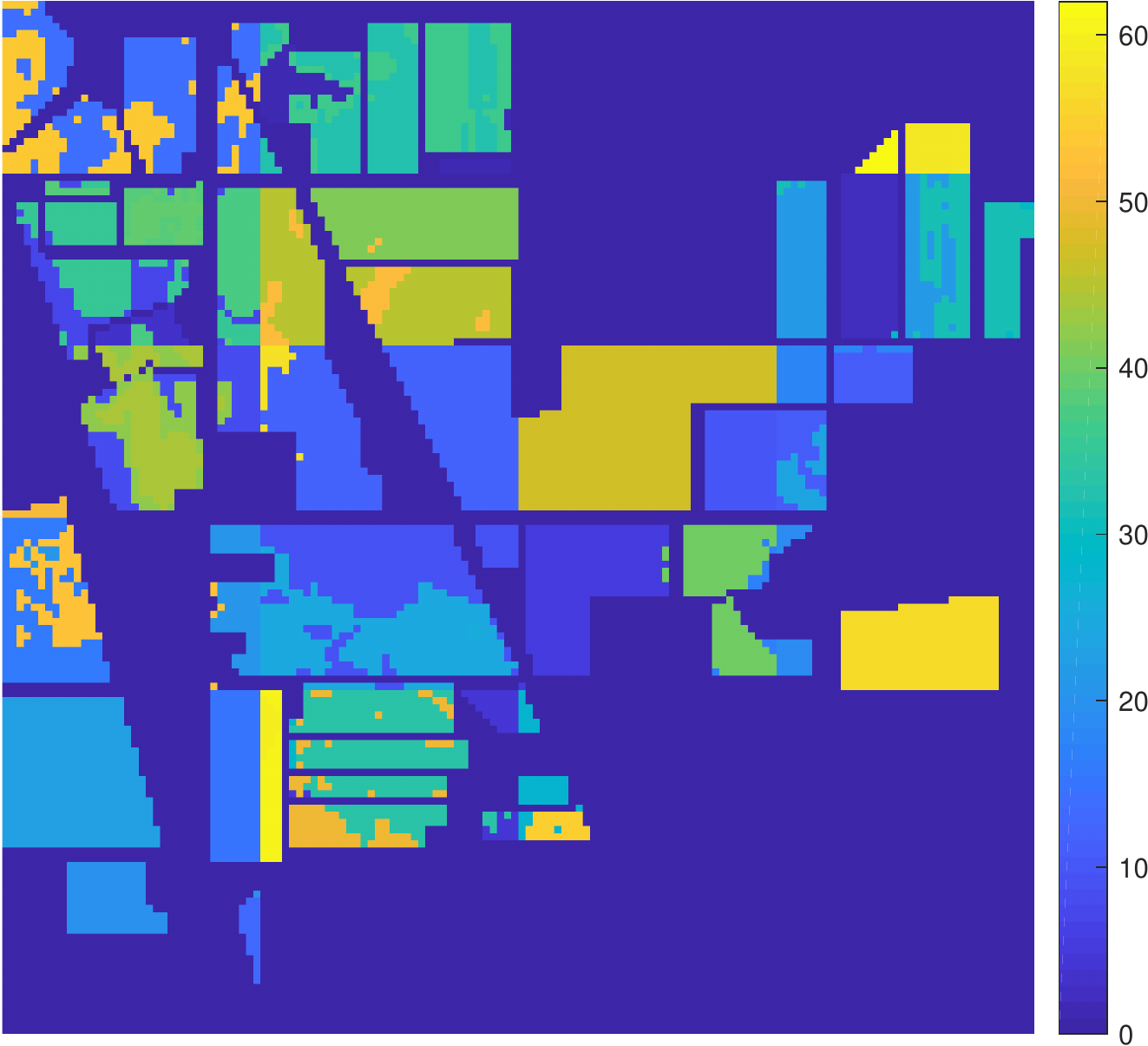}
\caption{$K$-means}
\end{subfigure}
\begin{subfigure}{.09\textwidth}
\includegraphics[width=\textwidth]{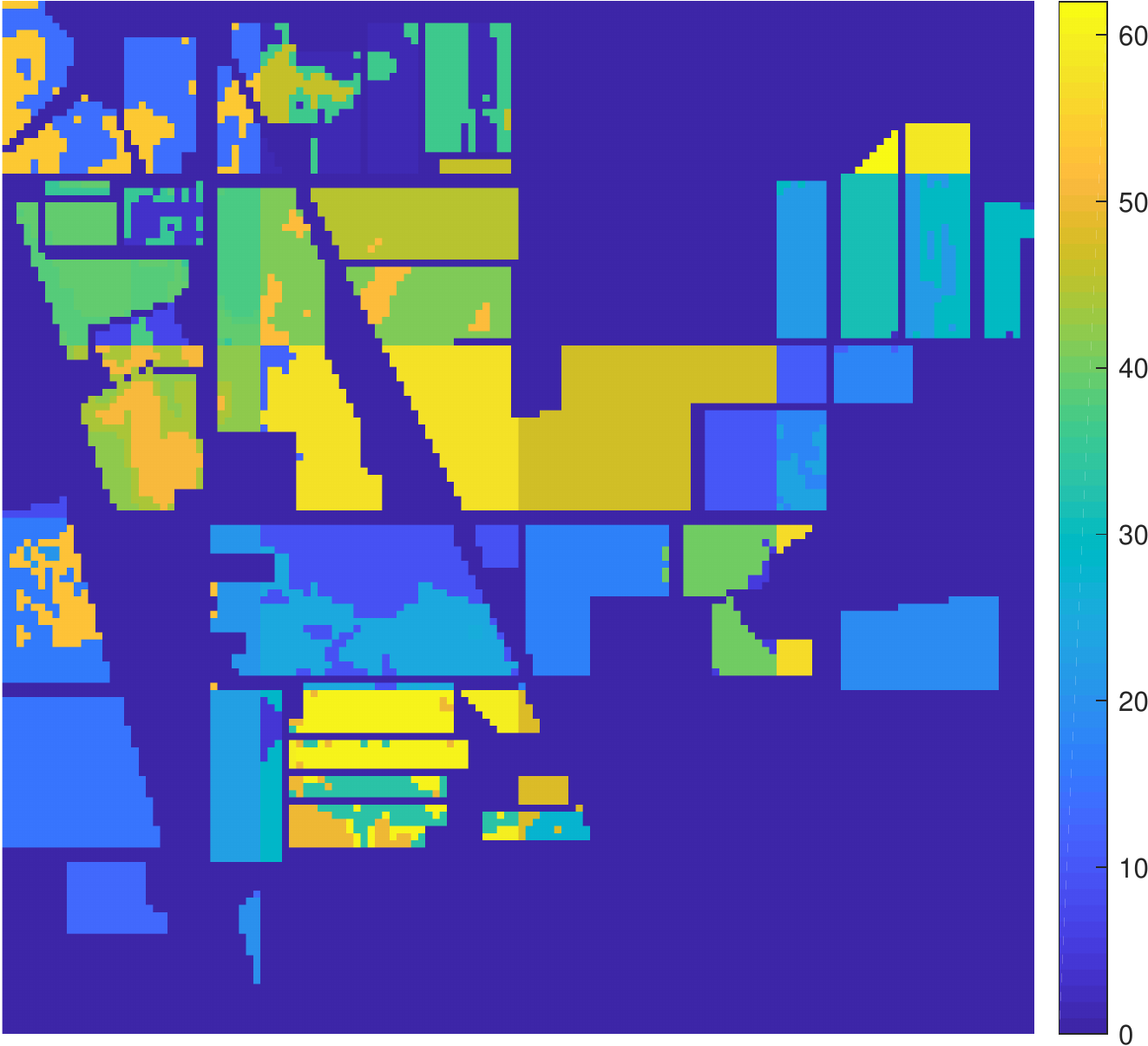}
\caption{PCA+$K$M}
\end{subfigure}
\begin{subfigure}{.09\textwidth}
\includegraphics[width=\textwidth]{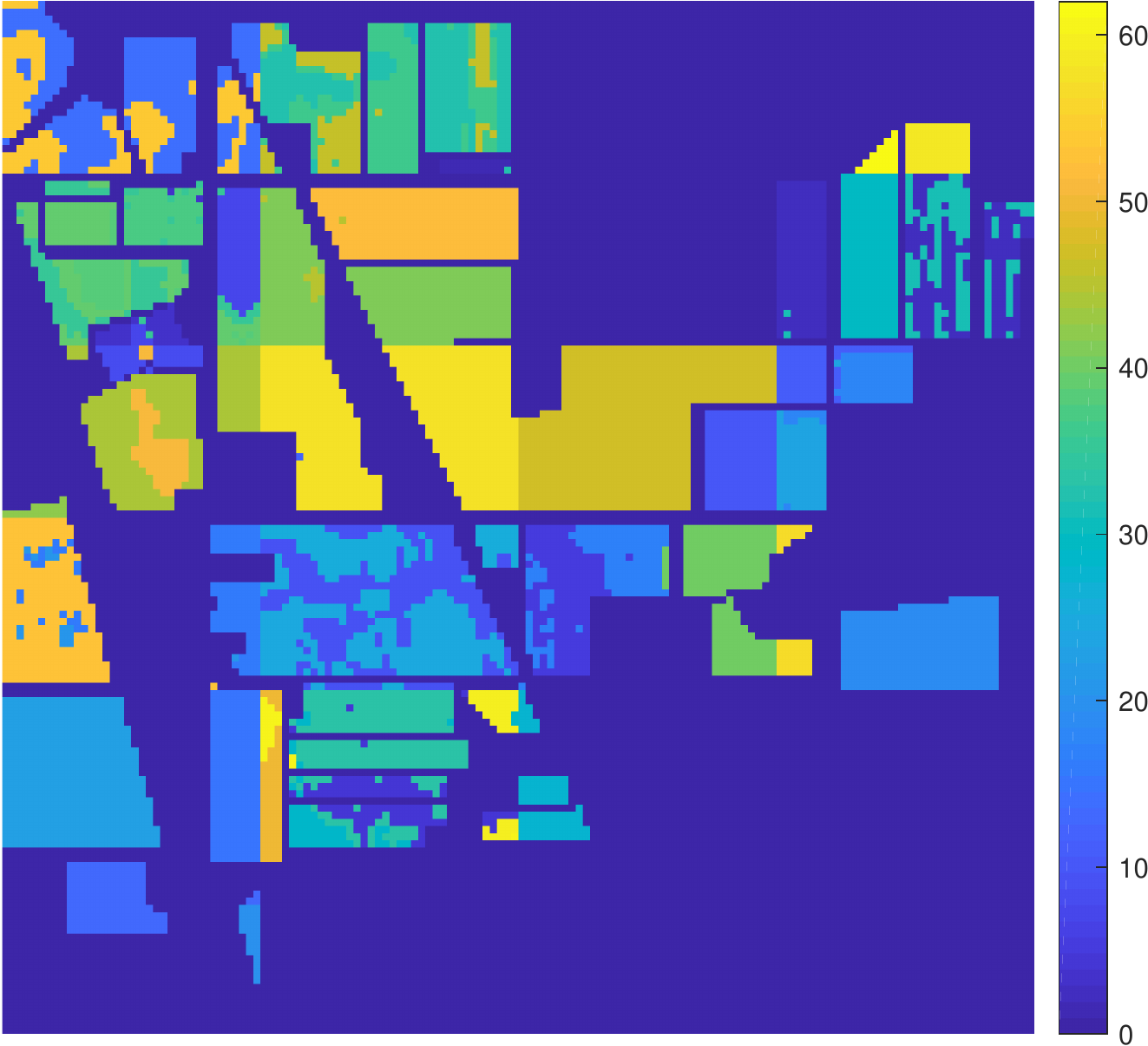}
\caption{ICA+$K$M}
\end{subfigure}
\begin{subfigure}{.09\textwidth}
\includegraphics[width=\textwidth]{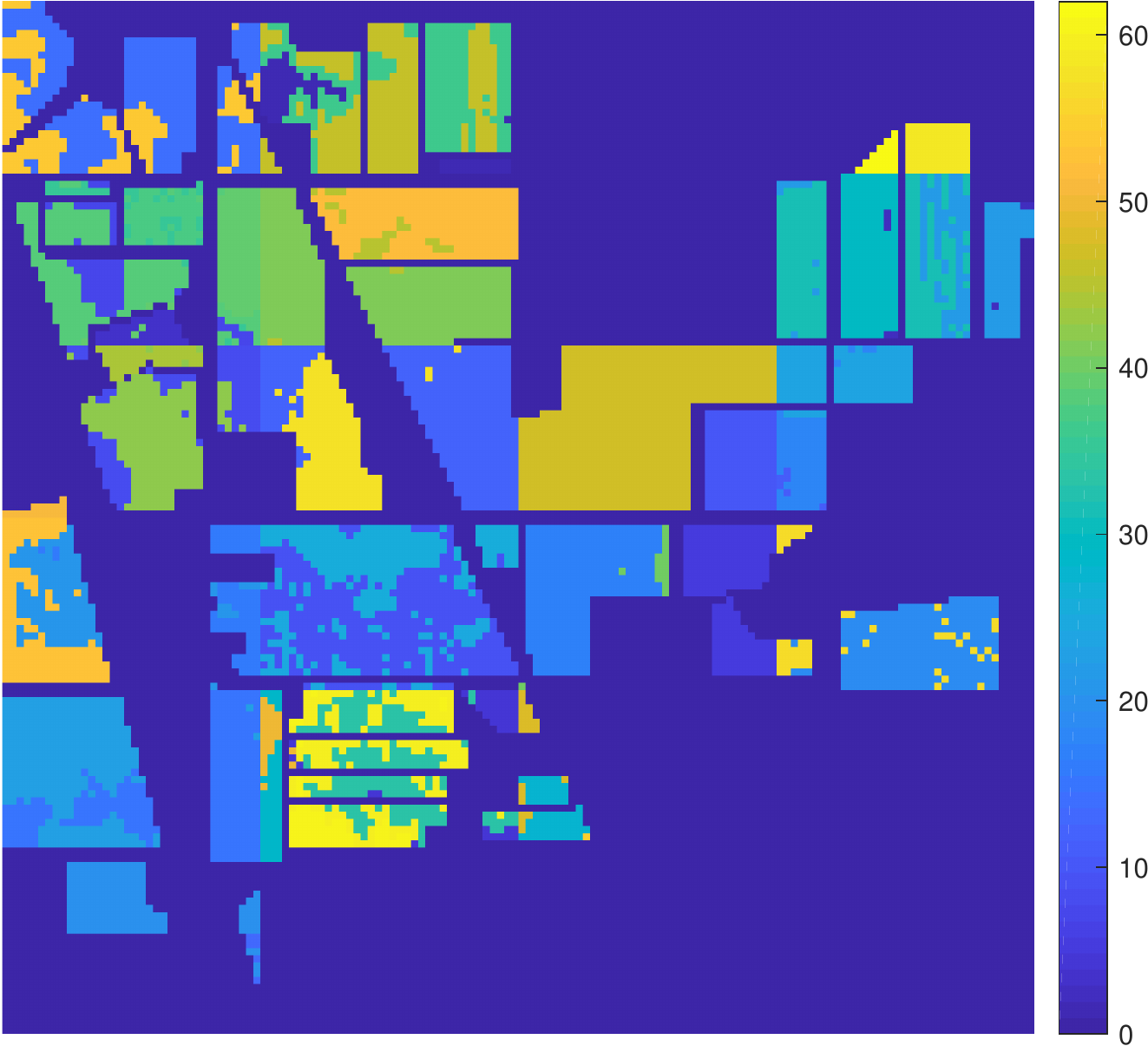}
\caption{RP+$K$M}
\end{subfigure}
\begin{subfigure}{ .09\textwidth}
\includegraphics[width=\textwidth]{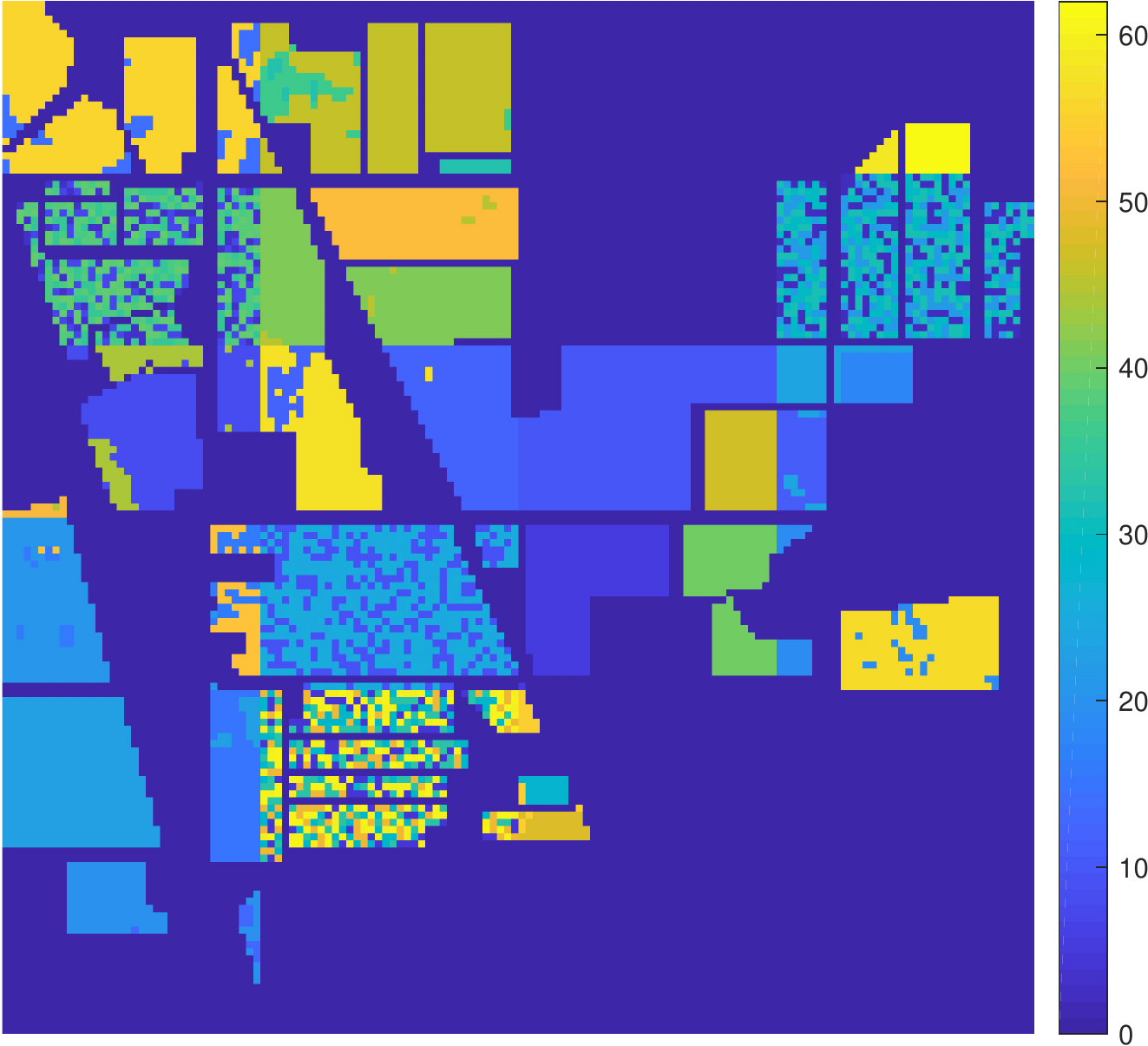}
\caption{DBSCAN}
\end{subfigure}
\begin{subfigure}{.09\textwidth}
\includegraphics[width=\textwidth]{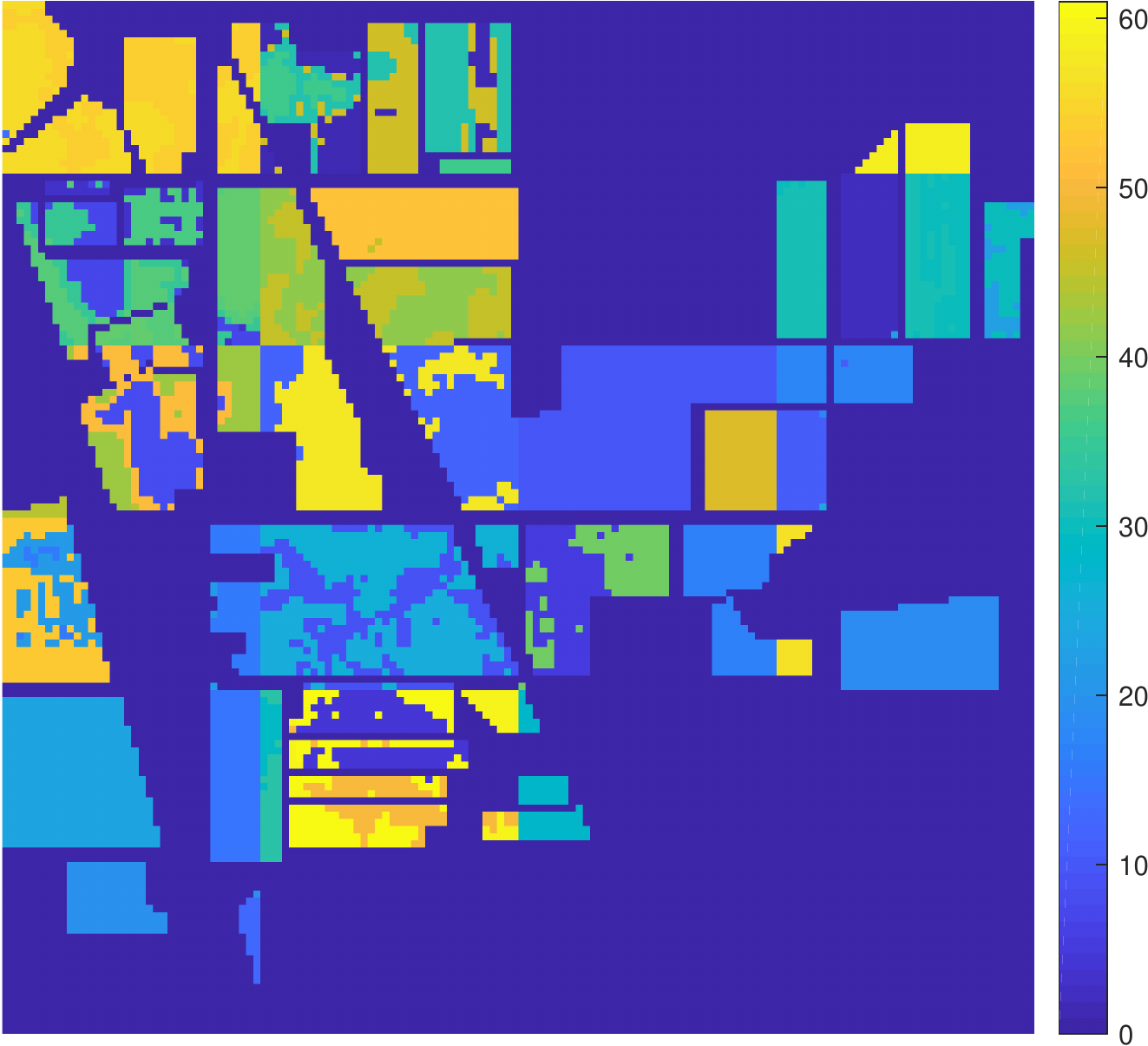}
\caption{SC}
\end{subfigure}
\begin{subfigure}{.09\textwidth}
\includegraphics[width=\textwidth]{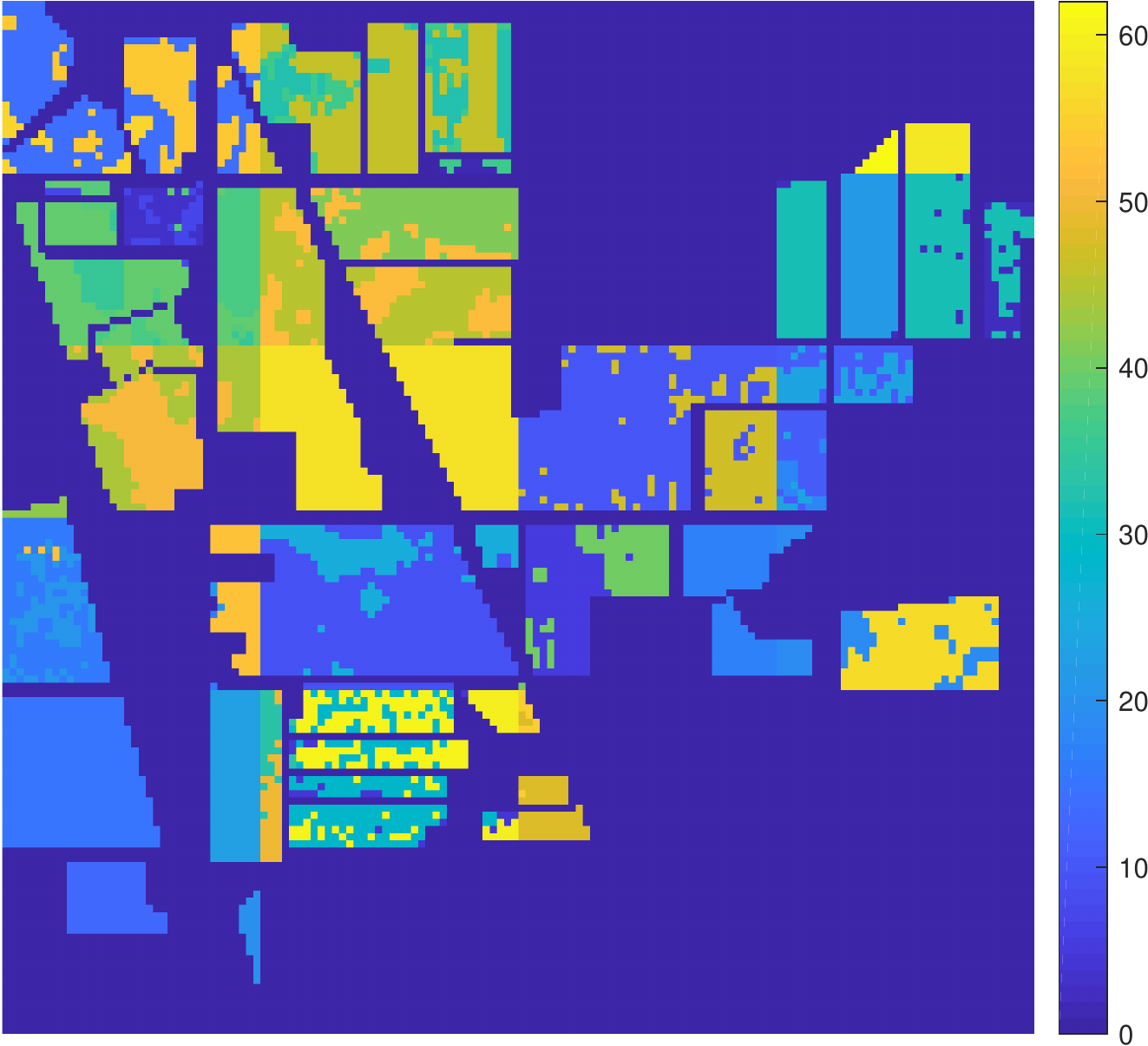}
\caption{GMM}
\end{subfigure}
\begin{subfigure}{.09\textwidth}
\includegraphics[width=\textwidth]{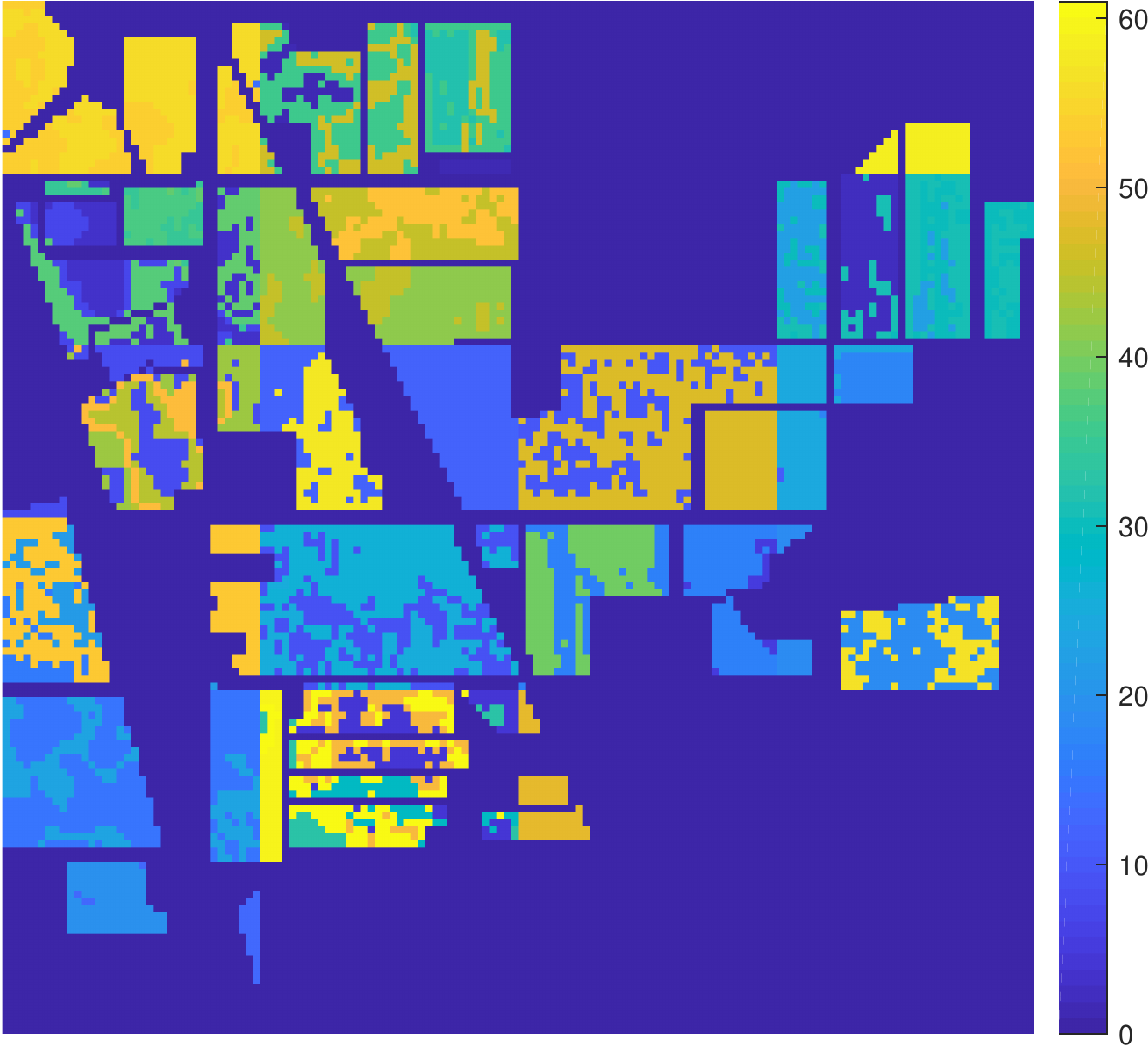}
\caption{SMCE}
\end{subfigure}
\begin{subfigure}{.09\textwidth}
\includegraphics[width=\textwidth]{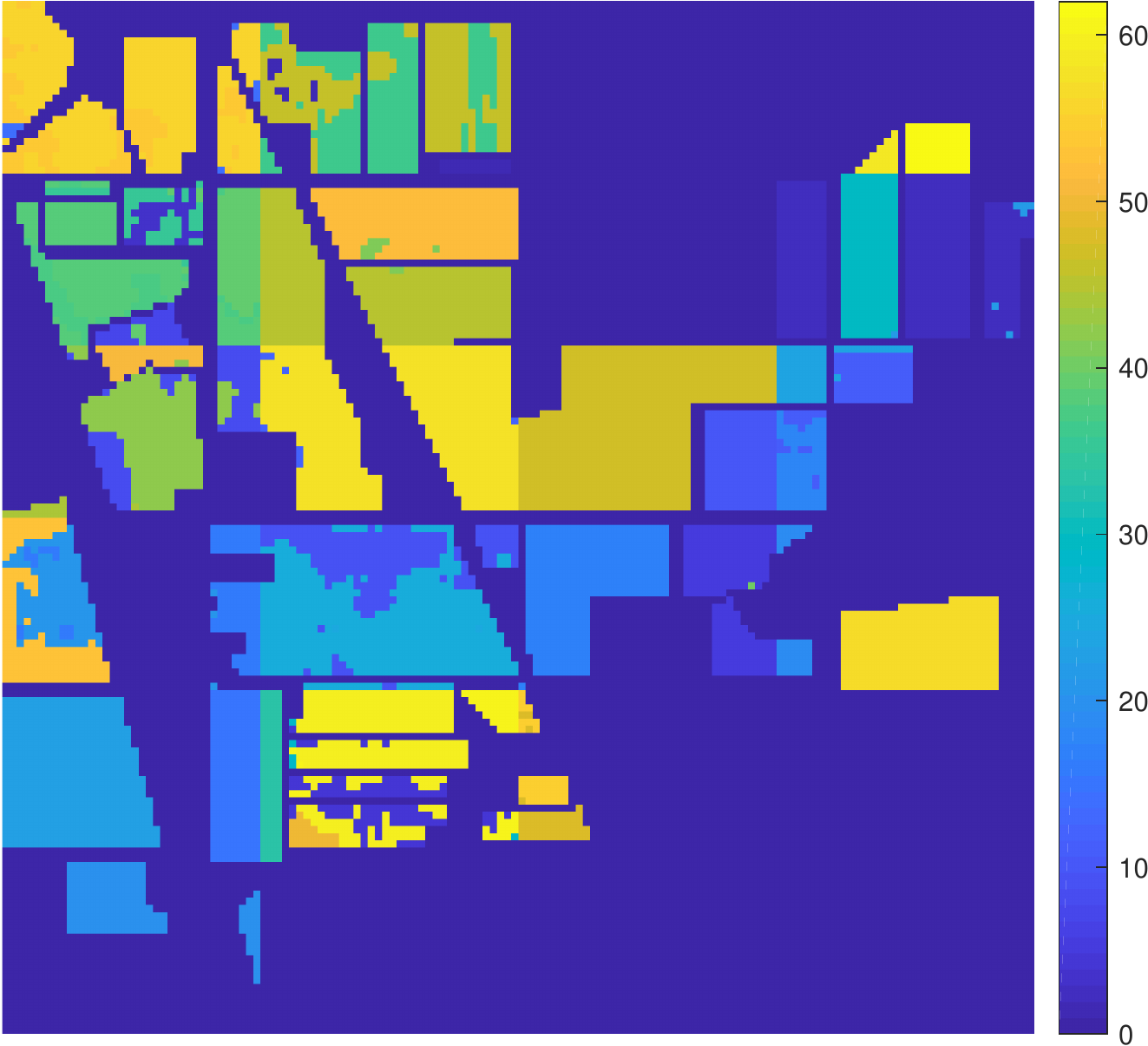}
\caption{HNMF}
\end{subfigure}
\begin{subfigure}{ .09\textwidth}
\includegraphics[width=\textwidth]{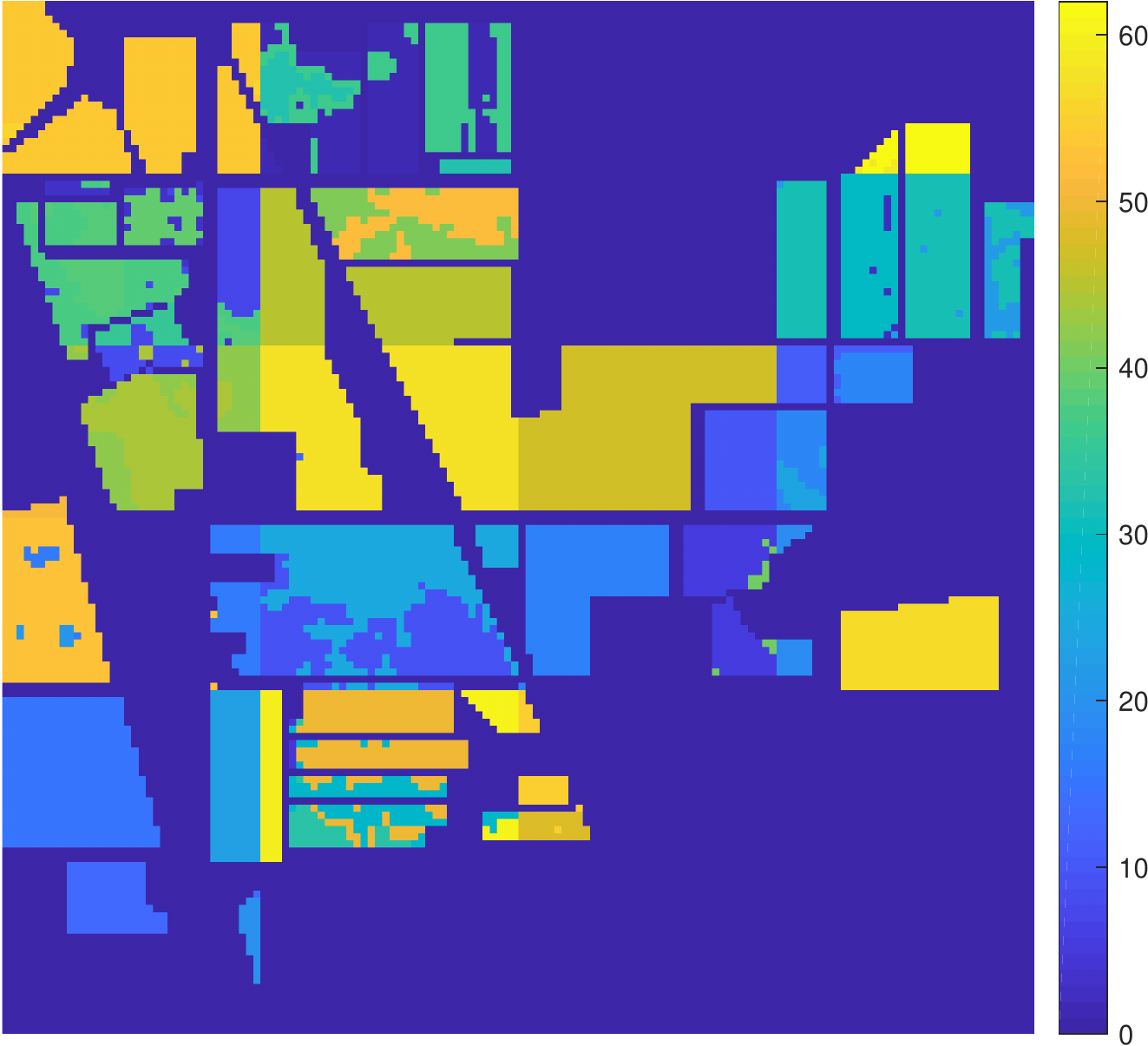}
\caption{FMS}
\end{subfigure}
\begin{subfigure}{ .09\textwidth}
\includegraphics[width=\textwidth]{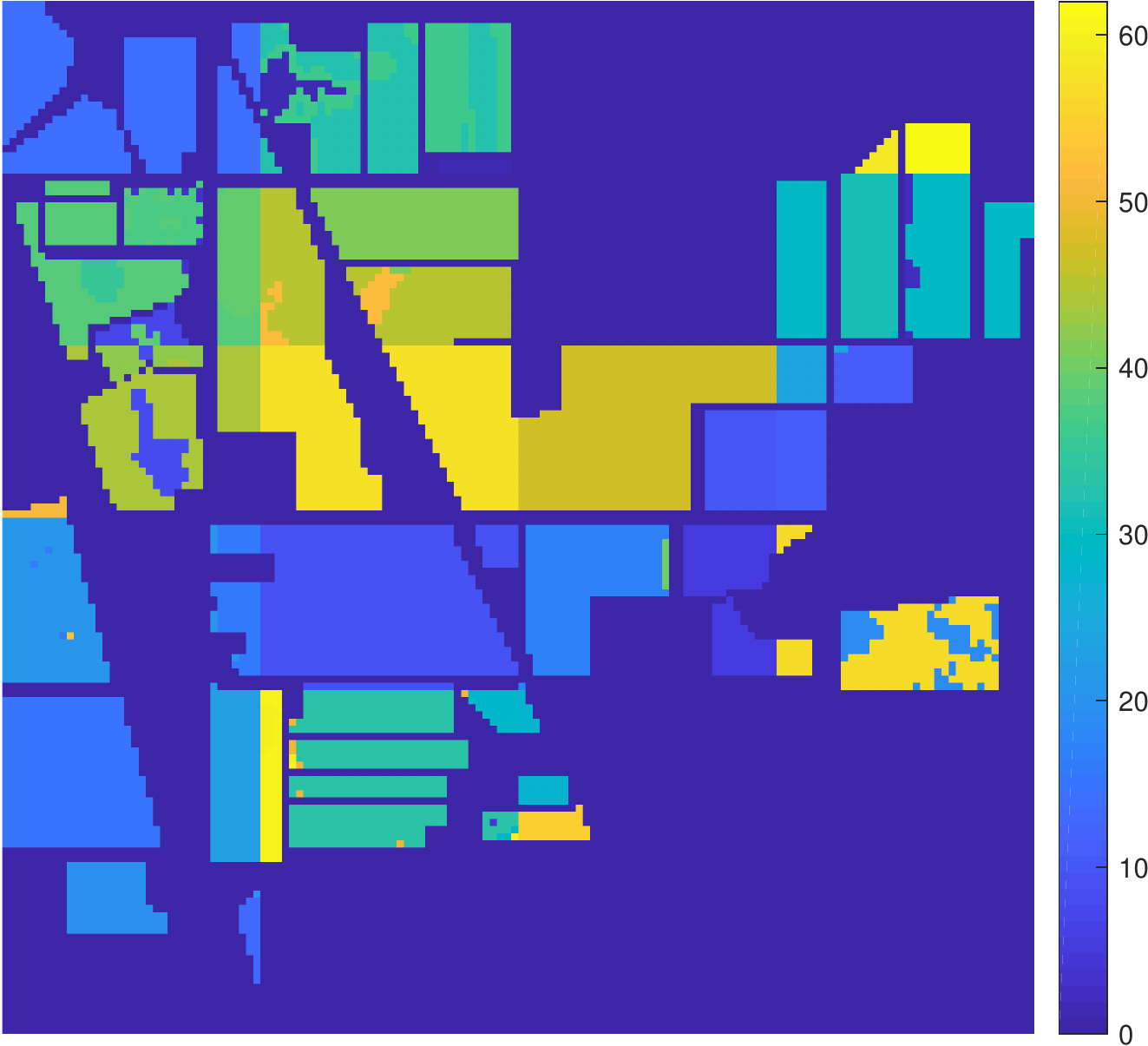}
\caption{FSFDPC}
\end{subfigure}
\begin{subfigure}{ .09\textwidth}
\includegraphics[width=\textwidth]{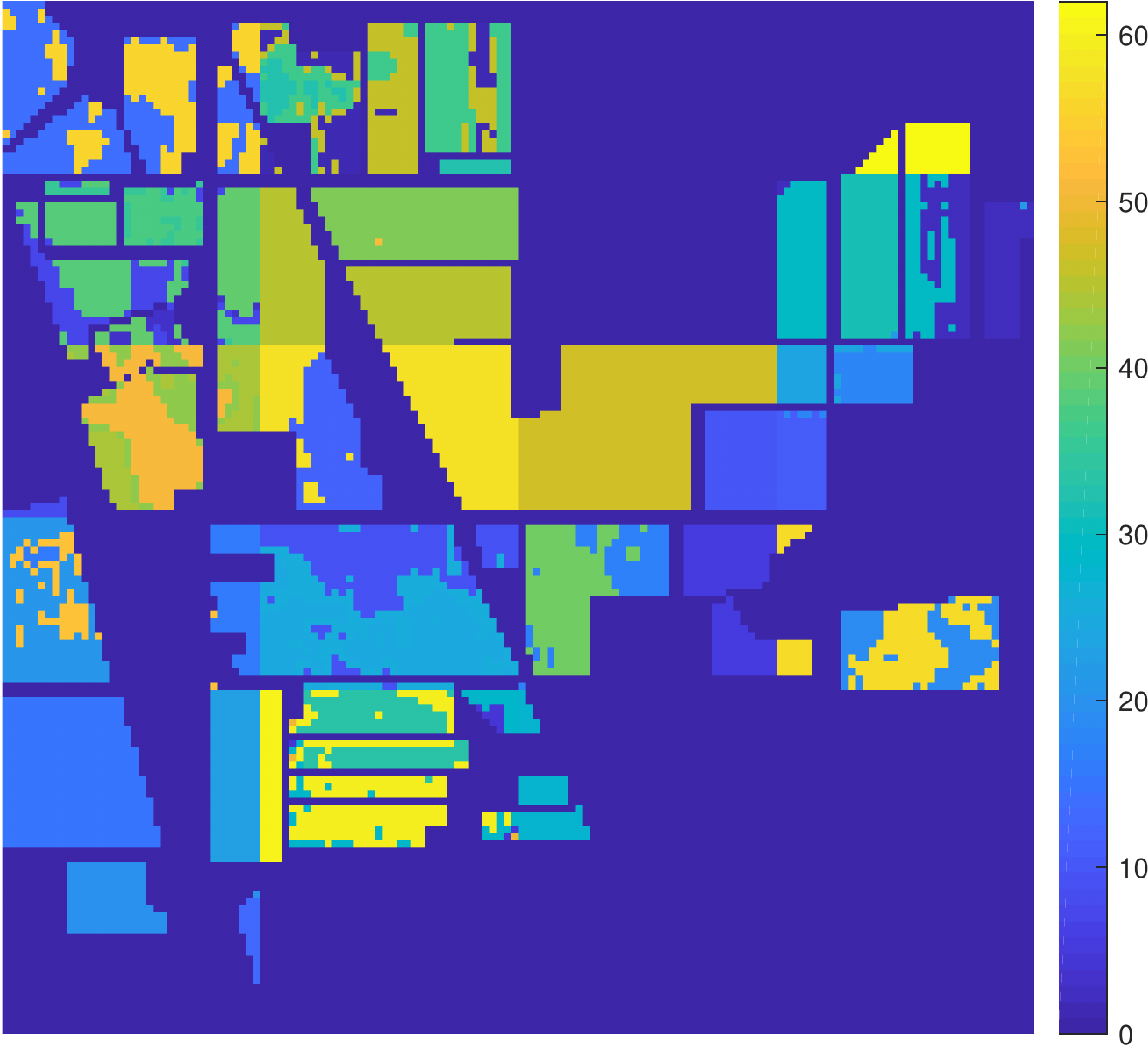}
\caption{DL}
\end{subfigure}
\begin{subfigure}{ .09\textwidth}
\includegraphics[width=\textwidth]{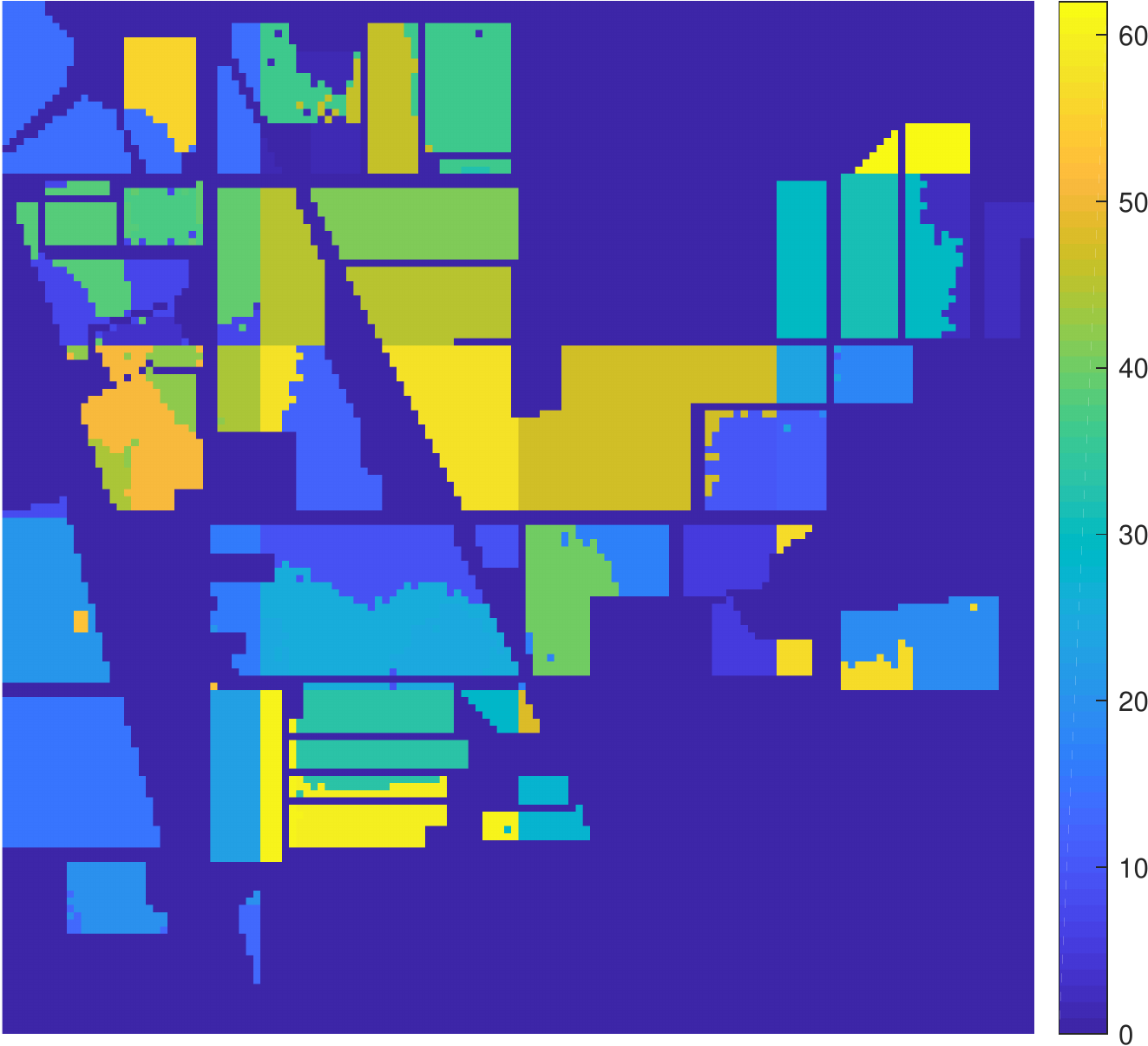}
\caption{DLSS}
\end{subfigure}
\caption{Results of clustering individual patches of the Indian Pines data, without synchronizing the labels.\label{fig:IP_Patches}} 
\end{figure}

\begin{figure}
\centering
\begin{subfigure}{.09\textwidth}
\includegraphics[width=\textwidth]{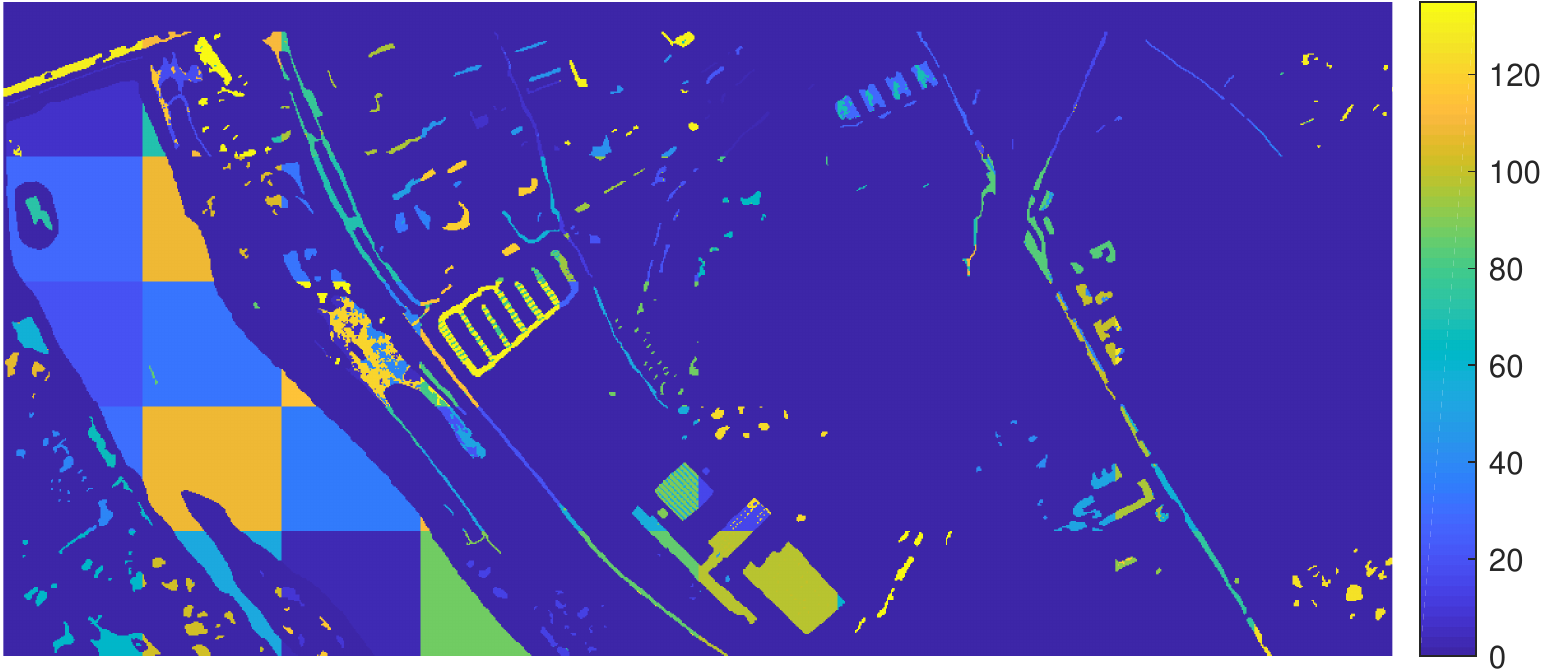}
\caption{$K$-means}
\end{subfigure}
\begin{subfigure}{.09\textwidth}
\includegraphics[width=\textwidth]{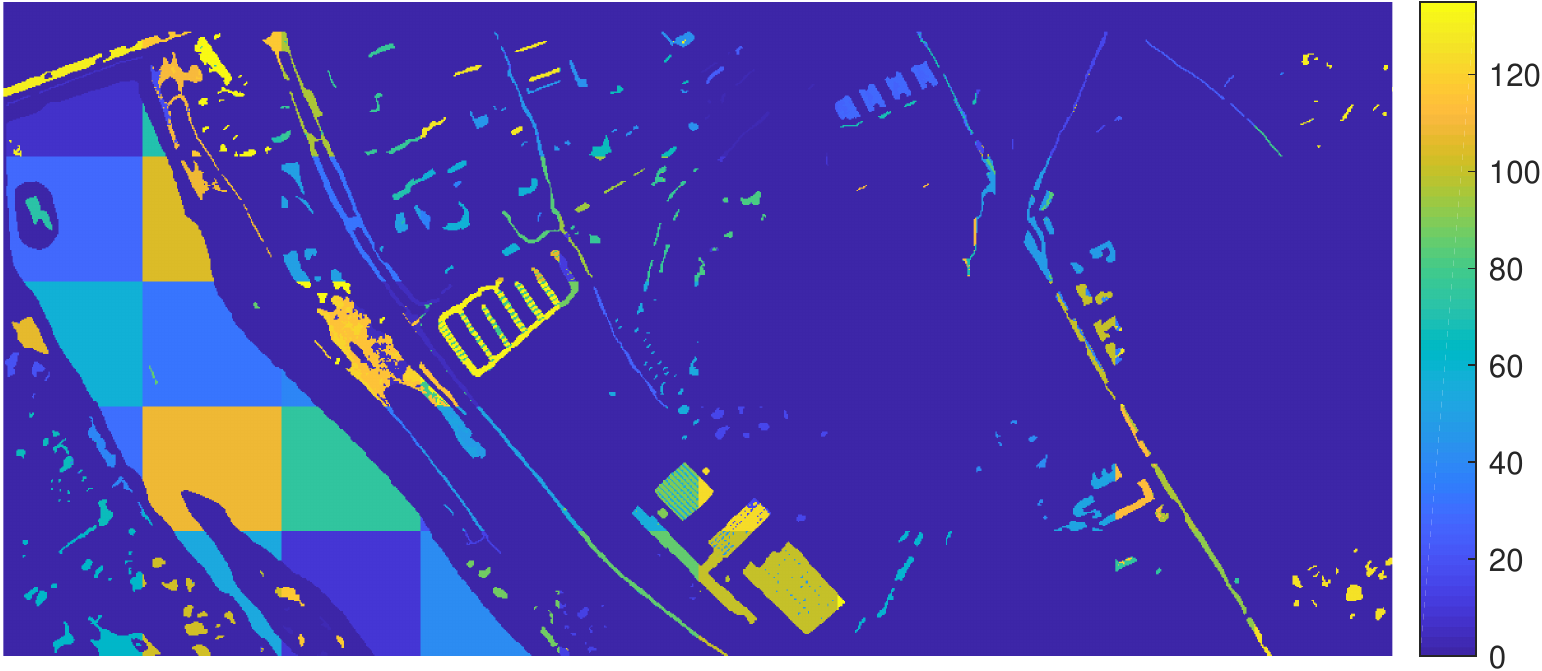}
\caption{PCA+$K$M}
\end{subfigure}
\begin{subfigure}{.09\textwidth}
\includegraphics[width=\textwidth]{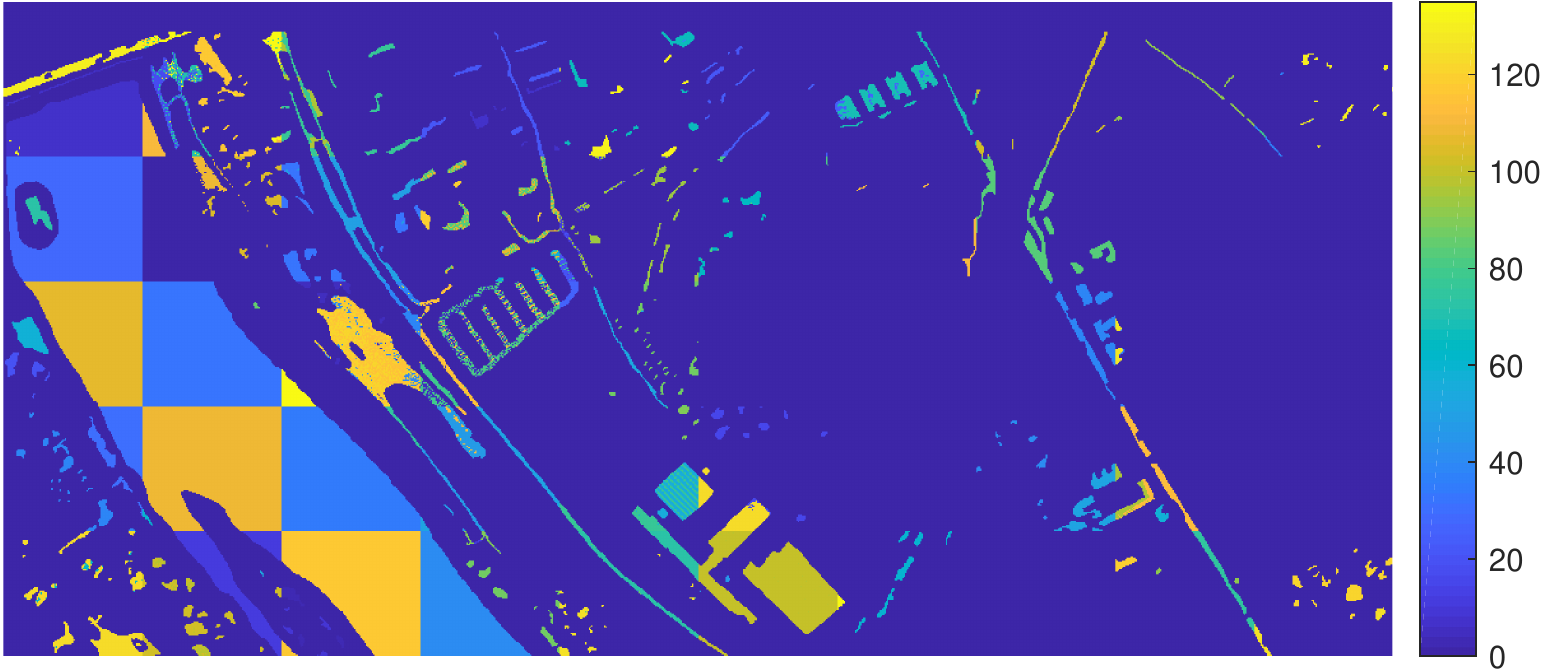}
\caption{ICA+$K$M}
\end{subfigure}
\begin{subfigure}{.09\textwidth}
\includegraphics[width=\textwidth]{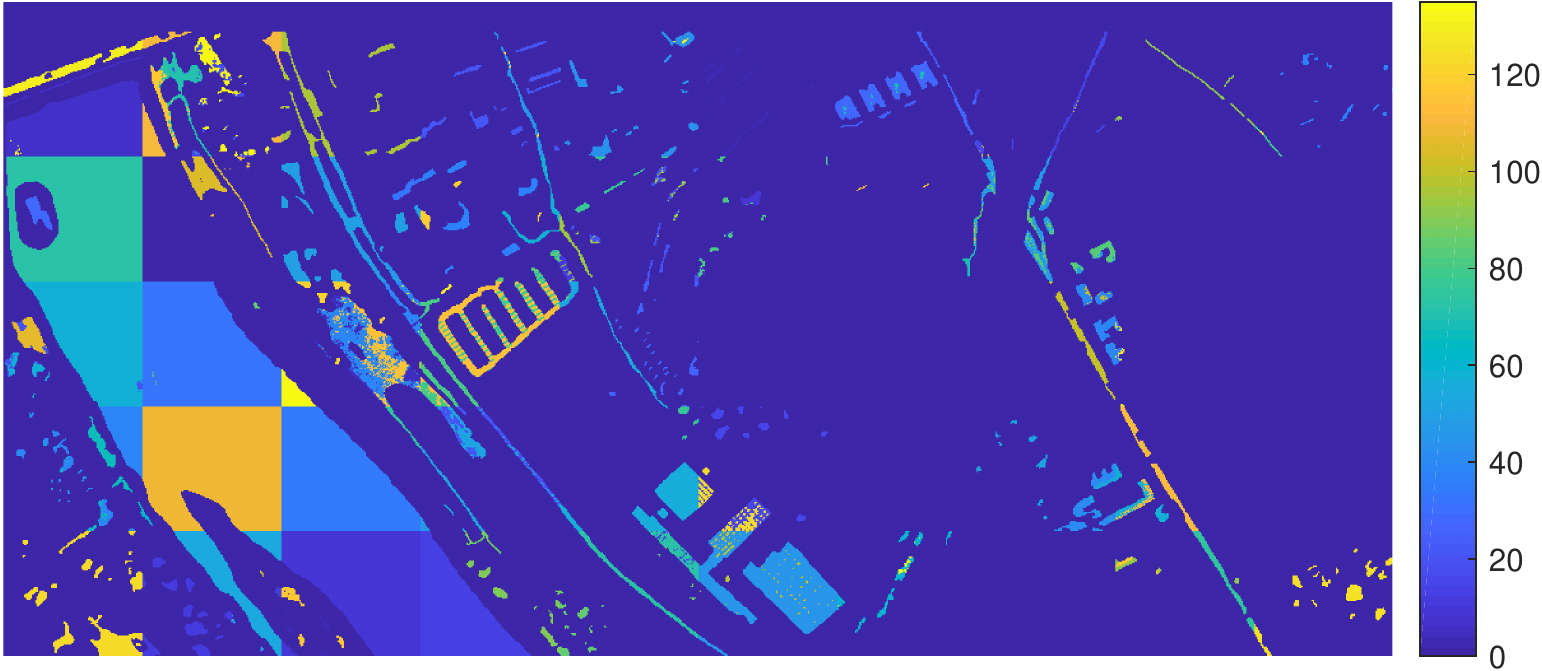}
\caption{RP+$K$M}
\end{subfigure}
\begin{subfigure}{ .09\textwidth}
\includegraphics[width=\textwidth]{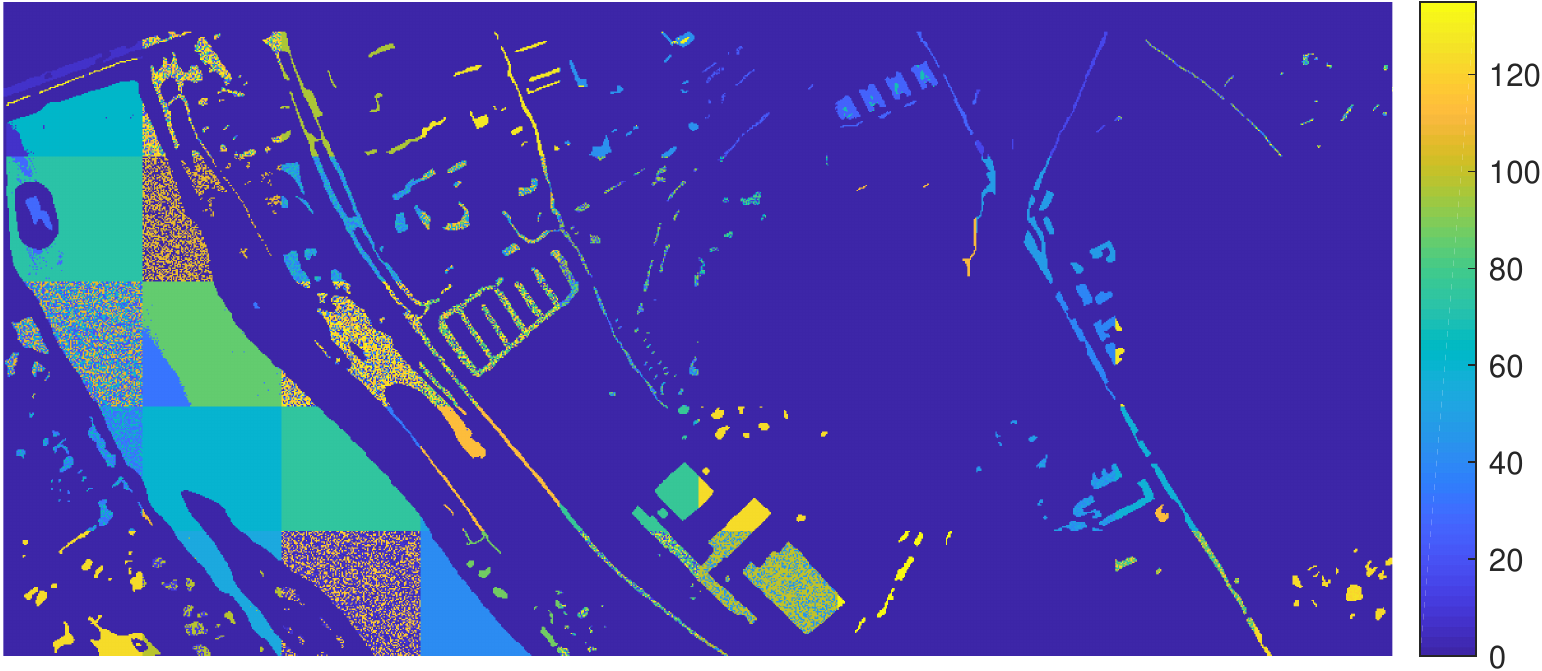}
\caption{DBSCAN}
\end{subfigure}
\begin{subfigure}{.09\textwidth}
\includegraphics[width=\textwidth]{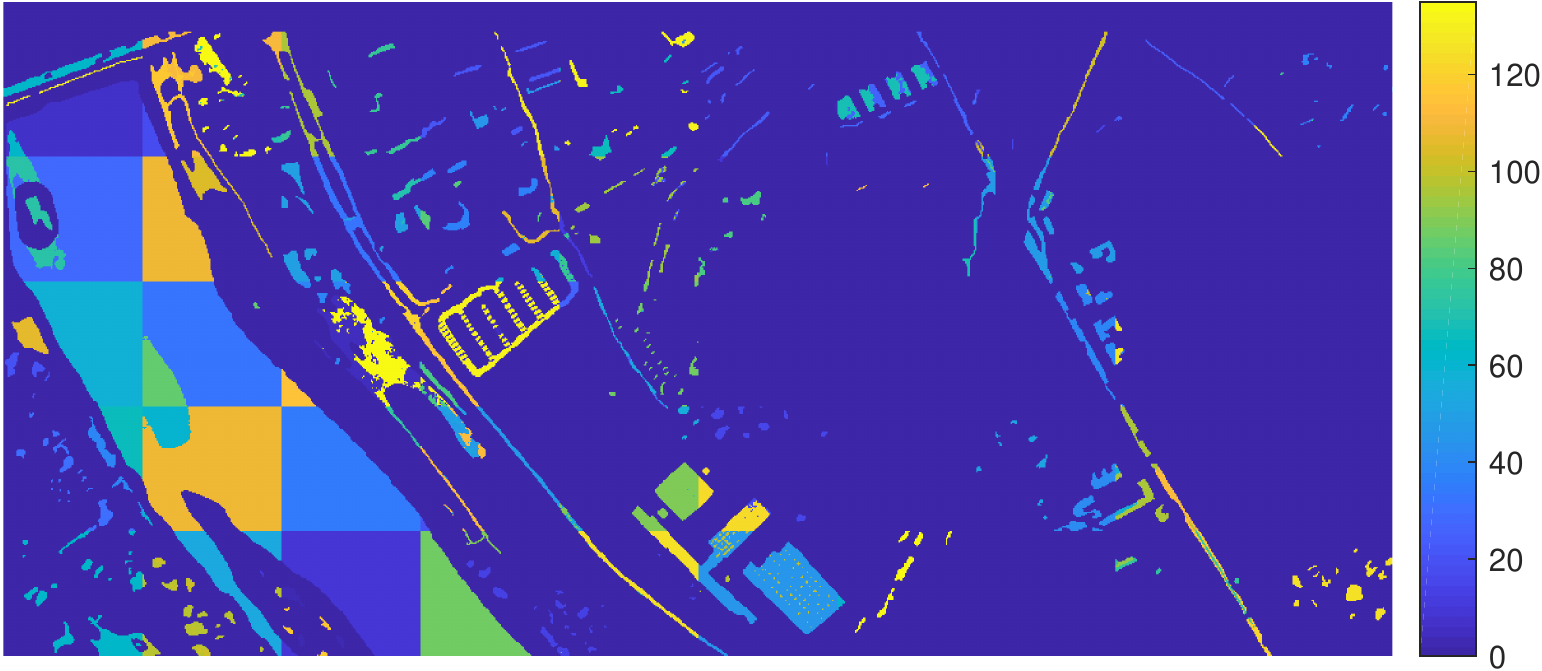}
\caption{SC}
\end{subfigure}
\begin{subfigure}{.09\textwidth}
\includegraphics[width=\textwidth]{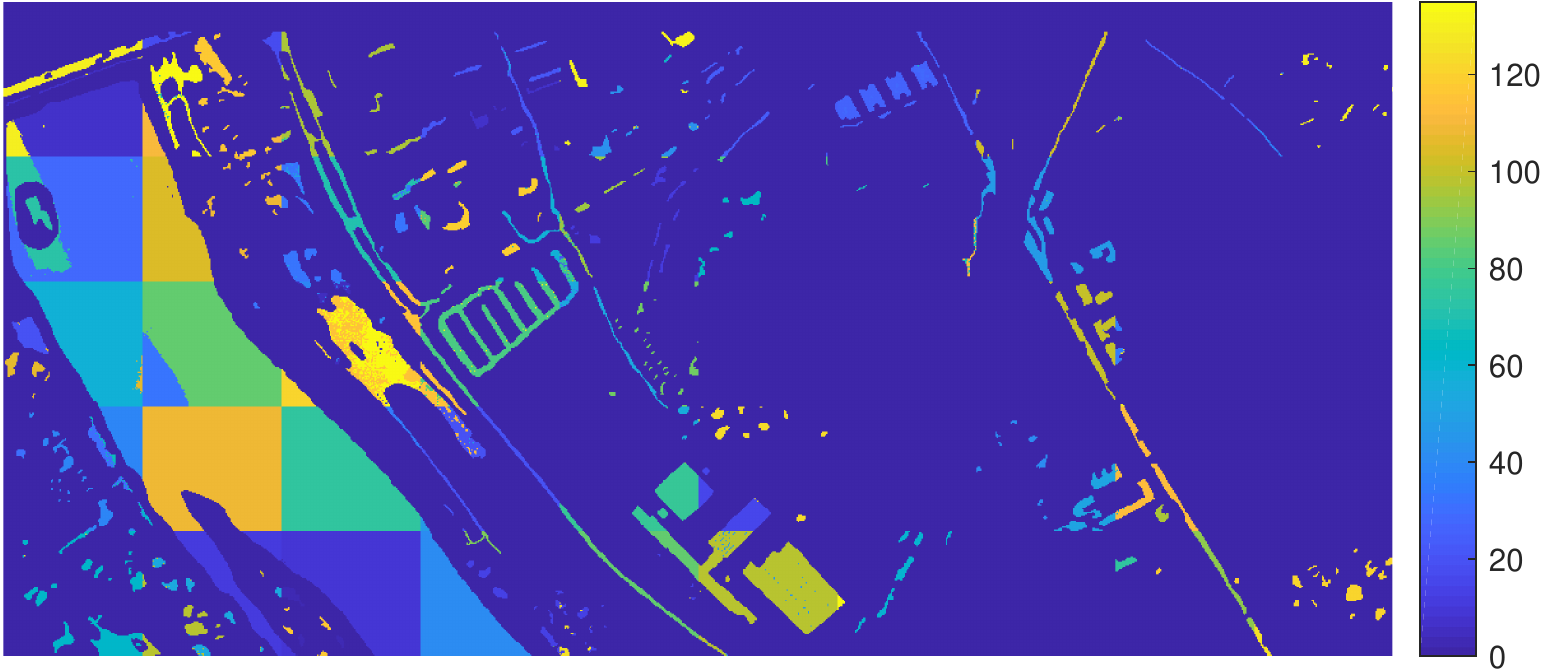}
\caption{GMM}
\end{subfigure}
\begin{subfigure}{.09\textwidth}
\includegraphics[width=\textwidth]{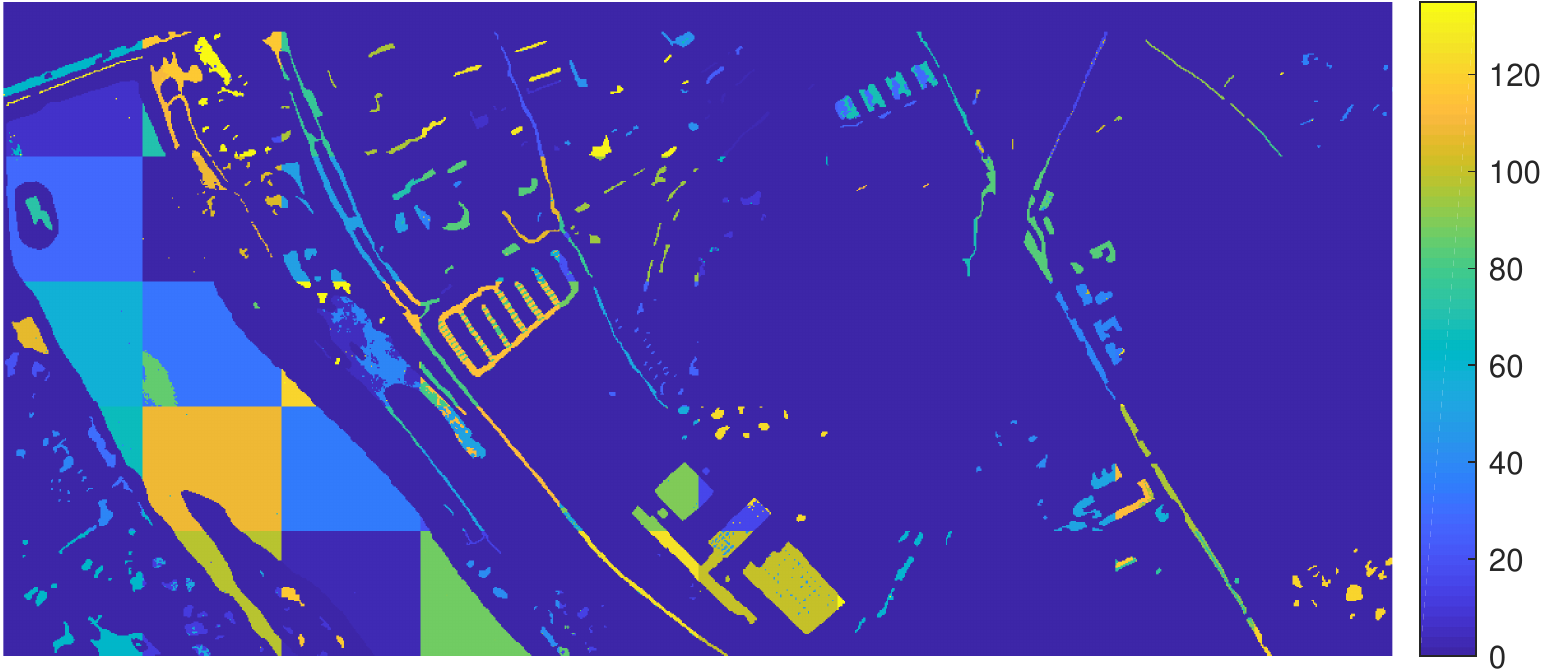}
\caption{SMCE}
\end{subfigure}
\begin{subfigure}{.09\textwidth}
\includegraphics[width=\textwidth]{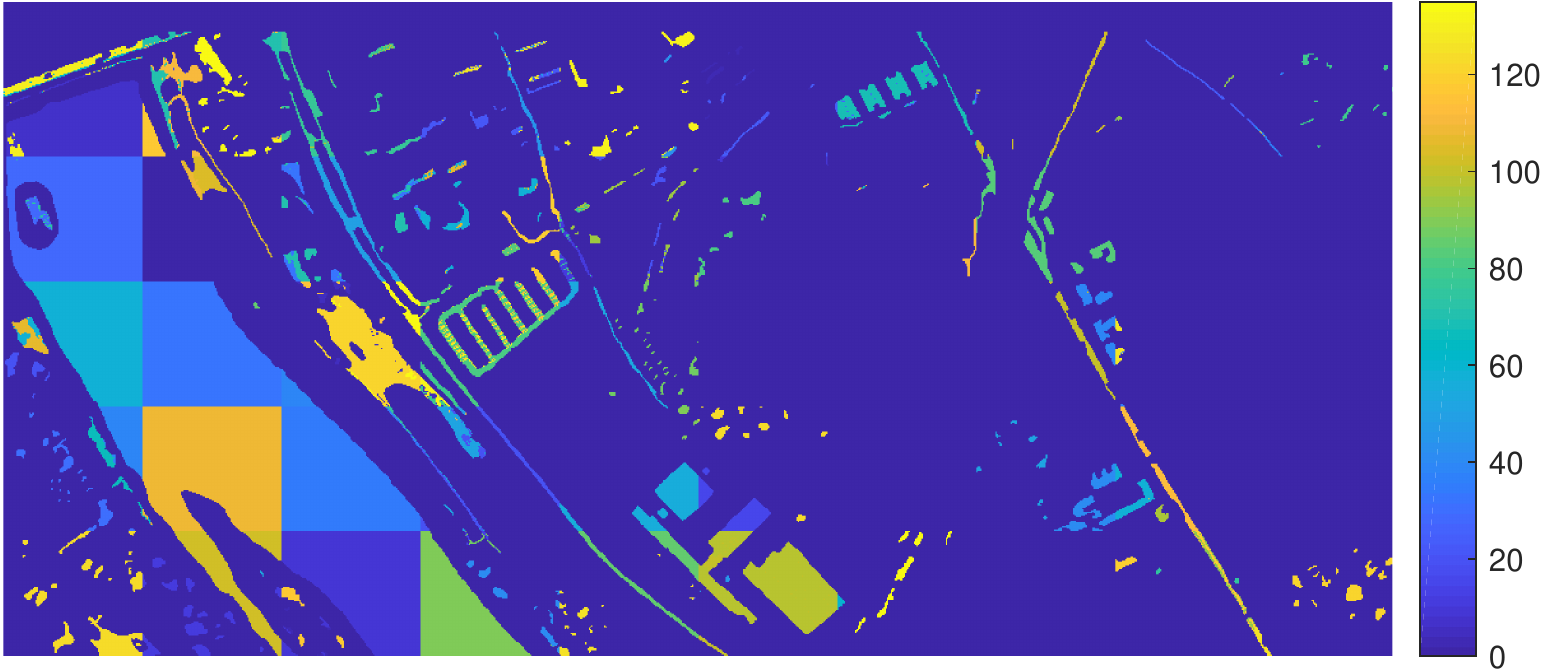}
\caption{HNMF}
\end{subfigure}
\begin{subfigure}{ .09\textwidth}
\includegraphics[width=\textwidth]{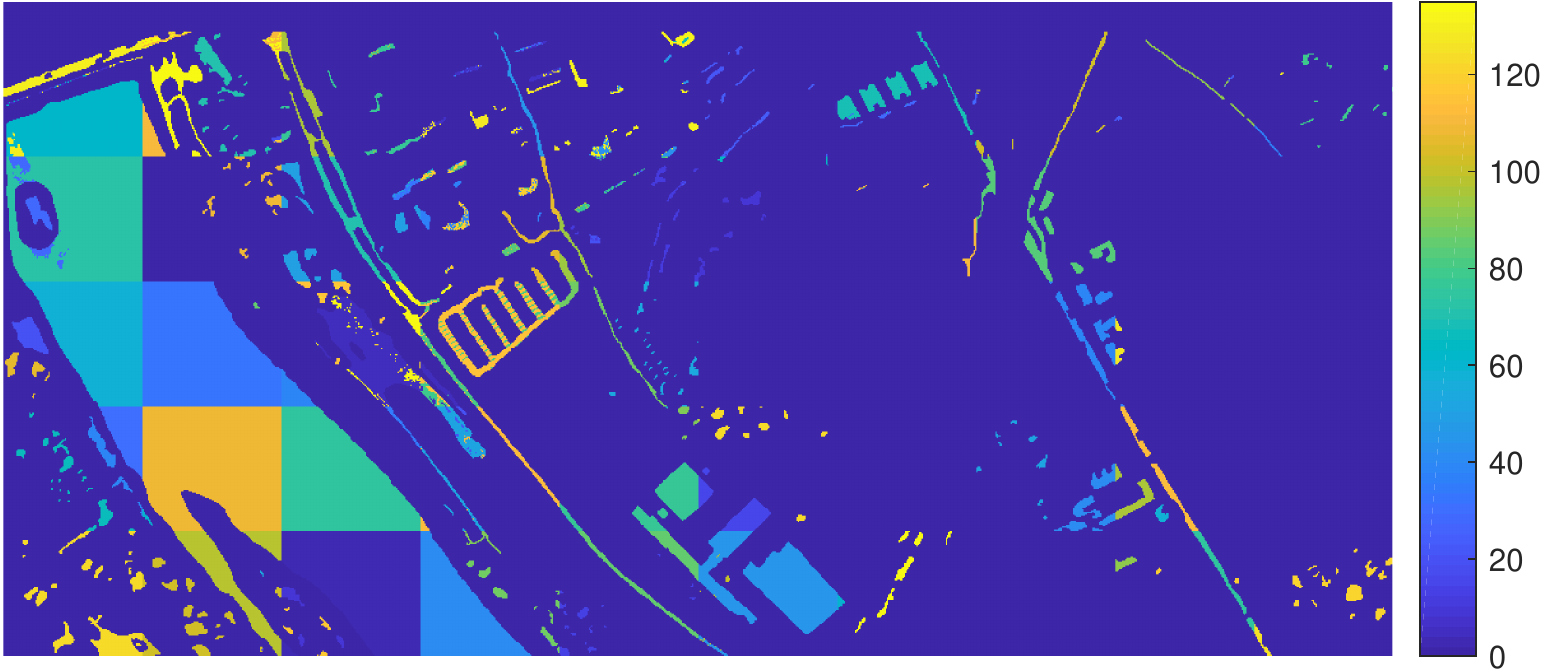}
\caption{FMS}
\end{subfigure}
\begin{subfigure}{ .09\textwidth}
\includegraphics[width=\textwidth]{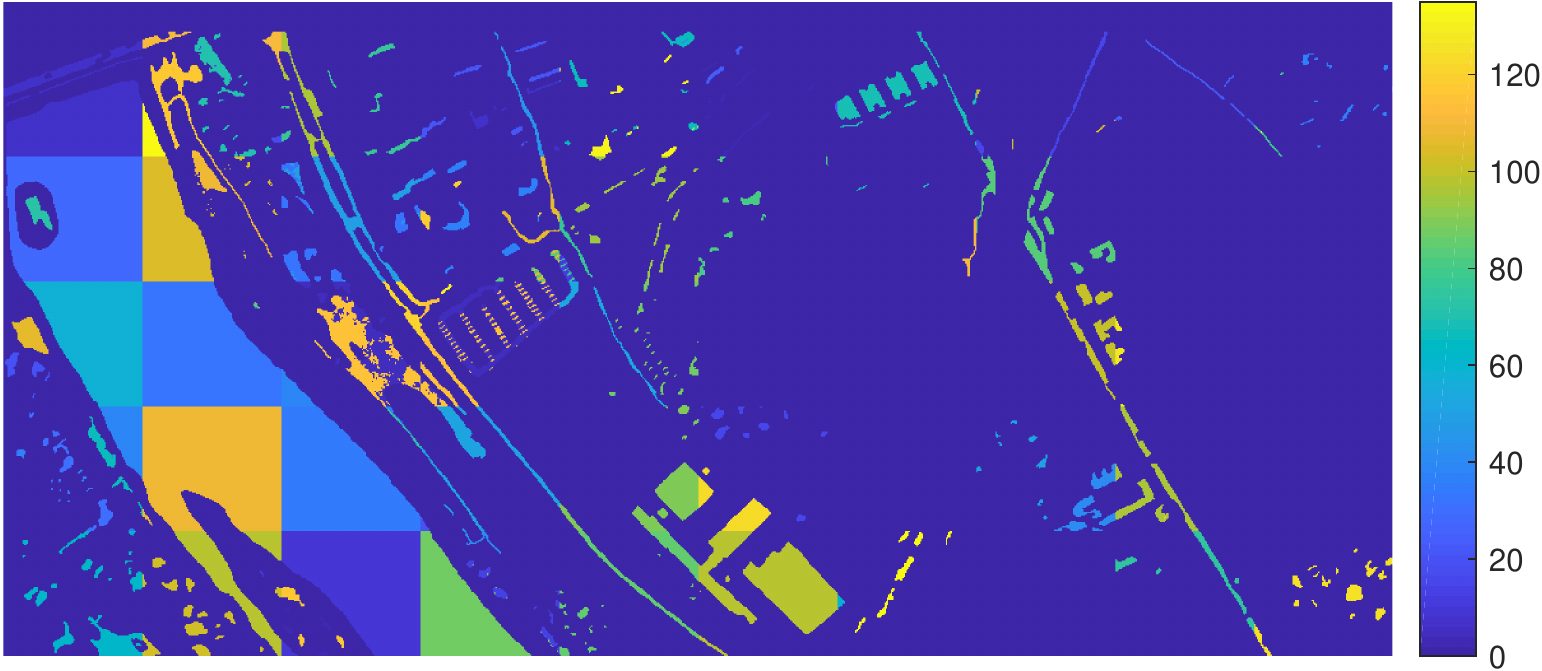}
\caption{FSFDPC}
\end{subfigure}
\begin{subfigure}{ .09\textwidth}
\includegraphics[width=\textwidth]{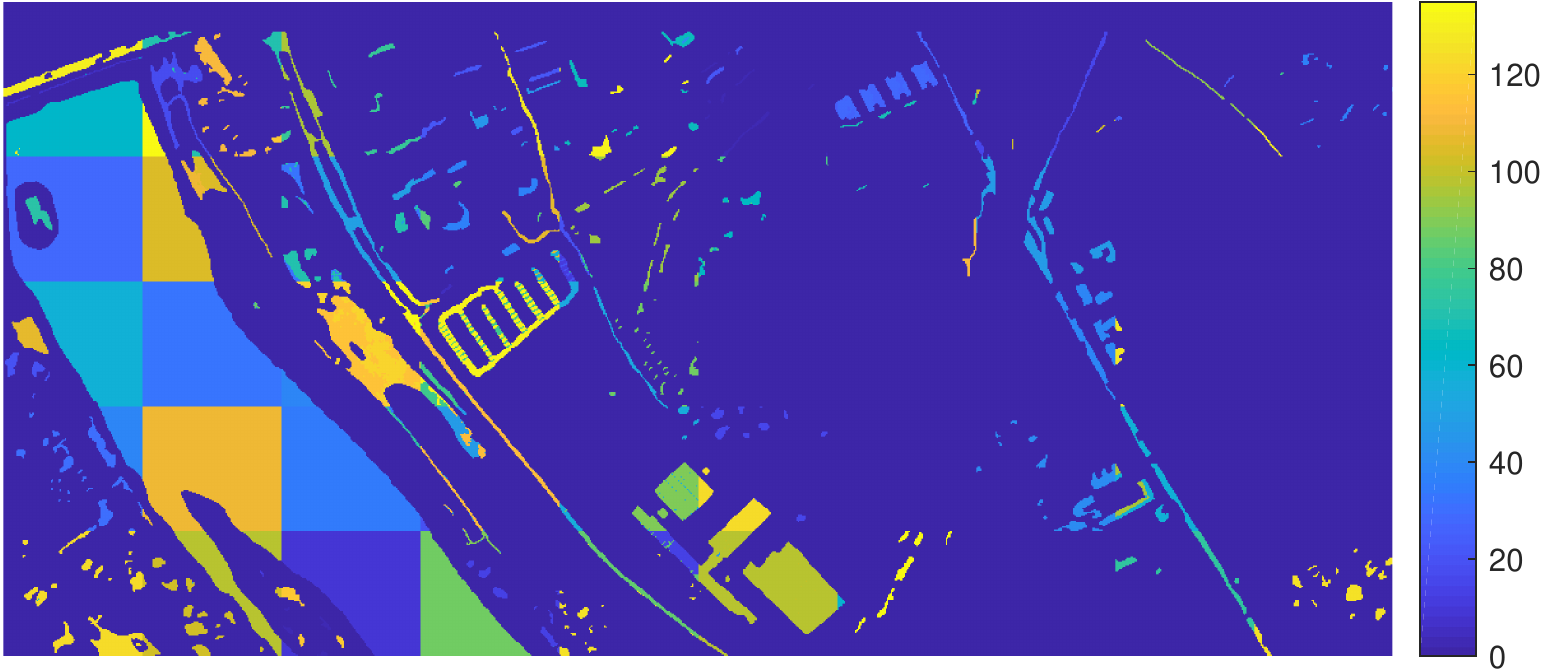}
\caption{DL}
\end{subfigure}
\begin{subfigure}{ .09\textwidth}
\includegraphics[width=\textwidth]{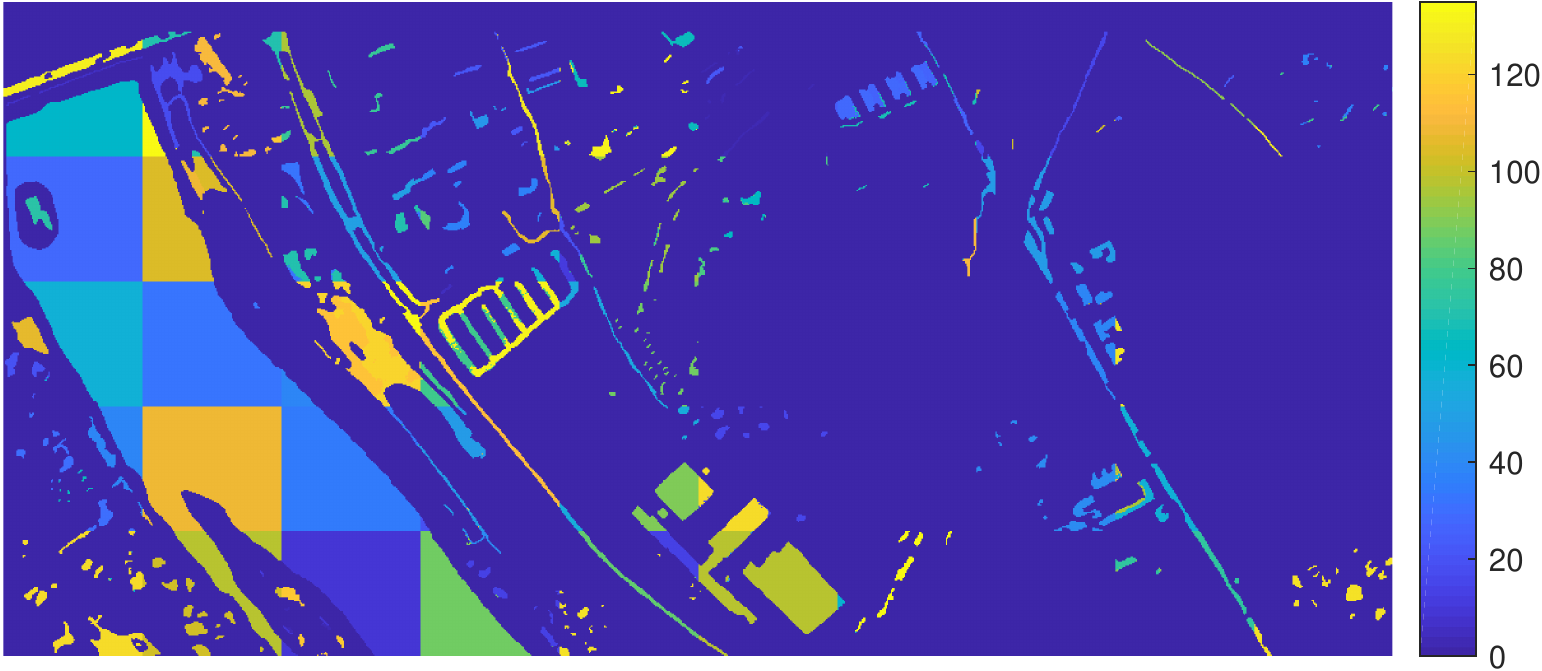}
\caption{DLSS}
\end{subfigure}
\caption{Results of clustering individual patches of the Pavia data, without synchronizing the labels.\label{fig:Pavia_Patches}} 
\end{figure}

\begin{figure}[t!]
\centering
\begin{subfigure}{.09\textwidth}
\includegraphics[width=\textwidth]{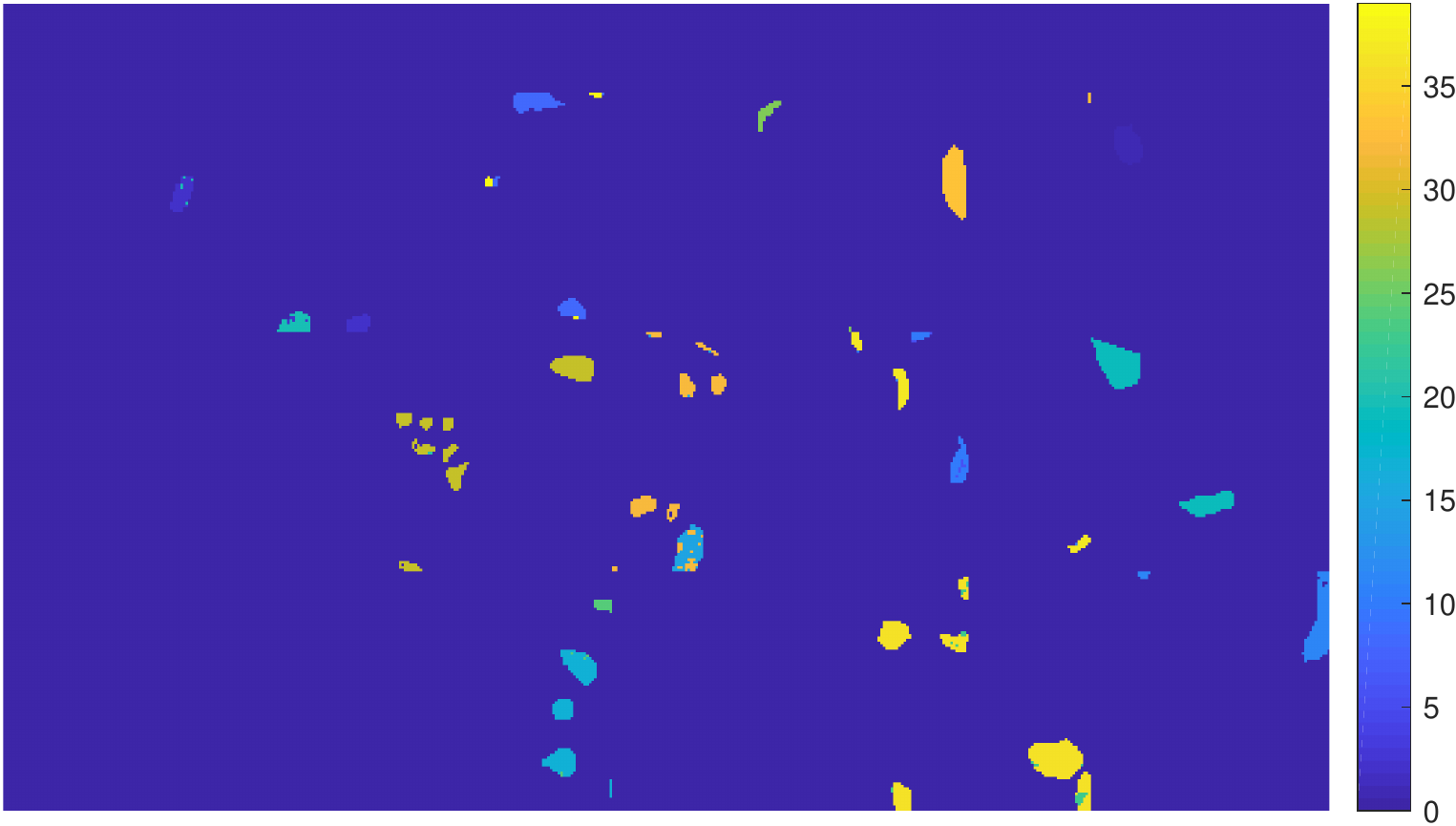}
\caption{$K$-means}
\end{subfigure}
\begin{subfigure}{.09\textwidth}
\includegraphics[width=\textwidth]{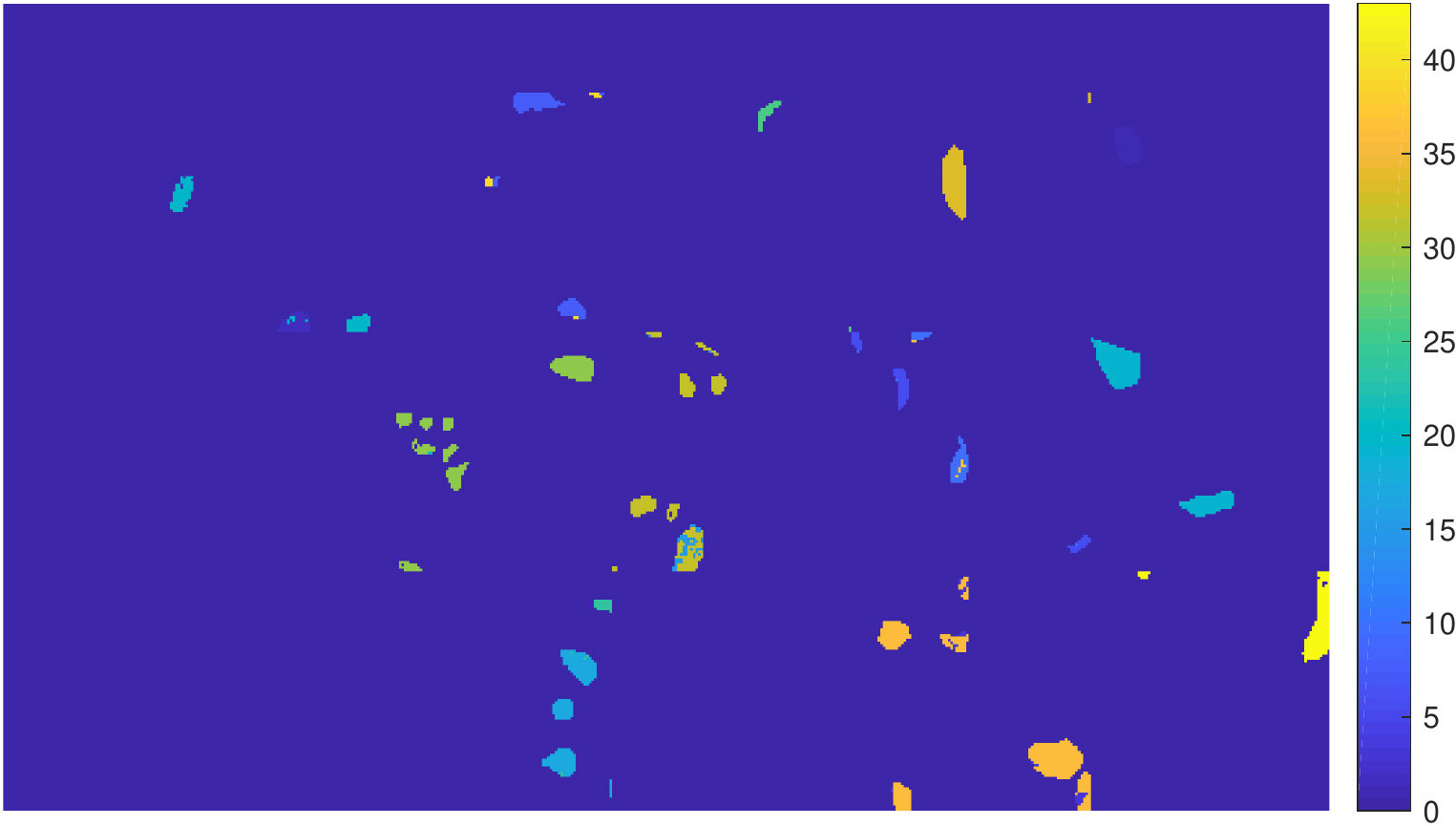}
\caption{PCA+$K$M}
\end{subfigure}
\begin{subfigure}{.09\textwidth}
\includegraphics[width=\textwidth]{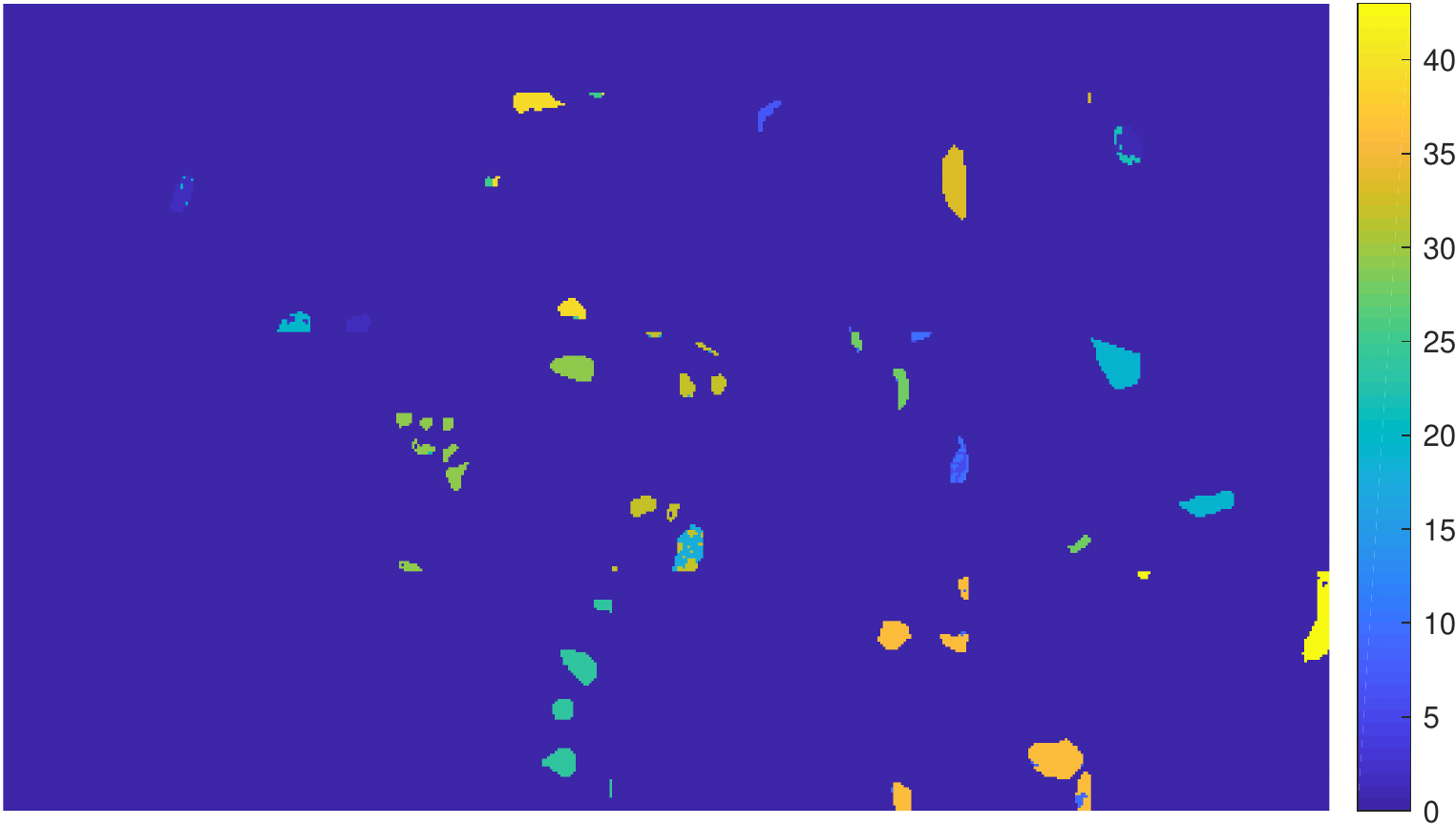}
\caption{ICA+$K$M}
\end{subfigure}
\begin{subfigure}{.09\textwidth}
\includegraphics[width=\textwidth]{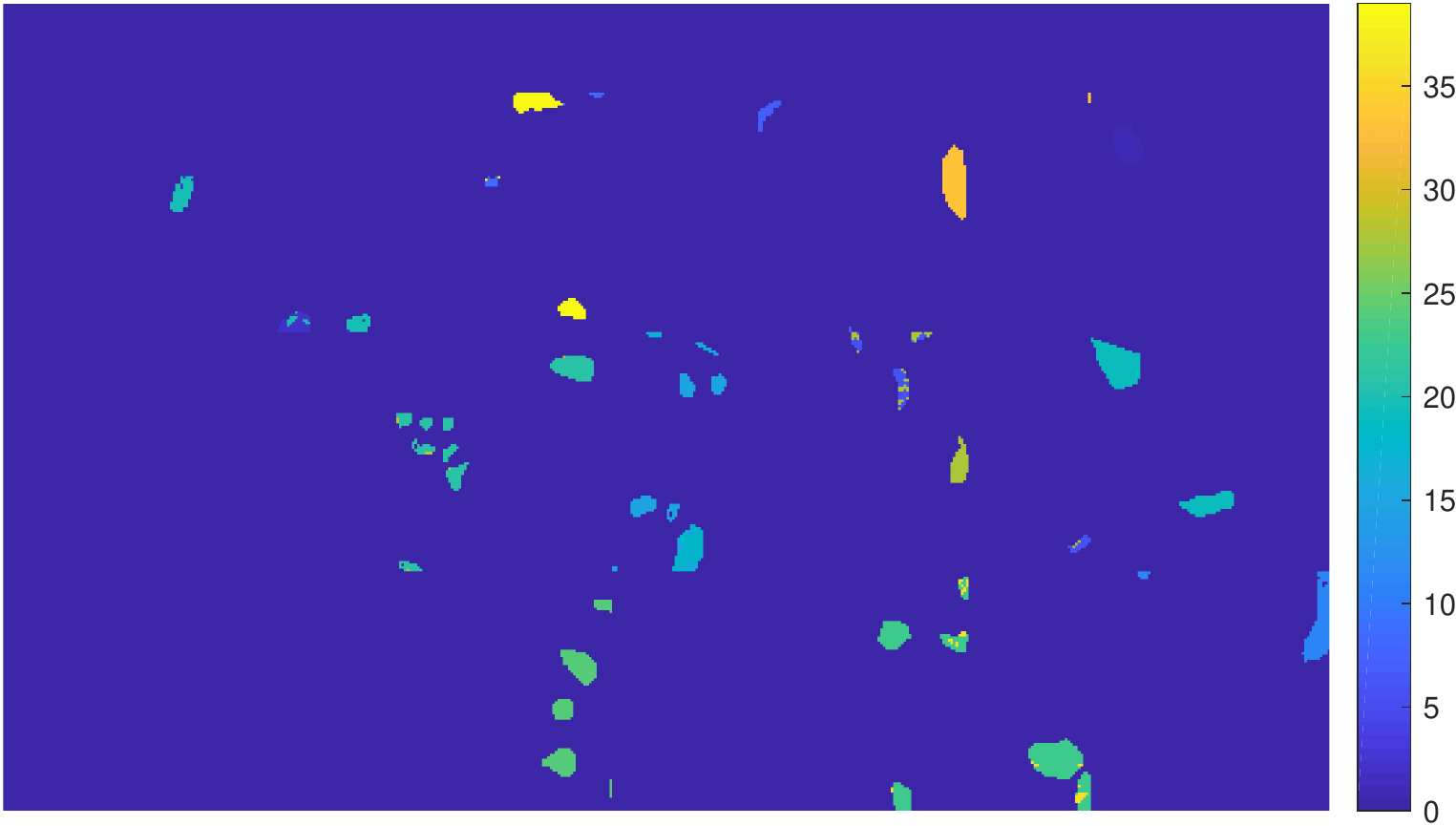}
\caption{RP+$K$M}
\end{subfigure}
\begin{subfigure}{ .09\textwidth}
\includegraphics[width=\textwidth]{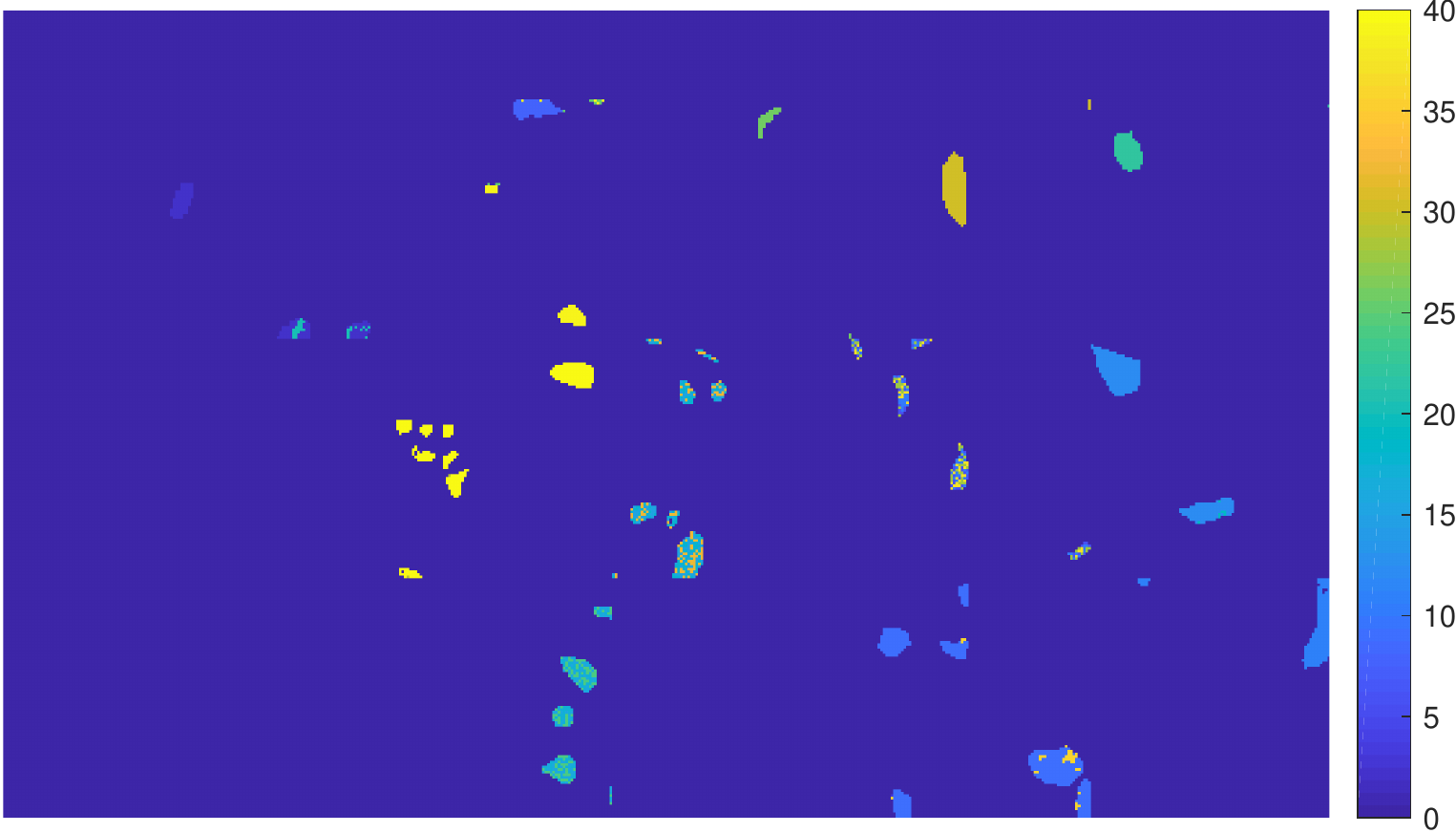}
\caption{DBSCAN}
\end{subfigure}
\begin{subfigure}{.09\textwidth}
\includegraphics[width=\textwidth]{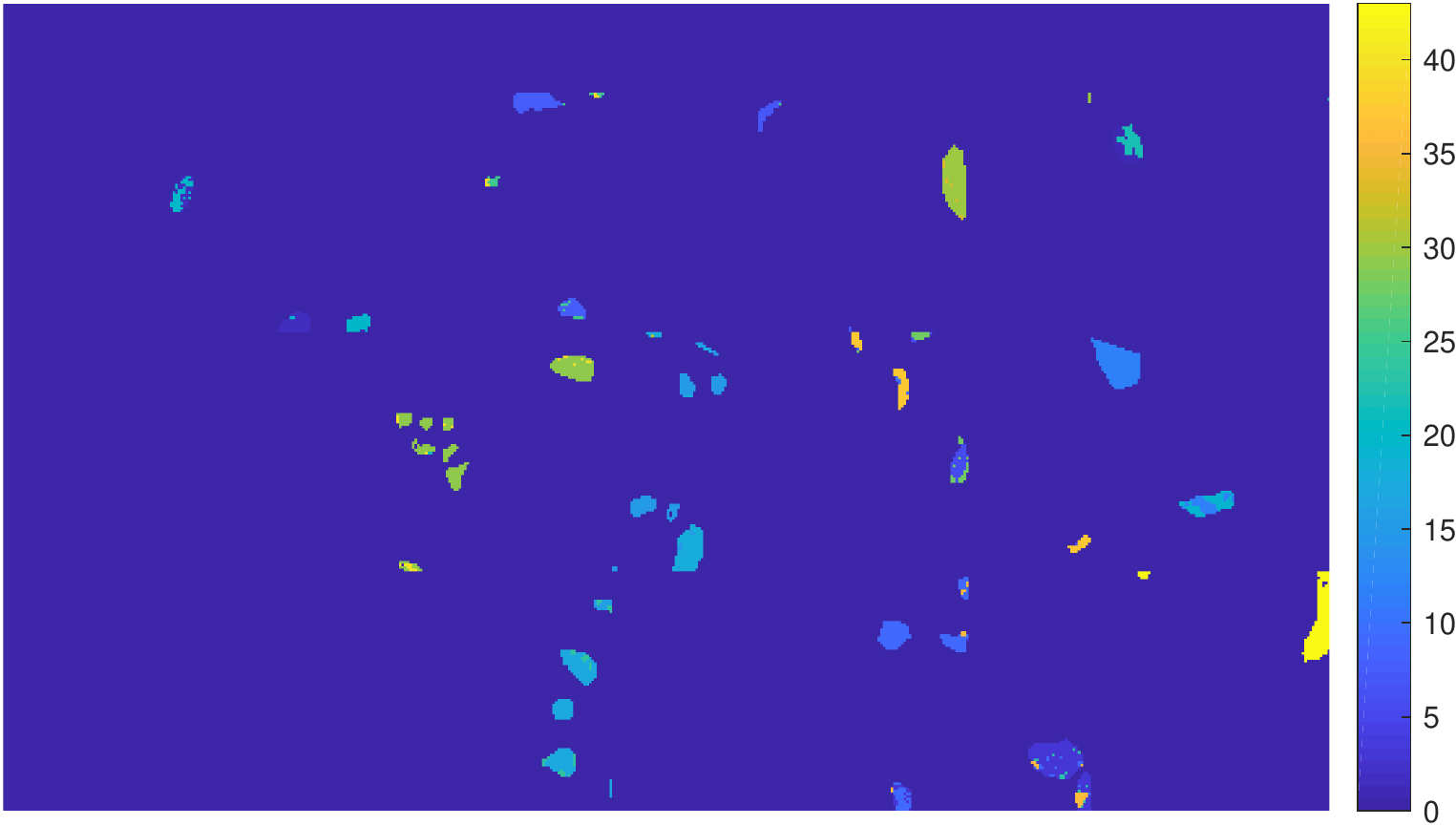}
\caption{SC}
\end{subfigure}
\begin{subfigure}{.09\textwidth}
\includegraphics[width=\textwidth]{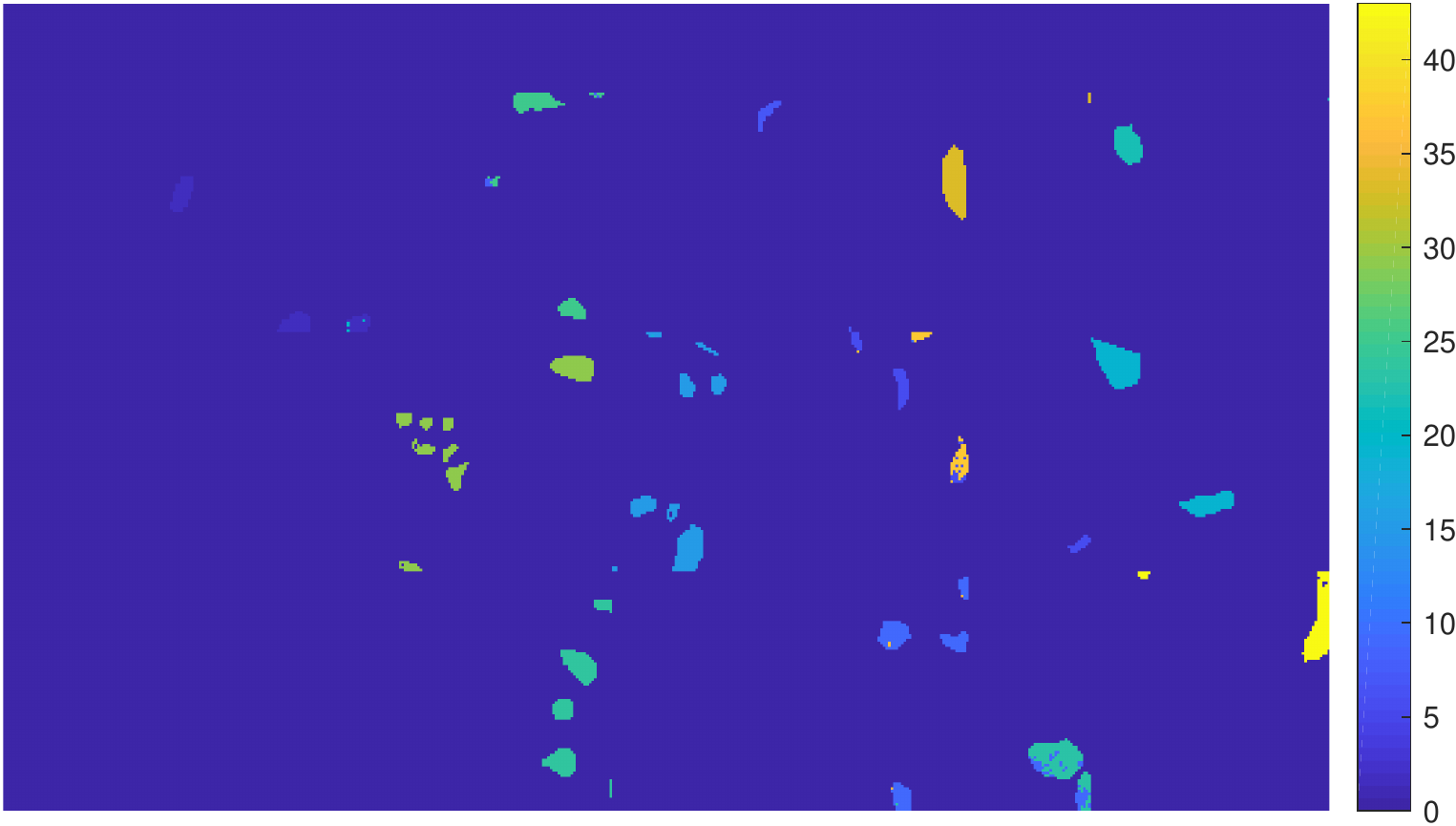}
\caption{GMM}
\end{subfigure}
\begin{subfigure}{.09\textwidth}
\includegraphics[width=\textwidth]{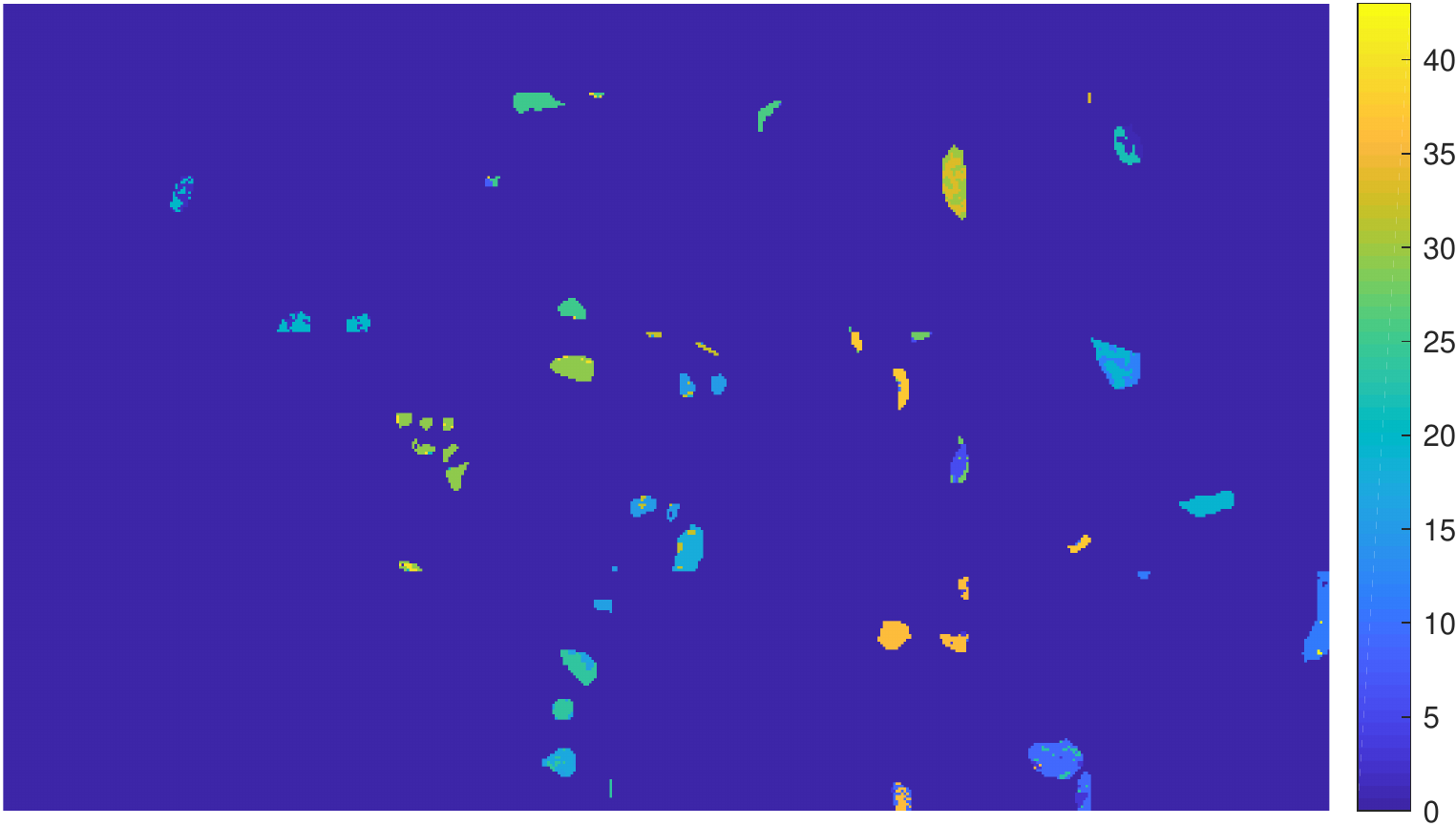}
\caption{SMCE}
\end{subfigure}
\begin{subfigure}{.09\textwidth}
\includegraphics[width=\textwidth]{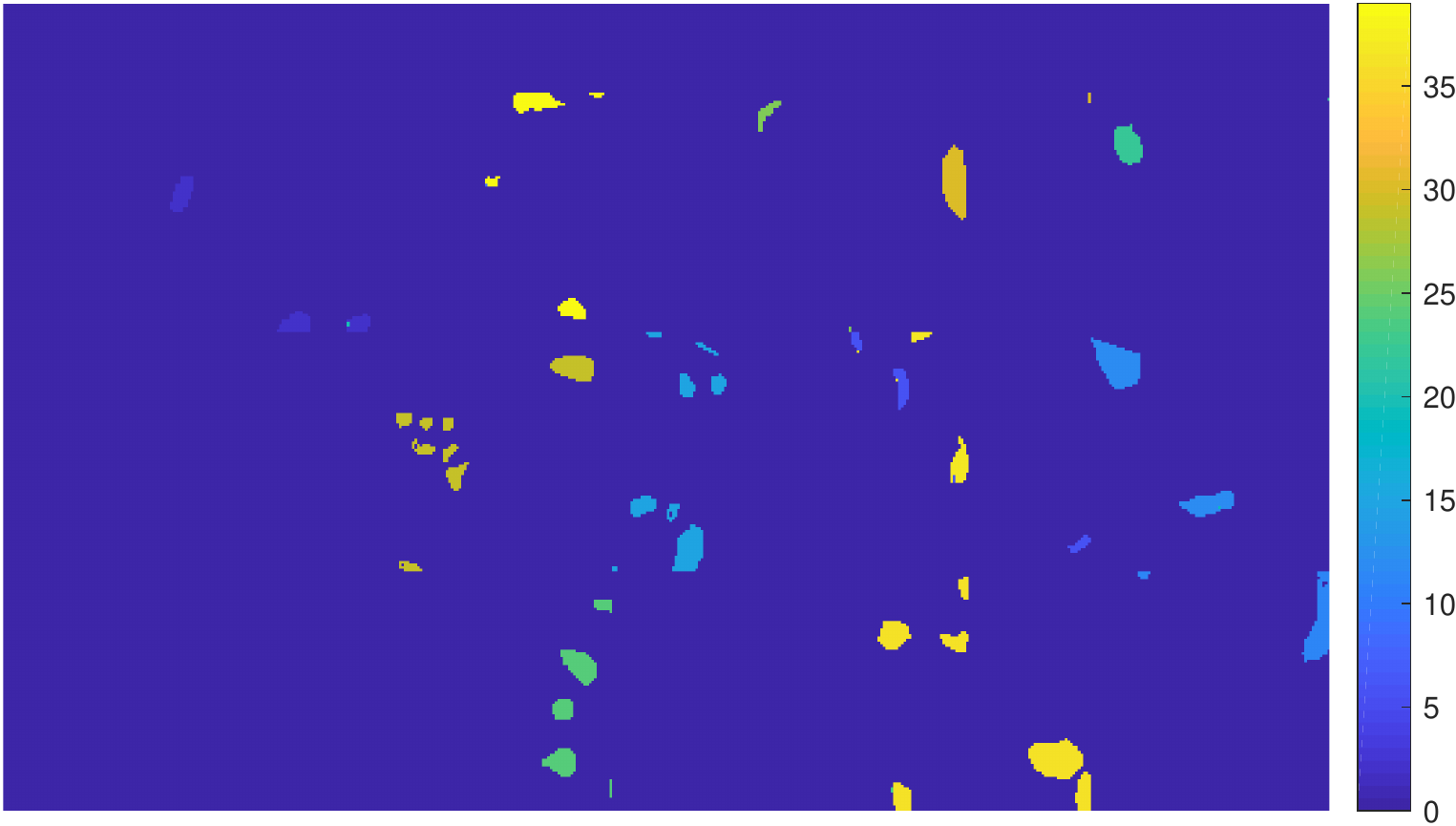}
\caption{HNMF}
\end{subfigure}
\begin{subfigure}{ .09\textwidth}
\includegraphics[width=\textwidth]{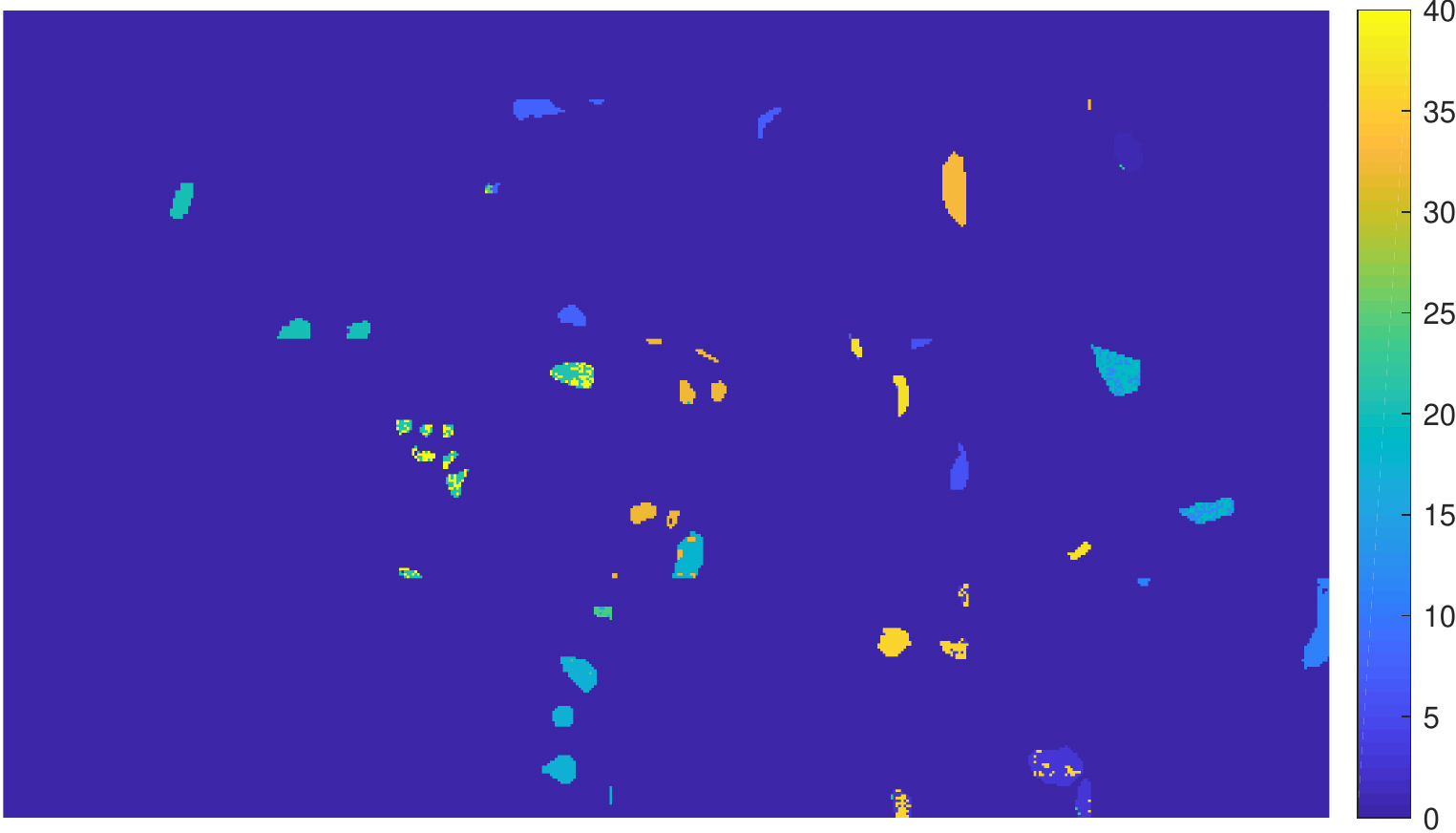}
\caption{FMS}
\end{subfigure}
\begin{subfigure}{ .09\textwidth}
\includegraphics[width=\textwidth]{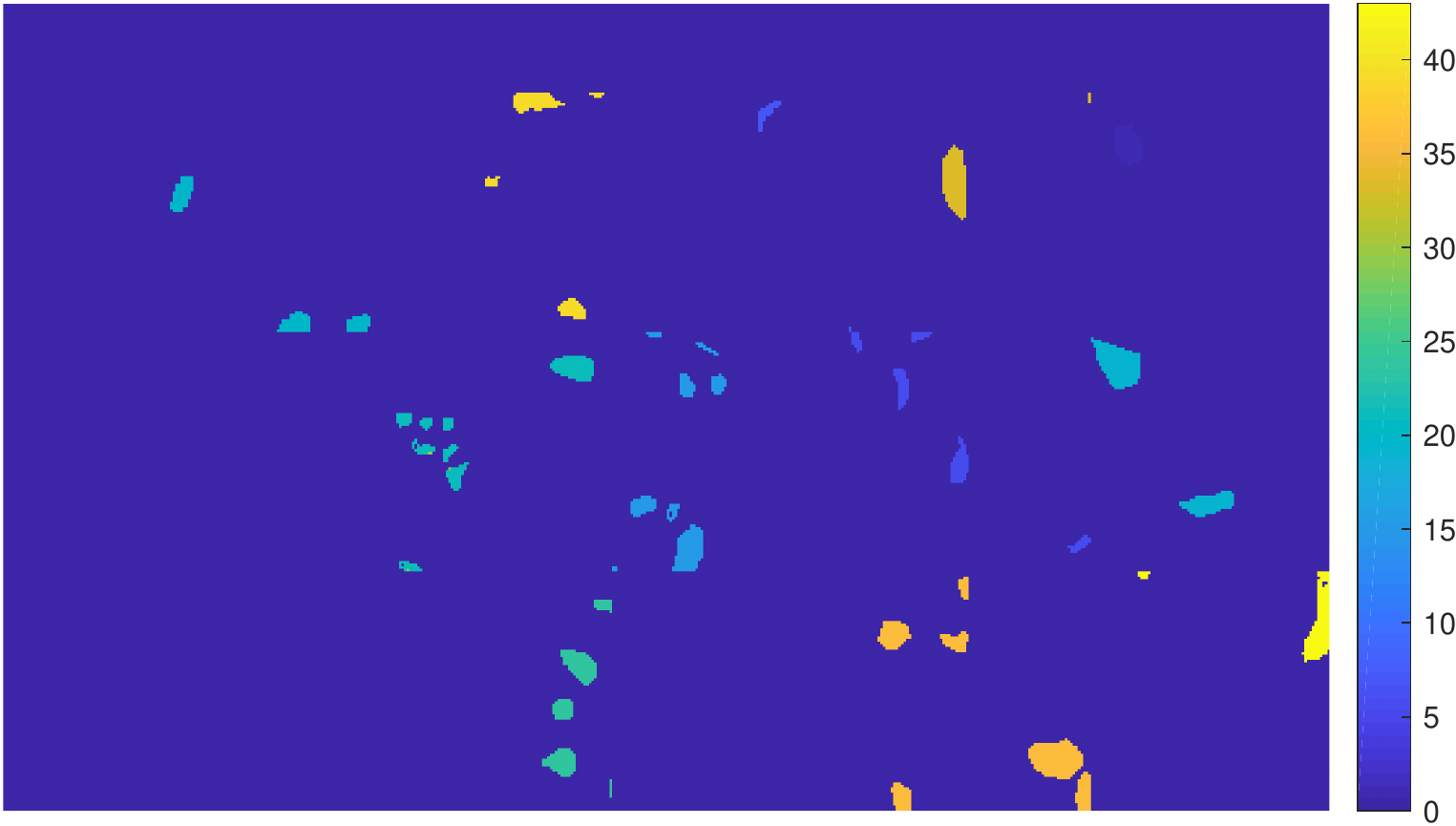}
\caption{FSFDPC}
\end{subfigure}
\begin{subfigure}{ .09\textwidth}
\includegraphics[width=\textwidth]{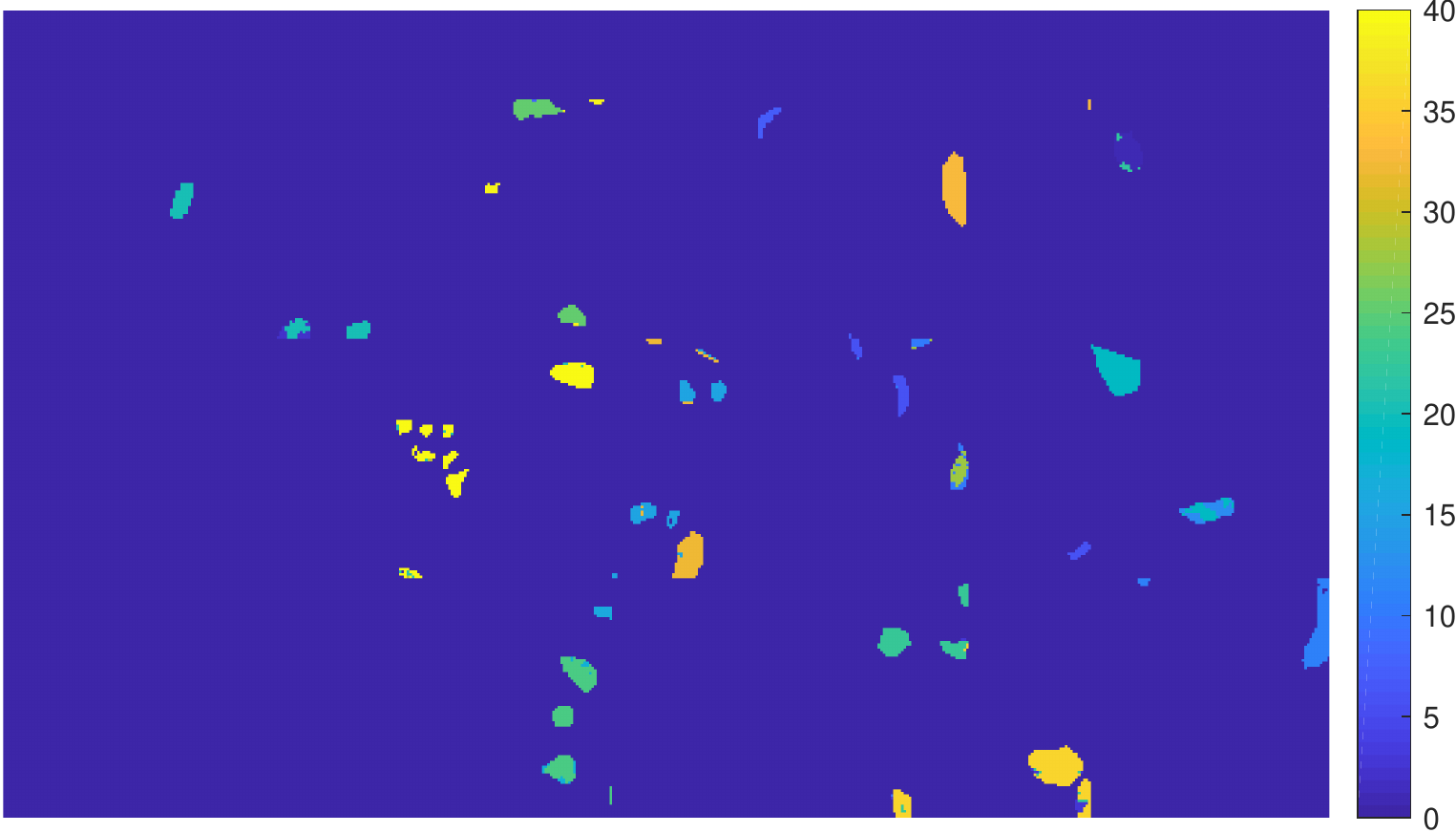}
\caption{DL}
\end{subfigure}
\begin{subfigure}{ .09\textwidth}
\includegraphics[width=\textwidth]{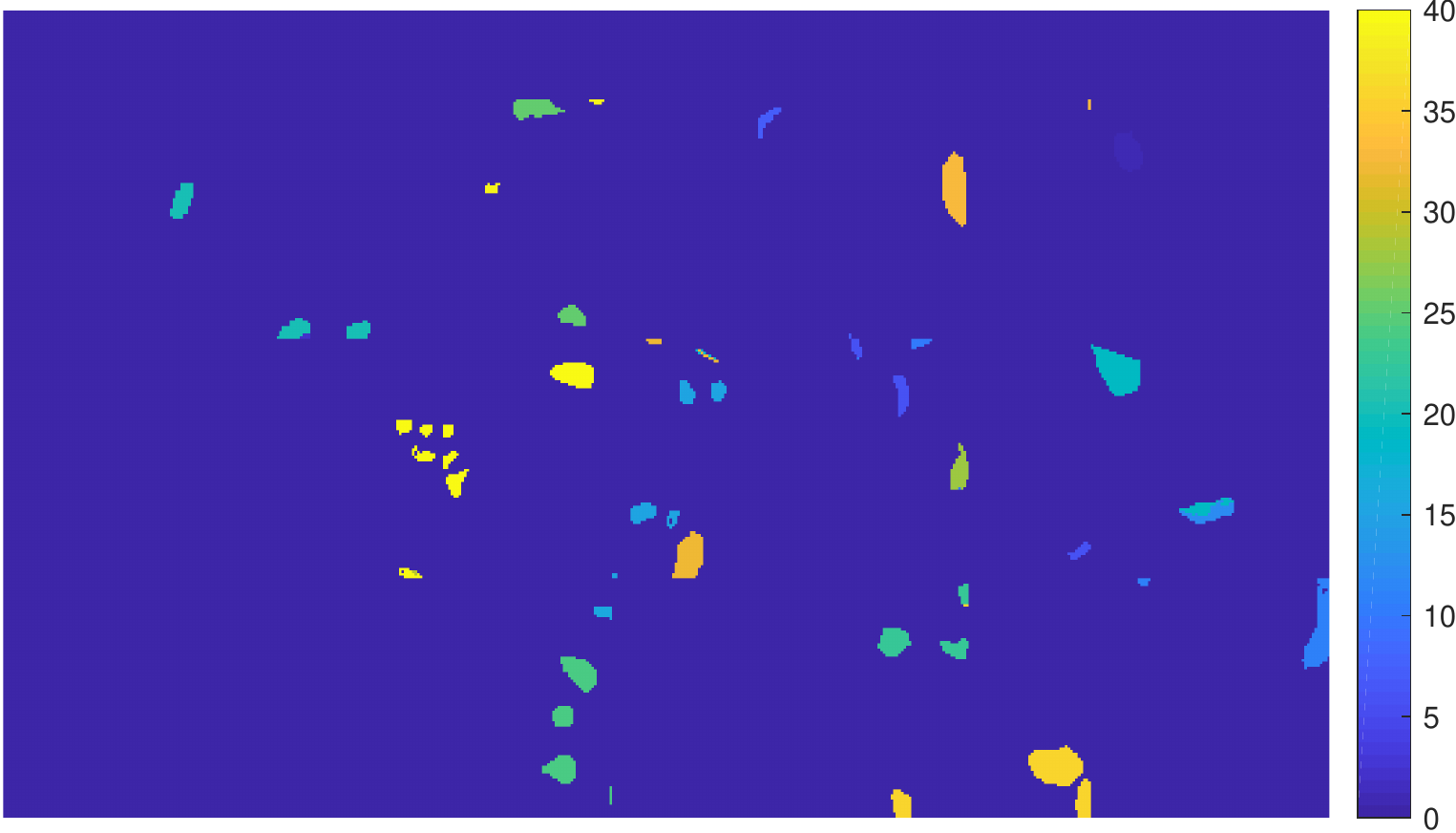}
\caption{DLSS}
\end{subfigure}
\caption{Results of clustering individual patches of Kennedy Space Center data, without synchronizing the labels.\label{fig:KSC_Patches}} 
\end{figure}

In addition to providing the basis for a statistical evaluation of the proposed algorithm, splitting large, complicated HSI into patches for clustering allows to over-segment the image.  Examples of the oversegmented maps, where we do not attempt to synchronize the labels across patches, appear in Figures \ref{fig:IP_Patches},  \ref{fig:Pavia_Patches},  \ref{fig:KSC_Patches}.  It is a topic of future research to combine these patches using the DLSS framework.  
\begin{figure*}
\centering
\begin{adjustbox}{max width=\textwidth}
\begin{tabular}{| c | c | c | c | c | c | c | c | c | c  | c | c | c | c ||}\hline
Method & $K$-means & PCA+$K$-means & ICA+$K$-means & RP+$K$-means & DBSCAN & SC  & GMM & SMCE & HNMF & FMS & FSFDPC & DL & DLSS\\ \hline
$t$-statistic & 2.0163 & 2.1039 & 2.7710 & 3.1630 & 6.44774 & 2.6301 & 3.4461 & 4.1093 & 2.2160 & 2.1810 & 2.1219 & 2.1357 & - \\ \hline
\end{tabular}
\end{adjustbox}
\caption{\label{tab:StatisticalAnalysis}For each of the $i$ comparison methods ($i=1,\dots,12$), DLSS ($j=13)$ is compared against method $i$ by computing the $t$-statistic score $\overline{\Delta^{i,j}}/(\sigma^{i,j}/\sqrt{\Npatches-1})$.  All of the $t$-statistics are significant enough to reject at the $95\%$ level the two-sided null hypothesis that the results of DLSS do not differ significantly from those of method $i$, corresponding to the $t$-statistic exceeding $1.9934$ when using $\Npatches-1=72$ degrees of freedom.} 
\end{figure*}

\section{Overall Comments on the Experiments and Conclusion}
\label{subsec:CC}

We proposed a novel unsupervised method for clustering HSI, using data-dependent diffusion distances to learn modes of clusters, followed by a spectral-spatial labeling scheme based on diffusion in both the spectral and spatial domains.  We demonstrated on various data sets that the proposed DLSS algorithm performs well compared to state-of-the-art techniques, and that the DLSS algorithm outperforms DL thanks to the incorporation of spatial information.  We remark that the methods which employ linear dimension reduction, including PCA, ICA, and random projections, generally outperform methods that use no dimension reduction, but do not perform as well as those which used nonlinear dimension reduction, including spectral clustering, SMCE, DL, and DLSS.  This indicates that while HSI data does exhibit intrinsically low-dimensional structure, the data lies close not to subspaces, but manifolds, i.e. nonlinear sets of low dimensionality.

The proposed DL method, consisting of the geometric learning of modes but only spectral assignment of labels, largely outperforms all comparison methods (see Table \ref{tab:Summary}).  In particular, it outperforms in all examples considered the very popular and recent FSFDPC algorithm.  This indicates that Euclidean distance is insufficient for learning the modes of complex HSI data.  Moreover, the joint spectral-spatial labeling scheme DLSS improves over DL in all instances.  In fact, DLSS gives the overall best performance for all datasets and all performance metrics.

The incorporation of active learning in the DLSS algorithm dramatically improves the accuracy of labeling of the Indian Pines, Pavia, and Salinas A datasets with very few label queries.  This parsimonious use of training labels has the potential to greatly improve the efficiency of machine learning tasks for HSI, in which the number of labels necessary to label a significant proportion of the image is very high.  The proposed active learning method can perform competitively with the state-of-the-art supervised EPF spectral-spatial classification algorithm, using a fraction of the number of labeled pixels.

\subsection{Computational Complexity and Runtime}
\label{subsec:CC}
Let the data be $X=\{x_{n}\}_{n=1}^{N}\subset\mathbb{R}^{D}$.  For the Indian Pines dataset, $N=1250, D=200$; for the Pavia dataset, $N=13500,D=102$; for the Salinas A dataset, $N=7138, D=224$; for the Kennedy Space Center dataset, $N=25000, D=176$.  The most expensive step in DLSS is the construction of the nearest neighbor graph: we achieve near-linear scaling in $N$, $O(C_{d}DN\log N+k_{1}DN)$, using the cover trees algorithm \cite{LangfordICML06-CoverTree} with $C_{d}$ a constant that depends exponentially on the intrinsic dimension $d$ of the data, which is quite small in all the data sets considered. Once the nearest neighbors are found, the kernel density estimator, the random walk, and its eigenvectors can all be quickly constructed in time $O(N\log N)$, assuming that the number of nearest neighbors used in the density estimator is $O(\log N)$ and that the number of eigenvectors needed is $O(1)$.  Computing the nearest spectral neighbor of higher empirical density, computing the spatial consensus labels, and active learning respectively have negligible computational complexity.  We show empirical runtimes in Table \ref{tab:RunTime}, which demonstrates that the proposed methods have superior runtimes to spectral clustering and DBSCAN, and are substantially faster than SMCE.

\begin{table}[h]
\centering
\begin{adjustbox}{max width=.49\textwidth}
\begin{tabular}{| c | c | c | c | c |}\hline
Method & IP & Pavia & Salinas A & KSC \\ \hhline{|=|=|=|=|=|}
$K$-means & 0.44 & 3.89 & 1.26 & 5.97\\ \hline
PCA+$K$-means & 0.01 & 0.01 & 0.01 & 0.16 \\ \hline
ICA+$K$-means & 0.22 & 0.87 & 0.30 & 1.29 \\ \hline
RP+$K$-means & 0.11 & 0.79 & 0.13 & 0.58 \\ \hline
DBSCAN  & 0.58 & 56.10 & 13.53 & 112.43 \\ \hline
SC & 0.77 & 101.20 & 13.87 & 483.56 \\ \hline
GMM &  0.40 & 3.07 & 1.91 & 2.16 \\ \hline
SMCE  & 11.74  & 466.90 & 222.40 & 1315.21 \\ \hline
HNMF & 0.52 & 0.93 & 0.74 & 1.41 \\ \hline
FMS & 0.15  & 0.73 & 0.29 & 0.89 \\ \hline
FSFDPC & 1.48 & 33.64 & 10.05 & 69.46 \\ \hline
DL  & 0.80 & 36.46 & 8.73 & 80.28  \\ \hline
DLSS  & 1.40 & 73.77 & 12.24 & 106.05 \\ \hline
\end{tabular}
\end{adjustbox}
\caption{\label{tab:RunTime}Run times for each method and each dataset, measured in seconds.  The linear dimension reduction methods are extremely fast, as are NMF and GMM.  The spectral clustering and FSFDPC algorithms are slower than DL, and DLSS is slightly slower is slightly slower than DL.  The SMCE algorithm is substantially slower.  Note that although FMS is quite fast, it is implemented in parallelized C++ code and ran on a machine with 24 cores and 48 threads.}
\end{table}


\section{Future Research Directions}\label{sec:Future}

\begin{figure}
\centering
\includegraphics[width=0.11\textwidth]{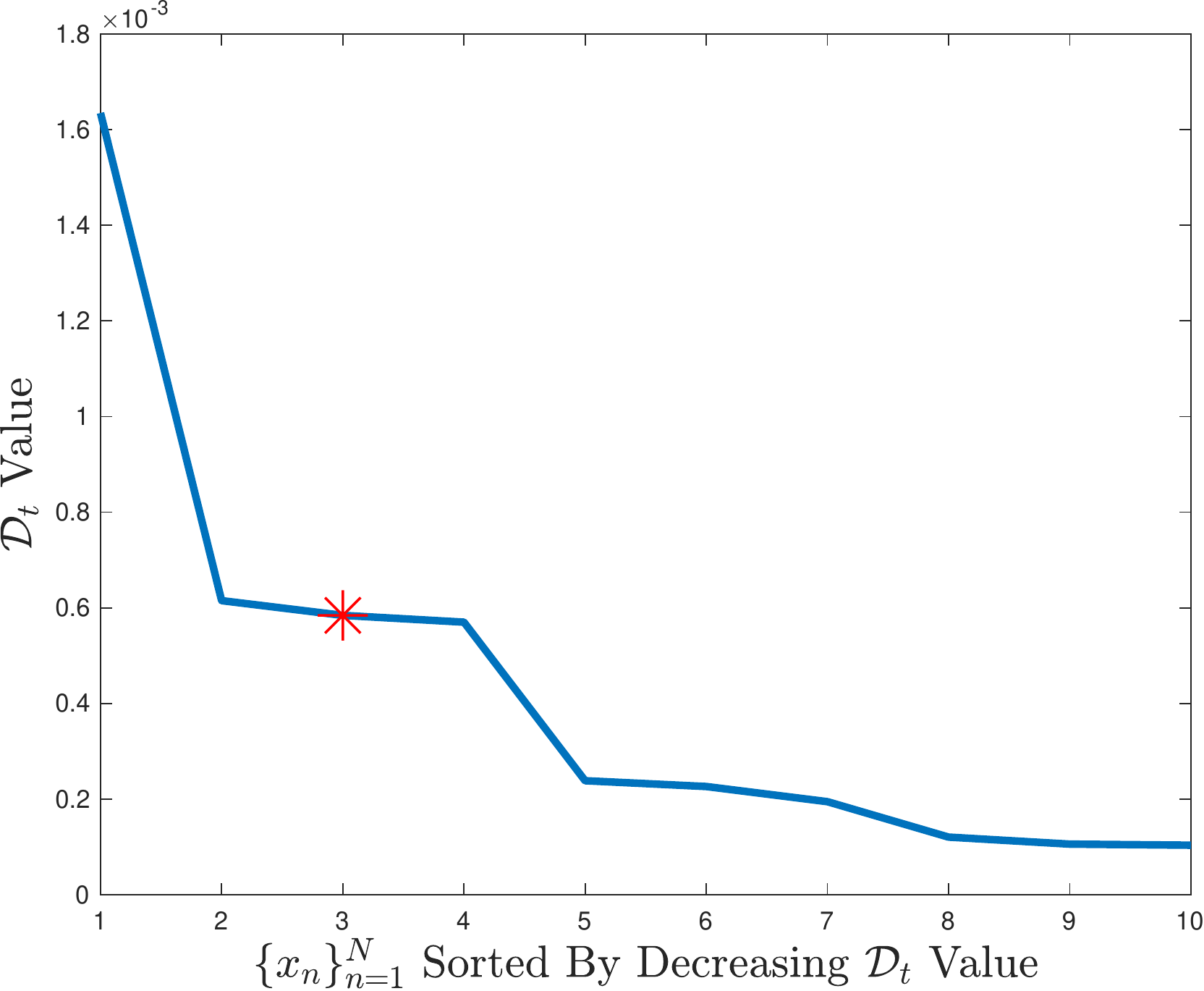}
\includegraphics[width=0.11\textwidth]{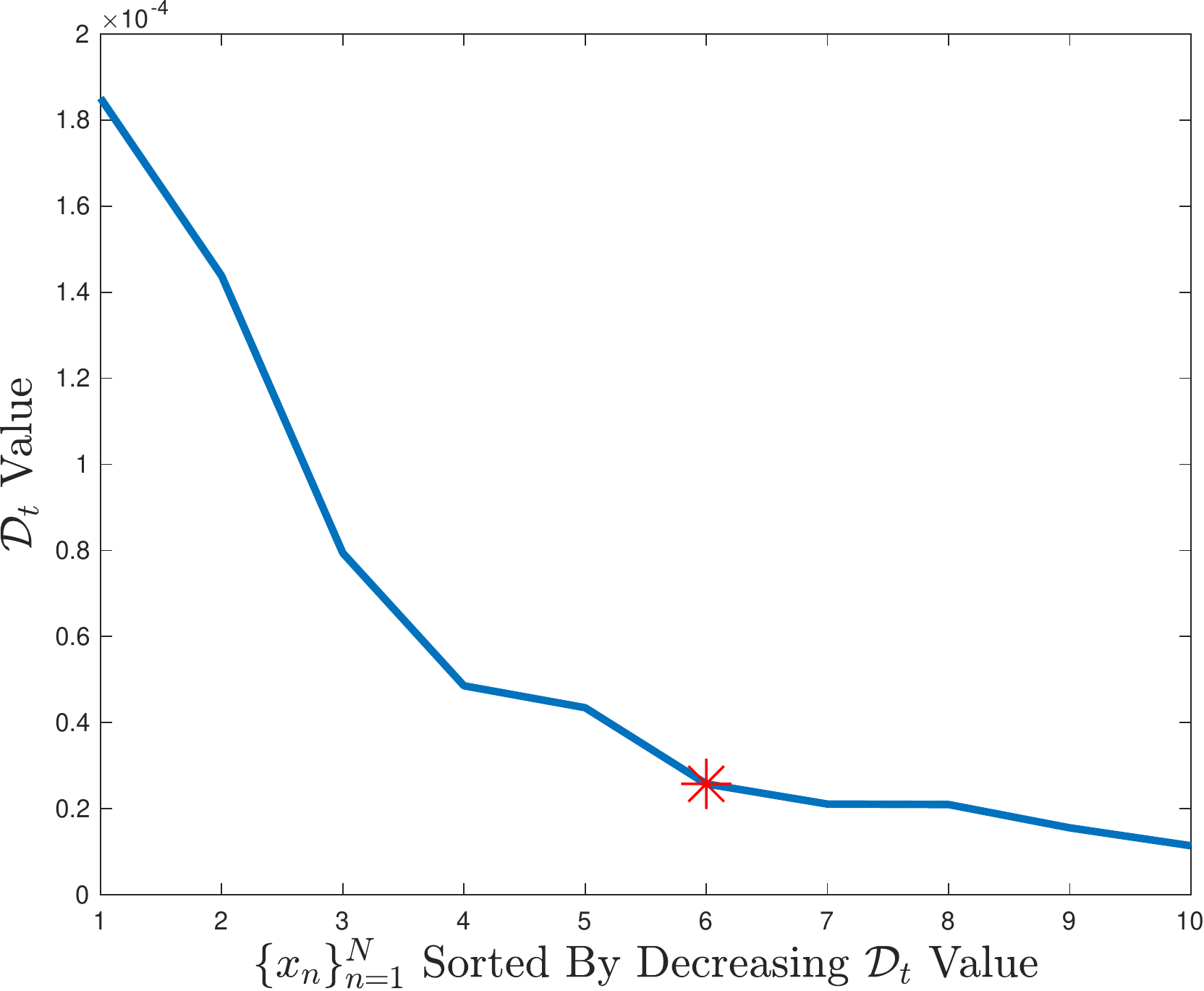}
\includegraphics[width=0.11\textwidth]{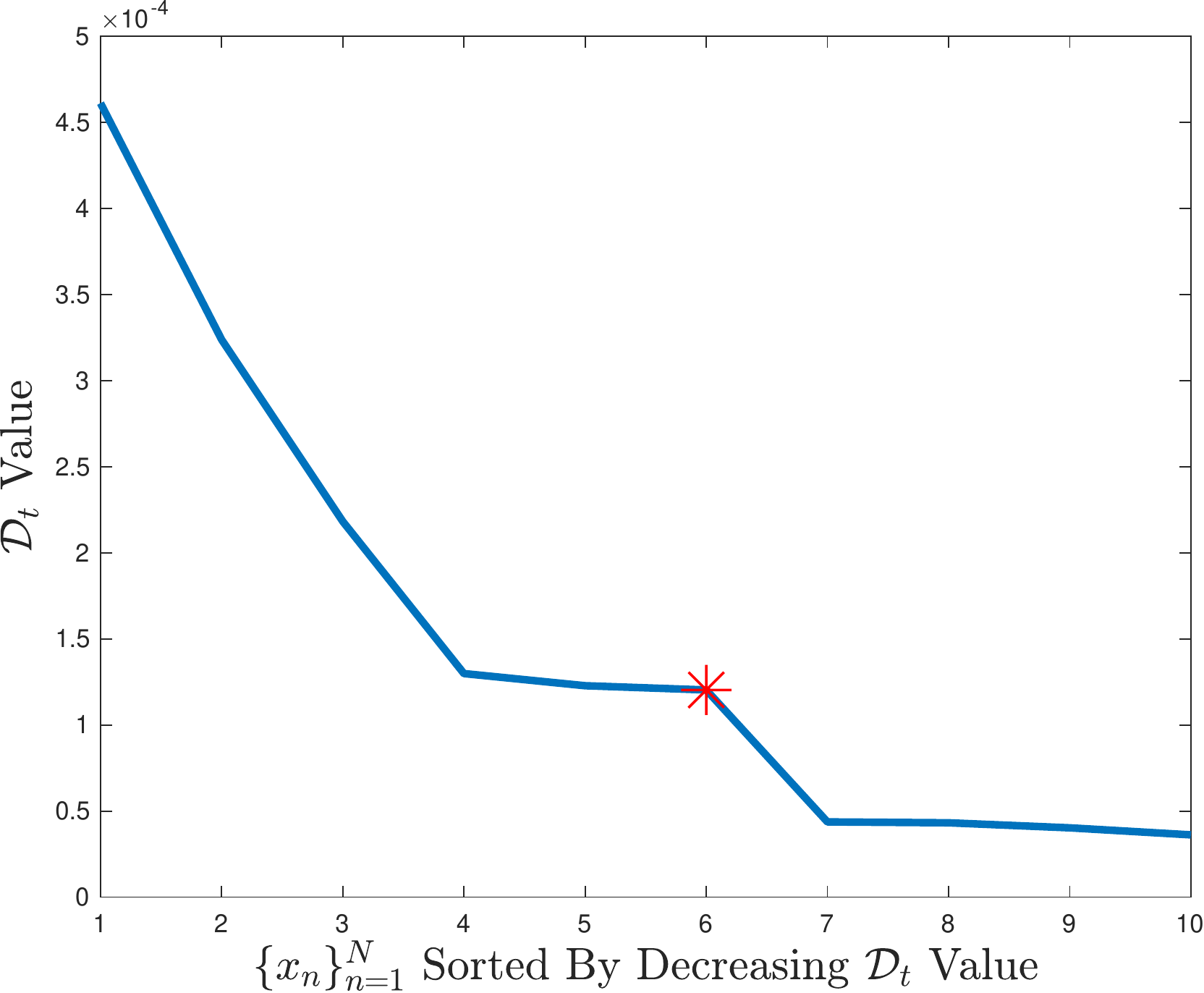}
\includegraphics[width=0.11\textwidth]{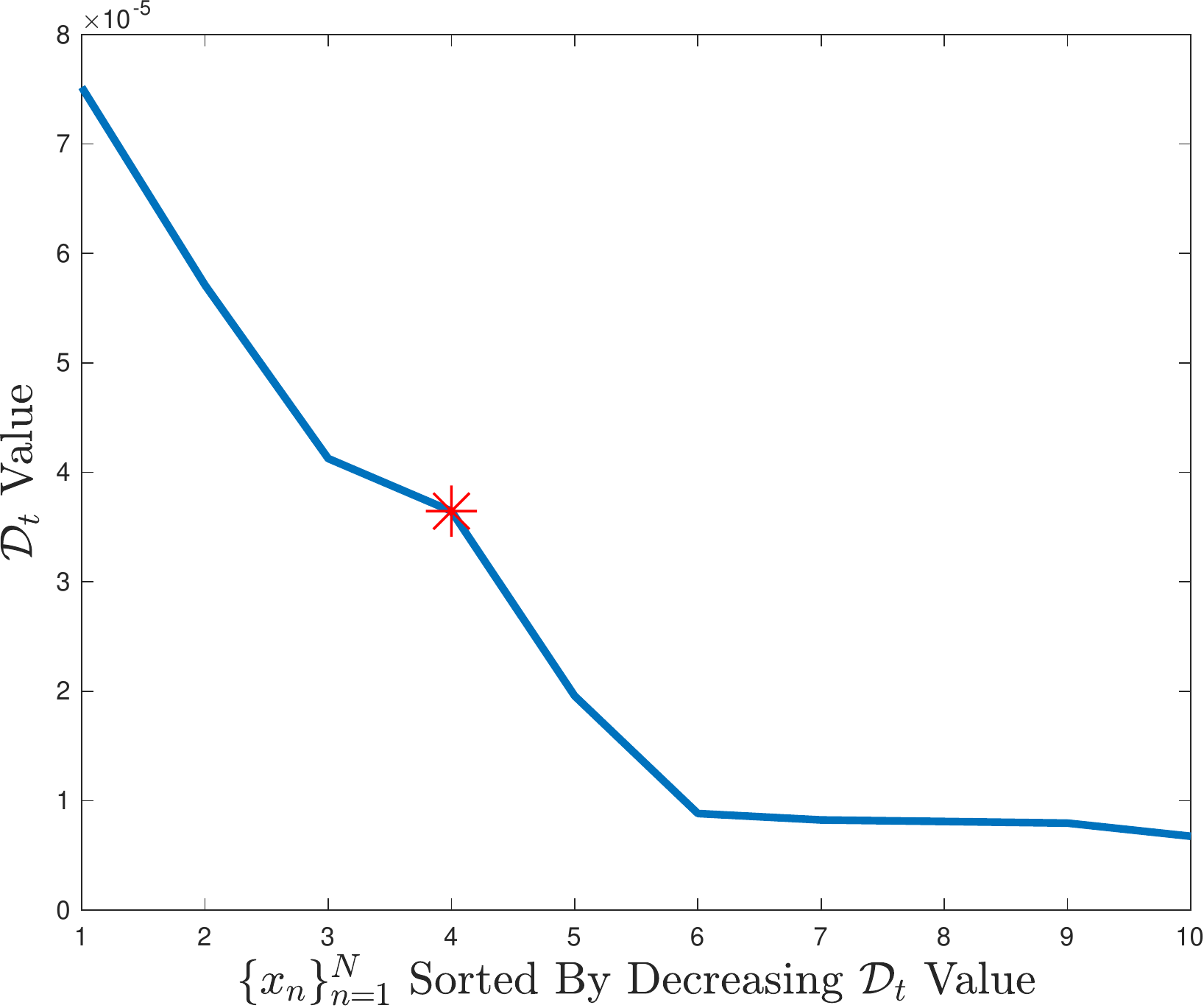}
\caption{\label{fig:KinkAnalysis}The sorted $\mathcal{D}_{t}(x_{i}^{\text{sorted}})$ values for each of the four datasets. From left to right: Indian Pines, Pavia, Salinas A, KSC.  The estimate $\hat{K}$ is shown as a red star.} 
\end{figure}

\begin{figure}
\centering
\includegraphics[width=0.11\textwidth]{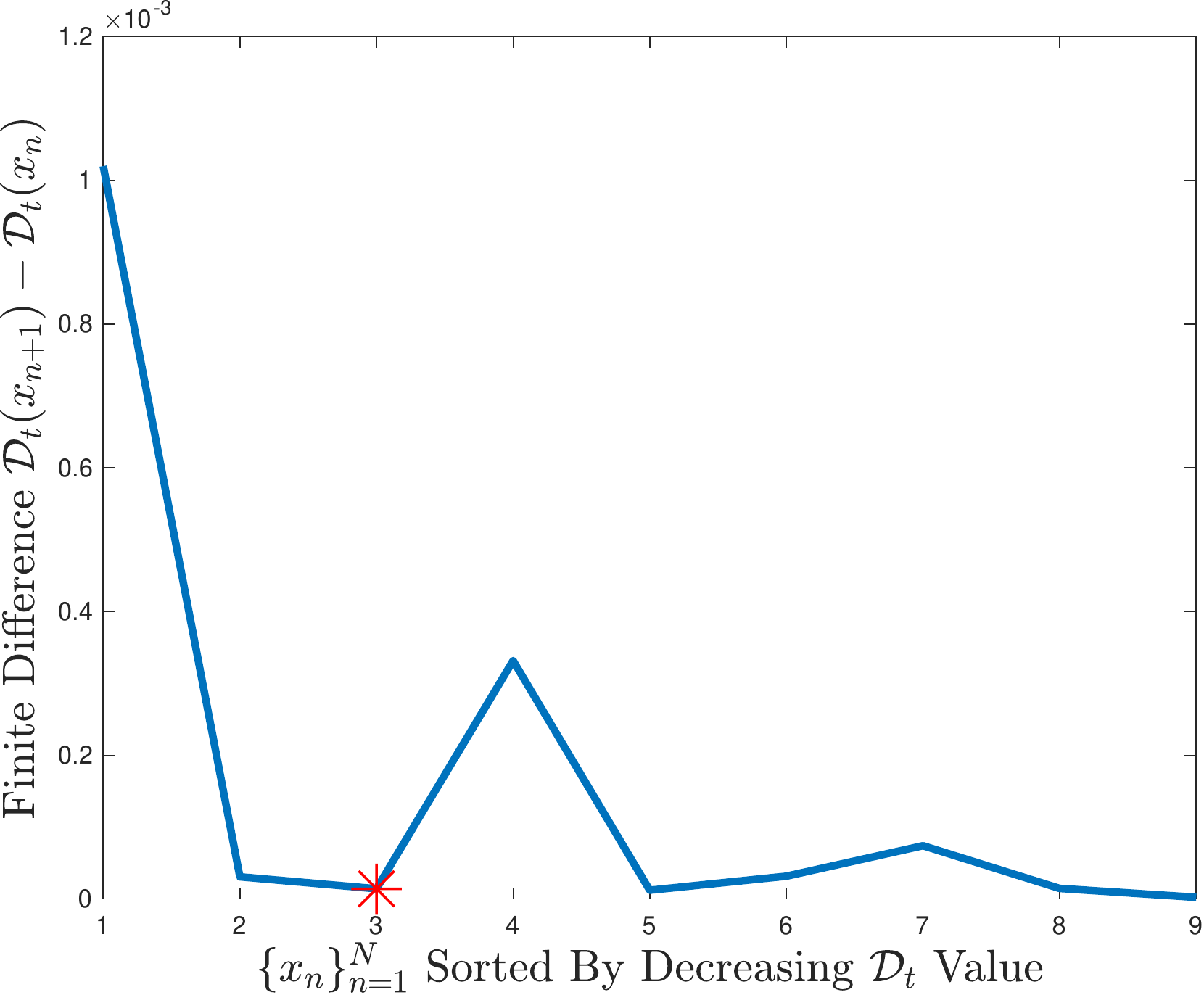}
\includegraphics[width=0.11\textwidth]{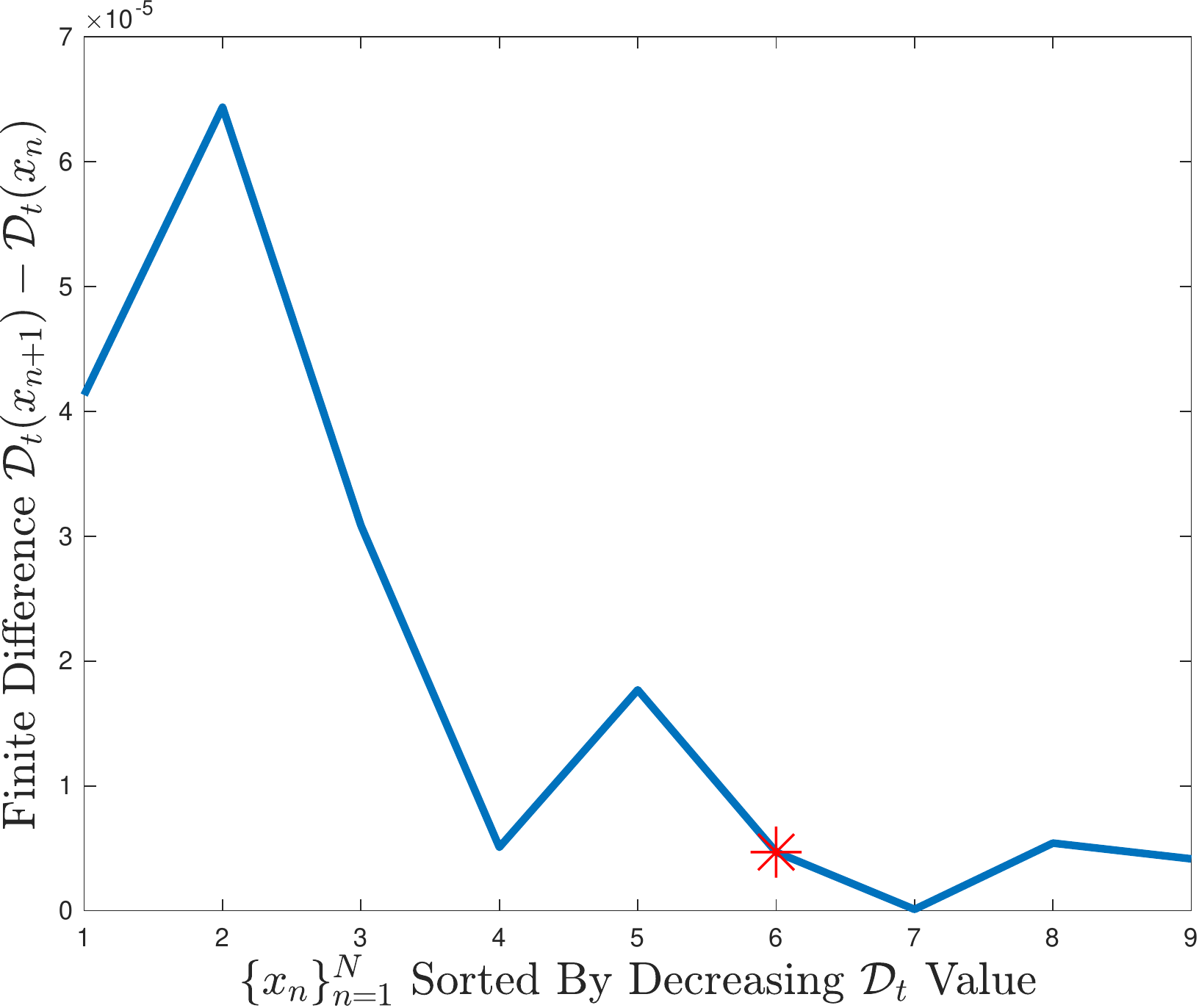}
\includegraphics[width=0.11\textwidth]{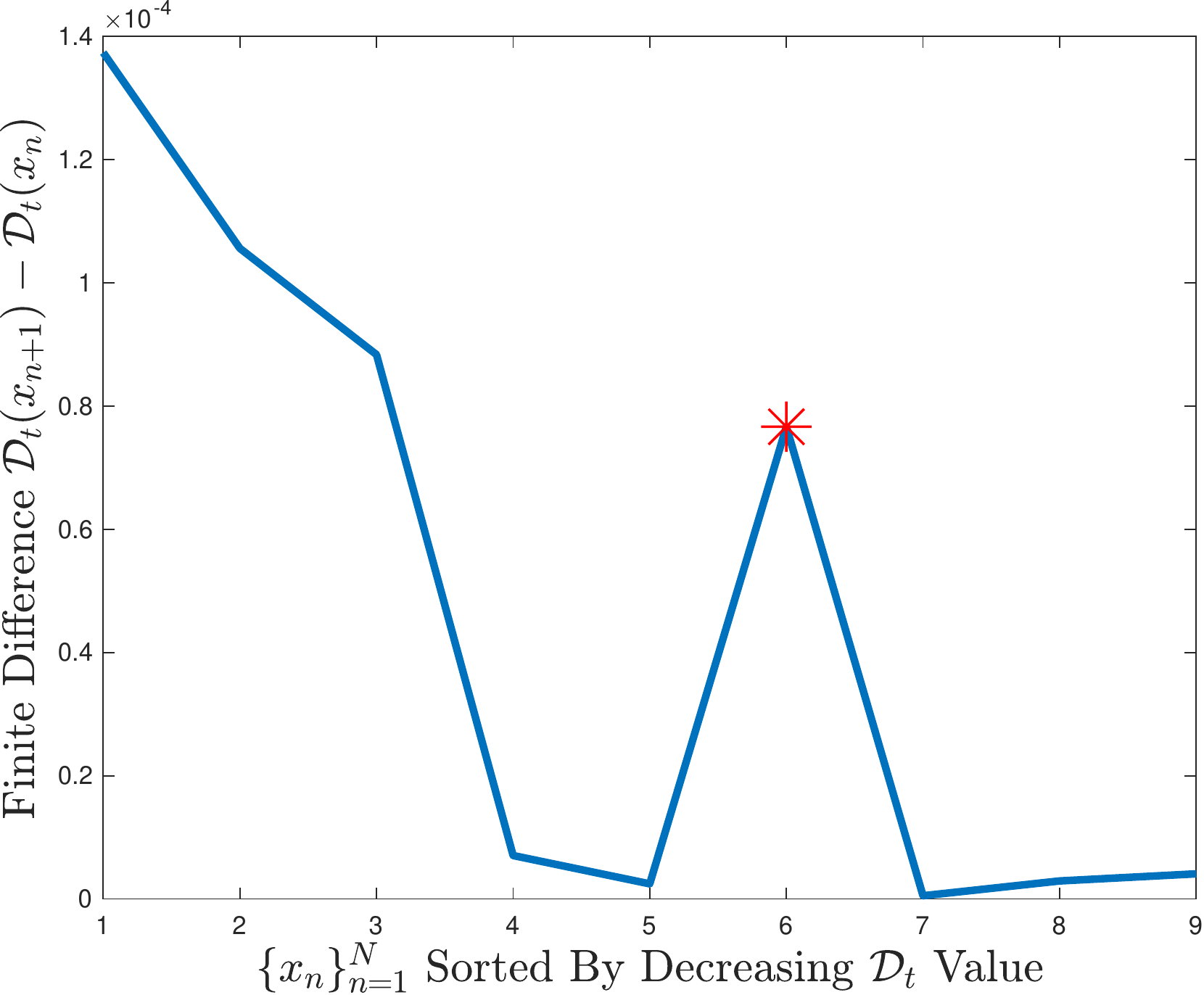}
\includegraphics[width=0.11\textwidth]{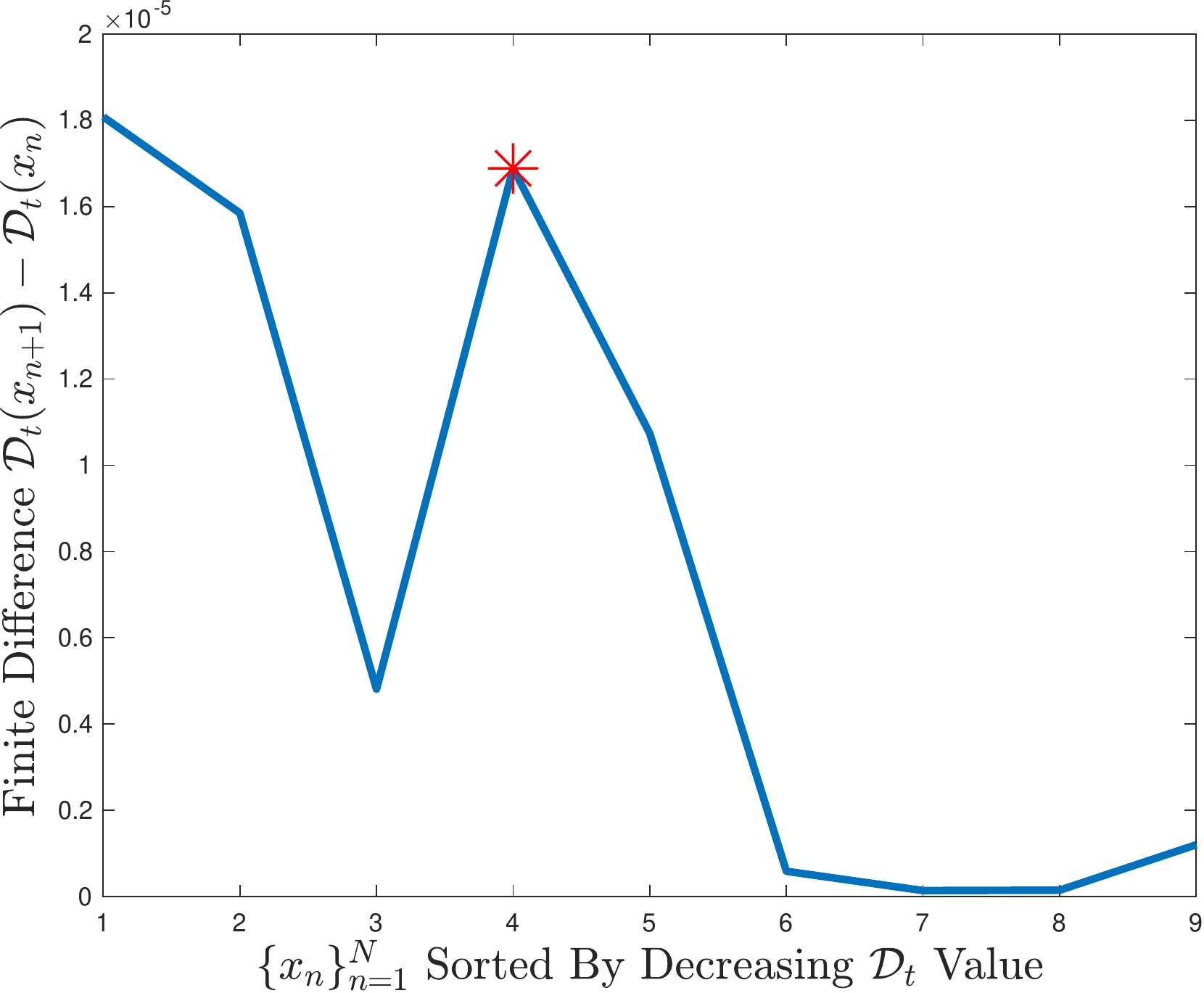}\\
\begin{subfigure}{ .11\textwidth}
\includegraphics[width=\textwidth]{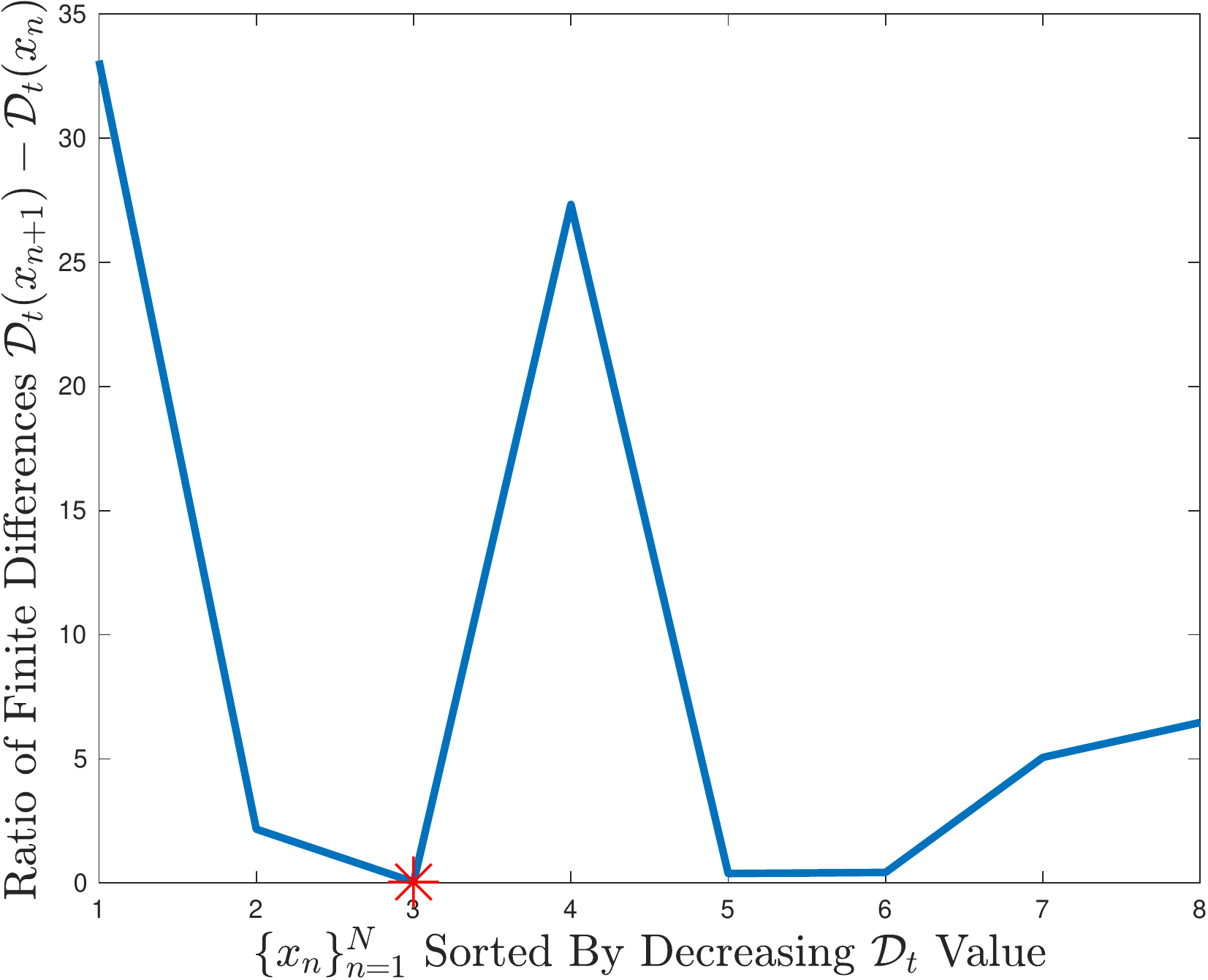}
\subcaption{Indian Pines: first order estimate $\hat K$ inconclusive (top); second order $\hat{K}=4$ incorrect (bottom).}
\end{subfigure}
\begin{subfigure}{ .11\textwidth}
\includegraphics[width=\textwidth]{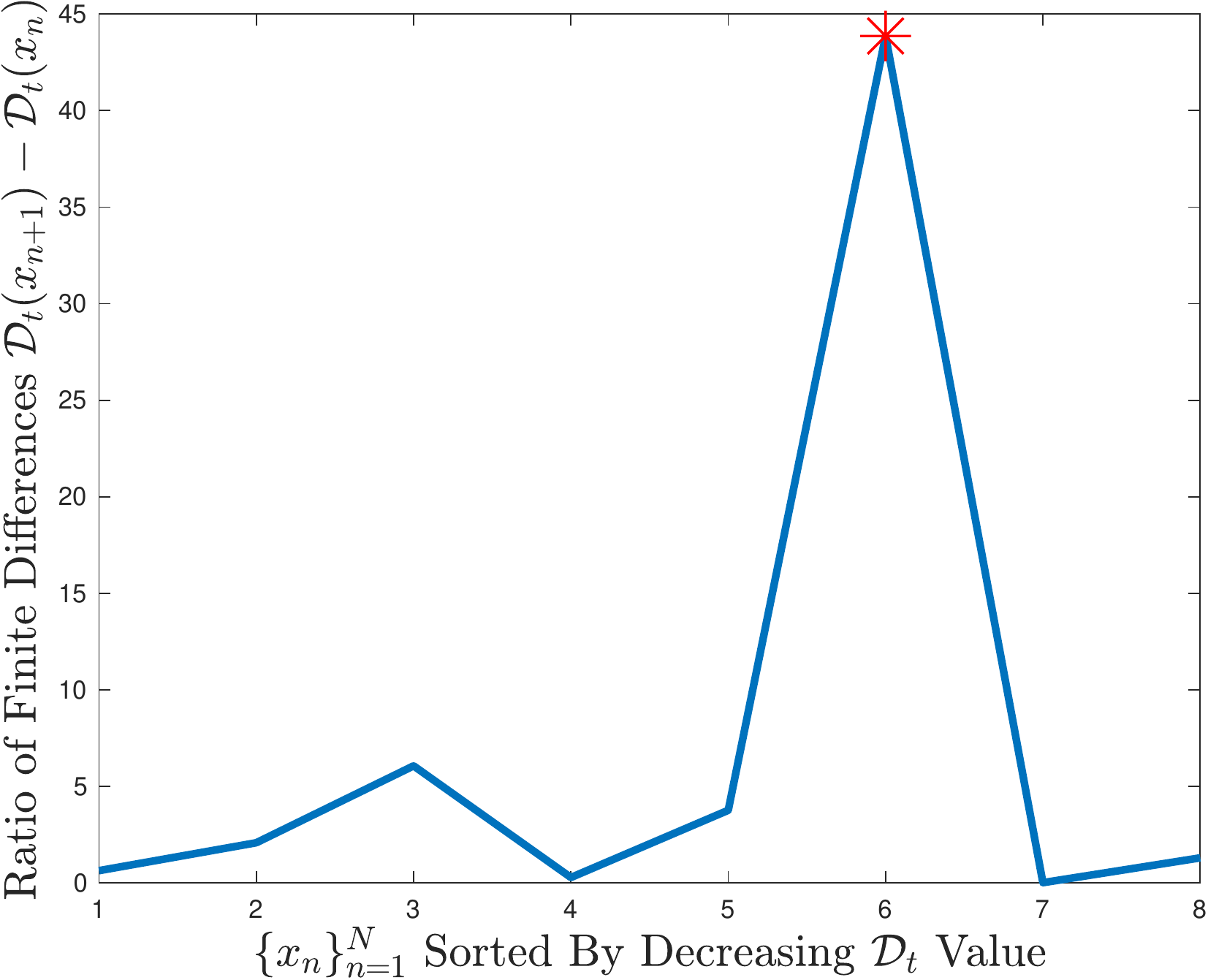}
\subcaption{Pavia: first order estimate $\hat K$ inconclusive (top); second order estimate $\hat{K}=6$ correct (bottom).}
\end{subfigure}
\begin{subfigure}{.11\textwidth}
\includegraphics[width=\textwidth]{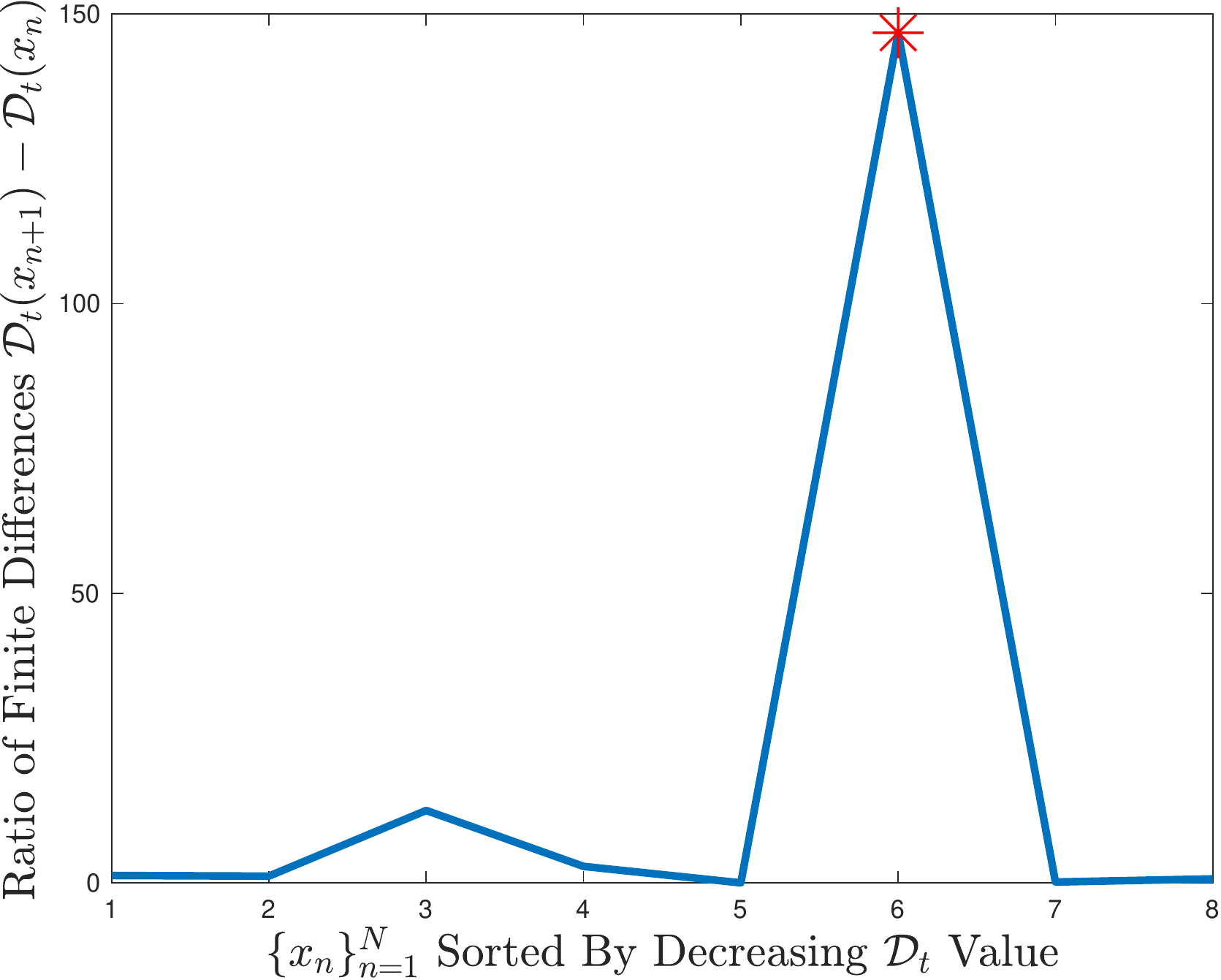}
\subcaption{Salinas A: first order $\hat{K}=6$ correct (top); second order $\hat{K}=6$ correct but not used (bottom).}
\end{subfigure}
\begin{subfigure}{.11\textwidth}
\includegraphics[width=\textwidth]{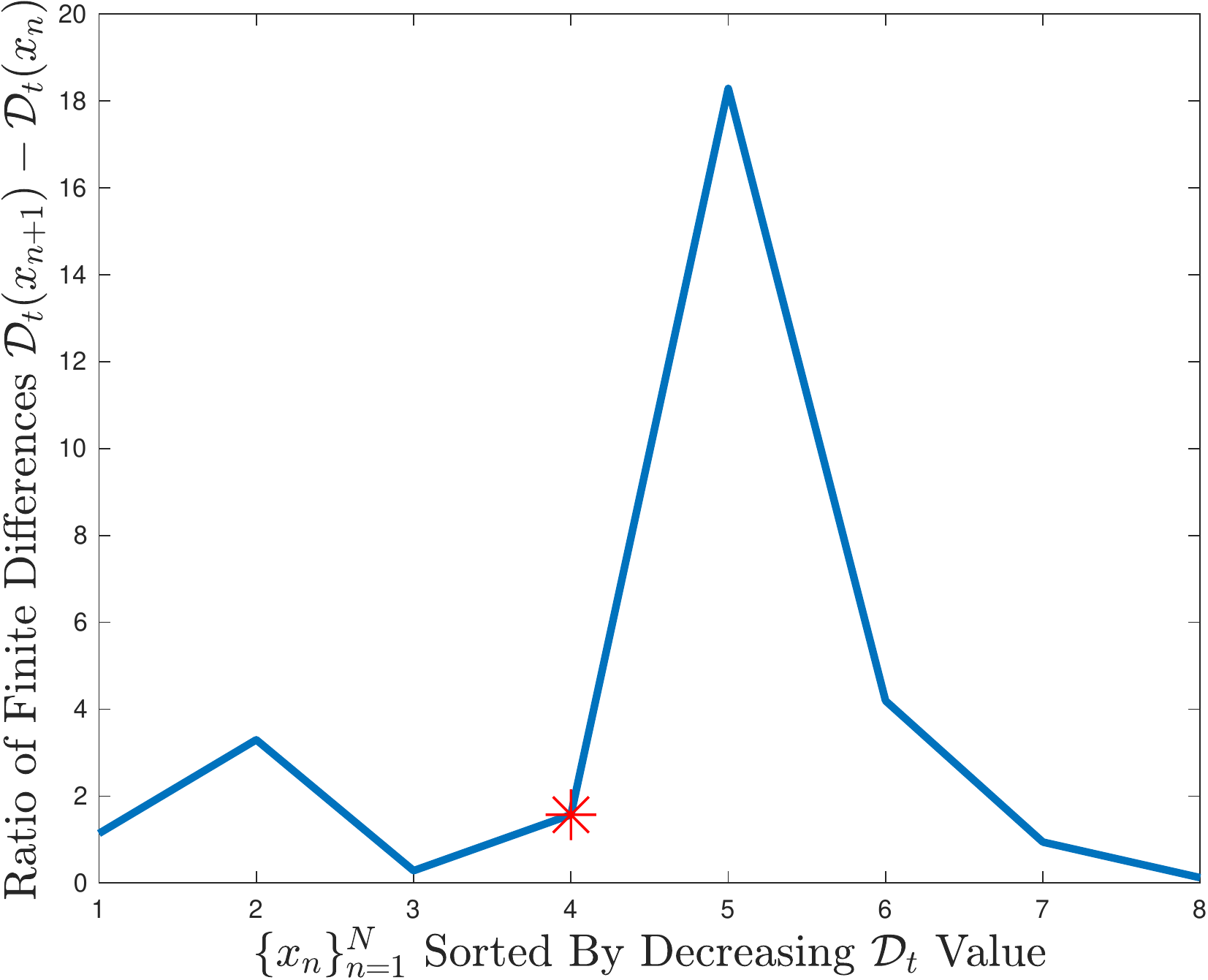}
\subcaption{KSC: first order $\hat{K}=4$ correct (top); second order $\hat{K}=5$ incorrect but not used (bottom).}
\end{subfigure}
\caption{\label{fig:DiffRatioAnalysis}We show the plots of $\{\mathcal{D}_{t}(x^{\text{sort}}_{i+1})-\mathcal{D}_{t}(x^{\text{sort}}_{i})\}$ (top row) and the ratios $\{({\mathcal{D}_{t}(x^{\text{sort}}_{i+1})-\mathcal{D}_{t}(x^{\text{sort}}_{i})})/({\mathcal{D}_{t}(x^{\text{sort}}_{i+2})-\mathcal{D}_{t}(x^{\text{sort}}_{i+1})}\})$ (bottom row) for each of the four HSI datasets considered in this article.  The true number of classes is shown with a red star: the proposed method of estimating $K$ is accurate for all the datasets except Indian Pines.  We see that the first order information correctly determines that there are 6 clusters in Salinas A and 4 clusters in the KSC HSI, owing to the prominent peaks.  The first order information is ambiguous for Indian Pines and Pavia, since there are no prominent peaks.  The second order information correctly estimates that there are 6 clusters in the Pavia data, and incorrectly estimates 4 clusters for Indian Pines.}
\end{figure}

A drawback of many clustering algorithms, including the ones presented in this paper, is the assumption that the number of clusters, $K$, is known a priori.  While unsupervised clustering experiments typically assume $K$ is known, it is of interest to develop methods that allow efficient and accurate estimation of $K$, in order to make a truly unsupervised clustering method.  Initial investigations suggest that looking for the ``kink" in the sorted plot of $\mathcal{D}_{t}(x_i)$ could be used to detect $K$ automatically.  More precisely, we check if there is a prominent peak in the value $\mathcal{D}_{t}(x^{\text{sort}}_{i+1})-\mathcal{D}_{t}(x^{\text{sort}}_{i})$, where where the points $\{x_{i}^{\text{sort}}\}_{i=1}^{n}$ are the data, sorted in decreasing order of their $\mathcal{D}_{t}$ values.  This is a discrete version of the gradient, so we are looking for a sharp drop-off in the sorted $\mathcal{D}_{t}$ curve.  If there is a prominent such maxima in $\mathcal{D}_{t}(x^{\text{sort}}_{i+1})-\mathcal{D}_{t}(x^{\text{sort}}_{i})$, precisely defined as a local maximum that is greater in magnitude than double the previous value, and also at least half the magnitude of the global maximum, we estimate $\hat{K}$ as this peak.  If there is no such prominent peak, then we proceed to examine the second order information $(\mathcal{D}_{t}(x^{\text{sort}}_{i+1})-\mathcal{D}_{t}(x^{\text{sort}}_{i}))/(\mathcal{D}_{t}(x^{\text{sort}}_{i+2})-\mathcal{D}_{t}(x^{\text{sort}}_{i+1}))$.  This is a discrete approximation to the second derivative of $\mathcal{D}_{t}$, to find when $\mathcal{D}_{t}$ begins to flatten.  Initial analysis on the Indian Pines, Pavia, Salinas A and Kennedy Space Center datasets used in this article confirm the promise of analyzing the decay of $\mathcal{D}_{t}(x_i)$; results showing plots of $\{\mathcal{D}_{t}(x_{i}^{\text{sort}})\}$ values appear in Figure \ref{fig:KinkAnalysis}, while the corresponding statistics $\{\mathcal{D}_{t}(x_{i+1})-\mathcal{D}_{t}(x_{i})\}$ and  $\{({\mathcal{D}_{t}(x^{\text{sort}}_{i+1})-\mathcal{D}_{t}(x^{\text{sort}}_{i})})/({\mathcal{D}_{t}(x^{\text{sort}}_{i+2})-\mathcal{D}_{t}(x^{\text{sort}}_{i+1})})\}$ are shown in Figure \ref{fig:DiffRatioAnalysis}.  The estimated number of clusters $\hat K$ appear in Table \ref{tab:EstimatedK}.

\begin{table}
\centering
\begin{tabular}{| c | c | c | c | c | c |}\hline
Dataset & IP & Pavia & Salinas A & KSC \\ \hline
Estimated Number of Classes $\hat{K}$ & 4 & 6 & 6 & 4 \\ \hline
Number of Labeled GT Classes $K$ & 3 & 6 & 6 & 4 \\ \hline
\end{tabular}
\caption{\label{tab:EstimatedK}We show the number of classes estimated by looking for the ``kink" in the sorted plot of $\mathcal{D}_{t}(x_n)$.  We see that for the Salinas A and Kennedy Space Center datasets, estimating based on  $\{\mathcal{D}_{t}(x^{\text{sort}}_{i+1})-\mathcal{D}_{t}(x^{\text{sort}}_{i})\}$ correctly estimates the number of labeled classes, while this statistic is inconclusive for Indian Pines and Salinas A.  For these data, we move to second order information, namely estimating $K$ by maximizing $({\mathcal{D}_{t}(x^{\text{sort}}_{i+1})-\mathcal{D}_{t}(x^{\text{sort}}_{i})})/({\mathcal{D}_{t}(x^{\text{sort}}_{i+2})-\mathcal{D}_{t}(x^{\text{sort}}_{i+1})})$.  This estimator overestimates the number of classes for Indian Pines, estimating 4 instead of 3, while it correctly estimates the number of classes in Pavia.  The Indian Pines dataset is the most challenging of the four labeled datasets analyzed, which suggests that the the proposed method for estimating $K$ may be insufficient for challenging HSI data.}
\end{table}

It is of interest to prove under what assumptions on the distributions and mixture model the plot $\mathcal{D}_{t}(x_{n})$ correctly determines $K$.  Moreover, in the case that one cluster is noticeably smaller or harder to detect than others, as in the case of the Indian Pines dataset, it may be advantageous to use a different statistic on the finite difference curve, rather than the proposed derivative conditions on $\mathcal{D}_{t}$.  Initial mathematical results and more subtle conjectures are proposed in an upcoming article \cite{Murphy2018}. 

Moreover, all unsupervised algorithms considered in this paper struggle with very large HSI scenes consisting of many classes.  This is due to the large variation within clusters compared to the differences between clusters, which leads to genuine classes being split incorrectly; this is a well-known challenge for unsupervised clustering of HSI \cite{Zhu2017}.  In Section \ref{sec:LargeScaleExperiments} it is shown that DLSS is very effective at clustering on different patches of a large HSI.  It remains an open question how to combine the results on these patches into a global clustering, which amounts to determining when to merge clusters learned in distinct patches.  Automatically implementing such mergers with the DLSS framework is a direction of future research.  

The present work is essentially empirical: it is not known mathematically under what constraints on the mixture model the method proposed for learning modes succeeds with high probability.  Besides being of mathematical interest, this would be useful for understanding the limitations of the proposed method for HSI.  To understand this phenomenon rigorously, a careful analysis of diffusion distances for data drawn from a non-parametric mixture model is required, which is related to investigating performance guarantees for spectral clustering and mode detection \cite{Genovese2016, Schiebinger2015}.

It is also of interest to explicitly incorporate spectral-spatial features into the diffusion construction.  It is known that use of spectral-spatial features is beneficial for supervised learning of HSI \cite{Kang2014,Ghamisi2016,Kang2017}, and their use in unsupervised learning is an exciting research direction.  Indeed, incorporating the spatial properties of the scene into the graph from which diffusion distances are generated may render the explicit spatial regularization step of the proposed algorithm redundant, thus improving runtime.  


\section*{Acknowledgments}
The authors would like to thank Ed Bosch for his helpful comments on a preliminary version of this paper. This research was partially supported by NSF-ATD-1222567, NSF-ATD-1737984, AFOSR FA9550-14-1-0033, AFOSR FA9550-17-1-0280, NSF-IIS-1546392, and ARO subcontract to W911NF-17-P-0039.  We would also like to thank the anonymous reviewers of this article, whose valuable comments significantly improved the presentation and content of this article.  


\bibliographystyle{unsrt}
\bibliography{HSI_LearningIEEE.bib}


\end{document}